\documentclass[preprint,12pt]{elsarticle}

\makeatletter
\def\ps@pprintTitle{%
  \let\@oddhead\@empty
  \let\@evenhead\@empty
  \def\@oddfoot{\hfil\thepage\hfil}%
  \let\@evenfoot\@oddfoot}
\makeatother
\usepackage[T1]{fontenc}
\usepackage[utf8]{inputenc}
\usepackage{lmodern}
\usepackage{microtype}

\usepackage[a4paper,left=2.3cm,right=2.3cm,top=2.6cm,bottom=2.6cm]{geometry}

\usepackage{graphicx}
\usepackage{epstopdf}          
\usepackage{float}             

\usepackage{amsmath}
\usepackage{amssymb}
\usepackage{amsfonts}
\usepackage{amsthm}
\usepackage{mathtools}
\usepackage{bm}

\usepackage{xcolor}
\usepackage{booktabs}
\usepackage{multirow}
\usepackage{array}
\usepackage{tabularx}
\usepackage{caption}
\usepackage{enumitem}

\usepackage{url}
\usepackage[hidelinks]{hyperref}   

\usepackage{tcolorbox}
\tcbuselibrary{skins,breakable}

\definecolor{greenA}{RGB}{88,150,65}
\definecolor{greenB}{RGB}{231,245,224}
\definecolor{blueA}{RGB}{54,125,205}
\definecolor{blueB}{RGB}{229,241,255}
\definecolor{redA}{RGB}{190,55,45}
\definecolor{redB}{RGB}{255,232,230}
\definecolor{purpleA}{RGB}{135,75,210}
\definecolor{purpleB}{RGB}{242,232,255}
\definecolor{orangeA}{RGB}{220,105,35}
\definecolor{orangeB}{RGB}{255,239,228}
\definecolor{yellowA}{RGB}{225,170,20}
\definecolor{yellowB}{RGB}{255,248,220}
\definecolor{cyanA}{RGB}{40,150,160}
\definecolor{cyanB}{RGB}{225,247,248}
\definecolor{tealA}{RGB}{47,127,116}
\definecolor{tealB}{RGB}{228,244,241}
\definecolor{roseA}{RGB}{192,90,138}
\definecolor{roseB}{RGB}{252,234,242}

\newtcolorbox{takeawaybox}[2][greenA]{%
enhanced,
colback=#1!8!white,
colframe=#1,
colbacktitle=#1,
coltitle=white,
title={\bfseries\sffamily #2},
fonttitle=\bfseries\large,
arc=5pt,
boxrule=1.2pt,
titlerule=0.8pt,
top=6pt, bottom=6pt,
left=10pt, right=10pt,
toptitle=4pt, bottomtitle=4pt,
attach boxed title to top left={yshift=-2mm, xshift=6mm},
boxed title style={
arc=3pt,
boxrule=0.8pt,
colframe=#1,
},
before upper={\setlength{\parskip}{2pt}},
}

\begin{document}

\begin{frontmatter}

\title{Fractional Optimizers Meet Fractal Activation Functions: An Empirical Study of Multi-Scale Optimization in Neural Networks}

\author[sba]{Sebastian Raubitzek\corref{cor1}}
\ead{sraubitzek2@sba-research.org}
\ead[url]{https://orcid.org/0000-0003-2206-9263}

\author[sba]{Georg Goldenits}
\ead{ggoldenits@sba-research.org}

\author[univie]{Sebastian Schrittwieser}
\ead{sebastian.schrittwieser@univie.ac.at}

\author[sba]{Philip K\"onig}
\ead{pkoenig@sba-research.org}

\author[sba]{Kevin Mallinger}
\ead{kmallinger@sba-research.org}

\cortext[cor1]{Corresponding author.}

\affiliation[sba]{organization={SBA Research gGmbH},
                  addressline={Floragasse 7, 5th Floor},
                  postcode={1040},
                  city={Vienna},
                  country={Austria}}

\affiliation[univie]{organization={Christian Doppler Laboratory for
                  Assurance and Transparency in Software Protection,
                  Research Group Security \& Privacy, Faculty of
                  Computer Science, University of Vienna},
                  addressline={Kolingasse 14--16},
                  postcode={1090},
                  city={Vienna},
                  country={Austria}}

\begin{abstract}
Fractional optimization methods and fractal activation functions are two independent directions for improving neural network training. Fractional optimizers extend first-order optimization through fractional derivatives and memory effects, whereas fractal activations introduce multi-scale nonlinear representations based on self-similar Weierstrass- and Blancmange-type functions. Here, we investigate their interaction within a unified experimental framework. We evaluate fractional optimizer families on Ackley and Himmelblau benchmark surfaces, in standard form and with additive Weierstrass-type perturbations, and then in feed-forward neural networks with conventional and fractal activations on ten classification datasets. The comparison includes standard methods, regularization-style optimizers, explicit and adaptive memory-based fractional optimizers, and other representative literature methods. Overall, fractional optimization and fractal activations show useful but selective pairings. Regularization-style fractional scaling performs well with selected fractal activations in network training, while Gr\"unwald--Letnikov memory is most relevant on perturbed surfaces. Adaptive memory improves plain memory substitution in several cases, supporting controlled fractional memory as a promising direction rather than a universal replacement.
\end{abstract}

\begin{keyword}
fractional calculus \sep
Gr\"unwald--Letnikov derivative \sep
Weierstrass function \sep
fractal activation functions \sep
fractional optimizers \sep
gradient descent \sep
neural networks \sep
multi-scale optimization
\end{keyword}

\end{frontmatter}

\section{Introduction}
\label{sec:introduction}
 
Irregular, multi-scale structure is common in nature. Coastlines, rough
surfaces, turbulent flows, and many physiological and financial signals
exhibit detail on every scale of observation, and fractal geometry was
developed precisely to describe such objects
\cite{mandelbrot1982fractal,falconer1990fractal,peitgen1992chaos,
barnsleynonlinear}. The mathematical prototypes of this behaviour predate
the term fractal by more than a century: Weierstrass constructed a
continuous function that is nowhere differentiable \cite{weierstrass1872},
Hardy determined the parameter range in which this pathology occurs
\cite{hardy1916weierstrass}, and Takagi gave a second, independent example
of the same phenomenon \cite{takagi1903continuous,allaart2011takagi}.
Berry and Lewis later connected the Weierstrass--Mandelbrot variant of
these constructions to physical modelling of scale-free signals
\cite{berry1980weierstrass}, and the dimension theory of their graphs has
been studied in detail \cite{hu1993weierstrass}. Functions of this family
are continuous but rough: each term of their defining series adds
oscillations at a finer scale, and the classical first derivative fails at
every point.
 
Fractional calculus provides the matching analytical concept of behaviour
\emph{between} the integer orders. A fractional derivative of order
$\nu\in(0,1)$ interpolates between the identity ($\nu=0$) and the ordinary
derivative ($\nu=1$); it is a nonlocal operator that weights function
increments over an entire interval rather than at a single point
\cite{oldham1974fractional,podlubny1999fractional,samko1993fractional,
oliveira2014review}. What's interesting for the presented research is that functions that possess no first-order derivative can possess fractional
derivatives of every order below one
\cite{ross1994functions}, and for Weierstrass-type functions Z\"ahle and
Ziezold proved that the fractional degree of differentiability equals the
roughness exponent of the construction: Weyl--Marchaud derivatives of all
lower orders exist, orders above the roughness level do not, and at the
critical order a logarithmically averaged \emph{gradual derivative in the
mean} exists and is constant almost everywhere, together with a
well-conditioned numerical procedure for computing it
\cite{zahle1996fractional}. In short, fractional
derivatives are the instruments that measure fractal roughness at the
correct scale, and this correspondence is developed in detail in
Section~\ref{sec:weierstrass_fractional}.
 
Both concepts have recently entered neural network research, but through
separate doors. On the representation side, earlier research introduced
\emph{fractal activation functions}, a class of computationally stable
activations derived from Blancmange- and Weierstrass-type series, and
showed on ten public classification benchmarks that such activations are
usable in standard training pipelines and can increase expressivity beyond
common choices such as ReLU and $\tanh$ \cite{raubitzek_fractals_2026,
nair2010relu,poole2016,raghu2017}. Independent evidence that fractality is
not foreign to neural networks comes from the observation that the
boundary between trainable and divergent hyper-parameter configurations is
itself a fractal set \cite{sohldickstein2024boundary}. On the optimization
side, a growing family of \emph{fractional optimizers} replaces or extends
the first-order gradient by an operator of non-integer order, either
through Caputo-type rescaling \cite{herrera2022fractional,
herrera2023pytorch} or through explicit Gr\"unwald--Letnikov gradient
memory \cite{zhou2023gl,shin2023accelerating,han2023adaptive}, with
adaptive and tempered variants forming an active research direction
\cite{xiang2025aofgd,chen2024adagl,huang2024mffgd,naifar2026tempered} and
several recent surveys documenting the field \cite{elnady2025survey,
fernandez2025role,VieraMartin2022,raubitzek_fractional_2023}.
 
Despite their shared mathematical root, these two lines have so far been
studied independently. This is the gap addressed here. The present work is
a mathematically motivated continuation of the fractal activation study of
Raubitzek et al.\ \cite{raubitzek_fractals_2026}: if fractal activations
inject multi-scale roughness into the representation and, through
backpropagation, into the gradients, then optimizers whose design
principle is multi-scale memory are the natural counterpart, and the
combination deserves a systematic evaluation. The correspondence is not
merely verbal. As shown in
Section~\ref{sec:weierstrass_fractional}, 
the same Gr\"unwald--Letnikov kernel that discretizes the fractional
derivative of a Weierstrass-type function defines, applied along the
iteration axis, the memory term of the fractional optimizers compared in
this article.
 
The objectives of this study are threefold. First, to evaluate how
fractional optimizers behave on objective landscapes with controlled
fractal structure, using the two standard benchmark surfaces Ackley and
Himmelblau \cite{jamil2013survey}, each in its standard form and in a
variant extended by an additive Weierstrass-type perturbation. Second, to
evaluate the same optimizer families in feed-forward neural networks that
use the strongest fractal activation functions from the predecessor
study, alongside conventional activations, on ten public
classification datasets \cite{vanschoren2014openml}. Third, to introduce
and test an \emph{adaptive memory-fractional} optimizer framework that
keeps the fractional order fixed and instead adapts a bounded trust
coefficient for the memory contribution, with an exact reduction to the
underlying classical optimizer at one boundary of that coefficient.
 
This work contributes the following:
\begin{enumerate}[label=(\roman*), leftmargin=2.2em]
  \item A unified presentation that connects the analysis of fractional
        derivatives of Weierstrass-type functions
        \cite{zahle1996fractional} with the discrete
        Gr\"unwald--Letnikov constructions used in fractional optimizers,
        making explicit why fractal activations and fractional optimizers
        are two realizations of one mathematical mechanism
        (Section~\ref{sec:weierstrass_fractional}).
  \item A new family of adaptive memory-based fractional optimizers
        (AdaptiveMemoryFSGD, -FRMSprop, -FAdam, -FAdadelta) that combines
        a norm-matched fractional-memory gradient with the ordinary
        gradient through a stability- and loss-controlled mixing
        coefficient
        (Section~\ref{sec:fractional_optimizers}).
  \item A systematic empirical comparison of 21 optimizers in five
        groups---classical baselines, Herrera-type fractional,
        memory-based fractional, adaptive memory-based fractional, and
        published related-work methods---first on fractally perturbed
        benchmark surfaces with known minima, and subsequently in neural
        network classification experiments with fractal and conventional
        activations
        (Sections~\ref{sec:surface_experiments}
        and~\ref{sec:nn_experiments}).
\end{enumerate}
 
The remainder of this article is organized as follows.
Section~\ref{sec:related_work} reviews related work.
Section~\ref{sec:intuition_fractals_fractional_derivatives}
gives an informal introduction to fractal functions and fractional
derivatives.
Section~\ref{sec:weierstrass_fractional} develops the mathematical
connection between Weierstrass-type functions and fractional derivatives,
from the continuous theory to the discrete Gr\"unwald--Letnikov form and
its role in automatic differentiation.
Section~\ref{sec:fractal_activations}
defines the four fractal activation functions used in the experiments.
Section~\ref{sec:fractional_optimizers} introduces the 21 employed optimizers.
Section~\ref{sec:surface_experiments} reports the controlled surface
optimization experiments, and Section~\ref{sec:nn_experiments} the neural
network classification experiments. Section~\ref{sec:discussion} discusses
the findings in context, and Section~\ref{sec:conclusion} concludes the article.

\,\par\noindent\textbf{We recognize that the present article is extensive.} However, this scope is necessary to examine the subject in sufficient depth, present the mathematical and intuitive foundations of the investigated techniques, and report the experimental results with appropriate nuance and detail, while concise summaries and highlighted takeaways are provided throughout to support readability and allow practitioners to skip theoretical explanations that are not required for applying the methods. We further provide the full code to reconstruct all experiments and reuse the developed optimizers and activation functions in a corresponding GitHub repository at \url{https://github.com/Raubkatz/FractalAndFractional2026}.

\begin{takeawaybox}[greenA]{Main Takeaways --- Introduction}
  \begin{itemize}[
    label={},
    leftmargin=0em,
    itemindent=0em,
    itemsep=3pt,
    topsep=2pt
  ]
    \item \textbf{Fractal roughness and fractional derivatives are
          mathematically linked.} Weierstrass-type functions provide
          controlled multi-scale roughness, while fractional derivatives
          describe change below the classical first-order threshold.
 
    \item \textbf{Fractal activations and fractional optimizers address
          the same multi-scale problem from different sides.} Fractal
          activations add multi-scale structure to the representation,
          while fractional optimizers introduce memory into the update
          direction.
 
    \item \textbf{This study evaluates the combined framework
          twofold.} It introduces adaptive memory-based fractional
          optimizers and compares 21 optimizers on fractally perturbed
          benchmark surfaces and neural-network classification tasks.
  \end{itemize}
\end{takeawaybox}

\section{Related Work}
\label{sec:related_work}

This section reviews the four research lines connected by the present study: fractal functions in classical analysis, fractional derivatives
of such functions, fractality in neural networks, and fractional
optimization methods for neural network training. Each paragraph closes
with the relation to the present work.

\,\par\noindent\textbf{Fractal functions in classical analysis. }
The Weierstrass function \cite{weierstrass1872} is the classical example
of a continuous, nowhere differentiable function; Hardy established the
precise parameter conditions under which non-differentiability holds
\cite{hardy1916weierstrass}, and the Takagi (Blancmange) function provides
a second canonical construction from a tent-type generator
\cite{takagi1903continuous,allaart2011takagi}. Mandelbrot placed these
objects at the centre of fractal geometry as models for irregular natural
structure \cite{mandelbrot1982fractal}, Berry and Lewis analysed the
Weierstrass--Mandelbrot function and its exact scale invariance for
physical modelling \cite{berry1980weierstrass}, and the dimension theory
of Weierstrass-type graphs was developed by Hu and Lau
\cite{hu1993weierstrass}; standard treatments are given in
\cite{falconer1990fractal,peitgen1992chaos,barnsleynonlinear}. In the
present study, exactly this function class is the raw material on
both experimental levels: the fractal perturbations of the benchmark
surfaces and the fractal activation functions inside the networks are
truncated Weierstrass- and Blancmange-type series.

\,\par\noindent\textbf{Fractional derivatives of Weierstrass-type functions. }
Fractional calculus, the theory of derivatives and integrals of
non-integer order, dates back to the nineteenth-century constructions of
Gr\"unwald, Letnikov, Riemann, Liouville, and Caputo's later
initial-value-friendly formulation \cite{oldham1974fractional,
podlubny1999fractional,caputo1967linear}; treatments and
comparisons of the inequivalent definitions are given in
\cite{samko1993fractional,oliveira2014review,rogosin2017letnikov,
ferrari2018weyl,ferrari2018weyl}. For rough functions, the decisive fact is that
fractional differentiability can survive where classical
differentiability fails: Ross, Samko, and Love showed that functions
without a first derivative may possess fractional derivatives of all
orders less than one \cite{ross1994functions}, and pointwise regularity of
this type is systematically captured by H\"older-scale analysis
\cite{jaffard1991pointwise}. The sharpest results for the function class
used here are due to Z\"ahle and Ziezold \cite{zahle1996fractional}: for
Weierstrass- and Weierstrass--Mandelbrot functions with roughness exponent
$\gamma_H$, all Weyl--Marchaud derivatives of order $\nu<\gamma_H$ exist
and map the function to another Weierstrass-type function with reduced
exponent; the fractional degree of differentiability equals $\gamma_H$;
and at the critical order the signed gradual derivative in the mean
vanishes almost everywhere while its absolute version is a positive
constant almost everywhere, computable by a well-conditioned Monte Carlo
procedure. The Gr\"unwald--Letnikov form, finally, is the discrete
computational representation of the same nonlocal idea
\cite{rogosin2017letnikov,michels2012grunwald}. The present study takes
this body of theory as its mathematical backbone: it explains in which
precise sense the objects we perturb and the operators we optimize with
belong together, and it is developed for our setting in
Section~\ref{sec:weierstrass_fractional}. 

\,\par\noindent\textbf{Fractality in neural networks. }
Fractal concepts have entered neural network research in several distinct
roles. At the level of training dynamics, Sohl-Dickstein demonstrated that
the boundary between trainable and divergent hyper-parameter
configurations is a fractal set, indicating that fractality emerges in
neural networks even when none is built in
\cite{sohldickstein2024boundary}. At the level of architecture and
features, fractal hierarchies of neurons \cite{zuev_fractalnn_2021},
fractal-feature ensembles for histology image classification
\cite{roberto_fractalnn_2021,DING2023118793}, fractal and entropy features
for EEG and financial forecasting \cite{HSU20121055,KARACA2020113098},
fractal decomposition for architecture search \cite{SOUQUET2023118947},
and fractal pooling for texture recognition \cite{FLORINDO2024122978} all
use fractal structure as a preprocessing, feature, or design device.
Fractal interpolation has further been shown to improve neural network
time-series prediction \cite{raubitzek_fractal_interpolation_2021,
raubitzek_taming_2021, raubitzek_LSTM_2022}. The direct predecessor of the present study is the work on 
fractal activation functions
\cite{raubitzek_fractals_2026}, which differs from all of the above by
building fractality into the network's nonlinearity itself: it provides a
general recipe for converting Weierstrass- and Blancmange-type series into
computationally stable activation functions, demonstrates their usability
on ten public classification datasets, quantifies their
expressivity through trajectory-length diagnostics with super-ReLU growth,
and analyses their gradient stability. A related but distinct idea is the
use of fractional calculus to construct adaptive activation functions
\cite{zamora2019adaptive}. The present study continues the predecessor
work on the optimization side: it adopts its strongest activation
candidates and its benchmark protocol, and asks how the choice of
optimizer interacts with the multi-scale gradients these activations
produce.

\,\par\noindent\textbf{Fractional optimizers for neural networks. }
Fractional-order optimization for neural networks began with fractional
backpropagation: Wang et al.\ derived a Caputo-based fractional gradient
descent rule with convergence guarantees \cite{wang2017fractional}, and
Bao et al.\ extended the approach to deeper, regularized networks
\cite{bao2018fractional}. Subsequent work produced fractional variants of
the standard deep-learning optimizers \cite{robbins1951stochastic,
polyak1964methods,tieleman2012rmsprop,kingma2015adam,zeiler2012adadelta,
ruder2016overview} along three lines. The first line rescales the current
gradient by a Caputo-type power law of the parameters, as in the families of Herrera-Alc\'antara and
collaborators \cite{herrera2022fractional,herrera2023pytorch, herrera3}. The second
line builds the update from an explicit Gr\"unwald--Letnikov gradient
history: Zhou et al.\ introduced short-memory G--L optimizers with a
stochastic perturbation of the history terms \cite{zhou2023gl}, Yu et al.\
applied the same substitution at the level of momentum \cite{yu2022fracm},
Shin et al.\ proposed Caputo-fractional gradient descent and Adam with
efficient large-scale implementations \cite{shin2023accelerating}, and Han
and Dong combined fractional gradients with adaptive momentum
\cite{han2023adaptive}. The third line makes the fractional order itself
adaptive: AOFGD adapts the order through a convergence evaluation factor
\cite{xiang2025aofgd}, 2SED-FOSGD through a curvature-aware effective
dimension \cite{partohaghighi2025twoscale}, MFFGD through an adaptive
Caputo formulation \cite{huang2024mffgd}, and further scheduler-, decay-,
and parameter-adaptive variants have been proposed
\cite{chen2024foadam,ma2025apfogdl,chen2025lambdafadamax}, alongside
tempered kernels for robust learning \cite{naifar2026tempered}. Between
the fixed-order and adaptive-order designs sits AdaGL, which combines a
G--L fractional gradient with a short-term step-size control coefficient
\cite{chen2024adagl}. Recent surveys organize this rapidly growing design
space and document its open problems, in particular the noise sensitivity
and hyperparameter cost of adaptive-order schemes \cite{elnady2025survey,
elnady2025survey,fernandez2025role,VieraMartin2022,
raubitzek_fractional_2023}. The adaptive memory-fractional framework
introduced in this article
(Section~\ref{sec:fractional_optimizers}) 
differs from all three lines: it keeps the fractional order fixed, treats
the memory contribution as an explicitly bounded, hysteresis-controlled
trust coefficient applied as a convex combination of the ordinary and the
norm-matched fractional gradient, and reduces exactly to the underlying
classical optimizer at one boundary of that coefficient.

\,\par\noindent\textbf{Other fractional optimization approaches.}
Beyond the optimizer families discussed above, fractional calculus has been introduced
at several other levels of optimization and learning. Vieira et al.\ formulated gradient
methods through the $\psi$-Hilfer derivative, including convergence analysis and
variable-order and step-size variants \cite{vieira2023fractional}; Yang et al.\ developed
stochastic fractional-order gradient methods for online optimization with standard,
adaptive, and momentum-based learning rates \cite{yang2023improved}; and Ye et al.\
combined variable fractional derivatives with variable step-size control
\cite{ye2023development}. Zhao et al.\ extended sequential minimal optimization to a
fractional-order support-vector-machine training scheme \cite{zhao2023fractional},
whereas Tan et al.\ proposed the self-organizing Caputo-gradient C-FOG method and
applied it to adversarial-sample generation \cite{tan2024selforganizing}. More recently,
Lee introduced FracGrad, which applies Riemann--Liouville-derived power-law weights to
accumulated microbatch gradients rather than replacing the derivative of the loss itself
\cite{lee2025fracgrad}, and Alaerjan employed tempered fractional gradient descent as
the meta-learner of a calibrated stacking framework for smart-grid stability prediction
\cite{alaerjan2026tempered}. These works illustrate the breadth of fractional
optimization, ranging from generalized derivative definitions and adaptive numerical
schemes to classifier-specific, accumulation-level, and application-oriented
formulations. The present study differs by examining fractional optimizer families
jointly with fractal activation functions on controlled multi-scale landscapes and
neural-network classification tasks.

\,\par\noindent\textbf{Positioning of the present study. }
Across the literature reviewed above, fractal structure in neural networks
and fractional structure in optimizers have been developed by largely
disjoint communities, although both are governed by the same mathematics
of multi-scale roughness and non-integer order. To our knowledge, no prior
work evaluates fractional optimizer families on objective landscapes with
controlled Weierstrass-type perturbations, and no prior work studies the
interaction between fractional optimizers and fractal activation
functions. The present article addresses both points within one
experimental framework: a controlled surface study in which the minima are
known and the fractal difficulty is dialled in explicitly, followed by a
neural network study on the benchmark protocol of
\cite{raubitzek_fractals_2026}, with 21 optimizers compared under
identical conditions throughout.

\begin{takeawaybox}[blueA]{Main Takeaways --- Related Work}
  \begin{itemize}[
    label={},
    leftmargin=0em,
    itemindent=0em,
    itemsep=3pt,
    topsep=2pt
  ]
    \item \textbf{Classical fractal functions provide the mathematical
          basis.} Weierstrass-, Weierstrass--Mandelbrot-, and
          Blancmange-type series underlie both the perturbed benchmark
          surfaces and the fractal activation functions studied here.

    \item \textbf{Fractional calculus provides the optimization link.}
          Fractional derivatives can describe nonlocal behaviour, and the
          Gr\"unwald--Letnikov form gives a discrete representation suitable
          for gradient-history based optimizers.

    \item \textbf{The present study connects fractal activations and
          fractional optimizers.} It evaluates fractional optimizer
          families on controlled fractal landscapes and combines fractal
          activations with 21 optimizers under identical classification
          conditions.
  \end{itemize}
\end{takeawaybox}

\section{Intuition Behind Fractal Functions and Fractional Derivatives}
\label{sec:intuition_fractals_fractional_derivatives}

This section does not contribute new mathematical results, algorithms, or
experimental evidence to the article. It also does not introduce additional
assumptions for the analysis in the following sections. Its purpose is
explanatory: it gives a visual and conceptual entry point to fractals,
fractal functions, and fractional derivatives. The aim is to make the later
use of Weierstrass-type activation functions and memory-based fractional
optimizers easier to interpret.

\subsection{Fractals and Fractal Functions}
\label{subsec:intuition_fractal_functions}

A fractal is a structure that is generated by repeating a rule across
scales. The rule is often simple, but repeated application produces detail
at increasingly fine resolutions. Many fractals also show self-similarity:
parts of the object resemble the whole object after rescaling. This idea is
central in fractal geometry \cite{mandelbrot1982fractal,falconer1990fractal}.
The following examples introduce the idea before it is transferred to
functions and activation functions.

\,\par
\begin{figure}[H]
    \centering

    \begin{minipage}{0.19\textwidth}
        \centering
        \includegraphics[width=\linewidth]{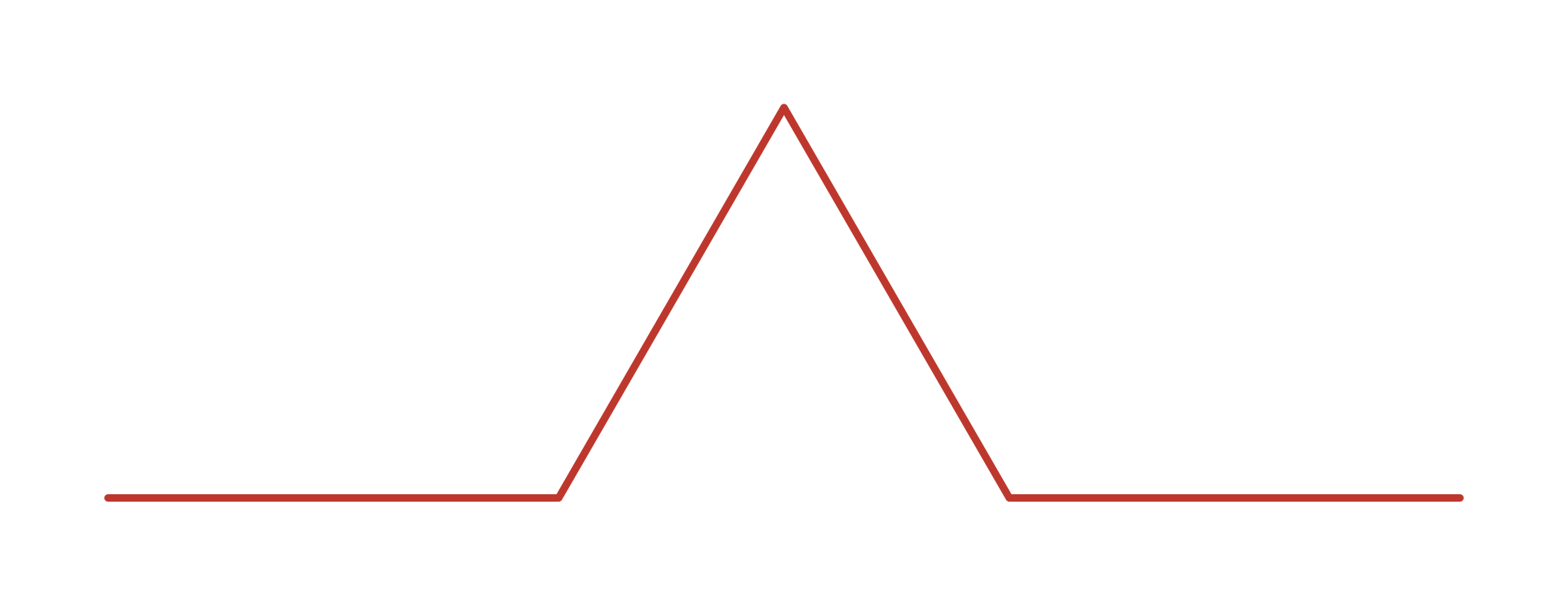}\\
        \small Depth 1
    \end{minipage}
    \hfill
    \begin{minipage}{0.19\textwidth}
        \centering
        \includegraphics[width=\linewidth]{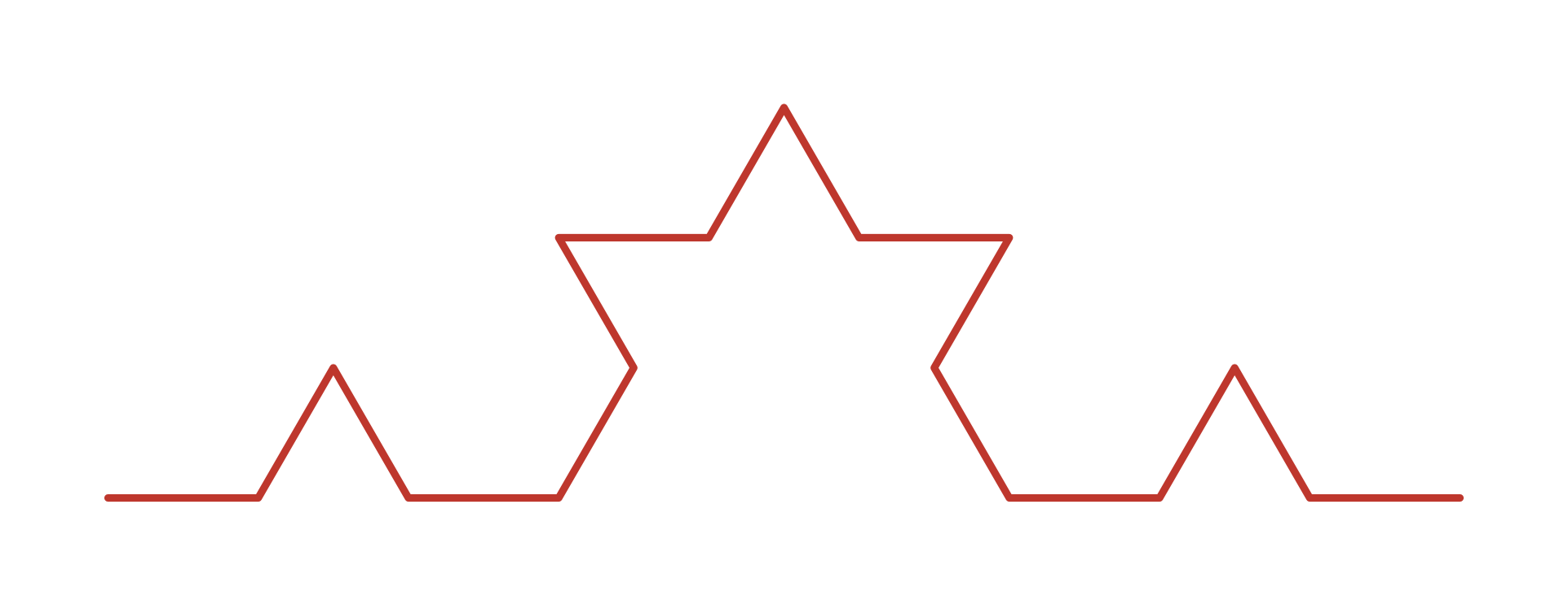}\\
        \small Depth 2
    \end{minipage}
    \hfill
    \begin{minipage}{0.19\textwidth}
        \centering
        \includegraphics[width=\linewidth]{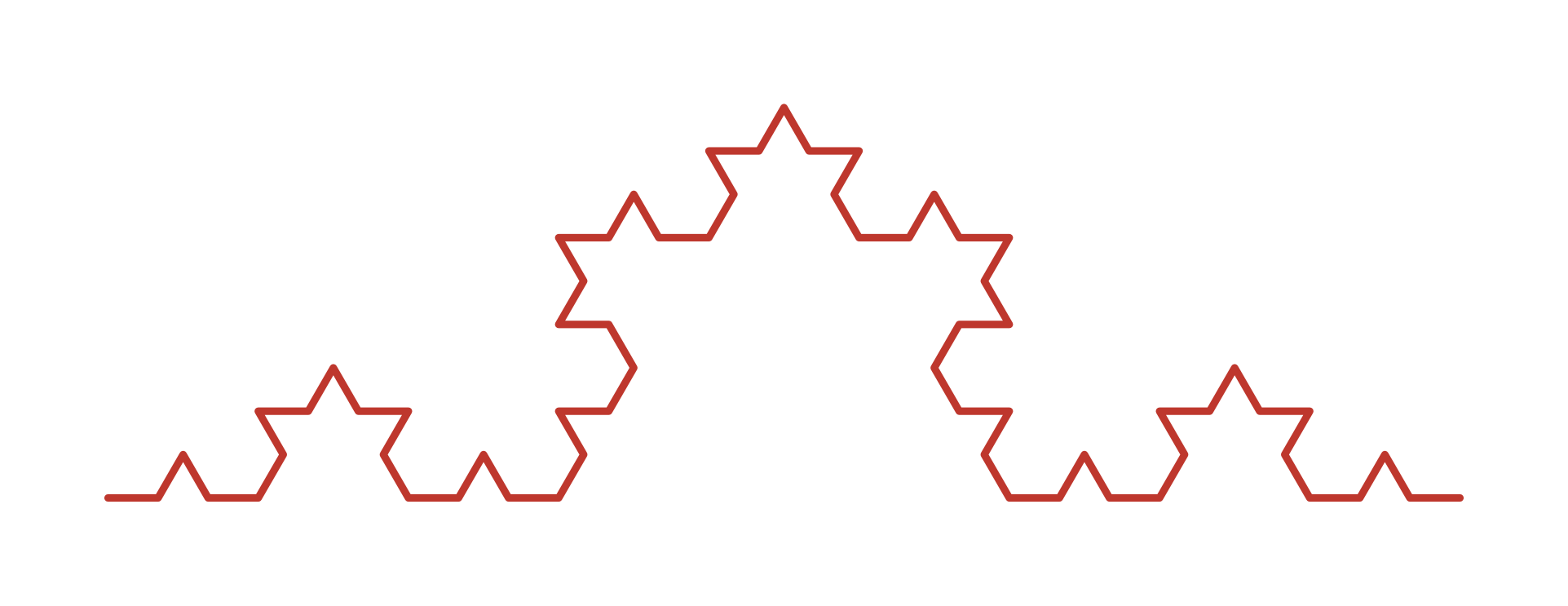}\\
        \small Depth 3
    \end{minipage}
    \hfill
    \begin{minipage}{0.19\textwidth}
        \centering
        \includegraphics[width=\linewidth]{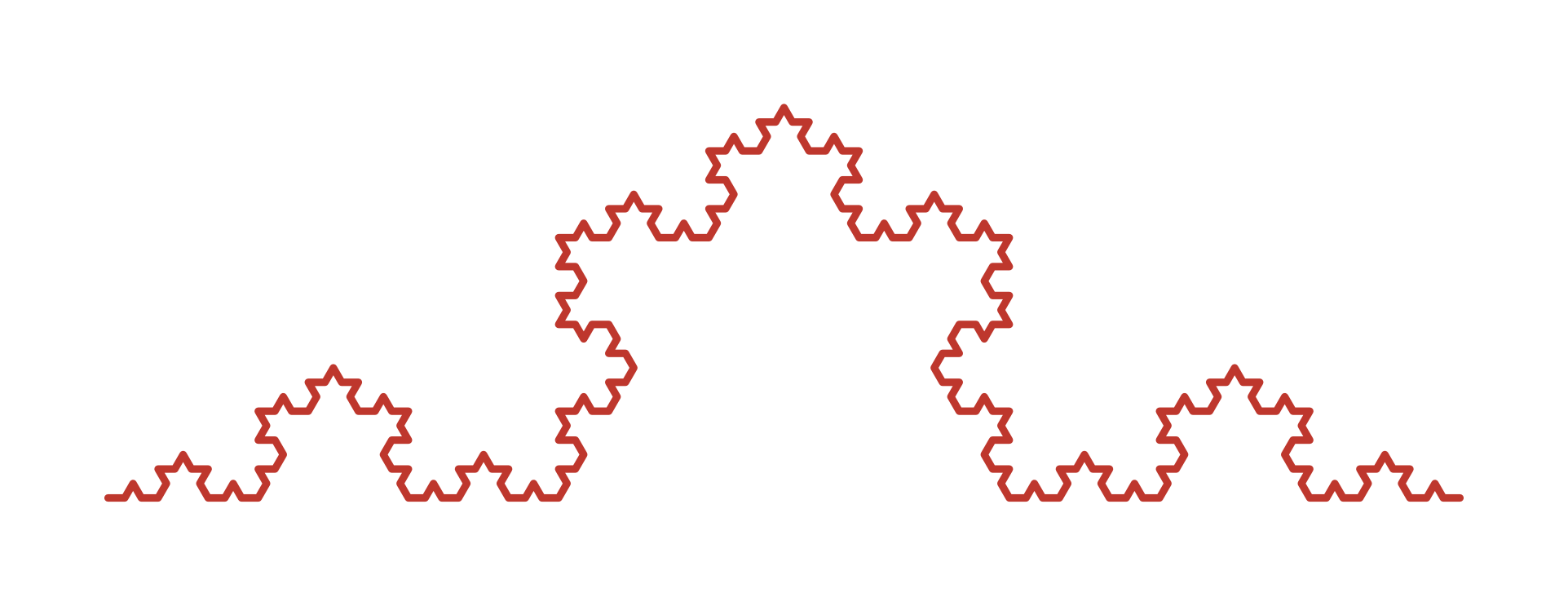}\\
        \small Depth 4
    \end{minipage}
    \hfill
    \begin{minipage}{0.19\textwidth}
        \centering
        \includegraphics[width=\linewidth]{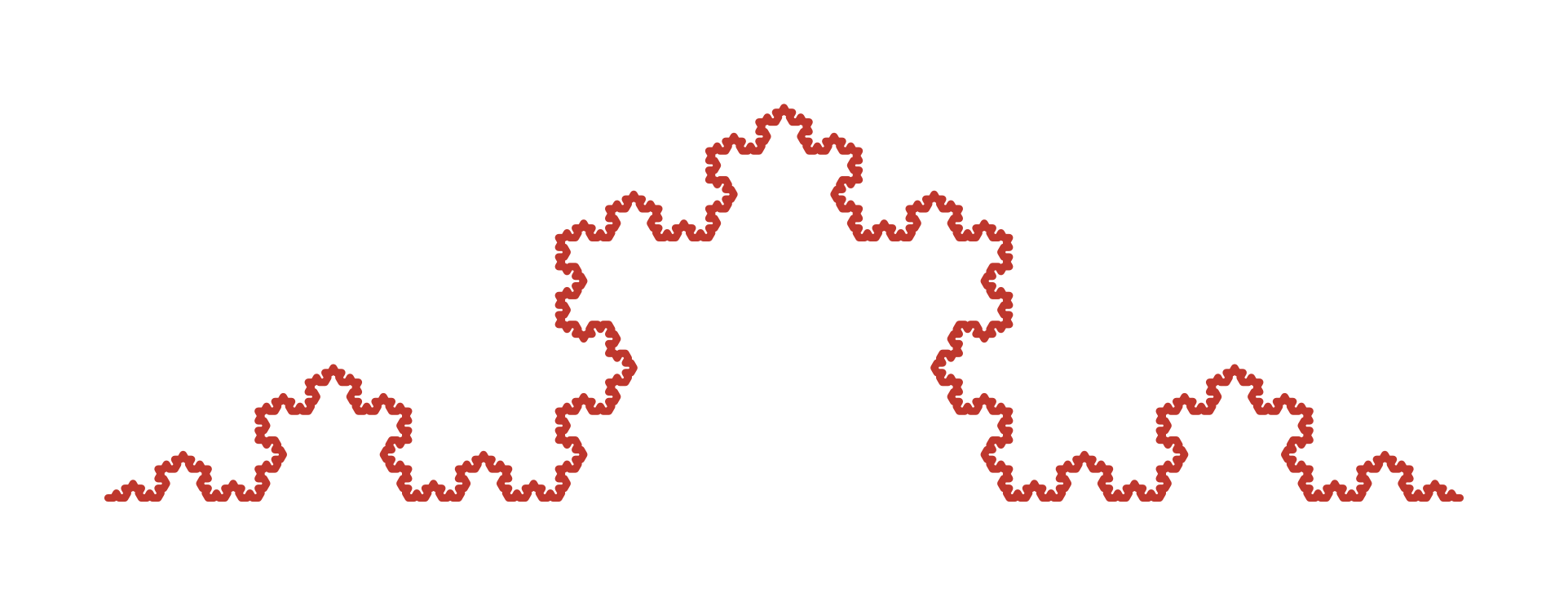}\\
        \small Depth 5
    \end{minipage}

    \caption{Development of the Koch curve over five iterations. The
    initial straight line is not shown. The first panel shows the result
    after applying the rule once; each following panel applies the same
    rule once more to every existing segment.}
    \label{fig:intuition_koch_curve_depths}
\end{figure}

\noindent\textbf{Koch curve.}
The Koch curve starts from a straight line segment. In one iteration, each
line segment is divided into three equal parts. The middle part is replaced
by two sides of an equilateral triangle, producing four shorter line
segments from one original segment. Repeating this rule gives a curve with
more and more small-scale detail. After $n$ iterations, the curve consists
of $4^n$ segments, each of length $3^{-n}$ times the original length. Its
fractal dimension is therefore
\begin{equation}
\label{eq:intuition_koch_dimension}
D_{\mathrm{frac}}
=
\frac{\log 4}{\log 3}.
\end{equation}
This dimension is larger than the dimension of a smooth curve, which is
$1$, but smaller than the dimension of a filled two-dimensional region,
which is $2$. The Koch curve is therefore a useful example of a curve that
is still one-dimensional in its construction, but too irregular to behave
like an ordinary smooth line. See Figure \ref{fig:intuition_koch_curve_depths}.

\,\par
\begin{figure}[H]
    \centering

    \begin{minipage}{0.19\textwidth}
        \centering
        \includegraphics[width=\linewidth]{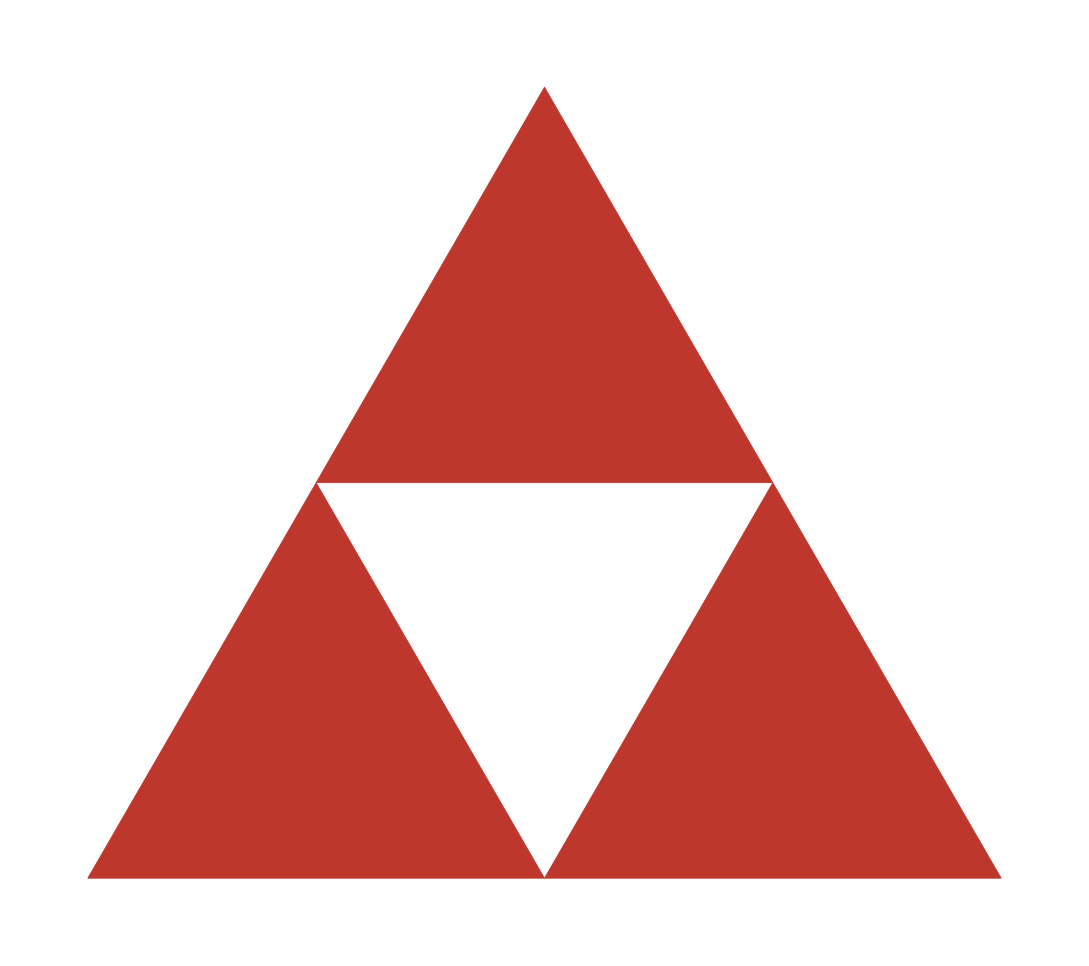}\\
        \small Depth 1
    \end{minipage}
    \hfill
    \begin{minipage}{0.19\textwidth}
        \centering
        \includegraphics[width=\linewidth]{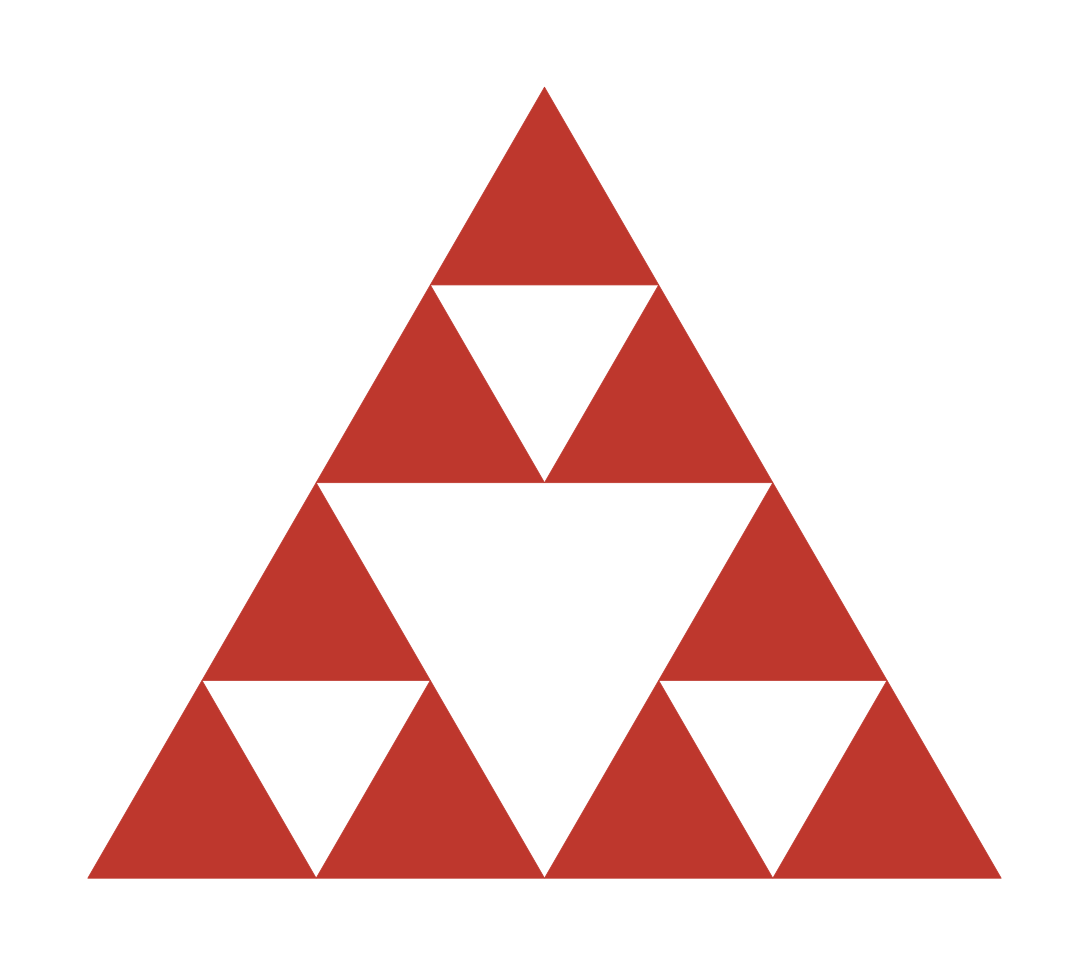}\\
        \small Depth 2
    \end{minipage}
    \hfill
    \begin{minipage}{0.19\textwidth}
        \centering
        \includegraphics[width=\linewidth]{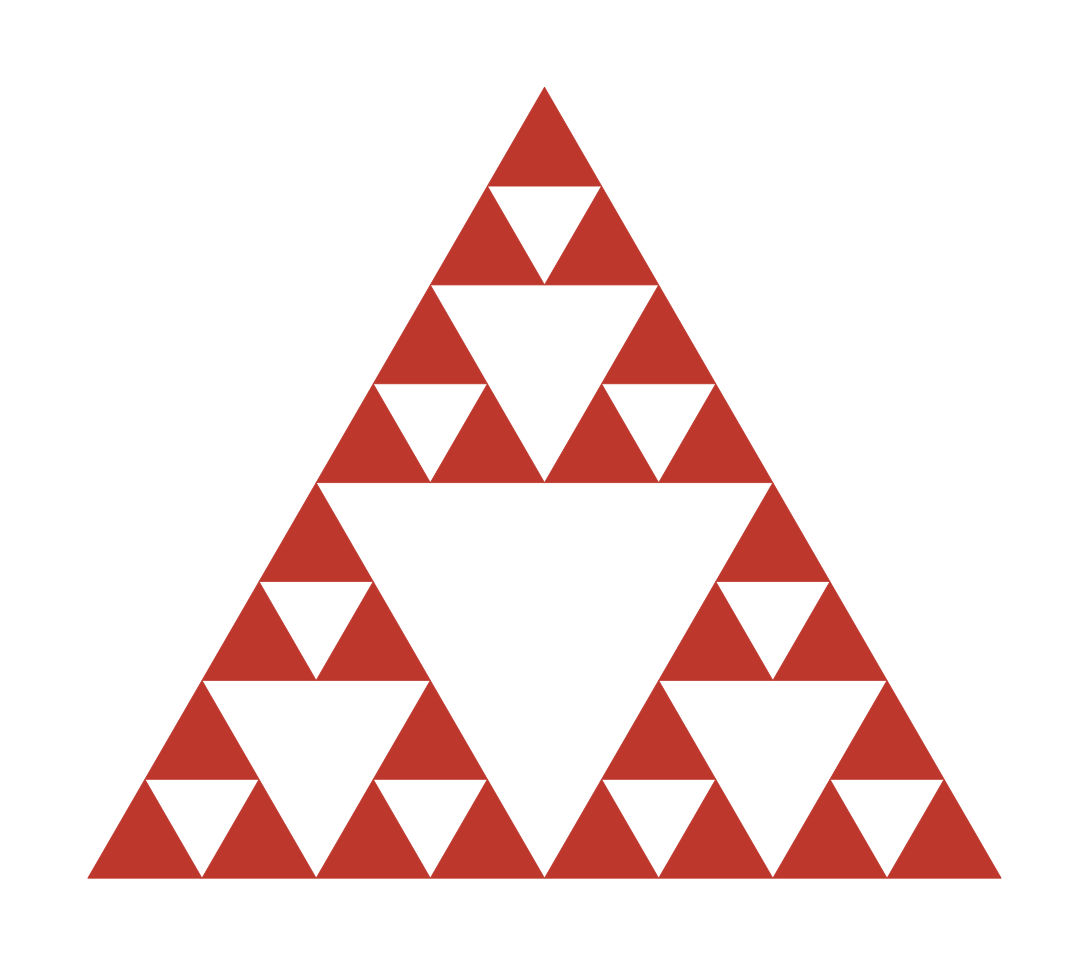}\\
        \small Depth 3
    \end{minipage}
    \hfill
    \begin{minipage}{0.19\textwidth}
        \centering
        \includegraphics[width=\linewidth]{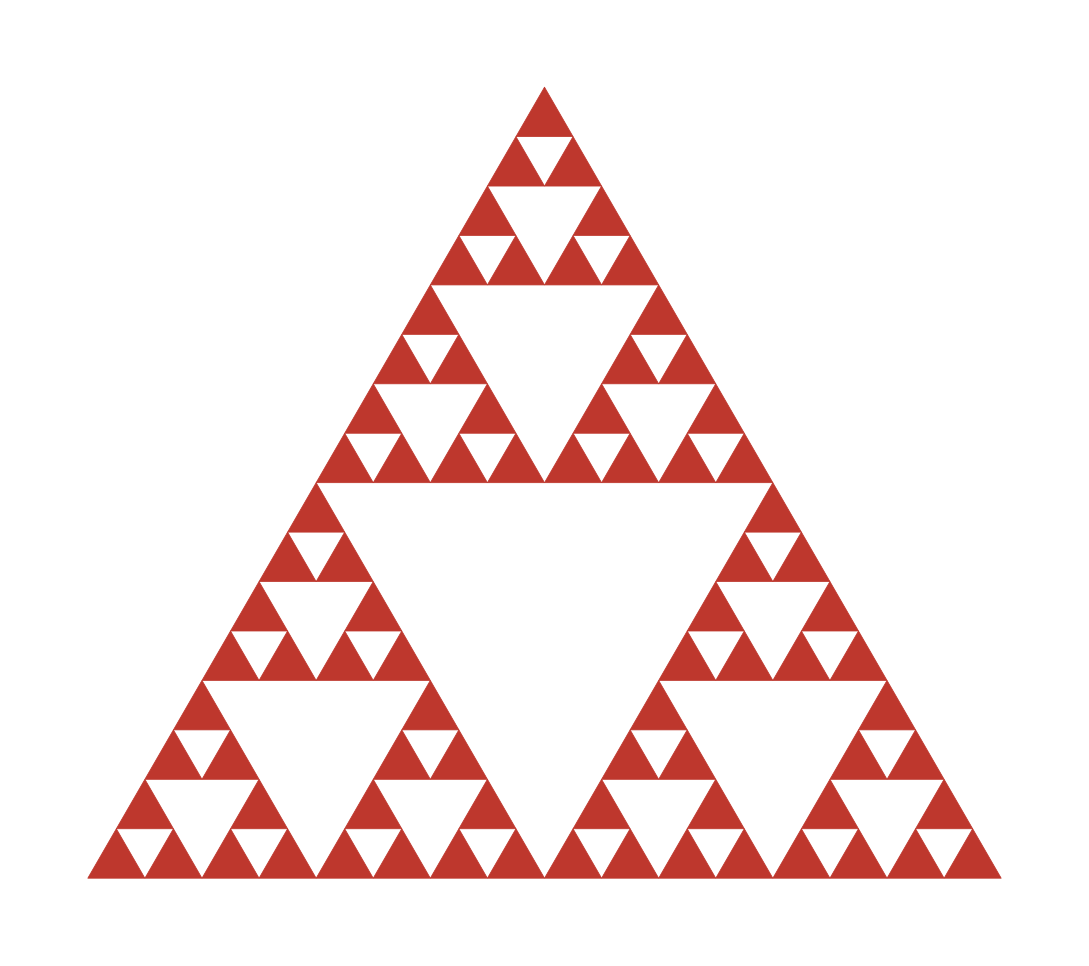}\\
        \small Depth 4
    \end{minipage}
    \hfill
    \begin{minipage}{0.19\textwidth}
        \centering
        \includegraphics[width=\linewidth]{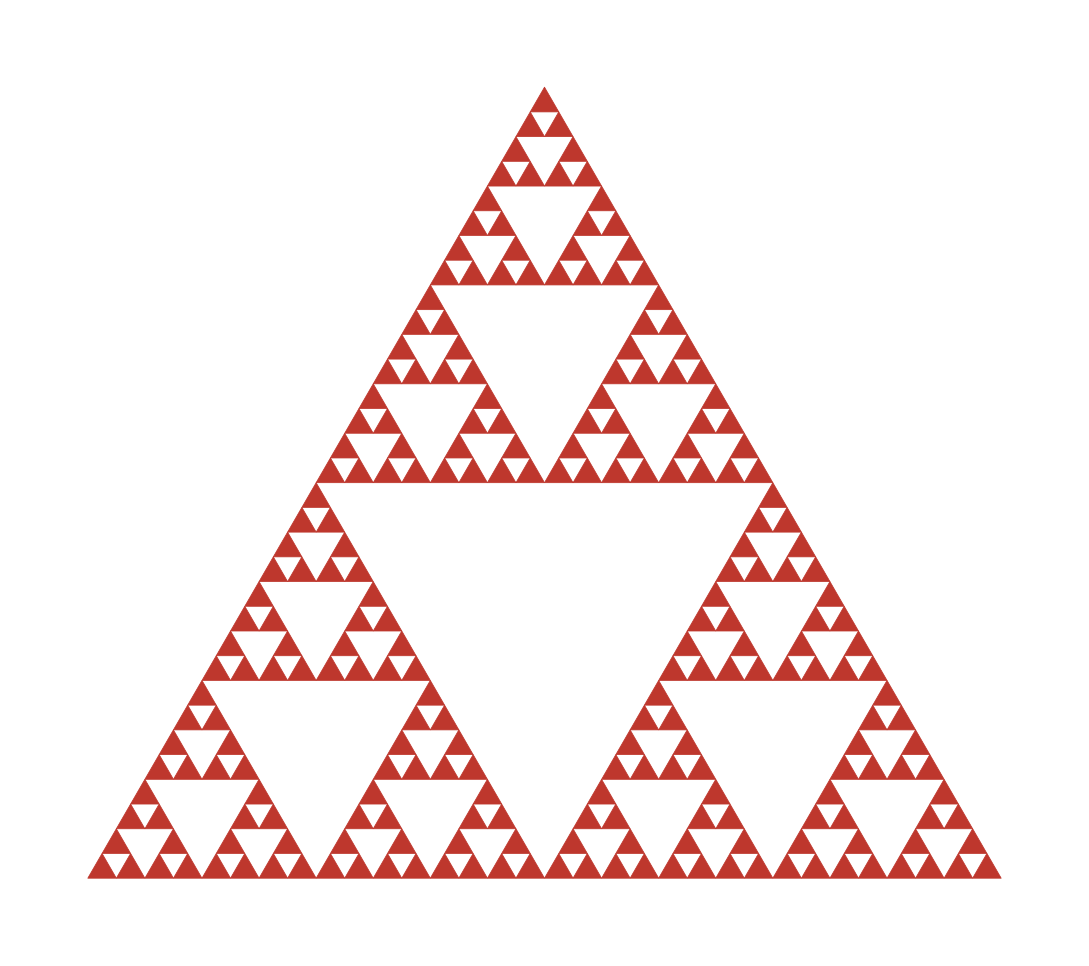}\\
        \small Depth 5
    \end{minipage}

    \caption{Development of the Sierpi\'nski triangle over five iterations.
    The first panel shows the result after one deletion step. Each following
    panel applies the same deletion rule to the remaining smaller
    triangles.}
    \label{fig:intuition_sierpinski_depths}
\end{figure}

\noindent\textbf{Sierpi\'nski triangle.}
The Sierpi\'nski triangle starts from a filled triangle. In one iteration,
the triangle is split into four congruent smaller triangles and the middle
triangle is removed. The same removal rule is then applied to each remaining
triangle. After $n$ iterations, there are $3^n$ remaining triangles, each
with side length $2^{-n}$ times the original side length. Its fractal
dimension is
\begin{equation}
\label{eq:intuition_sierpinski_dimension}
D_{\mathrm{frac}}
=
\frac{\log 3}{\log 2}.
\end{equation}
This value lies between $1$ and $2$. The object is not a curve in the usual
sense, because it branches over an area, but it also does not fill the
original triangle. This makes the Sierpi\'nski triangle a simple example of
how repeated deletion can produce a stable multi-scale structure. See Figure \ref{fig:intuition_sierpinski_depths}.

\,\par
\begin{figure}[H]
    \centering

    \begin{minipage}{0.19\textwidth}
        \centering
        \includegraphics[width=\linewidth]{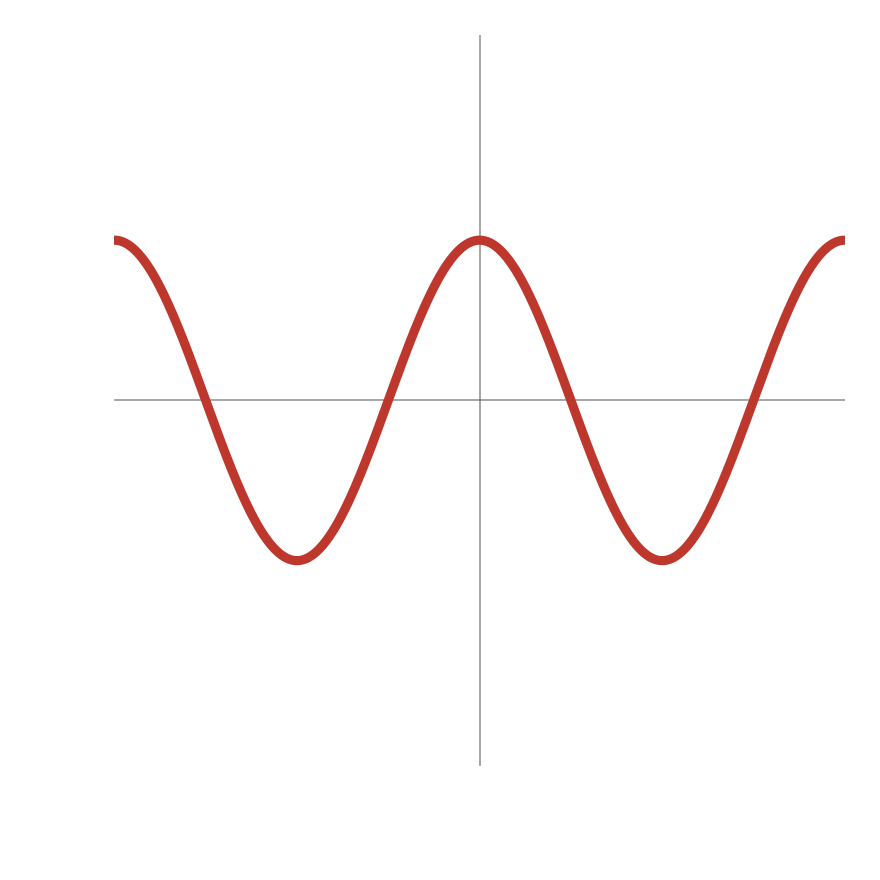}\\
        \small $W_1$
    \end{minipage}
    \hfill
    \begin{minipage}{0.19\textwidth}
        \centering
        \includegraphics[width=\linewidth]{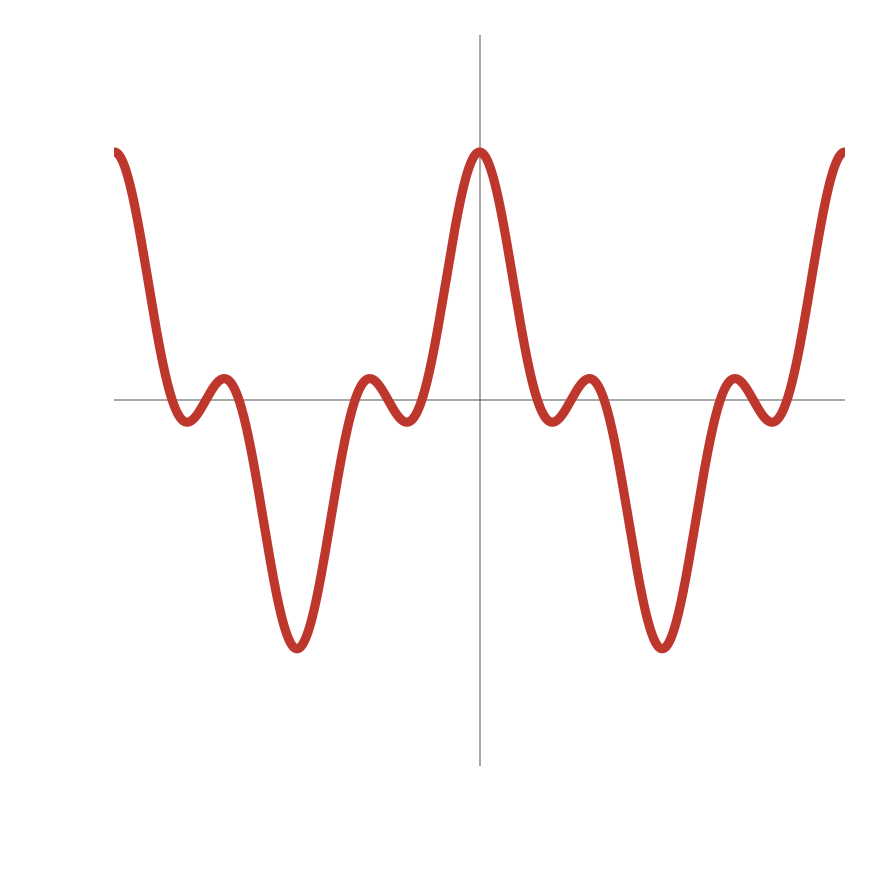}\\
        \small $W_2$
    \end{minipage}
    \hfill
    \begin{minipage}{0.19\textwidth}
        \centering
        \includegraphics[width=\linewidth]{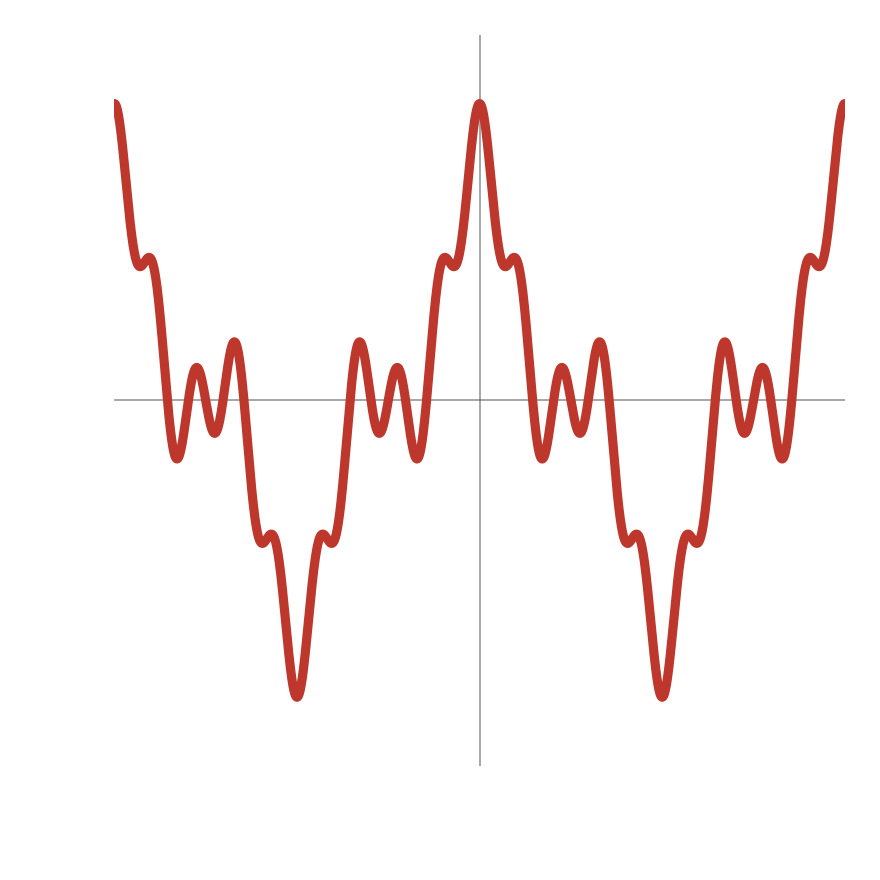}\\
        \small $W_3$
    \end{minipage}
    \hfill
    \begin{minipage}{0.19\textwidth}
        \centering
        \includegraphics[width=\linewidth]{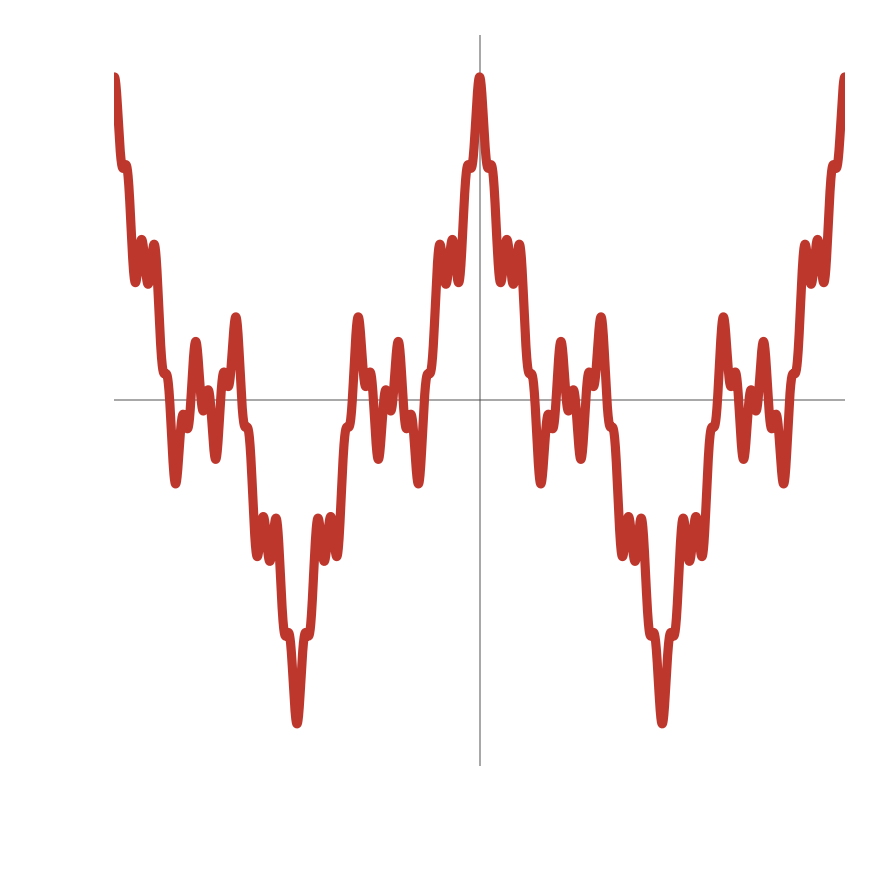}\\
        \small $W_4$
    \end{minipage}
    \hfill
    \begin{minipage}{0.19\textwidth}
        \centering
        \includegraphics[width=\linewidth]{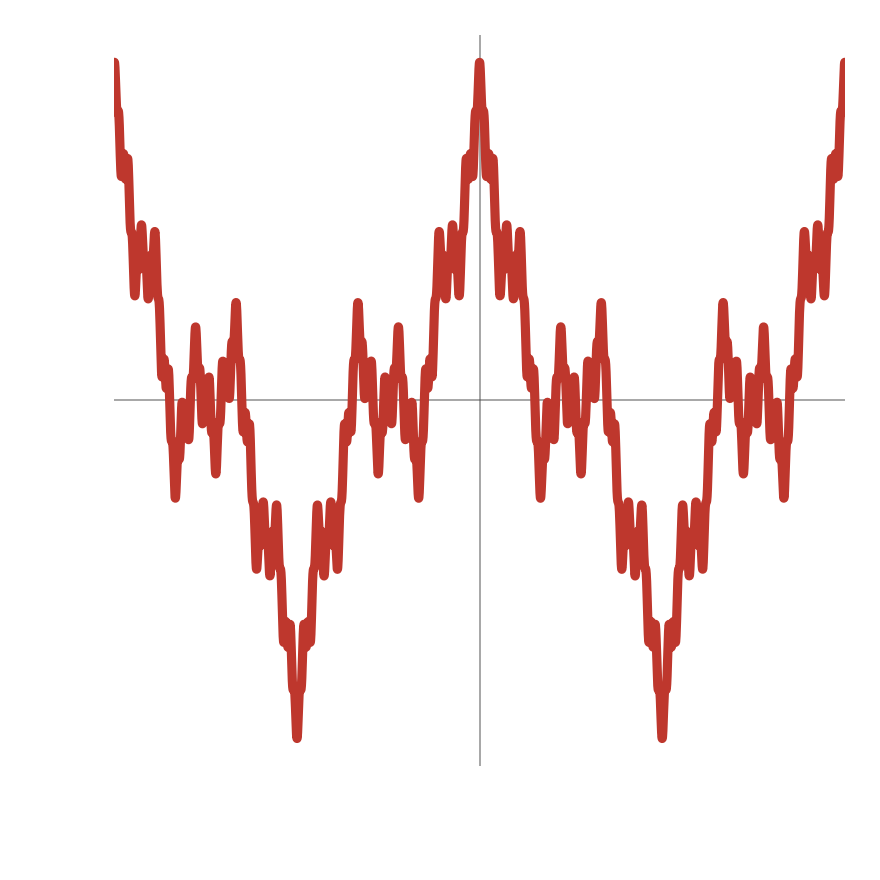}\\
        \small $W_5$
    \end{minipage}

    \caption{Finite partial sums of a Weierstrass-type function. Moving
    from left to right adds one additional oscillatory scale. The visual
    effect is analogous to the geometric examples above: a simple repeated
    rule creates increasing small-scale structure.}
    \label{fig:intuition_weierstrass_partial_sums}
\end{figure}

\noindent\textbf{From geometric fractals to fractal functions.}
The same idea can be applied to functions. Instead of repeatedly modifying
a line segment or a triangle, one repeatedly adds oscillatory terms at
smaller and smaller scales. A finite Weierstrass-type partial sum can be
written as
\begin{equation}
\label{eq:intuition_weierstrass_partial_sum}
W_M(x)
=
\sum_{m=0}^{M-1} a^m \cos\!\bigl(b^m\pi x\bigr),
\qquad
0<a<1,\qquad b>1.
\end{equation}
The parameter $a$ controls how fast the amplitudes decrease. The parameter
$b$ controls how fast the frequencies increase. Each new term adds a
smaller but faster oscillation. The first five partial sums are
\begin{align}
\label{eq:intuition_weierstrass_expansion}
W_1(x)
&=
\cos(\pi x),\\
W_2(x)
&=
\cos(\pi x)+a\cos(b\pi x),\\
W_3(x)
&=
\cos(\pi x)+a\cos(b\pi x)+a^2\cos(b^2\pi x),\\
W_4(x)
&=
W_3(x)+a^3\cos(b^3\pi x),\\
W_5(x)
&=
W_4(x)+a^4\cos(b^4\pi x).
\end{align}
For finite $M$, the function is an ordinary finite sum and can be plotted
directly. The fractal behavior is approached as more terms are added. The
classical Weierstrass construction is important because it gives continuous
functions whose graphs become increasingly rough and, in suitable parameter
ranges, nowhere differentiable in the classical sense
\cite{hardy1916weierstrass,zahle1996fractional}. See Figure \ref{fig:intuition_weierstrass_partial_sums}.

\,\par
\begin{figure}[H]
    \centering

    \begin{minipage}{0.19\textwidth}
        \centering
        \includegraphics[width=\linewidth]{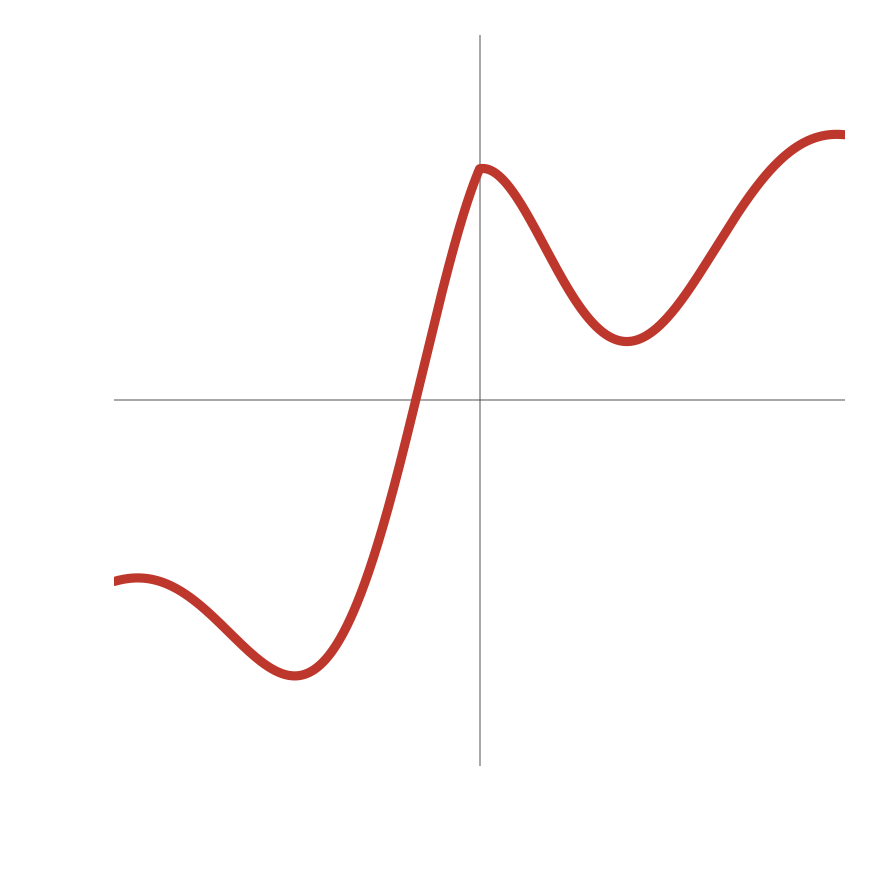}\\
        \small $M=1$
    \end{minipage}
    \hfill
    \begin{minipage}{0.19\textwidth}
        \centering
        \includegraphics[width=\linewidth]{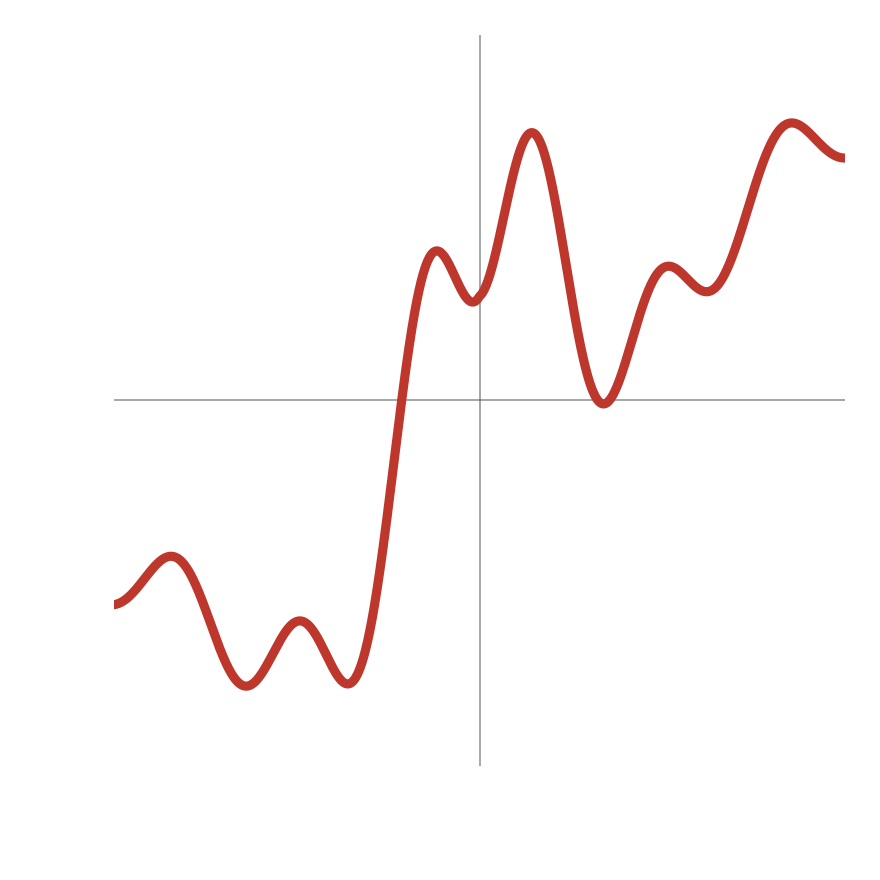}\\
        \small $M=2$
    \end{minipage}
    \hfill
    \begin{minipage}{0.19\textwidth}
        \centering
        \includegraphics[width=\linewidth]{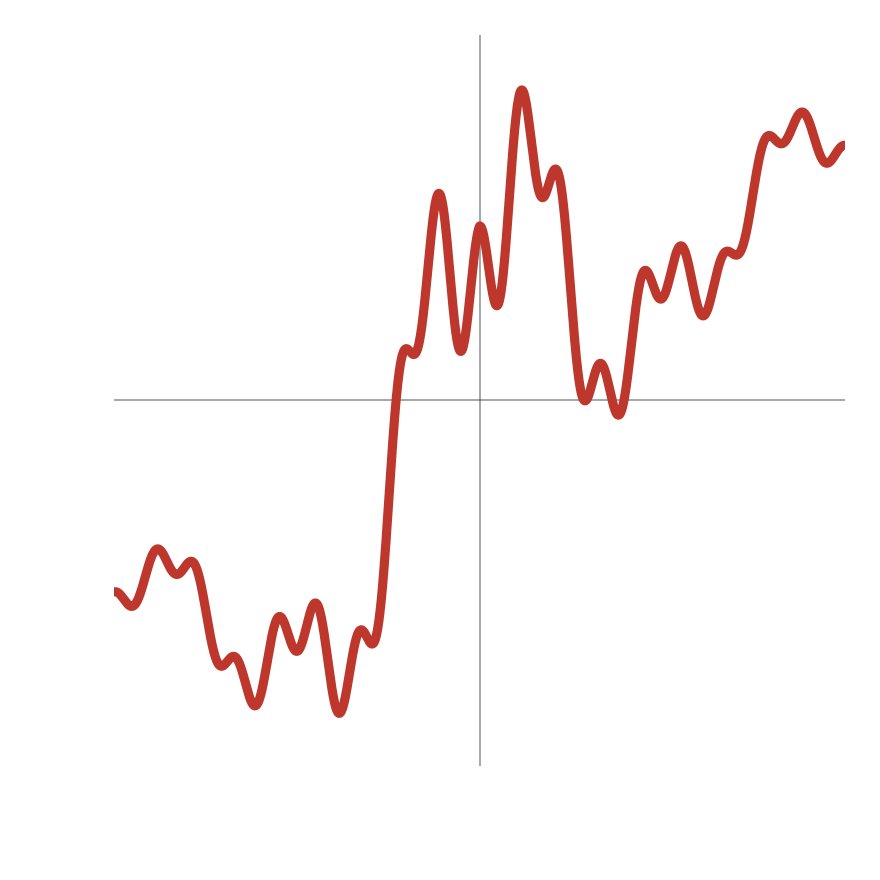}\\
        \small $M=3$
    \end{minipage}
    \hfill
    \begin{minipage}{0.19\textwidth}
        \centering
        \includegraphics[width=\linewidth]{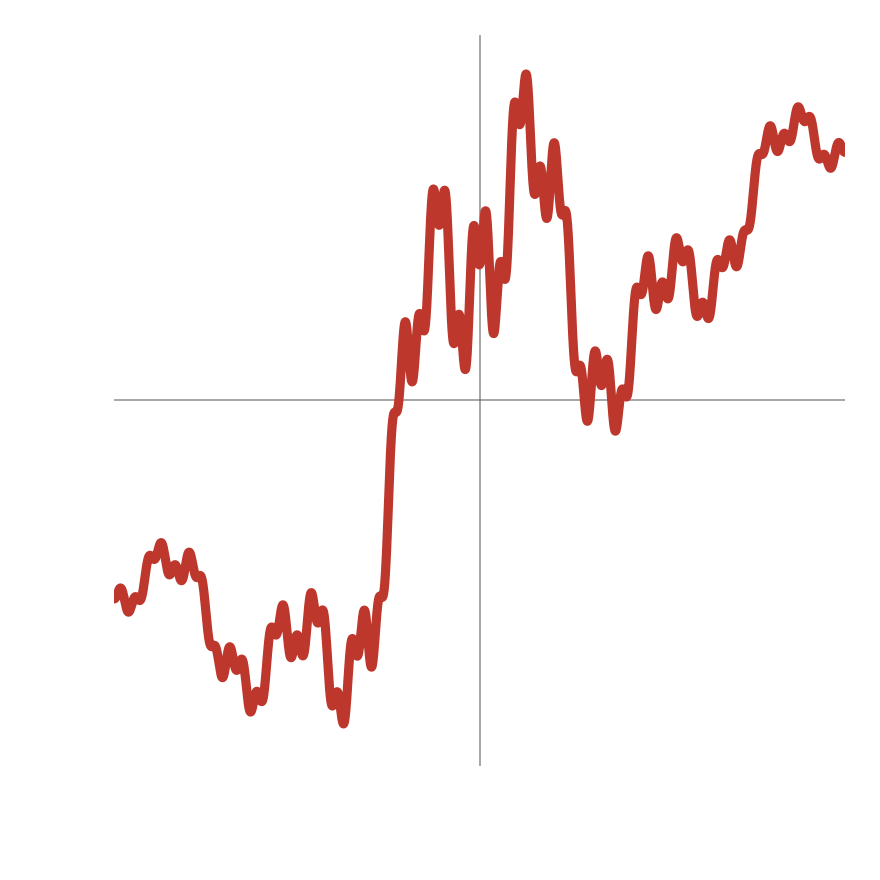}\\
        \small $M=4$
    \end{minipage}
    \hfill
    \begin{minipage}{0.19\textwidth}
        \centering
        \includegraphics[width=\linewidth]{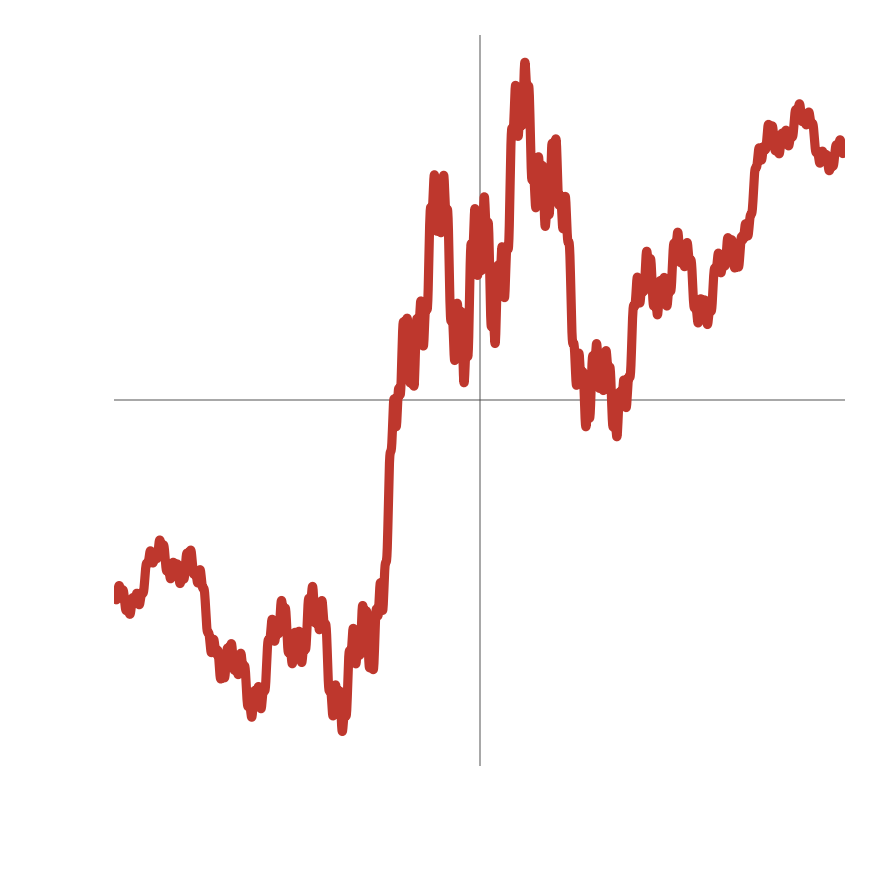}\\
        \small $M=5$
    \end{minipage}

    \caption{Depth-wise construction of the modified Weierstrass--Tanh
    activation. The smooth backbone is already present at every depth.
    Moving from left to right adds further oscillatory scales around that
    backbone.}
    \label{fig:intuition_weierstrass_tanh_depths}
\end{figure}

\noindent\textbf{From fractal functions to fractal activations.}
The same construction can be combined with a standard neural-network
activation. In this article, one representative example is the modified
Weierstrass--Tanh activation
\begin{equation}
\label{eq:intuition_weierstrass_tanh_activation}
\phi_{\mathrm{W}}(x)
=
\tanh(x)
+
\sum_{m=0}^{N-1}
(-1)^m a^m \cos\!\bigl(b^m\pi x\bigr)e^{-\mu_e |x|}.
\end{equation}
The term $\tanh(x)$ provides the ordinary activation backbone. The sum adds
a finite ladder of oscillations. The factor $a^m$ reduces the amplitude of
higher scales, $b^m$ increases their frequency, and
$e^{-\mu_e |x|}$ localizes the oscillatory correction around the origin.
The alternating sign $(-1)^m$ introduces alternating scales.
For visualization, define the depth-$M$ activation
\begin{equation}
\label{eq:intuition_weierstrass_tanh_depth}
\phi_{\mathrm{W},M}(x)
=
\tanh(x)
+
\sum_{m=0}^{M-1}
(-1)^m a^m \cos\!\bigl(b^m\pi x\bigr)e^{-\mu_e |x|}.
\end{equation}
Increasing $M$ adds additional scales around the same smooth backbone, Figure \ref{fig:intuition_weierstrass_tanh_depths}. This
is the functional analogue of increasing the iteration depth in the Koch
curve or the Sierpi\'nski triangle. The full activation family is discussed
in more detail in Section~\ref{sec:weierstrass_fractional} and in the
activation-function section of this article; see also
\cite{raubitzek_fractals_2026}.

\subsection{Fractional Derivatives}
\label{subsec:intuition_fractional_derivatives}

\begin{figure}[H]
    \centering

    \begin{minipage}{0.19\textwidth}
        \centering
        \includegraphics[width=\linewidth]{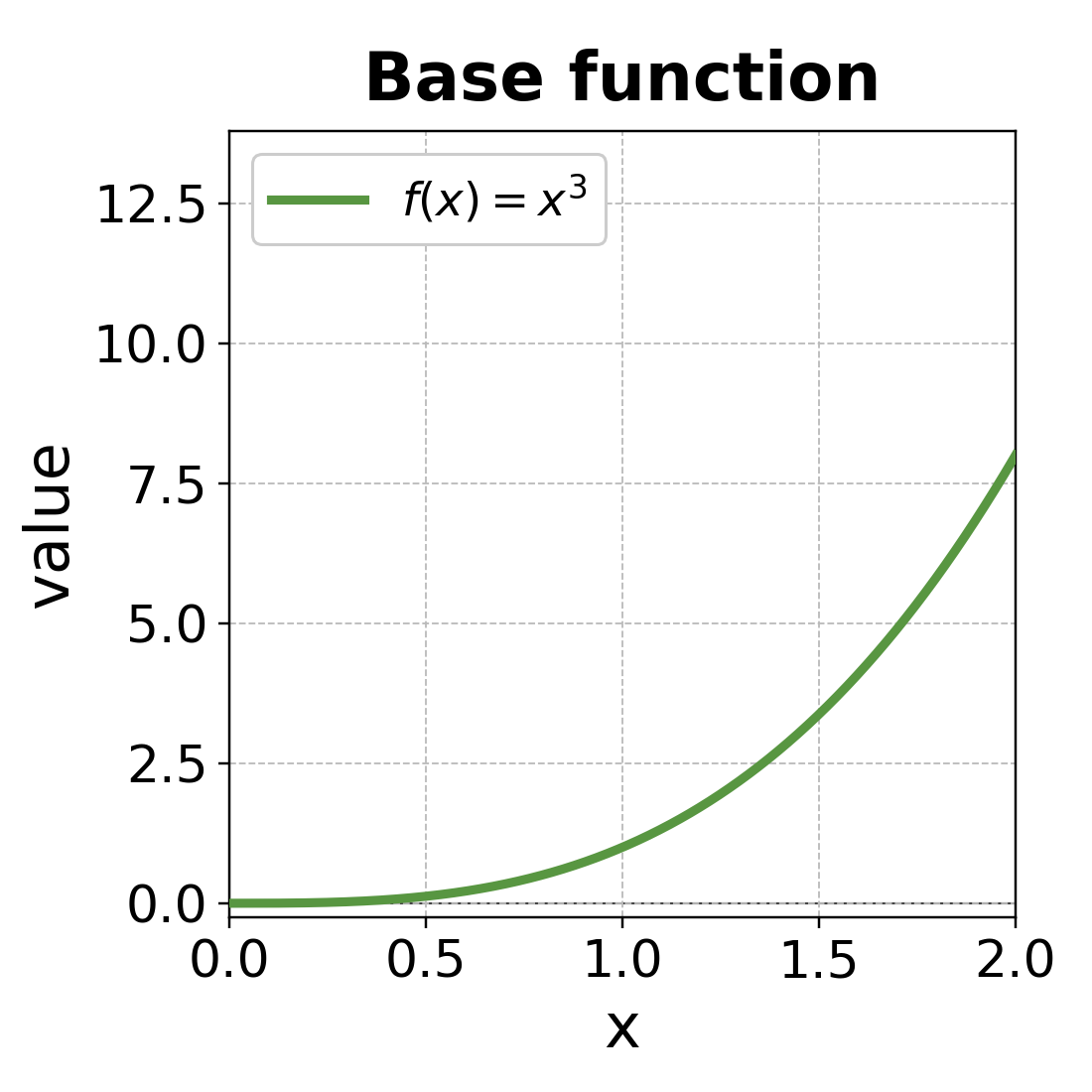}\\
        \small $f(x)=x^3$
    \end{minipage}
    \begin{minipage}{0.19\textwidth}
        \centering
        \includegraphics[width=\linewidth]{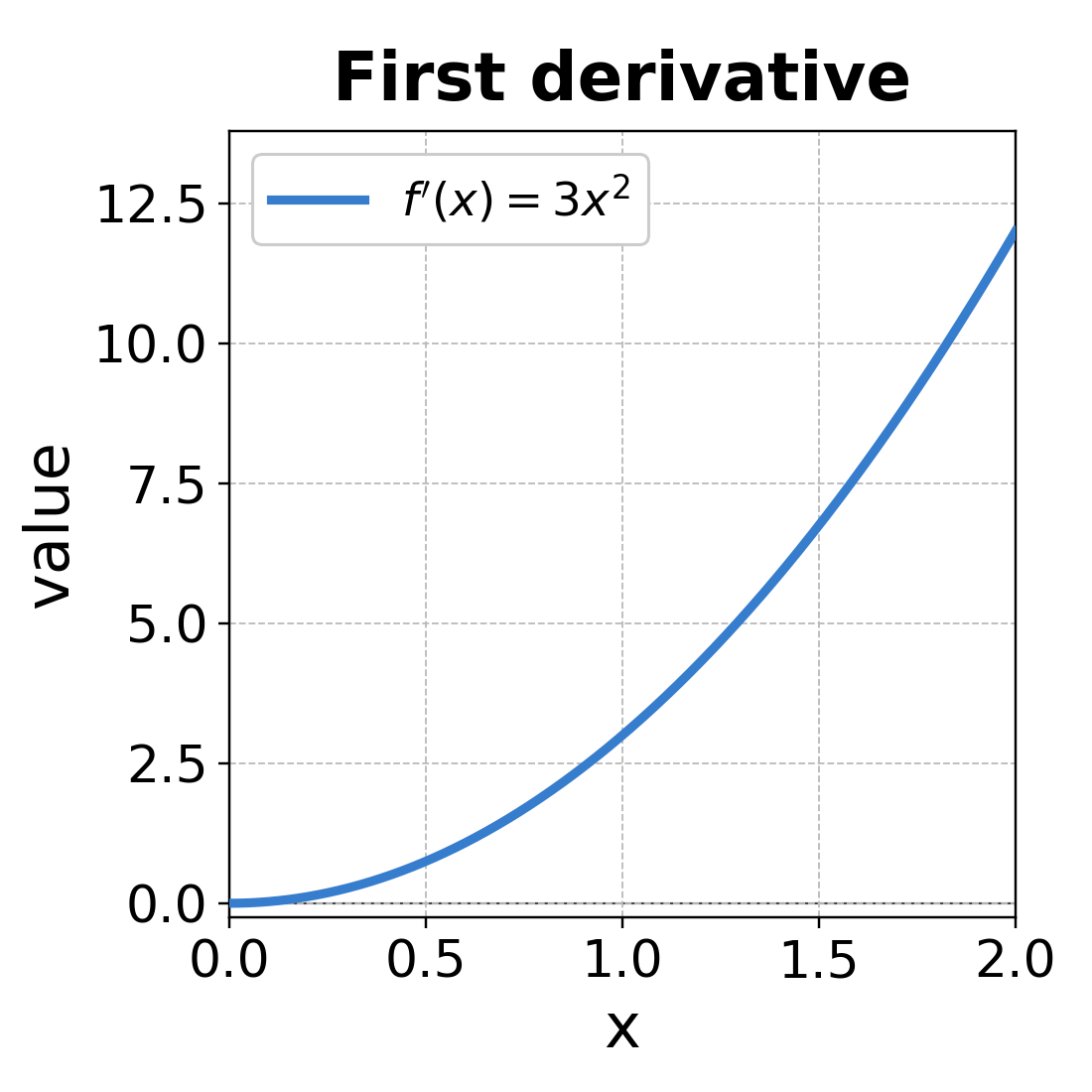}\\
        \small $f'(x)=3x^2$
    \end{minipage}
    \begin{minipage}{0.19\textwidth}
        \centering
        \includegraphics[width=\linewidth]{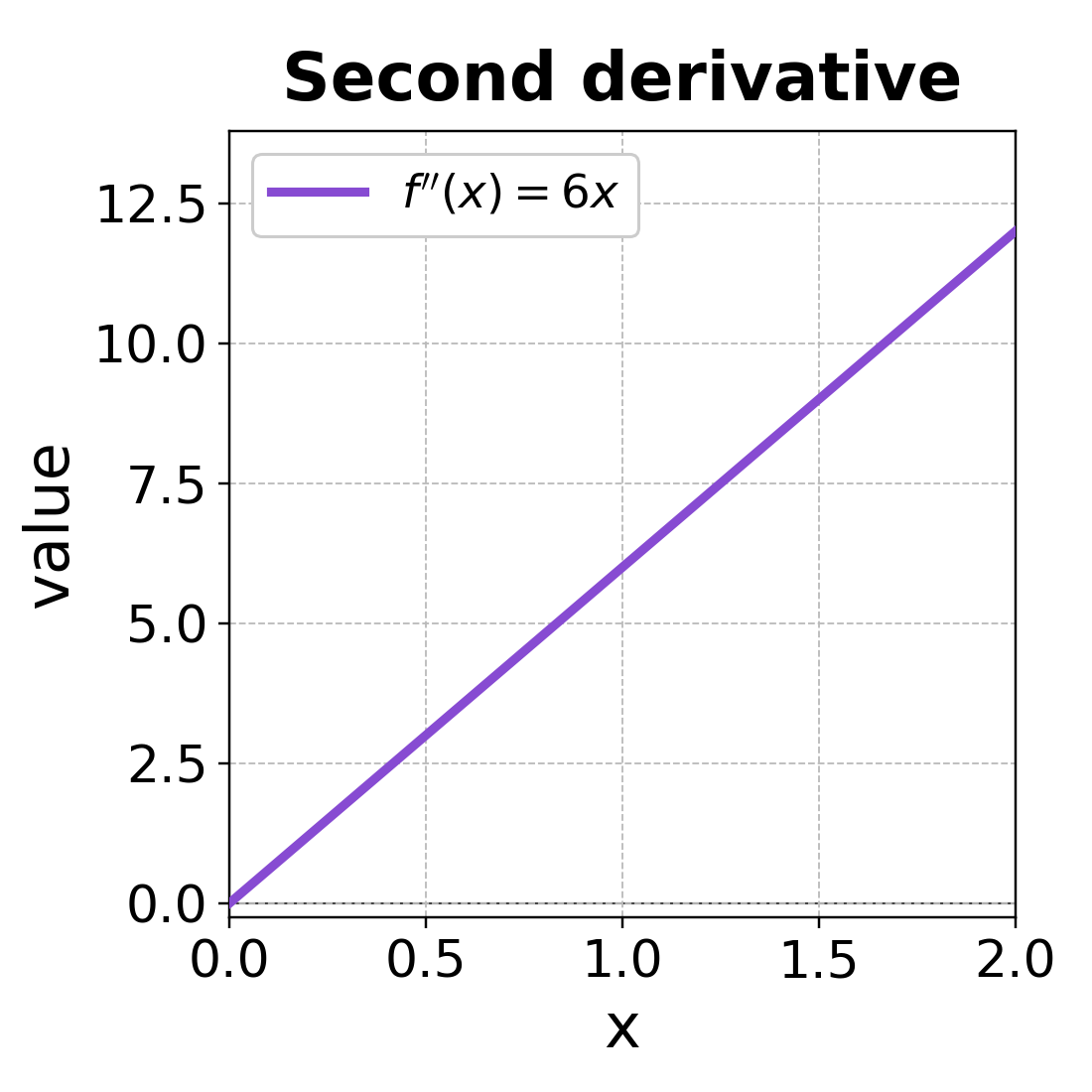}\\
        \small $f''(x)=6x$
    \end{minipage}
    \begin{minipage}{0.19\textwidth}
        \centering
        \includegraphics[width=\linewidth]{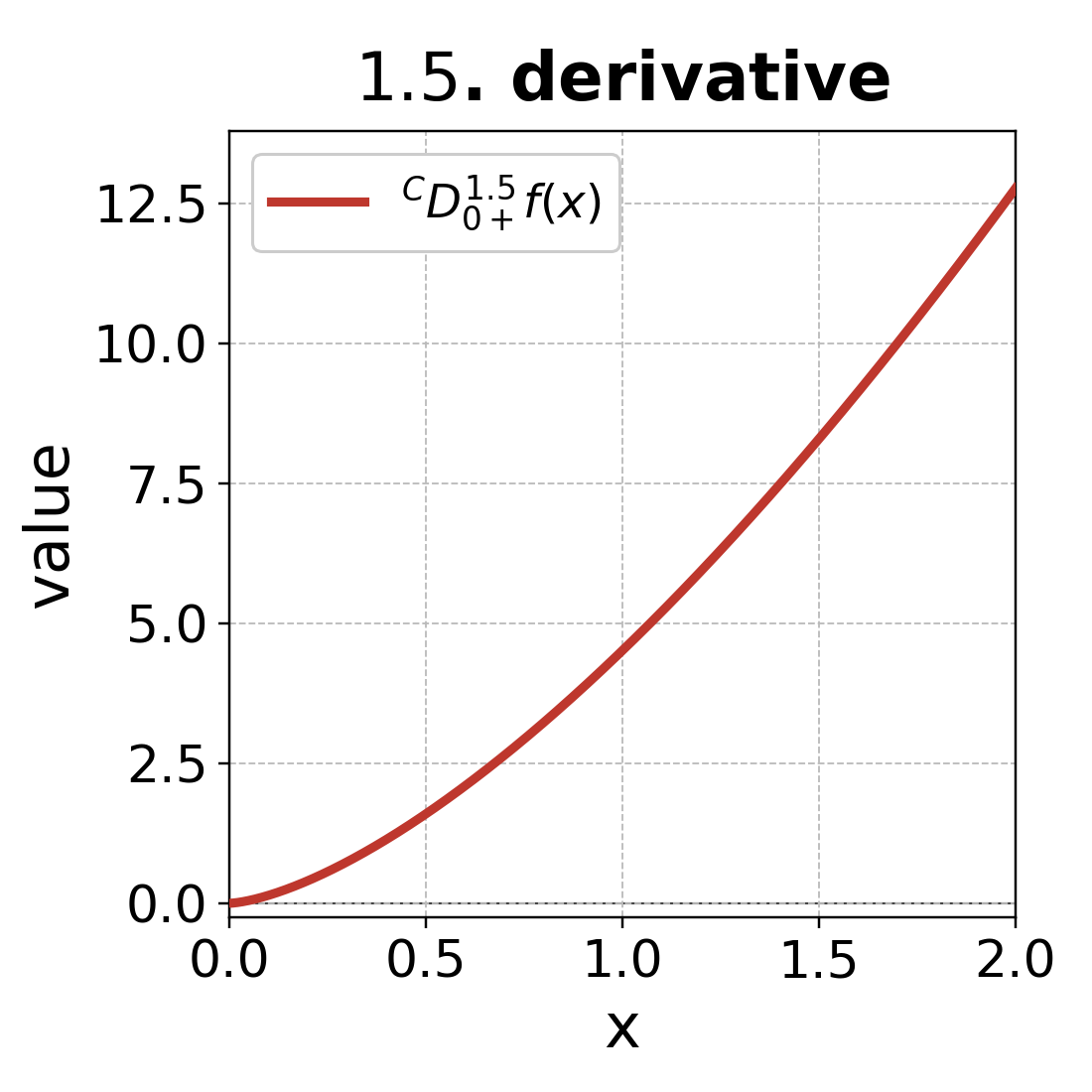}\\
        \small ${}^{C}D_{0+}^{3/2}f(x)$
    \end{minipage}
    \begin{minipage}{0.19\textwidth}
        \centering
        \includegraphics[width=\linewidth]{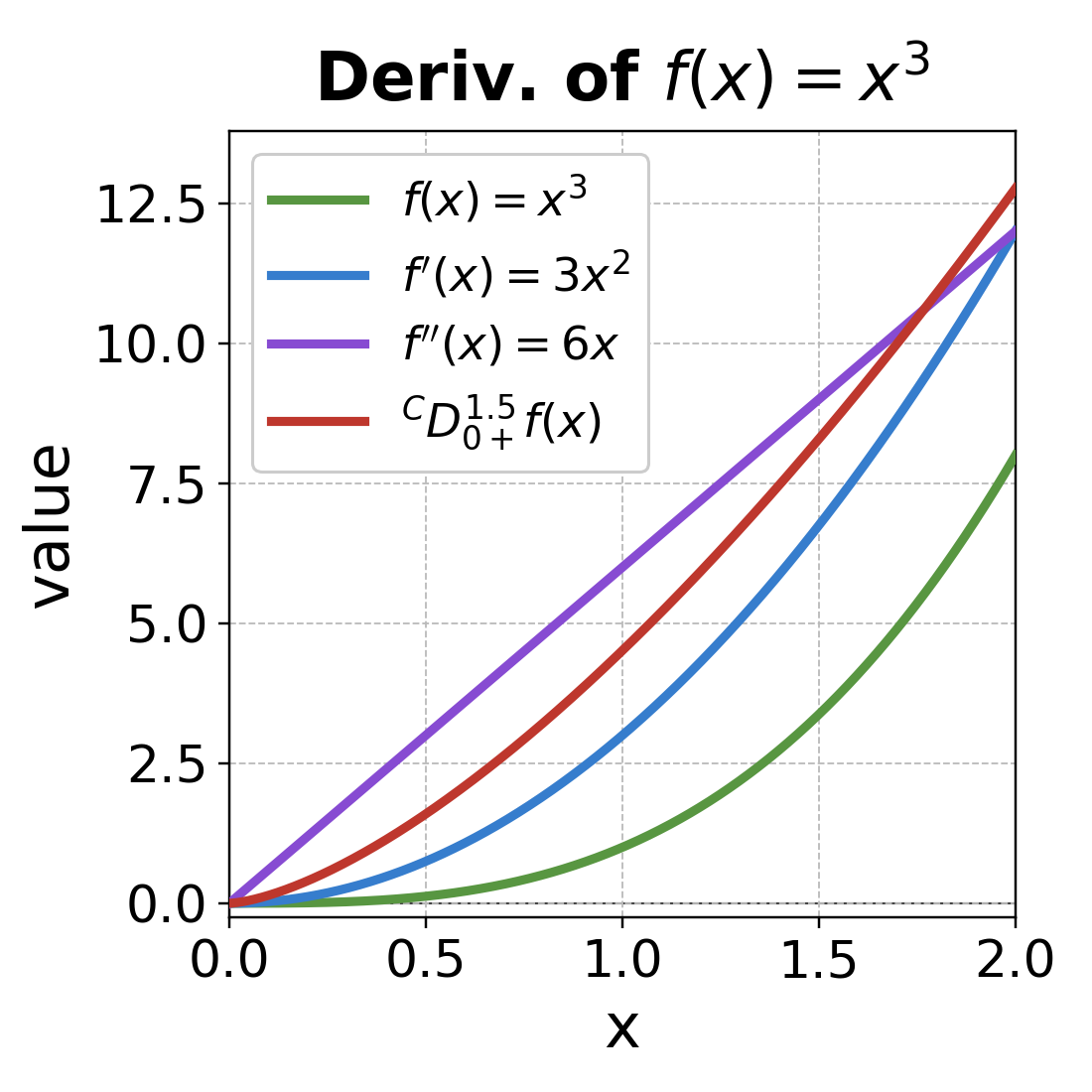}\\
        \small all together
    \end{minipage}

    \caption{Integer and fractional derivatives of the polynomial
    $f(x)=x^3$. The first derivative describes slope, the second derivative
    describes change of slope, and the derivative of order $3/2$ gives an
    intermediate response.}
    \label{fig:intuition_fractional_derivative_polynomial}
\end{figure}

Ordinary derivatives measure local change. The first derivative measures
slope, and the second derivative measures change of slope, or curvature.
For a smooth scalar function $f$, these are given by
\begin{equation}
\label{eq:intuition_first_second_derivative}
f'(x)
=
\lim_{h_{\mathrm{step}}\to 0}
\frac{f(x+h_{\mathrm{step}})-f(x)}{h_{\mathrm{step}}},
\qquad
f''(x)
=
\lim_{h_{\mathrm{step}}\to 0}
\frac{f'(x+h_{\mathrm{step}})-f'(x)}{h_{\mathrm{step}}}.
\end{equation}
These formulas use local information near $x$. A fractional derivative
keeps the idea of measuring change, but it does so at a noninteger order
and usually with a memory term. Instead of asking only for the slope at one
point, it combines information over an interval or over a history of
sampled values \cite{oldham1974fractional,podlubny1999fractional,samko1993fractional}.
For an accessible closed-form example, consider the polynomial
\begin{equation}
\label{eq:intuition_polynomial_example}
f(x)=x^3,
\qquad x\ge 0.
\end{equation}
The ordinary first and second derivatives are
\begin{equation}
\label{eq:intuition_polynomial_integer_derivatives}
f'(x)=3x^2,
\qquad
f''(x)=6x.
\end{equation}
A derivative of order $3/2$ lies between these two integer orders. Using
the Caputo derivative with lower terminal $0$, and using the standard power
rule \cite{diethelm2010analysis,kilbas2006theory}
\begin{equation}
\label{eq:intuition_caputo_power_rule}
{}^{C}D_{0+}^{\nu}x^p
=
\frac{\Gamma(p+1)}{\Gamma(p+1-\nu)}
x^{p-\nu},
\qquad
p\ge \lceil \nu\rceil,
\end{equation}
one obtains
\begin{equation}
\label{eq:intuition_polynomial_fractional_derivative}
{}^{C}D_{0+}^{3/2}x^3
=
\frac{\Gamma(4)}{\Gamma(5/2)}x^{3/2}
=
\frac{8}{\sqrt{\pi}}x^{3/2}.
\end{equation}
The result is neither the first derivative nor the second derivative. It
has an intermediate order: it differentiates more strongly than
$f'(x)$, but less strongly than applying two full derivatives. This is the
basic intuition behind noninteger differentiation, Figure \ref{fig:intuition_fractional_derivative_polynomial}.

The Caputo formula is useful for this polynomial example because it gives a
simple closed form. The optimizers in this article use the discrete
Gr\"unwald--Letnikov point of view, because optimization proceeds in
iterations. For a sampled sequence $u_t=u(t\,h_{\mathrm{step}})$, the
backward Gr\"unwald--Letnikov derivative of order $\nu$ is
\begin{equation}
\label{eq:intuition_gl_derivative}
D_{h_{\mathrm{step}}}^{\nu}u_t
=
\frac{1}{h_{\mathrm{step}}^{\nu}}
\sum_{k=0}^{t} c_k^{(\nu)}u_{t-k},
\qquad
c_0^{(\nu)}=1,
\qquad
c_k^{(\nu)}
=
\left(1-\frac{\nu+1}{k}\right)c_{k-1}^{(\nu)}.
\end{equation}
This formula is important for the intuition of the memory-based optimizers.
For $\nu=1$, the coefficient sequence collapses to the ordinary first
difference. For noninteger $\nu$, more past values receive nonzero weights.
The derivative is therefore not only a local operation; it carries a
weighted memory of previous values. Replacing the sequence $u_t$ by the
gradient sequence $g_t$ gives the finite-history fractional gradient
$g_t^{(\nu)}$ used later in the optimizer definitions. In this sense,
fractional derivatives and fractal functions are linked by the same
multi-scale idea: both describe behavior that cannot be reduced to one
single local scale \cite{michels2012grunwald,zahle1996fractional}.

\subsection{Fractional Derivatives of Fractal Functions}
\label{subsec:intuition_fractional_derivatives_fractal_functions}

The previous subsection introduced fractional derivatives with a smooth
polynomial. This subsection applies the same idea to a finite
Weierstrass-type function. The purpose is again explanatory. It shows why
ordinary pointwise derivatives can become difficult to use on functions
with strong small-scale structure, and why a finite-memory fractional
derivative can provide a more stable object for optimization.

Consider again the finite Weierstrass-type partial sum
\begin{equation}
\label{eq:intuition_weierstrass_derivative_example}
W_N(x)
=
\sum_{m=0}^{N-1} a^m \cos\!\bigl(b^m\pi x\bigr),
\qquad
0<a<1,\qquad b>1.
\end{equation}
For finite $N$, this is a finite sum of smooth functions. Its ordinary
first derivative therefore exists and is given by
\begin{equation}
\label{eq:intuition_weierstrass_classical_derivative}
\frac{d}{dx}W_N(x)
=
-\pi
\sum_{m=0}^{N-1}
a^m b^m \sin\!\bigl(b^m\pi x\bigr).
\end{equation}
The important point is the factor $a^m b^m=(ab)^m$. If $ab>1$, then the
amplitude of the derivative contribution from the higher-frequency terms
grows with $m$. The original function can still look bounded and structured,
because the amplitudes $a^m$ decrease. Its ordinary derivative, however,
can become large and strongly oscillatory as more scales are added. This is
one reason why fractal or near-fractal functions are difficult for methods
that rely only on local first-order information.

\begin{figure}[H]
    \centering

    \begin{minipage}{0.24\textwidth}
        \centering
        \includegraphics[width=\linewidth]{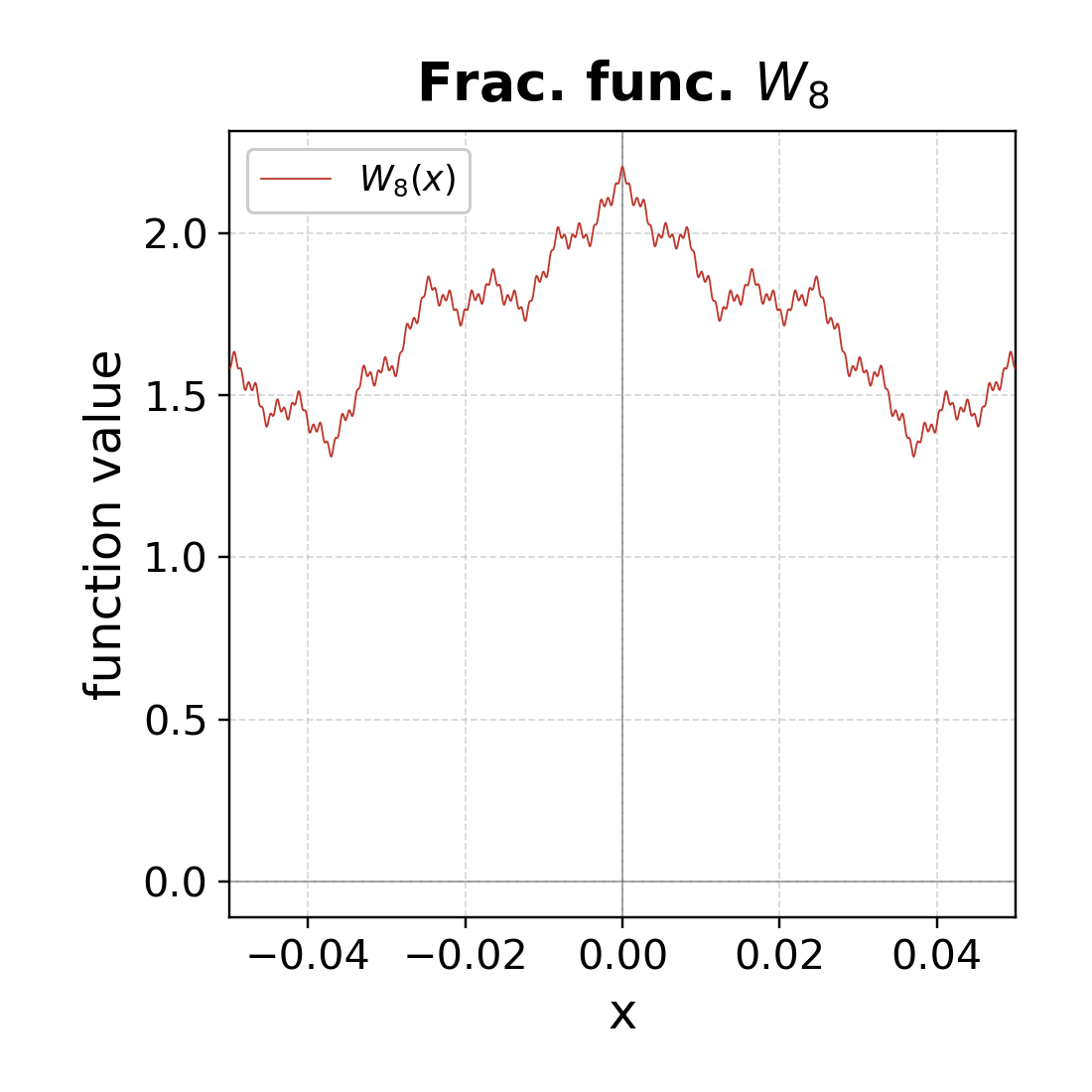}\\
        \small $W_8(x)$
    \end{minipage}
    \hfill
    \begin{minipage}{0.24\textwidth}
        \centering
        \includegraphics[width=\linewidth]{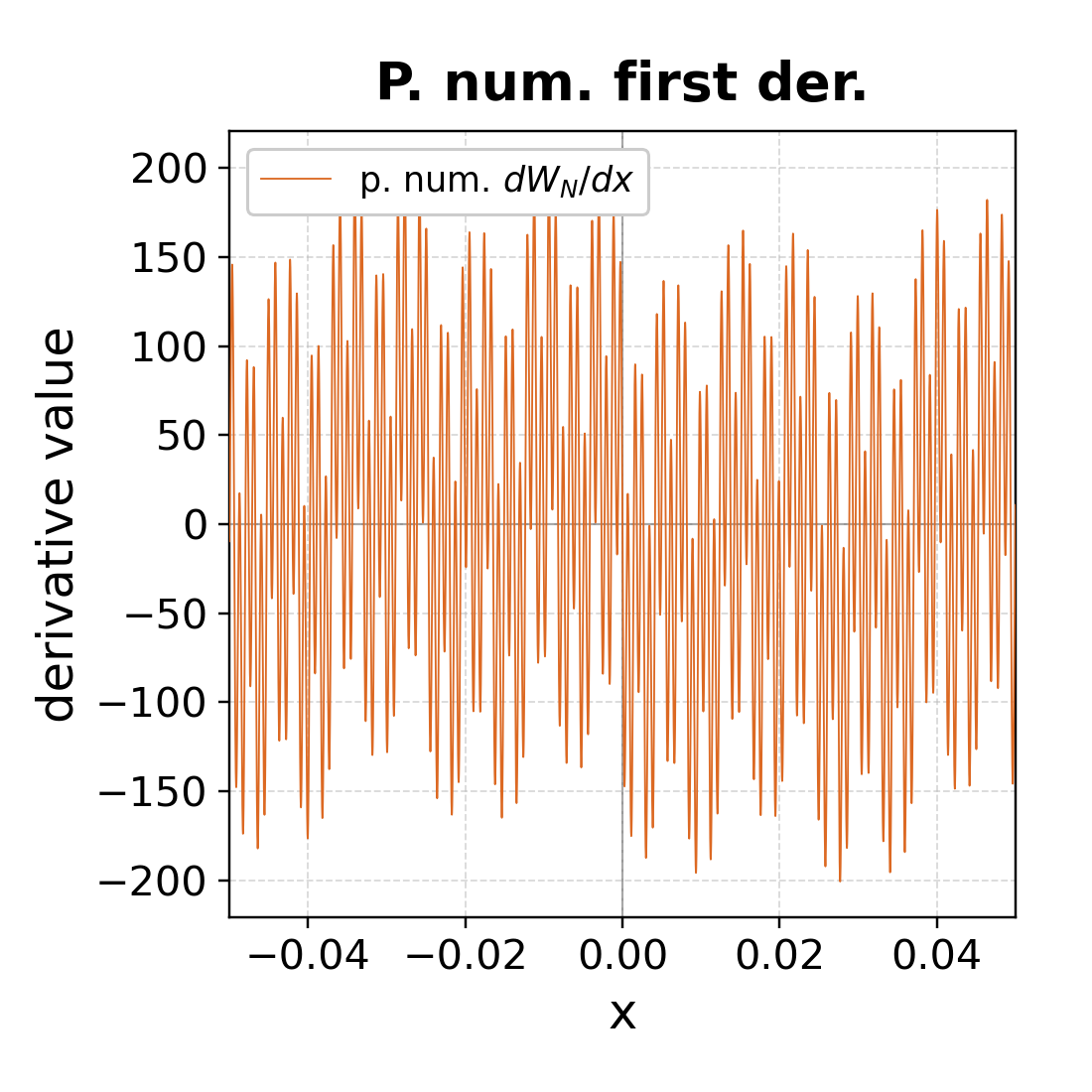}\\
        \small pointwise derivative
    \end{minipage}
    \hfill
    \begin{minipage}{0.24\textwidth}
        \centering
        \includegraphics[width=\linewidth]{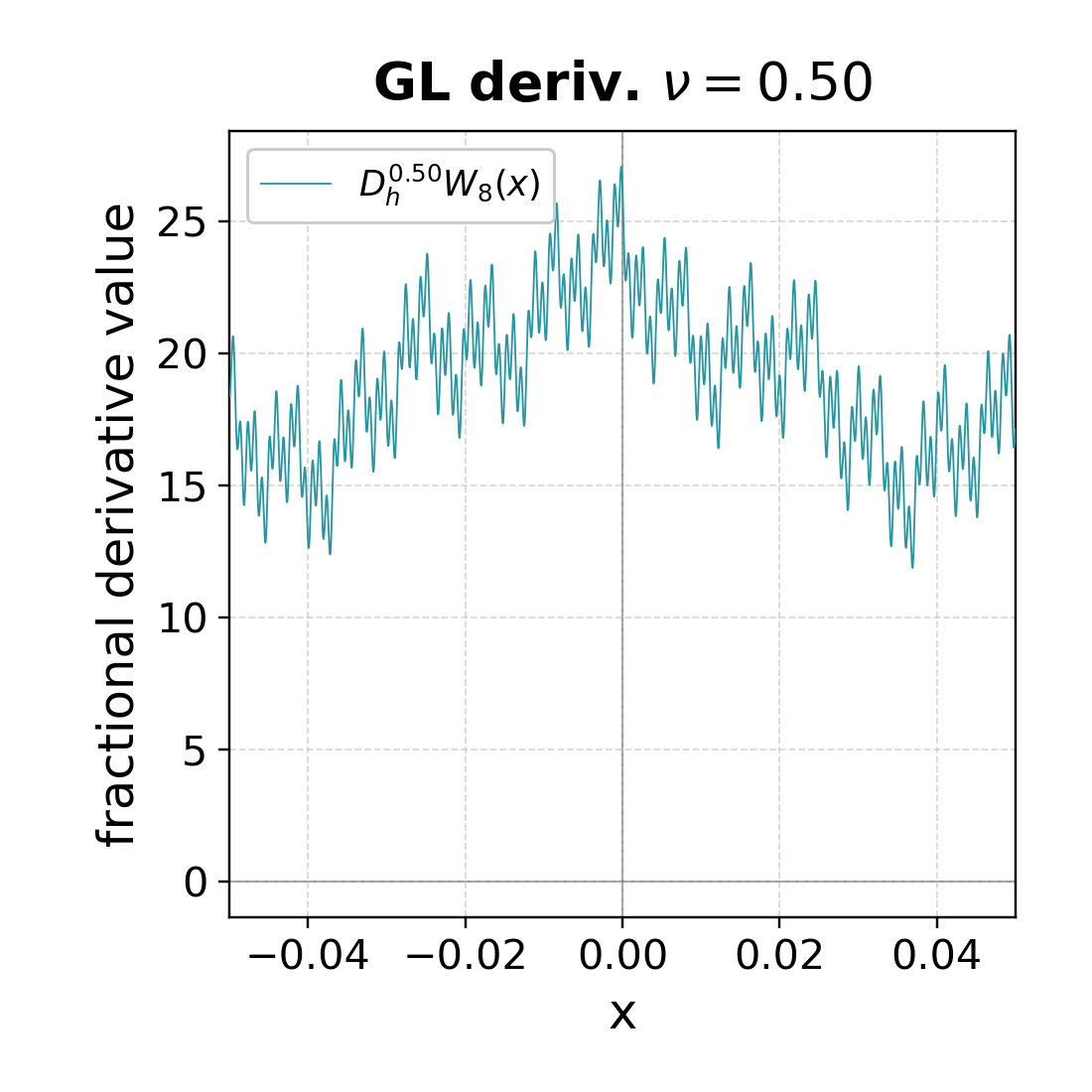}\\
        \small $D_h^{0.50}W_8(x)$
    \end{minipage}
    \hfill
    \begin{minipage}{0.24\textwidth}
        \centering
        \includegraphics[width=\linewidth]{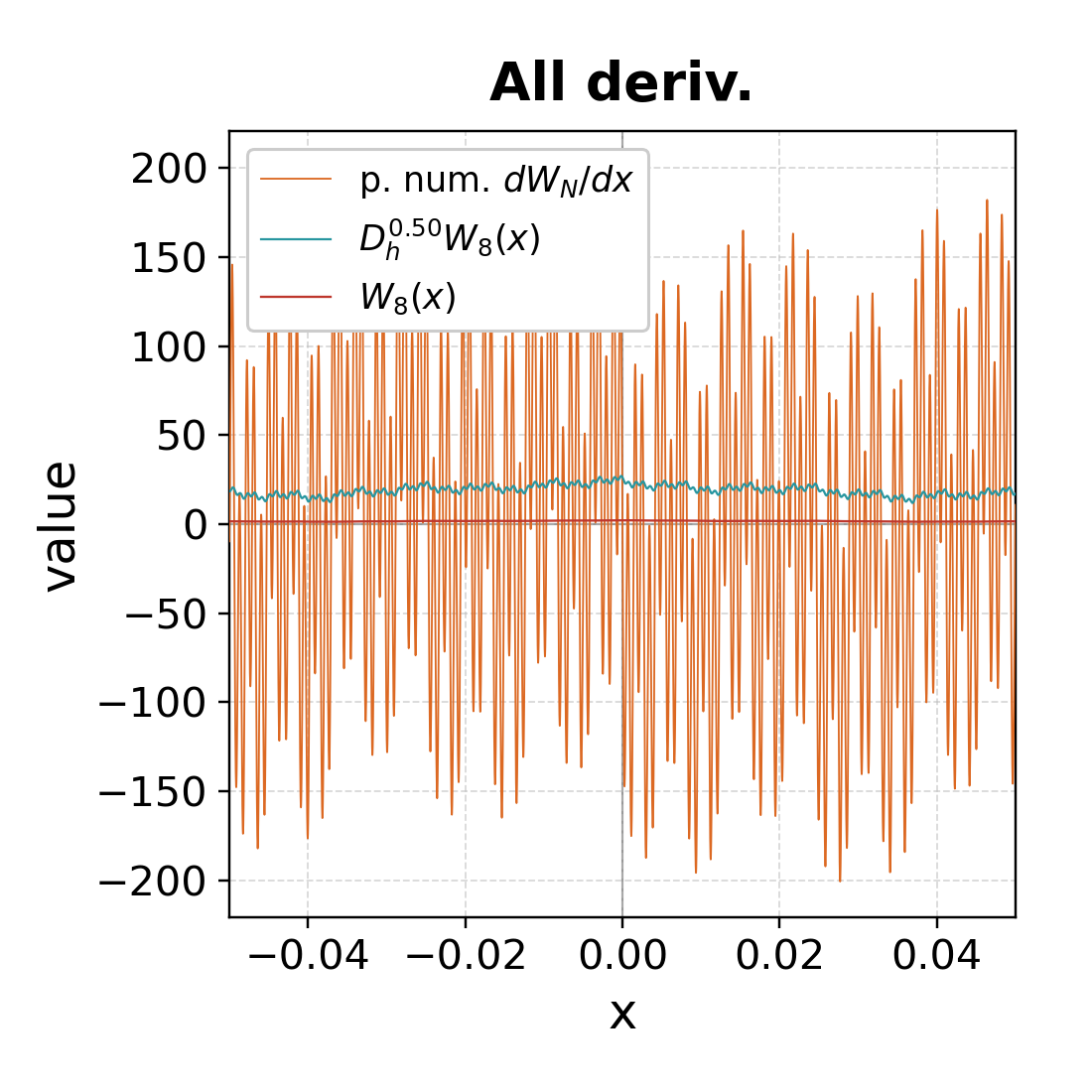}\\
        \small all together
    \end{minipage}

    \caption{Ordinary and fractional-memory derivatives of a finite
    Weierstrass-type function. The function $W_8$ remains bounded on the
    displayed interval, while its pointwise numerical derivative is much
    more oscillatory. The finite-memory Gr\"unwald--Letnikov derivative of
    order $\nu=0.50$ uses a weighted history and therefore gives a different
    response to the same small-scale structure.}
    \label{fig:intuition_weierstrass_fractional_derivatives}
\end{figure}

The finite-memory Gr\"unwald--Letnikov derivative gives a different view.
For sampled values $u_i=W_N(x_i)$ with spacing $h_{\mathrm{step}}$, it is
computed as
\begin{equation}
\label{eq:intuition_weierstrass_gl_derivative}
D_{h_{\mathrm{step}}}^{\nu}u_i
=
\frac{1}{h_{\mathrm{step}}^{\nu}}
\sum_{k=0}^{\min(i,K-1)}
c_k^{(\nu)}u_{i-k},
\qquad
c_0^{(\nu)}=1,
\qquad
c_k^{(\nu)}
=
\left(1-\frac{\nu+1}{k}\right)c_{k-1}^{(\nu)}.
\end{equation}
For $\nu=1$, this construction reduces to a first-difference form. For
$0<\nu<1$, it combines the current value with a finite history of previous
values. In the visual example below, Figure \ref{fig:intuition_weierstrass_fractional_derivatives}, the same finite Weierstrass-type
function $W_8$ is shown together with a pointwise numerical first derivative
and a finite-memory Gr\"unwald--Letnikov derivative of order $\nu=0.50$.
The ordinary numerical derivative reacts strongly to the small-scale
oscillations. The fractional-memory derivative still reflects the local
structure, but its response is shaped by the finite history.

This example gives the intuition behind the coupling of fractal functions
and fractional optimizers. Fractal functions introduce structure across
several scales. A purely local derivative can be dominated by the finest
resolved oscillations. A fractional-memory construction does not remove the
multi-scale structure, but it evaluates it through a weighted history. When
the same construction is applied to gradient sequences in optimization, the
update direction can therefore include information from previous steps
instead of reacting only to the current local gradient. This is the basic
reason why fractal activations and memory-based fractional optimizers are a
natural pair in the experiments of this article.

\begin{takeawaybox}[redA]{Main Takeaways --- Fractals and Fractional Derivatives}
  \begin{itemize}[
    label={},
    leftmargin=0em,
    itemindent=0em,
    itemsep=3pt,
    topsep=2pt
  ]
    \item \textbf{Fractals and fractal functions build structure across scales.}
          Geometric fractals repeat construction rules, while
          Weierstrass-type functions transfer this idea to curves by adding
          oscillations with decreasing amplitudes and increasing frequencies.

    \item \textbf{Fractal activations add multi-scale structure to smooth
          backbones.} A standard activation such as tanh is supplemented by
          localized oscillatory terms at different frequencies.

    \item \textbf{Fractional derivatives provide the optimizer connection.}
          The Gr\"unwald--Letnikov form combines local information with a
          weighted finite history and motivates the memory-based fractional
          updates used later.
  \end{itemize}
\end{takeawaybox}

\section{Weierstrass-Type Functions and Fractional Derivatives}
\label{sec:weierstrass_fractional}

This section explains the mathematical connection between the two central
objects of this article: fractal structures built from Weierstrass-type
functions, and optimization methods built from fractional derivatives.
Both objects appear throughout the experiments. Weierstrass-type
constructions generate the fractal perturbations of the benchmark surfaces
and the fractal activation functions inside the neural networks. Fractional
derivatives generate the memory-based optimizers that train on these
surfaces and networks. The purpose of this section is to show that this
pairing is not arbitrary. Both objects are organized around the same idea,
namely structure that repeats across many scales, and fractional
differentiation is the analytical tool that measures such structure at
exactly the right scale.

The section proceeds in three steps. Section~\ref{subsec:weierstrass_fractional_discussion}
introduces Weierstrass-type functions, defines their natural regularity
measure, the roughness exponent $\gamma_H$, and explains in which precise
sense fractional derivatives of order $\nu$ exist for these functions.
Section~\ref{subsec:discrete_fractional_weierstrass_autodiff} translates
the continuous theory into the discrete Gr\"unwald--Letnikov formulation,
which is the form in which fractional calculus becomes computable and in
which it enters the optimizers of
Section~\ref{sec:fractional_optimizers}. 
Section~\ref{subsec:fractional_gd_fractal_activation} then makes the
interaction concrete for one specific activation function, the modified
Weierstrass--Tanh activation, and derives two optimizer variants for it:
a hereditary variant with explicit gradient memory and a non-hereditary
surrogate without memory. Readers who are mainly interested in the
algorithms can read the summary paragraphs at the end of each subsection
and continue with the experiments.

\subsection{Weierstrass-Type Functions and Their Fractional Differentiability}
\label{subsec:weierstrass_fractional_discussion}

Here we introduce Weierstrass-type functions and the parameters that determine their oscillatory structure and regularity. It then explains how their fractional differentiability depends on the relation between the derivative order and the roughness exponent.

\,\par\noindent\textbf{The building blocks. }
A Weierstrass-type function is assembled from three named ingredients.
The first ingredient is a \emph{periodic generator function}
$h:\mathbb{R}\to\mathbb{R}$ with period $1$ and $h(0)=0$; the generator
determines the shape of a single oscillation. The second ingredient is a
\emph{geometric frequency base} $b\in\{2,3,\dots\}$; it determines how fast
the oscillation frequencies grow from one term of the construction to the
next. The third ingredient is a \emph{roughness exponent}
$\gamma_H\in(0,1)$; it determines how fast the oscillation amplitudes decay
from one term to the next. With these three ingredients, the
Weierstrass-type function $W_h^{\gamma_H}$ is defined as the series
\begin{equation}
\label{eq:weierstrass_type_def}
W_h^{\gamma_H}(x)
=
\sum_{k=1}^{\infty} b^{-k\gamma_H}\, h(b^k x),
\qquad
b \in \{2,3,\dots\},
\qquad
0<\gamma_H<\beta_H\leq 1,
\end{equation}
where the generator $h$ is assumed to be H\"older continuous of order
$\beta_H$ (H\"older continuity is defined below). Each term of the series
adds an oscillation that is $b$ times faster and $b^{\gamma_H}$ times
smaller than the previous one. The classical Weierstrass function
\cite{hardy1916weierstrass} corresponds to the generator
$h(u)=\sin(2\pi u)$, and the Takagi function is obtained from a periodic
tent-type generator. A closely related object is the
\emph{Weierstrass--Mandelbrot function}
\begin{equation}
\label{eq:weierstrass_mandelbrot_def}
M_h^{\gamma_H}(x)
=
\sum_{k=-\infty}^{\infty} b^{-k\gamma_H}\, h(b^k x),
\end{equation}
in which the sum runs over all integers, so that arbitrarily coarse and
arbitrarily fine scales are both present. This two-sided sum satisfies the
exact scaling relation
\begin{equation}
\label{eq:wm_scaling}
M_h^{\gamma_H}(x)=b^{-\gamma_H}\, M_h^{\gamma_H}(b x),
\end{equation}
which states that zooming into the graph by the factor $b$ reproduces the
same graph, rescaled in amplitude by $b^{-\gamma_H}$. Relation
\eqref{eq:wm_scaling} is the cleanest formal expression of self-similarity
in this family and explains why these functions are called fractal.

\,\par\noindent\textbf{Why these functions are rough. }
The functions $W_h^{\gamma_H}$ and $M_h^{\gamma_H}$ are continuous, because
the amplitudes $b^{-k\gamma_H}$ decay geometrically and the series converges
uniformly. At the same time they are rough, because each term adds
oscillations at a finer scale, and the amplitude decay is slow relative to
the frequency growth. Finer scales become smaller, but they do not become
negligible fast enough to produce a classical tangent line. For
$\gamma_H<1$ and suitable parameter ranges, the resulting functions are
continuous but nowhere differentiable in the classical sense
\cite{hardy1916weierstrass}. In other words, the limit
that defines an ordinary first derivative fails at every point, not because
the function jumps, but because ever finer oscillations keep changing the
local slope.

\,\par\noindent\textbf{The correct regularity scale: H\"older continuity. }
Since the ordinary derivative does not exist, differentiability is the
wrong measurement tool for these functions. The appropriate tool is
\emph{H\"older continuity} \cite{jaffard1991pointwise}. A function $f$ is
called H\"older continuous of
order $\gamma_H\in(0,1)$ if there exists a constant $C>0$ such that
\begin{equation}
\label{eq:hoelder_def}
|f(x)-f(y)| \le C\,|x-y|^{\gamma_H}
\qquad\text{for all } x,y.
\end{equation}
Condition \eqref{eq:hoelder_def} states that increments of $f$ over a
distance $|x-y|$ are at most of size $|x-y|^{\gamma_H}$. For $\gamma_H=1$
this is Lipschitz continuity, which is close to differentiability; for
$\gamma_H<1$ it is strictly weaker. For Weierstrass-type functions the
exponent $\gamma_H$ in the definition \eqref{eq:weierstrass_type_def} is
exactly this H\"older exponent, which is why we call it the roughness
exponent. The heuristic is simple: at the scale $y\approx b^{-k}$, the term
with index $k$ dominates the increment, and this term has amplitude
$b^{-k\gamma_H}\approx y^{\gamma_H}$. So the increment
$W_h^{\gamma_H}(x+y)-W_h^{\gamma_H}(x)$ behaves like $y^{\gamma_H}$: the
graph is rough, but the roughness is controlled by a single number,
$\gamma_H$.

\,\par\noindent\textbf{Why fractional derivatives are the natural tool. }
A fractional derivative of order $\nu\in(0,1)$ asks a weaker question than
an ordinary derivative. An ordinary derivative asks whether increments
behave like $|y|^{1}$, that is, linearly. A fractional derivative of order
$\nu$ asks whether increments behave like $|y|^{\nu}$, and it asks this
question in a nonlocal way, by integrating increments over all scales at
once. Comparing with \eqref{eq:hoelder_def}, one expects the following
picture: a function with roughness exponent $\gamma_H$ should admit
fractional derivatives of every order $\nu<\gamma_H$, should be borderline
at $\nu=\gamma_H$, and should fail to be differentiable of any order
$\nu>\gamma_H$. This expectation is correct, and making it precise requires
choosing the right definition of the fractional derivative.
The suitable definition for globally defined rough functions is the
\emph{Weyl--Marchaud derivative}
\cite{samko1993fractional,ferrari2018weyl}. Its left- and right-sided
versions of order $\nu\in(0,1)$ are given formally by
\begin{equation}
\label{eq:weyl_marchaud_def}
D^{\nu}_{-} f(x)
=
\frac{\nu}{\Gamma(1-\nu)}
\int_0^\infty \frac{f(x)-f(x-y)}{y^{1+\nu}}\,dy,
\qquad
D^{\nu}_{+} f(x)
=
\frac{\nu}{\Gamma(1-\nu)}
\int_0^\infty \frac{f(x)-f(x+y)}{y^{1+\nu}}\,dy,
\end{equation}
whenever the integrals converge in the required sense. Here the subscript
$-$ labels the backward increment $f(x)-f(x-y)$ and the subscript $+$ the
forward increment $f(x)-f(x+y)$; this labelling is the mirror image of the
one used in \cite{samko1993fractional,ferrari2018weyl} and is adopted for
consistency with the gradual derivatives
\eqref{eq:gradual_derivative_signed}--\eqref{eq:gradual_derivative_absolute}
below. The decisive
structural feature of \eqref{eq:weyl_marchaud_def} is that it is built from
\emph{increments} $f(x)-f(x\mp y)$ and not from the ordinary derivative
$f'(x)$. The definition therefore remains meaningful for functions that are
nowhere differentiable in the classical sense. The kernel $y^{-1-\nu}$
weights small scales strongly: the smaller the increment scale $y$, the
larger its weight. Convergence of the integral is therefore a competition
between the decay of the increments, which is governed by $\gamma_H$, and
the growth of the kernel, which is governed by $\nu$.

\,\par\noindent\textbf{The subcritical regime $\nu<\gamma_H$. }
For Weierstrass-type functions this competition can be resolved exactly.
If the generator $h$ is H\"older continuous of order $\beta_H$ and
$0<\nu<\gamma_H<\beta_H\le 1$, then the Weyl--Marchaud derivative of
$W_h^{\gamma_H}$ exists and satisfies the identity
\cite{zahle1996fractional}
\begin{equation}
\label{eq:wm_mapping_identity}
D^{\nu}_{\pm}\, W_h^{\gamma_H}(x)
=
W_{D^{\nu}_{\pm} h}^{\,\gamma_H-\nu}(x).
\end{equation}
Identity \eqref{eq:wm_mapping_identity} should be read as a mapping
statement: fractional differentiation of order $\nu$ maps a
Weierstrass-type function with roughness exponent $\gamma_H$ to another
Weierstrass-type function, built from the differentiated generator
$D^{\nu}_{\pm}h$, with the reduced roughness exponent $\gamma_H-\nu$. Three
consequences are worth stating explicitly. First, the multiscale structure
survives differentiation; the result is again a fractal object of the same
family. Second, differentiation of order $\nu$ consumes exactly $\nu$ units
of regularity, which is the fractional generalization of the familiar fact
that one full derivative consumes one unit of smoothness. Third, as long as
$\nu<\gamma_H$, the reduced exponent $\gamma_H-\nu$ is positive, so the
derivative is still a continuous function. This is the subcritical regime,
in which fractional differentiation of a Weierstrass-type function makes
strongest classical sense.

\,\par\noindent\textbf{The critical order $\nu=\gamma_H$. }
The roughness exponent acts as a threshold. Below the threshold the
fractional derivative exists pointwise; above the threshold the kernel
$y^{-1-\nu}$ demands more increment decay than the function possesses, and
the derivative generally ceases to exist as a pointwise object. The
critical case $\nu=\gamma_H$ is the most informative one. At this order the
small-scale part of the integral in \eqref{eq:weyl_marchaud_def} diverges,
but only logarithmically. Z\"ahle and Ziezold \cite{zahle1996fractional}
therefore introduce a \emph{gradual fractional derivative in the mean},
obtained by cutting the integral off at a small scale $\delta$ and
normalizing by the logarithm of the cutoff:
\begin{equation}
\label{eq:gradual_derivative_signed}
d^{\gamma_H}_{-} f(x)
=
\lim_{\delta\to 0^+}
\frac{1}{|\ln\delta|}
\int_\delta^\infty
\frac{f(x)-f(x-y)}{y^{1+\gamma_H}}\,dy,
\end{equation}
together with the corresponding absolute version
\begin{equation}
\label{eq:gradual_derivative_absolute}
|d^{\gamma_H}_{-}| f(x)
=
\lim_{\delta\to 0^+}
\frac{1}{|\ln\delta|}
\int_\delta^\infty
\frac{|f(x)-f(x-y)|}{y^{1+\gamma_H}}\,dy.
\end{equation}
The logarithmic normalization $1/|\ln\delta|$ reflects the borderline
nature of the singularity: the divergence is logarithmic, so dividing by
the logarithm extracts a stable limit. For periodic Weierstrass-type
functions and for the associated Weierstrass--Mandelbrot functions, the
result is the following. The signed gradual derivative
\eqref{eq:gradual_derivative_signed} vanishes almost everywhere, while the
absolute gradual derivative \eqref{eq:gradual_derivative_absolute} exists
almost everywhere and equals a positive constant
\cite{zahle1996fractional}. The interpretation is direct: at the critical
order, the oscillations of the function are perfectly balanced in sign on
average, but their size does not vanish. The critical-order fractional
derivative therefore does not produce a slope; it measures roughness. This
gives a precise meaning to the statement that the fractional order of
differentiability of $W_h^{\gamma_H}$ is exactly $\gamma_H$.

\,\par\noindent\textbf{Which fractional derivative to use. }
Several inequivalent definitions of fractional derivatives exist, and it is
useful to state why the Weyl--Marchaud form is the appropriate one here
\cite{oliveira2014review,samko1993fractional}. The \emph{Caputo} derivative
is defined through an integral of the ordinary derivative $f'$; since a
Weierstrass-type function has no ordinary derivative, the Caputo framework
is not aligned with the regularity of the object. The
\emph{Riemann--Liouville} derivative can be defined under suitable interval
and boundary conventions and coincides almost everywhere with the Marchaud
derivative on sufficiently regular classes, but it is tied to a fixed base
point, which fits poorly with global periodicity and self-similarity. The
\emph{Weyl--Marchaud} derivative \eqref{eq:weyl_marchaud_def} is built
directly from increments, is compatible with periodicity, and interacts
cleanly with the scaling relation \eqref{eq:wm_scaling}; this is why the
exact identity \eqref{eq:wm_mapping_identity} holds in this framework. The
\emph{Gr\"unwald--Letnikov} derivative, finally, is best understood in this
context as the discrete computational representation of the same nonlocal
idea \cite{rogosin2017letnikov,podlubny1999fractional}; it is the form we
use in all algorithms and is developed in the next subsection.

\,\par\noindent\textbf{Summary in plain terms. }
A Weierstrass-type function is rough because it contains oscillations on
infinitely many scales, organized by the frequency base $b$ and damped by
the roughness exponent $\gamma_H$. Ordinary differentiation fails because
it asks for a single local tangent, and the finer scales keep destroying
that tangent. Fractional differentiation asks a weaker, scale-weighted
question about increments. If the order $\nu$ is below the roughness level
$\gamma_H$, the answer is finite and is again a Weierstrass-type function
with roughness $\gamma_H-\nu$. If $\nu=\gamma_H$, the correct quantity is a
logarithmically averaged one, and it returns a constant that quantifies the
roughness itself. If $\nu>\gamma_H$, the function is too rough for that
order. Fractional derivatives do not bypass the roughness of the
Weierstrass function; they measure it at the correct scale. This is the
property that motivates everything that follows: if roughness across scales
is the phenomenon, then operators of fractional order are the matched
instruments.

\subsection{The Discrete Gr\"unwald--Letnikov Formulation and Automatic
Differentiation}
\label{subsec:discrete_fractional_weierstrass_autodiff}

The previous subsection explains why Weierstrass-type functions are natural
test objects for fractional differentiation. For numerical work, however,
one does not evaluate singular integrals such as
\eqref{eq:weyl_marchaud_def} directly. Instead, one replaces the continuum
operator by a discrete memory operator. This is where the
Gr\"unwald--Letnikov formulation becomes central: it already has the
structure of a weighted backward difference, and this is exactly the
structure that numerical algorithms and tensor libraries evaluate
most efficiently \cite{podlubny1999fractional,michels2012grunwald}. It is also
the structure that reappears, unchanged, inside the memory-based optimizers
of Section~\ref{sec:fractional_optimizers}. 

\,\par\noindent\textbf{The discrete fractional derivative. }
Let $u:\mathbb{R}\to\mathbb{R}$ be sampled on a uniform grid
$t_n = n\,h_{\mathrm{step}}$, where $h_{\mathrm{step}}>0$ is the
discretization step and $u_n := u(t_n)$ denotes the sample
at index $n$. For an order $\nu\in(0,2)$, the backward
Gr\"unwald--Letnikov derivative is written as
\begin{equation}
\label{eq:gl_discrete_def}
D_{h_{\mathrm{step}}}^{\nu} u_n
:=
\frac{1}{h_{\mathrm{step}}^{\nu}}
\sum_{k=0}^{n} c_k^{(\nu)}\, u_{n-k},
\qquad
c_k^{(\nu)} = (-1)^k \binom{\nu}{k},
\qquad
\binom{\nu}{k}
=
\frac{\Gamma(\nu+1)}{\Gamma(k+1)\,\Gamma(\nu-k+1)}.
\end{equation}
The generalized binomial coefficient is well defined for noninteger $\nu$.
The coefficient sequence $c_k^{(\nu)}$ is the single most important object
of this subsection, because it is shared by the analysis and by
the optimizers: the same sequence defines the finite-history fractional
gradient $g_t^{(\nu)}$ used in
Sections~\ref{sec:fractional_optimizers} and~\ref{subsec:adaptive_memory_optimizers}.
It can be generated recursively, without any evaluation of Gamma
functions, through
\begin{equation}
\label{eq:gl_coefficient_recursion}
c_0^{(\nu)} = 1,
\qquad
c_k^{(\nu)} = \left(1-\frac{\nu+1}{k}\right) c_{k-1}^{(\nu)}
\quad\text{for } k\geq 1.
\end{equation}
Two properties of this sequence carry all of the intuition. First, for
noninteger $\nu$ the coefficients decay only \emph{algebraically} in $k$
(like a power law, not like a geometric sequence). Second, at the integer
order $\nu=1$ the recursion gives $c_0^{(1)}=1$, $c_1^{(1)}=-1$, and
$c_k^{(1)}=0$ for $k\ge 2$, so that \eqref{eq:gl_discrete_def} collapses to
the ordinary backward difference
$(u_n-u_{n-1})/h_{\mathrm{step}}$. The classical derivative is therefore
contained in the fractional family as the special case in which all memory
weights beyond the immediate past vanish.

\,\par\noindent\textbf{Memory as the defining feature. }
The algebraic decay of $c_k^{(\nu)}$ is the computational expression of
memory. In an ordinary first difference, only the nearest neighbor matters.
In the fractional difference \eqref{eq:gl_discrete_def}, the whole past
matters: distant samples are down-weighted, but never discarded. The order
$\nu$ controls how strongly the remote past is remembered. A smaller $\nu$
produces a heavier tail and hence longer memory; a $\nu$ closer to
$1$ produces behavior closer to an ordinary derivative. In plain terms, a
discrete fractional derivative is not a local slope computed from one small
increment; it is a history-aware weighted sum of many increments.

The same statement can be made in linear-algebra form. Collecting the
samples in a vector $\mathbf{u}=(u_0,\dots,u_T)^\top$, the operator
\eqref{eq:gl_discrete_def} is the lower-triangular Toeplitz matrix
\begin{equation}
\label{eq:gl_toeplitz}
\mathbf{D}^{(\nu)}_{h_{\mathrm{step}}}\, \mathbf{u}
=
h_{\mathrm{step}}^{-\nu}
\begin{bmatrix}
c_0^{(\nu)} & 0 & 0 & \cdots & 0\\
c_1^{(\nu)} & c_0^{(\nu)} & 0 & \cdots & 0\\
c_2^{(\nu)} & c_1^{(\nu)} & c_0^{(\nu)} & \cdots & 0\\
\vdots & \vdots & \vdots & \ddots & \vdots\\
c_T^{(\nu)} & c_{T-1}^{(\nu)} & c_{T-2}^{(\nu)} & \cdots & c_0^{(\nu)}
\end{bmatrix}
\mathbf{u},
\end{equation}
that is, a discrete convolution with a fixed long-memory kernel. In the
discrete setting, a fractional derivative is a concrete linear map, and its
lower-triangular shape encodes causality: the value at index $n$ depends
only on the present and the past.

\,\par\noindent\textbf{Truncation and the memory window $K$. }
A full evaluation of \eqref{eq:gl_discrete_def} has cost growing with the
history length, so truncation is unavoidable in applications. One
introduces a \emph{memory window} $K$ and uses the short-memory
approximation
\begin{equation}
\label{eq:gl_truncated}
D_{h_{\mathrm{step}},K}^{\nu} u_n
:=
\frac{1}{h_{\mathrm{step}}^{\nu}}
\sum_{k=0}^{\min(n,\,K-1)} c_k^{(\nu)}\, u_{n-k}.
\end{equation}
This truncation changes the operator: it is no longer the full fractional
derivative, but a finite-memory approximation of it. In optimization
language, the method now has a controllable memory horizon. In analysis
language, a genuinely nonlocal operator has been replaced by a regularized
one. The trade-off is standard: a larger $K$ is closer to the intended
fractional model but costs more computation and can amplify numerical
sensitivity. Short-memory truncation of exactly this type is used in the
Gr\"unwald--Letnikov deep-learning optimizers of Zhou et al.
\cite{zhou2023gl}, and it is the form used by every memory-based optimizer
in this article.

\,\par\noindent\textbf{What the grid does to a Weierstrass-type function. }
The discrete operator \eqref{eq:gl_truncated} connects naturally to
Weierstrass-type functions because such functions are themselves multiscale
sums. Sampling $W_h^{\gamma_H}$ from \eqref{eq:weierstrass_type_def} on a
spatial grid $x_n=n\,\Delta x$ gives
$W_n := W_h^{\gamma_H}(x_n)=\sum_{m=1}^{\infty} b^{-m\gamma_H}
h(b^m n\,\Delta x)$, and applying the discrete derivative and exchanging
the (absolutely convergent) sums yields
\begin{equation}
\label{eq:gl_on_weierstrass}
D_{\Delta x}^{\nu} W_n
=
\sum_{m=1}^{\infty} b^{-m\gamma_H}\,
\frac{1}{(\Delta x)^{\nu}}
\sum_{k=0}^{n} c_k^{(\nu)}\, h\!\left(b^m (n-k)\,\Delta x\right).
\end{equation}
Equation \eqref{eq:gl_on_weierstrass} displays the central structure
of the discrete theory: the fractional difference interacts with each scale
$m$ separately, but the same long-memory stencil $c_k^{(\nu)}$ is applied
to all scales. In the continuous setting, differentiation of order $\nu$
lowers the roughness exponent from $\gamma_H$ to $\gamma_H-\nu$, by
identity \eqref{eq:wm_mapping_identity}. In the discrete setting the same
tendency appears, but filtered by the grid: oscillations finer than the
grid scale are invisible or misrepresented (aliasing), so the observed
fractional derivative depends not only on $\nu$ and $\gamma_H$ but also on
the relation between the frequency ladder $b^m$ and the sampling scale
$\Delta x$. In plain terms, the discrete derivative does not see the exact
mathematical fractal; it sees the part of the fractal that fits on the
grid. This matters because the fractional derivative is designed to be
sensitive to multiscale structure: if the grid removes some scales, the
computed derivative changes accordingly.

\,\par\noindent\textbf{Two distinct uses of fractional ideas in machine learning. }
For the machine-learning part of this article it is essential to separate
two constructions that are both described with fractional language but are
algorithmically different components.

The first construction is a \emph{fractionalized scalar activation}. Let
$z$ denote the pre-activation of a layer. Given a base activation
$\phi_0$ (for example a sigmoid or hyperbolic tangent), one defines
\begin{equation}
\label{eq:fractional_activation_abstract}
\phi_{\nu}(z)
=
\mathcal{F}_{\nu}[\phi_0](z),
\end{equation}
where $\mathcal{F}_{\nu}$ is a fractional operator acting with respect to
the scalar argument of $\phi_0$, implemented in practice through a
truncated stencil on a local argument grid or through a closed-form
approximation. The point is that $\phi_\nu(z)$ depends on the single value
$z$ only. It modifies the pointwise nonlinearity of the network, that is,
it acts on \emph{signal geometry in feature space}.

The second construction is a \emph{fractional history operator} applied
along the iteration axis of training. Given the gradient sequence
$g_0,g_1,\dots$, with $g_t=\nabla_\theta L(\theta_t)$, one forms the
finite-history fractional gradient
\begin{equation}
\label{eq:gl_history_operator}
g_t^{(\nu)}
=
\sum_{k=0}^{\min(t,\,K-1)} c_k^{(\nu)}\, g_{t-k},
\end{equation}
and uses it in a parameter update such as
$\theta_{t+1}=\theta_t-\eta\, h_{\mathrm{step}}^{-\nu}\, g_t^{(\nu)}$. This
object depends on a whole sequence of past gradients. It resembles momentum, but the kernel is not exponential; it decays algebraically,
which is the defining signature of fractional memory. It acts on
\emph{memory in parameter space}. Both constructions originate from the
same nonlocal idea, but the first changes what the network computes and the
second changes how training aggregates information over time. In this
article, the fractal activation functions realize (a truncated, explicit
version of) the first mechanism, and the memory-based optimizers realize
the second.

\,\par\noindent\textbf{What automatic differentiation actually computes. }
The distinction above becomes clear when automatic differentiation is
considered. Suppose the fractionalized activation
\eqref{eq:fractional_activation_abstract}, or any explicit truncated
Weierstrass-type activation, is implemented pointwise as a tensor
expression. Then the training framework differentiates it by the ordinary
chain rule of the computation graph: for $y=\phi_\nu(z)$ and loss $L(y)$,
reverse-mode automatic differentiation computes
\begin{equation}
\label{eq:autodiff_chain_rule}
\frac{\partial L}{\partial z}
=
\frac{\partial L}{\partial y}\,
\frac{\partial \phi_\nu(z)}{\partial z},
\end{equation}
provided the implemented map $z\mapsto\phi_\nu(z)$ is differentiable as a
tensor program. Nothing fractional happens inside the differentiation
engine. One is not asking the framework to compute a fractional derivative
with respect to $z$; one is asking it to compute the ordinary derivative of
a function whose \emph{formula} was built from fractional calculus. The
backpropagation rule is the usual one; only the forward function has
changed.

The same principle applies to the history operator
\eqref{eq:gl_history_operator}. If the operator is coded as a finite
convolution or triangular recurrence of tensor operations, it is
differentiable through the standard mechanisms (in TensorFlow,
reverse-mode differentiation via \texttt{tf.GradientTape}
\cite{tensorflow_autodiff_guide,tensorflow_gradienttape_api} and forward-mode
differentiation via \texttt{tf.autodiff.ForwardAccumulator}
\cite{tensorflow_forwardaccumulator_api}), and a custom
backward rule can be supplied where stability or efficiency requires it
\cite{tensorflow_custom_gradient_api}. Two practical consequences follow.
First, a Gr\"unwald--Letnikov optimizer stores a gradient history of length
$K$, so its state is larger than that of SGD or Adam, which store only one
or two exponentially weighted moments; the memory window $K$ is therefore
also a computational budget. Second, fractional calculus has its own
chain-rule-type identities, but these are \emph{not} what the framework
applies; the framework differentiates the discrete surrogate model that was
actually implemented. The discrete formulation is therefore not a numerical
afterthought. It is the object that the optimization software sees.

\,\par\noindent\textbf{Summary in plain terms. }
Fractional calculus becomes computable once it is rewritten as a weighted
history sum with the kernel $c_k^{(\nu)}$ of
\eqref{eq:gl_coefficient_recursion}. This kernel has a power-law tail,
which distinguishes fractional memory from the exponential forgetting of
momentum methods, and it reduces exactly to the classical one-step
difference at $\nu=1$. Applied to a sampled Weierstrass-type function, the
kernel probes the multiscale ladder $b^m$ as far as the grid resolves it.
Applied along the training, the same kernel defines the
finite-history fractional gradient $g_t^{(\nu)}$ that powers the
memory-based optimizers of this article. In both cases automatic
differentiation proceeds by the ordinary chain rule over the implemented
discrete operations. One kernel, two uses: as an analytical probe of
roughness and as a memory mechanism for optimization.

\subsection{Fractional Gradient Descent With and Without Memory for a
Weierstrass-Based Activation}
\label{subsec:fractional_gd_fractal_activation}

The two preceding subsections establish, first, that fractional derivatives
are the matched instruments for Weierstrass-type roughness and, second,
that their discrete Gr\"unwald--Letnikov form is a history-weighted sum
that can serve as an optimizer update. This subsection combines the two
threads in the most concrete setting available in this article: one
specific fractal activation function inside a neural network, trained by
fractional gradient descent with and without memory.

\,\par\noindent\textbf{The chosen activation. }
Among the fractal activation functions considered in this work, the most
suitable candidate for a detailed optimizer-level discussion is the
\emph{modified Weierstrass--Tanh activation} from the activation study of
Raubitzek et al. \cite{raubitzek_fractals_2026}. In the experimental
results of that study it is one of the most reliable fractal activations
across datasets, and in its gradient-stability analysis it remains in a
substantially milder regime than the more extreme Weierstrass-based
variants, in particular the ReLU-based modification.
We consider the finite truncation
\begin{equation}
\label{eq:mw_tanh_activation}
\phi(x)
=
\tanh(x)
+
\sum_{m=0}^{N-1}
(-1)^m a^m \cos(b^m \pi x)\,e^{-\mu_e |x|},
\qquad
0<a<1,\quad b>1,\quad \mu_e>0,
\end{equation}
with the representative parameter choice $a=0.5$, $b=1.5$, $\mu_e=0.75$,
and truncation length $N=100$. Every ingredient of
\eqref{eq:mw_tanh_activation} has a named role. The term $\tanh(x)$ is the
\emph{backbone}: it provides a stable low-frequency nonlinearity and
prevents the activation from being purely oscillatory. The sum is the
\emph{oscillatory ladder}: term $m$ oscillates with frequency proportional
to $b^m$ and amplitude $a^m$, so the ladder has the same
geometric-frequency, geometric-amplitude structure as the Weierstrass-type
series \eqref{eq:weierstrass_type_def}, with amplitude decay written
through the base $a$ rather than through an exponent. The factor
$e^{-\mu_e|x|}$ is the \emph{envelope}: it localizes the oscillations
around the origin and suppresses them for large $|x|$. Because the series
is truncated at $N$ terms, $\phi$ is an ordinary, implementable activation
function; the infinite parent series is a Weierstrass-type object, but the
network only ever sees the truncated function. This repeats, at the level
of a single activation, the grid-filtering principle of
Section~\ref{subsec:discrete_fractional_weierstrass_autodiff}: the
implemented object retains only the scales that are numerically present.

The same geometric ladder also appears in the fractal surface
perturbations of the surface experiments
(Section~\ref{sec:surface_experiments}), 
where the perturbation term
$P_J(x,y)=\sum_{k=0}^{J}a_P^{\,k}[\cos(b_P^{\,k}\pi x)+\cos(b_P^{\,k}\pi y)]$ is added to or
multiplied with the benchmark objectives. The surface experiments and the
activation experiments therefore probe one and the same mechanism from two
sides: in the first case the multiscale ladder sits directly in the loss
landscape, in the second case it sits in the network and reaches the loss
landscape through composition.

\,\par\noindent\textbf{The gradient field induced by the activation. }
For $x\neq 0$, the ordinary derivative of the truncated activation
\eqref{eq:mw_tanh_activation} is
\begin{equation}
\label{eq:mw_tanh_derivative}
\phi'(x)
=
\operatorname{sech}^2(x)
+
\sum_{m=0}^{N-1}
(-1)^m a^m e^{-\mu_e |x|}
\left[
-\,b^m\pi \sin(b^m\pi x)
-\mu_e\,\operatorname{sign}(x)\cos(b^m\pi x)
\right].
\end{equation}
Formula \eqref{eq:mw_tanh_derivative} makes the optimization issue visible.
Differentiating term $m$ multiplies its amplitude $a^m$ by its frequency
$b^m\pi$, so the derivative ladder carries the combined factor $(ab)^m$. If
$ab<1$, which holds for the representative choice ($ab=0.75$), this
amplification is controlled and the truncated derivative remains bounded in
a practical sense. Even then, the local slope varies rapidly across nearby
inputs: two almost identical pre-activations can produce noticeably
different local Jacobians. In ordinary gradient descent this appears as a
more irregular gradient field. In a fractional optimizer, the same
irregularity is filtered through the memory kernel $c_k^{(\nu)}$. In plain
terms, $\tanh$ bends the signal once; $\phi$ bends it repeatedly across
many scales, and the optimizer therefore sees gradients that carry a coarse
trend with superimposed fine oscillations.

\,\par\noindent\textbf{Setting and the role of backpropagation. }
Consider a network with parameter vector $\theta_t\in\mathbb{R}^d$ at
iteration $t$ and loss $L(\theta)$, and let
\begin{equation}
g_t := \nabla_\theta L(\theta_t)
\end{equation}
denote the ordinary gradient computed by backpropagation. The first
important point repeats the conclusion of
Section~\ref{subsec:discrete_fractional_weierstrass_autodiff}:
backpropagation itself remains standard. The automatic-differentiation
engine computes $g_t$ by the ordinary chain rule applied to the implemented
computation graph, which includes the truncated activation
\eqref{eq:mw_tanh_activation}. The fractional aspect enters only when the
gradient is \emph{used} for the parameter update, not when the graph
derivative is formed.

\,\par\noindent\textbf{The hereditary optimizer: fractional gradient descent with
memory. }
The genuinely fractional optimizer, in the Gr\"unwald--Letnikov sense, is
the \emph{hereditary} variant. It replaces the instantaneous gradient by
the finite-history fractional gradient of
\eqref{eq:gl_history_operator},
\begin{equation}
\label{eq:gl_memory_gradient}
g_t^{(\nu)}
=
\sum_{k=0}^{\min(t,\,K-1)}
c_k^{(\nu)}\, g_{t-k},
\qquad
c_k^{(\nu)} = (-1)^k \binom{\nu}{k},
\qquad
0<\nu<2,
\end{equation}
and updates the parameters by
\begin{equation}
\label{eq:hereditary_fractional_update}
\theta_{t+1}
=
\theta_t
-
\eta\, h_{\mathrm{step}}^{-\nu}\, g_t^{(\nu)}.
\end{equation}
Here $\eta>0$ is the learning rate, $K$ is the memory length, and
$h_{\mathrm{step}}>0$ is the algorithmic step scale whose power
$h_{\mathrm{step}}^{-\nu}$ is the discrete analogue of the scaling in
\eqref{eq:gl_discrete_def}. The truncation to the most recent $K$ gradients
is the short-memory approximation \eqref{eq:gl_truncated}; storing the full
history is rarely feasible, and short histories have been reported to
suffice in practice \cite{zhou2023gl}. By the coefficient recursion
\eqref{eq:gl_coefficient_recursion}, the case $\nu=1$ gives
$c_0^{(1)}=1$ and $c_1^{(1)}=-1$, so that for $K\ge 2$ the memory sum at
$\nu=1$ equals the gradient difference $g_t-g_{t-1}$ rather than the
gradient itself, and for $\nu\to 1$ only the memory weights beyond the
immediate past vanish. The classical method is therefore recovered through
the one-term window $K=1$, in which the update reduces to plain gradient
descent with learning rate $\eta\, h_{\mathrm{step}}^{-\nu}$; the
memory-based optimizers additionally enforce the exact classical fallback
$g_t^{(\nu)}=g_t$ for $|\nu-1|\le\tau_\nu$
(Section~\ref{subsec:memory_fractional_optimizers}). 

The update \eqref{eq:hereditary_fractional_update} is the natural partner
for the activation \eqref{eq:mw_tanh_activation}. The activation creates
repeated oscillatory perturbations in the local gradient field through
\eqref{eq:mw_tanh_derivative}; the Gr\"unwald--Letnikov update forms a
weighted average over the recent gradient history with a power-law kernel
rather than an exponential one. Small but persistent oscillatory patterns
are therefore not discarded immediately; they remain visible in the update
for longer than under ordinary SGD or short-memory momentum. This has two
opposite consequences, and both are observed in the experiments. On the
positive side, if the fine-scale oscillations carry stable directional
content, the hereditary optimizer can use it instead of washing it out. On
the negative side, if the oscillations produce mainly noise-like sign
changes, too much memory preserves unhelpful fluctuations and slows
stabilization. The memory length $K$ and the order $\nu$ therefore act as
regularity parameters of the optimizer itself: a smaller $\nu$ corresponds
to a heavier memory tail, so older gradients remain influential longer,
while a larger $\nu$ makes the update more local and closer to an ordinary
first-order method.

\,\par\noindent\textbf{The interaction at the level of one neuron. }
The interaction can be read off explicitly for a single neuron. Let
$z_t = w_t^\top x + b_t$ be the pre-activation of the neuron, with
per-neuron weight vector $w_t$, input $x$, and bias $b_t$, and let
$a_t=\phi(z_t)$ be its output. For a loss contribution $\ell_t$, the
gradient with respect to the weight vector decomposes as
\begin{equation}
\label{eq:single_neuron_gradient}
\nabla_{w_t}\ell_t
=
e_t\,\phi'(z_t)\,x,
\end{equation}
where $e_t$ is the backpropagated error term. The factor
$\phi'(z_t)$ is where the multiscale oscillation enters: by
\eqref{eq:mw_tanh_derivative} it is a superposition of terms with
frequencies $b^m$, so the scalar multiplier of $x$ oscillates across scales
as $z_t$ moves during training. The hereditary update
\eqref{eq:hereditary_fractional_update} then combines several past values
of $e_t\,\phi'(z_t)\,x$, not just the current one. In effect, the optimizer
sees a temporally filtered version of the activation-induced oscillation
pattern: the activation shapes the fine structure of the gradient, and the
fractional optimizer decides how long that fine structure continues to
affect the update.

\,\par\noindent\textbf{The non-hereditary surrogate: fractional scaling without
memory. }
There is no single canonical memoryless fractional gradient descent,
because a true Gr\"unwald--Letnikov derivative is inherently
history-dependent. A memoryless variant is therefore best understood as a
\emph{fractional surrogate update}: it preserves a fractional scaling law
but stores no gradient history. A practical and transparent form is the
componentwise preconditioned update
\begin{equation}
\label{eq:memoryless_fractional_surrogate}
\theta_{t+1}
=
\theta_t
-
\eta\,\Lambda_t^{(\nu)} \odot g_t,
\end{equation}
where $\odot$ denotes the componentwise product and $\Lambda_t^{(\nu)}$ is
the positive diagonal scaling
\begin{equation}
\label{eq:memoryless_fractional_scaling}
\Lambda_t^{(\nu)}
=
\operatorname{diag}\!\left(
\frac{1}{\Gamma(2-\nu)}
\bigl(|\theta_t-\theta_{\mathrm{ref}}|+\varepsilon\bigr)^{1-\nu}
\right),
\qquad
0<\nu<2,
\end{equation}
with a reference point $\theta_{\mathrm{ref}}$ and a small stabilizer
$\varepsilon>0$. The scaling law
$|\theta-\theta_{\mathrm{ref}}|^{1-\nu}/\Gamma(2-\nu)$ is the Caputo-type
power law that also underlies the gradient-scaled optimizers of
Section~\ref{sec:fractional_optimizers} 
\cite{herrera2022fractional}. The update
\eqref{eq:memoryless_fractional_surrogate} is \emph{non-hereditary}: it
rescales the current step by a noninteger-order law but integrates nothing
over time. Its advantage is computational simplicity; its cost is close to
that of an ordinary first-order method and no history buffer is required.
Its disadvantage is conceptual: it does not reproduce the nonlocal
character of fractional calculus. For a Weierstrass-based activation this
means that it reacts to the current local roughness of the gradient field,
but it cannot integrate oscillatory behavior over several past steps in the
way the hereditary method \eqref{eq:hereditary_fractional_update} can.

\,\par\noindent\textbf{Both variants in one notation. }
It is convenient to state both optimizers for the same activation and the
same backpropagated gradient in a single display. With the layer map
\begin{equation}
\label{eq:layer_with_fractal_activation}
a^{(\ell)}_t
=
\phi\!\left(\mathbf{W}_t^{(\ell)} a_t^{(\ell-1)} + b_t^{(\ell)}\right),
\end{equation}
where $\mathbf{W}_t^{(\ell)}$ and $b_t^{(\ell)}$ are the weight matrix and
bias of layer $\ell$ and $\phi$ is given by
\eqref{eq:mw_tanh_activation}, the two updates are
\begin{equation}
\label{eq:optimizer_split_notation}
g_t = \nabla_\theta L(\theta_t),
\qquad
u_t =
\begin{cases}
h_{\mathrm{step}}^{-\nu} \displaystyle\sum_{k=0}^{\min(t,\,K-1)}
c_k^{(\nu)} g_{t-k}, & \text{hereditary},\\[1.6em]
\Lambda_t^{(\nu)} \odot g_t, & \text{non-hereditary},
\end{cases}
\qquad
\theta_{t+1} = \theta_t - \eta\, u_t,
\end{equation}
with the combined update vector $u_t$ separating the two design choices.
The hereditary branch assumes that recent gradient history carries useful
information that should decay by a power-law kernel. The non-hereditary
branch assumes that the main benefit of fractionalization is the
noninteger rescaling of the current step. The division of roles is
explicit: the activation determines the nonlinear geometry of the forward
and backward signal, and the optimizer determines how that signal is
accumulated across iterations.

\,\par\noindent\textbf{A remark on magnitudes and normalization. }
The finite-history sum in \eqref{eq:optimizer_split_notation} changes not
only the direction of the update but also its magnitude, because the
coefficients $c_k^{(\nu)}$ do not sum to one and partially cancel. When the
fractional gradient is combined with, rather than substituted for, the
ordinary gradient, this uncontrolled scale is undesirable. The adaptive
framework of Section~\ref{subsec:adaptive_memory_optimizers} 
therefore uses the norm-matched, descent-safeguarded fractional gradient
$\hat g_t^{(\nu)} = \tilde g_t^{(\nu)}\,\|g_t\|/(\|\tilde g_t^{(\nu)}\|+\varepsilon)$,
where the descent-safeguarded fractional gradient $\tilde g_t^{(\nu)}$
equals $g_t^{(\nu)}$ if $\langle g_t^{(\nu)},g_t\rangle>0$ and $g_t$
otherwise (Section~\ref{subsec:memory_fractional_optimizers}), 
which preserves the directional information of the memory term while
keeping its norm comparable to that of the current gradient. This detail
matters precisely in the Weierstrass setting: the oscillatory ladder in
\eqref{eq:mw_tanh_derivative} can make consecutive gradients partially
cancel inside the memory sum, so matching norms separates the question
``in which direction does the history point'' from the question ``how
large should the step be.''

\,\par\noindent\textbf{Implementation pattern. }
Both variants fit one computational pattern. One first computes the
ordinary gradient $g_t$ with the standard reverse-mode machinery
(\texttt{tf.GradientTape} in TensorFlow
\cite{tensorflow_gradienttape_api}). One then either assembles the
history-filtered quantity $g_t^{(\nu)}$ from a buffer of the $K$ most
recent gradients, or assembles the local scaling $\Lambda_t^{(\nu)}$ from
the current parameter state, and finally applies the update
\eqref{eq:optimizer_split_notation}. If the optimizer step itself must
remain part of a differentiable meta-loop, the state update has to be
written in tensor form; for ordinary training the optimizer state can be
treated as external, and custom gradient rules can be supplied where a more
stable backward pass is needed \cite{tensorflow_custom_gradient_api}.

\,\par\noindent\textbf{Summary and what to expect. }
The conclusion of this section is deliberately conditional, not universal.
It is not the case that every fractal activation should be paired with a
fractional optimizer. The precise statement is the following. A moderately
irregular, bounded, truncated Weierstrass-based activation such as the
modified Weierstrass--Tanh function \eqref{eq:mw_tanh_activation} produces
gradients with multiscale structure, controlled by the ladder condition
$ab<1$, without entering an extreme instability regime. In that regime, the
hereditary optimizer \eqref{eq:hereditary_fractional_update} is the natural
choice when medium-range gradient memory should be preserved, because its
power-law kernel is matched to the power-law organization of the
activation's scales; the non-hereditary surrogate
\eqref{eq:memoryless_fractional_surrogate} is the natural choice when a
lightweight method is wanted that still reflects fractional scaling. For
substantially more aggressive activations, in particular ReLU-based
Weierstrass variants with very large gradient magnitudes, long memory can
amplify exactly the oscillations one wants to control, and the memory must
be shortened or additional stabilization (clipping, normalization,
norm matching) must be introduced. The same expectations transfer to the
surface experiments: on objectives carrying the fractal perturbation ladder
$P_J$, memory-based methods should show their advantage where the
perturbation adds structured, persistent oscillation to the gradient field,
and should lose it where the perturbation acts as effectively uncorrelated
noise. Both effects are visible in the results of
Section~\ref{sec:surface_experiments}. 
In one sentence: the activation and the surface perturbations place
multiscale roughness into representation space and loss space, and the
fractional optimizer decides, through its order $\nu$ and memory window
$K$, whether and for how long that roughness is remembered in update space.

\begin{takeawaybox}[purpleA]{Main Takeaways --- Fractal Functions and Derivatives}
  \begin{itemize}[
    label={},
    leftmargin=0em,
    itemindent=0em,
    itemsep=3pt,
    topsep=2pt
  ]
    \item \textbf{The roughness exponent sets the fractional scale.}
          A Weierstrass-type function with exponent \(\gamma_H\) admits
          fractional derivatives for \(\nu<\gamma_H\); in this regime, the
          result remains in the same multi-scale family with reduced
          exponent \(\gamma_H-\nu\).

    \item \textbf{The Gr\"unwald--Letnikov form makes this computable.}
          Its algebraically decaying coefficients define a weighted history
          operator; \(\nu=1\) recovers the ordinary difference, while the
          memory window \(K\) controls cost and history length.

    \item \textbf{Fractal activations and fractional optimizers meet in the
          gradient history.} The activation shapes the multi-scale gradient
          structure, while the optimizer decides how present and past
          gradients are combined; the benefit of memory is therefore
          conditional.
  \end{itemize}
\end{takeawaybox}

\section{Fractal Activation Functions}
\label{sec:fractal_activations}

This section describes the fractal activation functions used in the
neural-network experiments of this article. The construction and the full
catalogue of nine fractal activations were introduced in the predecessor
study of Raubitzek et al.\ \cite{raubitzek_fractals_2026}; here we recall
the general building principle in the notation of the present article and
then define the four activations that are actually used. In contrast to
the predecessor study, which evaluated the complete catalogue, the present
work deliberately restricts attention to these four, because they form a
representative subset containing the most reliable performer, strong
dataset-specific variants, and activations with different levels of
multiscale irregularity. Restricting the activation set keeps the
optimizer--activation grid of the experiments in
Section~\ref{sec:nn_experiments} 
tractable while retaining a representative subset of the class.

\subsection{The General Construction}
\label{subsec:fractal_activation_construction}

This subsection presents the general construction used to derive fractal activation functions from multi-scale series. It defines the roles of amplitude decay, frequency growth, and the resulting roughness exponent before describing their practical implementation in neural networks.

\,\par\noindent\textbf{From fractal series to activations. }
Conventional activation functions bend the signal once: ReLU introduces a
single kink, and $\tanh$ or the sigmoid a single saturating bend. A
fractal activation instead superimposes self-similar oscillations of
shrinking amplitude and growing frequency. The generic construction is the
geometric ladder
\begin{equation}
\label{eq:generic_fractal_activation}
\mathcal{F}(x)
=
\sum_{m=0}^{\infty} a^{m}\,h\bigl(b^{m}x\bigr),
\qquad
0<a<1,\quad b>1,
\end{equation}
where $h(\cdot)$ is a bounded periodic generator, $a$ is the amplitude
decay rate, and $b$ is the geometric frequency base. Term $m$ oscillates
$b$ times faster and is $1/a$ times smaller than term $m-1$. This is the
same construction as the Weierstrass-type series of
Section~\ref{sec:weierstrass_fractional}, 
written through the pair $(a,b)$ instead of the pair $(b,\gamma_H)$: the
correspondence is $a=b^{-\gamma_H}$, so the formal roughness exponent of
the ladder is
\begin{equation}
\label{eq:roughness_from_ab}
\gamma_H = -\frac{\ln a}{\ln b}.
\end{equation}
The geometric weights guarantee uniform convergence of
\eqref{eq:generic_fractal_activation} for continuous $h$, so $\mathcal{F}$
is always continuous. Weierstrass introduced the original continuous
nowhere-differentiable construction \cite{weierstrass1872}. Its
differentiability is governed by the product $ab$, which controls the
differentiated ladder: term-wise differentiation multiplies the amplitude
$a^m$ by the frequency $b^m$. If $h$ is continuously differentiable with
bounded derivative and $ab<1$ (equivalently $\gamma_H>1$), the
differentiated series converges uniformly and the infinite sum is
continuously differentiable. For the classical sine and cosine
Weierstrass series, Hardy established nowhere differentiability in the
critical and supercritical regime $ab\geq 1$ (equivalently
$\gamma_H\leq 1$) \cite{hardy1916weierstrass}. The Takagi (Blancmange)
function is the corresponding construction from a sawtooth generator,
sitting exactly on the boundary $ab=1$
\cite{takagi1903continuous,allaart2011takagi}.

\,\par\noindent\textbf{Truncation. }
In practice the series is truncated at a fixed number of terms $N$. This
step is essential, not cosmetic. The infinite classical parent ladders in
the critical and supercritical regimes are nowhere differentiable, and
their classical derivatives are undefined, which would make direct
backpropagation through the ideal infinite objects inoperable. The
truncated sums, by contrast, are finite tensor expressions: automatic
differentiation applies the standard chain rule wherever they are
differentiable and the framework's derivative conventions at explicit
nonsmooth points, as discussed in
Section~\ref{subsec:discrete_fractional_weierstrass_autodiff}. 
The truncation retains the self-similar oscillations on the $N$ coarsest
scales, which are the scales that are numerically representable anyway;
harmonics beyond $N\approx 30$--$100$ are below or near machine precision
for the parameter choices used here. The network therefore never sees the
ideal mathematical fractal, but a scale-limited version of it, in exact
analogy to the grid-filtering principle of the discrete fractional
operators in Section~\ref{sec:weierstrass_fractional}. 

\,\par\noindent\textbf{Why such activations, and why these four. }
The rationale, established in \cite{raubitzek_fractals_2026}, is that
structured high-frequency detail injected at the activation level
increases the functional richness of a network without adding parameters
or architectural complexity, while envelopes and backbone terms can
moderate outputs and gradients in the corresponding constructions. The
predecessor study constructed nine such activations from Blancmange- and
Weierstrass-type series, evaluated them on ten public classification
datasets, and analysed their expressivity and gradient stability. The four
selected here form a representative subset containing the most reliable
performer, strong dataset-specific variants, and activations with different
levels of multiscale irregularity: the modulated Blancmange curve, the
decaying cosine function, the modified Weierstrass--Tanh function, and the
Weierstrass--Mandelbrot $x{+}\sin$ variant. These four are used in this
article. Conveniently for the purposes of this study, their underlying
oscillatory ladders also span the formal regularity spectrum of the
construction: as shown below, one is subcritical ($ab<1$), two sit on the
critical boundary ($ab=1$), and one is supercritical ($ab>1$) with a
genuinely nowhere-differentiable parent of roughness exponent
$\gamma_H=1/2$, belonging to the class of Weierstrass--Mandelbrot
functions analysed by Z\"ahle and Ziezold \cite{zahle1996fractional} and
recalled in Section~\ref{sec:weierstrass_fractional}. 

\subsection{The Four Activation Functions Used in This Study}
\label{subsec:four_fractal_activations}

Each activation below is stated as the finite truncated sum that is
actually implemented, with its parameters and its structural ingredients
named explicitly. The naming convention (\texttt{typewriter} identifiers)
follows the code and the result tables. Throughout, $m$ denotes the series
index, $N$ the truncation length, $a$ the amplitude decay rate, $b$ the
frequency base, and $\mu_e$ the envelope decay constant.

\subsubsection{Modulated Blancmange Curve
(\texttt{modulated\_blancmange\_curve})}
\label{subsubsec:fab_blancmange}

\begin{equation}
\label{eq:fab_blancmange}
\phi_{\mathrm{B}}(x)
=
\sum_{m=0}^{N-1}
\tanh\!\bigl(a_{\mathrm{mod}}\,2^{m}x\bigr)\,
\frac{\bigl|\,(2^{m}x \bmod 2)-a_{\mathrm{mod}}\sqrt{|x|}\,\bigr|}{2^{m}},
\qquad
a_{\mathrm{mod}}=0.75,\quad N=30.
\end{equation}
The construction has three named ingredients. The \emph{sawtooth profile}
$|(2^m x\bmod 2)-a_{\mathrm{mod}}\sqrt{|x|}|$ is a piecewise-linear
periodic-type term in the spirit of the Takagi generator
$\operatorname{dist}(x,\mathbb{Z})$
\cite{takagi1903continuous,allaart2011takagi}, with an input-dependent
offset $a_{\mathrm{mod}}\sqrt{|x|}$ that breaks exact periodicity and
shifts the corner locations with $|x|$. The \emph{gate}
$\tanh(a_{\mathrm{mod}}2^m x)$ makes each term odd around the origin and
suppresses the excessively flat region near zero that the plain Takagi sum
would produce. The \emph{ladder} uses amplitude decay $2^{-m}$ against
frequency growth $2^{m}$, that is, $a=1/2$ and $b=2$.

The underlying Takagi ladder with $a=1/2$ and $b=2$ lies exactly on the
critical boundary $ab=1$, equivalently $\gamma_H=1$ by
\eqref{eq:roughness_from_ab}, and the standard infinite Takagi function is
continuous and nowhere differentiable
\cite{takagi1903continuous,allaart2011takagi}. Because
Equation~\eqref{eq:fab_blancmange} additionally contains a scale-dependent
$\tanh$ gate and an input-dependent offset, however, the regularity of its
infinite parent does not follow directly from the standard Takagi results
and would require a separate proof. The practical consequence of the
finite implementation is a large multiscale set of corner points: the
local slope changes at locations with spacings of order $2^{-m}$ for every
retained scale $m$, which provides sharp, frequently changing gradient
signals during training while the $2^{-m}$ amplitude decay keeps their
total contribution bounded.

\subsubsection{Decaying Cosine Function
(\texttt{decaying\_cosine\_function})}
\label{subsubsec:fab_decaying_cosine}

\begin{equation}
\label{eq:fab_decaying_cosine}
\phi_{\mathrm{D}}(x)
=
\sum_{m=0}^{N-1}
\zeta\Bigl(
0.05\,\tanh(\pi x)
+a^{m}\cos\!\bigl(b^{m}\pi x\bigr)\,
e^{-|x|/2}\,\operatorname{sign}(x)
\Bigr),
\end{equation}
\begin{equation}
a=0.5,\; b=2,\; \zeta=0.2666,\; N=75.
\end{equation}
Here the envelope decay constant is $\mu_e=1/2$, and the global scale
$\zeta$ multiplies every term. Two structural observations explain the
constants. First, because the smooth term $0.05\tanh(\pi x)$ sits inside
the sum, it is added $N$ times; the aggregated \emph{backbone} is
therefore $\zeta N\cdot 0.05\,\tanh(\pi x)\approx 1.00\cdot\tanh(\pi x)$
for $\zeta=0.2666$ and $N=75$, so the constant $\zeta$ is precisely the
normalization that makes the activation a unit-scale $\tanh(\pi x)$ plus
an oscillatory correction. Second, the correction ladder has total
amplitude $\zeta\sum_m a^m\approx 2\zeta\approx 0.53$, is localized by the
envelope $e^{-|x|/2}$, and is made odd-signed by the factor
$\operatorname{sign}(x)$, so the oscillations respect the overall
antisymmetric character of the backbone.

The underlying unwindowed cosine ladder satisfies $ab=1$ and falls
under Hardy's critical nowhere-differentiable case
\cite{hardy1916weierstrass}. This classification applies to the cosine
ladder, not to the full sign-modulated activation, which is discontinuous
at $x=0$; every finite truncation is smooth away from that discontinuity.
Of the four activations, this one has the strongest smooth backbone
relative to its oscillatory part, which made it one of the most stable
fractal activations in the gradient analysis of
\cite{raubitzek_fractals_2026}.

\subsubsection{Modified Weierstrass--Tanh
(\texttt{modified\_weierstrass\_tanh})}
\label{subsubsec:fab_weierstrass_tanh}

\begin{equation}
\label{eq:fab_weierstrass_tanh}
\phi_{\mathrm{W}}(x)
=
\tanh(x)
+
\sum_{m=0}^{N-1}
(-1)^{m}a^{m}\cos\!\bigl(b^{m}\pi x\bigr)\,e^{-\mu_e|x|},
\qquad
a=0.5,\; b=1.5,\; \mu_e=0.75,\; N=100.
\end{equation}
This is the activation analysed in detail in
Section~\ref{subsec:fractional_gd_fractal_activation}, 
and its ingredients are repeated here only briefly: the \emph{backbone}
$\tanh(x)$ provides a stable low-frequency nonlinearity, the
\emph{oscillatory ladder} adds scales with amplitudes $a^m$ and
frequencies $b^m\pi$, the alternating sign $(-1)^m$ (equivalently, a
ladder with negative ratio $-a$) makes adjacent scales partially cancel,
and the \emph{envelope} $e^{-\mu_e|x|}$ localizes the oscillations around
the origin.

The regularity classification is the decisive point. The unwindowed
oscillatory ladder satisfies $ab=0.75<1$, equivalently
$\gamma_H=\ln 2/\ln 1.5\approx 1.71>1$ by
\eqref{eq:roughness_from_ab}. This ladder is therefore
\emph{subcritical}: differentiating term by term multiplies each amplitude
by $(ab)^m\to 0$, and the differentiated series converges uniformly to a
bounded, though rapidly varying, derivative. The complete implemented
activation additionally contains the envelope $e^{-\mu_e|x|}$, which is
not differentiable at $x=0$; because the oscillatory ladder is generally
nonzero there, the full activation is generally not differentiable at the
origin, although it is smooth elsewhere. The subcritical classification
therefore refers to the multiscale ladder rather than to the global
regularity of the enveloped activation. The modified Weierstrass--Tanh
activation is thus fractal-inspired rather than fractal in the strict
analytic sense: it carries a full multiscale oscillation ladder, but the
ladder stays strictly below the non-differentiability threshold. This is
exactly the ``balanced regime'' identified in
\cite{raubitzek_fractals_2026}, where it was among the most reliable
fractal activations across datasets while remaining far from the extreme
gradient magnitudes of the ReLU-based Weierstrass variant, and it is the
reason this activation serves as the reference case for the
optimizer-level analysis of
Section~\ref{subsec:fractional_gd_fractal_activation}. 

\subsubsection{Weierstrass--Mandelbrot
(\texttt{weierstrass\_mandelbrot\_xpsin})}
\label{subsubsec:fab_wm_xpsin}

\begin{equation}
\label{eq:fab_wm_xpsin}
\phi_{\mathrm{M}}(x)
=
\sum_{k=1}^{N}
2^{-k\gamma_H}\Bigl(x+\sin\!\bigl(2\pi\,b^{k}x\bigr)\Bigr),
\qquad
\gamma_H=0.5,\; b=2,\; N=100.
\end{equation}
This activation is written directly in the $(b,\gamma_H)$ parametrization
of Section~\ref{sec:weierstrass_fractional} 
because it \emph{is} a classical Weierstrass-type object. The
Weierstrass--Mandelbrot terminology follows the broader fractal-geometric
treatment of Mandelbrot \cite{mandelbrot1982fractal}. Splitting the sum
gives
\begin{equation}
\label{eq:fab_wm_xpsin_split}
\phi_{\mathrm{M}}(x)
=
c_{\mathrm{lin}}\,x + W_h^{\gamma_H}(x),
\qquad
c_{\mathrm{lin}}=\sum_{k=1}^{N}2^{-k\gamma_H}
\;\xrightarrow{\,N\to\infty\,}\;
\frac{1}{2^{\gamma_H}-1}=\frac{1}{\sqrt{2}-1}\approx 2.414,
\end{equation}
that is, a \emph{linear drift} with aggregated slope $c_{\mathrm{lin}}$
plus the Weierstrass-type sum $W_h^{\gamma_H}$ with generator
$h(u)=\sin(2\pi u)$, base $b=2$, and roughness exponent $\gamma_H=0.5$,
the classical sine-generated case studied by Hardy, Berry and Lewis, and
Hu and Lau \cite{hardy1916weierstrass,berry1980weierstrass,
hu1993weierstrass}. In ladder terms, $a=2^{-1/2}$ and
$ab=\sqrt{2}>1$: the activation is \emph{supercritical}. Its parent
oscillatory part is continuous and nowhere differentiable with H\"older
exponent exactly $1/2$. Z\"ahle and Ziezold derived
fractional-derivative formulas for Weierstrass and
Weierstrass--Mandelbrot functions and developed a numerical procedure for
computing derivatives in the mean \cite{zahle1996fractional}.

Structurally, the linear drift plays the role that $\tanh$ or ReLU
backbones play in the other activations: it guarantees a nonvanishing
average slope $c_{\mathrm{lin}}$, so that gradient signal propagates even
where the oscillatory part locally cancels, while the unbounded linear
growth is kept in check in practice by the bounded pre-activation ranges
of standardized inputs. Among the four activations used here, this is the
roughest one: it carries genuine critical-order fractality into the
network, and it is therefore the case in which the multiscale gradient
structure discussed in
Sections~\ref{sec:weierstrass_fractional} 
and~\ref{sec:fractional_optimizers} 
is most pronounced.

\subsection{Summary and Role in the Experiments}
\label{subsec:fractal_activation_summary}

\begin{figure}[H]
    \centering
    \includegraphics[width=\textwidth]{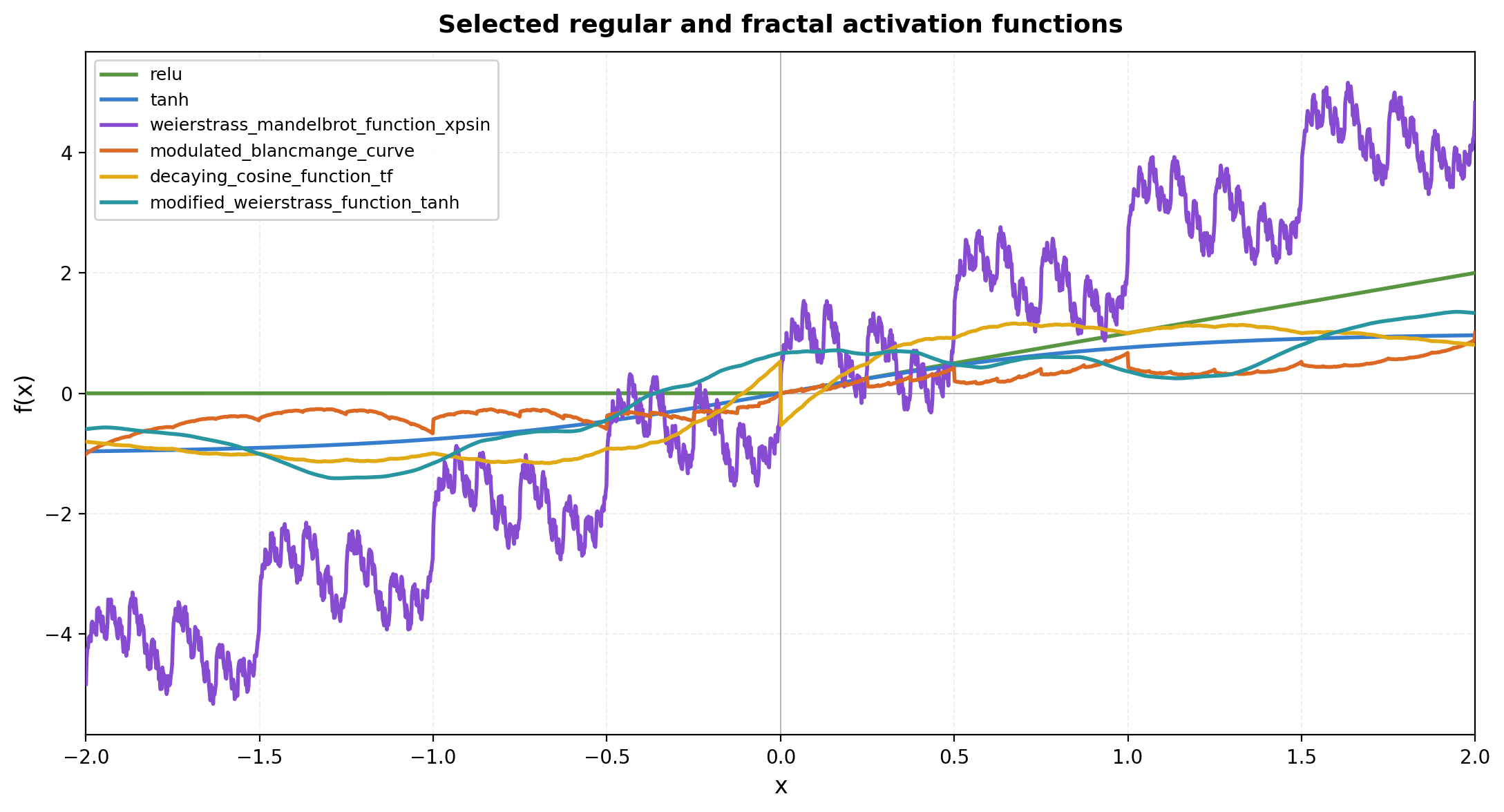}
    \caption{Selected regular and fractal activation functions used in the neural-network experiments. The plot shows the two regular baselines, \texttt{relu} and \texttt{tanh}, together with the four fractal activations considered in this study: \texttt{weierstrass\_mandelbrot\_function\_xpsin}, \texttt{modulated\_blancmange\_curve}, \texttt{decaying\_cosine\_function\_tf}, and \texttt{modified\_weierstrass\_function\_tanh}.}
    \label{fig:activations}
\end{figure}

Table~\ref{tab:fractal_activation_summary} collects the four activations
with their structural parameters and regularity regimes. Further, all activations used in this article are depicted in Figure \ref{fig:activations}.

\begin{table}[H]
\caption{The four fractal activation functions used in this study.
``$ab$'' is the amplitude--frequency product of the oscillatory ladder,
$\gamma_H=-\ln a/\ln b$ the formal roughness exponent of the parent
series, and ``Regime'' the formal regularity class of the underlying
oscillatory ladder: subcritical ($ab<1$, differentiable ladder under the
stated smoothness assumptions), critical ($ab=1$, classical
Takagi/Hardy boundary), and supercritical ($ab>1$, classical
nowhere-differentiable ladder with H\"older exponent
$\gamma_H<1$). These classifications do not by themselves determine the
global regularity of the complete activations after gates, envelopes,
offsets, or sign factors are introduced.
\label{tab:fractal_activation_summary}}
\small
\begin{tabularx}{\textwidth}{lXccccc}
\toprule
\textbf{Activation} & \textbf{Backbone / envelope} & $a$ & $b$ &
$ab$ & $\gamma_H$ & \textbf{Regime} \\
\midrule
\texttt{modulated\_blancmange\_curve} &
$\tanh$ gate; no envelope & $1/2$ & $2$ & $1$ & $1$ & critical \\
\texttt{decaying\_cosine\_function} &
$\approx\tanh(\pi x)$ backbone; $e^{-|x|/2}$ & $1/2$ & $2$ & $1$ & $1$ &
critical \\
\texttt{modified\_weierstrass\_tanh} &
$\tanh(x)$ backbone; $e^{-0.75|x|}$ & $1/2$ & $1.5$ & $0.75$ &
$\approx1.71$ & subcritical \\
\texttt{weierstrass\_mandelbrot\_xpsin} &
linear drift $c_{\mathrm{lin}}x$; no envelope & $2^{-1/2}$ & $2$ &
$\sqrt{2}$ & $0.5$ & supercritical \\
\bottomrule
\end{tabularx}
\end{table}

In plain terms, all four activations follow one recipe: a stable backbone
(a $\tanh$-type bend or a linear drift) plus a geometric ladder of
oscillations, truncated at $N$ terms so that automatic differentiation
can be applied to the finite implementation. They differ in where their
underlying oscillatory ladders sit relative to the differentiability
threshold $ab=1$, and this is deliberate: the selection covers a
subcritical smooth-but-oscillatory ladder, two borderline ladders, and one
genuinely rough ladder with the critical H\"older exponent
$\gamma_H=1/2$. For the purposes of this article, the
four activations therefore serve a double role. As in the predecessor
study \cite{raubitzek_fractals_2026}, they are competitive activation
functions in their own right; in addition, they act as controlled sources
of multiscale gradient structure of graded intensity, against which the
fractional optimizers of
Section~\ref{sec:fractional_optimizers} 
are evaluated in Section~\ref{sec:nn_experiments}. 

\begin{takeawaybox}[orangeA]{Main Takeaways --- Fractal Activation Functions}
  \begin{itemize}[
    label={},
    leftmargin=0em,
    itemindent=0em,
    itemsep=3pt,
    topsep=2pt
  ]
    \item \textbf{Fractal activations add multi-scale structure without
          adding parameters.} A stable backbone is combined with oscillations
          of decreasing amplitude and increasing frequency.

    \item \textbf{Truncation makes the functions usable in neural networks.}
          A finite number of retained scales gives a finite tensor
          expression that standard automatic differentiation can process,
          including framework conventions at explicit nonsmooth points.

    \item \textbf{The selected activations provide controlled gradient
          roughness.} Their underlying oscillatory ladders cover subcritical,
          critical, and supercritical regimes and serve as graded sources of
          multi-scale gradient structure for evaluating the fractional
          optimizers.
  \end{itemize}
\end{takeawaybox}

\section{Fractional Optimizers}
\label{sec:fractional_optimizers}

This section introduces the optimization methods compared in this article.
A fractional optimizer is a gradient-based optimization method in which the
ordinary first-order derivative is replaced, rescaled, or extended by an
operator of non-integer order $\nu$. The motivation follows directly from
Section~\ref{sec:weierstrass_fractional}: fractional derivatives are
nonlocal operators. Whereas an ordinary derivative depends only on the
local behaviour of a function, a fractional derivative depends on function
values over an interval \cite{oldham1974fractional,podlubny1999fractional}.
Transferred to optimization, this nonlocality can introduce history
dependence into an update rule, while the order $\nu$ can affect convergence
speed, robustness, and sensitivity to noisy or oscillatory gradients
\cite{elnady2025survey,fernandez2025role}. Early work established fractional
gradient descent for neural network training and analysed its convergence
\cite{wang2017fractional,bao2018fractional}; more recent work has produced
fractional variants of the standard deep-learning optimizers
\cite{herrera2022fractional,zhou2023gl,shin2023accelerating,
han2023adaptive,huang2024mffgd,naifar2026tempered}.

The comparison in this article contains \emph{21 optimizers in five
groups}. The first group contains the four classical baselines SGD,
RMSprop, Adam, and Adadelta
(Section~\ref{subsec:classical_baselines}). The second group contains the
four Herrera-type fractional optimizers FSGD, FRMSprop, FAdam, and
FAdadelta, which rescale the current gradient by a Caputo-type factor and
store no history (Section~\ref{subsec:herrera_fractional_optimizers}). The
third group contains the four memory-based fractional optimizers
MemoryFSGD, MemoryFRMSprop, MemoryFAdam, and MemoryFAdadelta, which build
their update direction from the finite-history Gr\"unwald--Letnikov
gradient $g_t^{(\nu)}$ of Section~\ref{sec:weierstrass_fractional},
stabilized by a descent safeguard and norm matching
(Section~\ref{subsec:memory_fractional_optimizers}). The fourth group
contains the four adaptive memory-based optimizers AdaptiveMemoryFSGD,
AdaptiveMemoryFRMSprop, AdaptiveMemoryFAdam, and AdaptiveMemoryFAdadelta,
which keep the order $\nu$ fixed and adapt only a mixing coefficient
$\lambda_t$ that controls how much the memory term is trusted
(Section~\ref{subsec:adaptive_memory_optimizers}). The fifth group contains five related-work optimizers, including two from a
recent preprint: AdaGL \cite{chen2024adagl}, FCSGD\_GL and FCAdam\_GL
\cite{zhou2023gl}, and AOFGD\_SGD and AOFGD\_Adam
\cite{xiang2025aofgd} (Section~\ref{subsec:related_work_optimizers}).
Section~\ref{subsec:optimizer_summary} summarizes all 21 methods in one
table.

The five groups differ in which quantities are swept and which are held
fixed, and this distinction matters for reading the result tables. Only
the Herrera-type and the memory-based groups sweep the fractional order;
they are evaluated at every $\nu\in\{0.75,1.25,1.50\}$. The
adaptive memory-based group and the five related-work methods are
evaluated at a single, fixed internal configuration, because their
defining mechanism is an adaptive control quantity rather than the order
itself. Where the result tables report $\nu=1.00$ for a method of these
two groups, that entry is a display placeholder for a swept order that
does not exist for the method; the internally used orders are stated in
the corresponding subsections below and are \emph{not} equal to one.

Throughout, $\theta_t\in\mathbb{R}^d$ denotes the parameter vector at
iteration $t$, $L(\theta)$ the loss function,
$g_t=\nabla_\theta L(\theta_t)$ the ordinary gradient computed by
backpropagation, $\eta>0$ the learning rate, and $\varepsilon>0$ a small
numerical stabilizer. Squares, absolute values, and powers of vectors are
taken componentwise; $\langle\cdot,\cdot\rangle$ denotes the Euclidean
inner product and $\|\cdot\|$ the Euclidean norm, both taken per parameter
tensor.

\subsection{Classical First-Order Baselines: SGD, RMSprop, Adam, Adadelta}
\label{subsec:classical_baselines}

All optimizers in this article follow the generic template
\begin{equation}
\label{eq:generic_update_template}
\theta_{t+1} = \theta_t - \eta\, u_t,
\end{equation}
where $u_t$ is an update direction assembled from the gradient $g_t$ and
the internal state of the method. The four baselines differ in how they
assemble $u_t$; they are stated here explicitly because every fractional
optimizer below is obtained from one of them by a controlled modification.

\,\par\noindent\textbf{SGD with momentum. }
Stochastic gradient descent \cite{robbins1951stochastic} takes the raw
gradient as the update, $\theta_{t+1}=\theta_t-\eta\,g_t$. With momentum
\cite{polyak1964methods}, a velocity term $v_t^{(\mathrm{mom})}$ with decay
rate $\beta_m\in[0,1)$ accumulates an exponentially weighted average of
past steps,
\begin{equation}
\label{eq:sgd_momentum}
v_t^{(\mathrm{mom})} = \beta_m\, v_{t-1}^{(\mathrm{mom})} - \eta\, g_t,
\qquad
\theta_{t+1} = \theta_t + v_t^{(\mathrm{mom})}.
\end{equation}

\,\par\noindent\textbf{RMSprop. }
RMSprop \cite{tieleman2012rmsprop} maintains a squared-gradient
accumulator $v_t^{(2)}$ with decay rate $\gamma_v\in[0,1)$ and divides the
gradient by its root,
\begin{equation}
\label{eq:rmsprop}
v_t^{(2)} = \gamma_v\, v_{t-1}^{(2)} + (1-\gamma_v)\, g_t^2,
\qquad
\theta_{t+1} = \theta_t - \eta\,
\frac{g_t}{\sqrt{v_t^{(2)}+\varepsilon}}.
\end{equation}
This normalizes the step size per coordinate by the recent gradient
magnitude.

\,\par\noindent\textbf{Adam. }
Adam \cite{kingma2015adam} combines a first-moment estimate $m_t$ (decay
rate $\beta_1$) with a squared-gradient accumulator $v_t^{(2)}$ (decay rate
$\beta_2$) and applies bias correction:
\begin{equation}
\label{eq:adam}
\begin{aligned}
m_t &= \beta_1\, m_{t-1} + (1-\beta_1)\, g_t,
&\qquad
v_t^{(2)} &= \beta_2\, v_{t-1}^{(2)} + (1-\beta_2)\, g_t^2,\\[0.2em]
\hat m_t &= \frac{m_t}{1-\beta_1^t},
&\qquad
\hat v_t^{(2)} &= \frac{v_t^{(2)}}{1-\beta_2^t},
\end{aligned}
\qquad
\theta_{t+1} = \theta_t - \eta\,
\frac{\hat m_t}{\sqrt{\hat v_t^{(2)}}+\varepsilon}.
\end{equation}

\,\par\noindent\textbf{Adadelta. }
Adadelta \cite{zeiler2012adadelta} maintains two accumulators with a shared
decay rate $\gamma_v$: the squared-gradient accumulator $v_t^{(2)}$ and an
accumulator $v_t^{(\Delta)}$ of squared past updates,
\begin{equation}
\label{eq:adadelta}
\begin{aligned}
v_t^{(2)} &= \gamma_v\, v_{t-1}^{(2)} + (1-\gamma_v)\, g_t^2,\\
\Delta\theta_t &= -\,
\frac{\sqrt{v_{t-1}^{(\Delta)}+\varepsilon}}
     {\sqrt{v_t^{(2)}+\varepsilon}}\; g_t,\\
v_t^{(\Delta)} &= \gamma_v\, v_{t-1}^{(\Delta)}
                 + (1-\gamma_v)\, (\Delta\theta_t)^2,
\end{aligned}
\qquad
\theta_{t+1} = \theta_t + \eta\,\Delta\theta_t.
\end{equation}
The implementation used in this article applies the additional global
learning-rate multiplier $\eta$ to the original Adadelta increment; the
original formulation uses $\theta_{t+1}=\theta_t+\Delta\theta_t$ without
this extra multiplier.

\,\par\noindent\textbf{The common feature. }
All four baselines are first-order and short-memory methods
\cite{ruder2016overview}. Their state consists of at most two
exponentially weighted moments; the influence of a gradient from $k$ steps
ago decays geometrically, like $\beta^k$. In the terminology of
Section~\ref{sec:weierstrass_fractional}, they use exponential forgetting.
Fractional optimizers change exactly this point: either they rescale the
current gradient by a law of non-integer order, or they replace exponential
forgetting by the algebraically decaying Gr\"unwald--Letnikov kernel
$c_k^{(\nu)}$.

\subsection{From Integer-Order to Fractional-Order Updates}
\label{subsec:from_integer_to_fractional}

The starting point of every method below is the observation that the
gradient $g_t$ in \eqref{eq:generic_update_template} plays the role of a
first-order derivative of the loss with respect to the parameters. A
fractional optimizer replaces this first-order object by an operator of
order $\nu\neq 1$. Two structurally different routes exist, and both were
prepared in Section~\ref{sec:weierstrass_fractional}.

The first route starts from the \emph{Caputo} derivative. For a smooth
function, the Caputo derivative of order $\nu$ admits closed-form
expressions on power functions, and the corresponding power law can be
used as a Caputo-inspired factor that rescales the current gradient. The
result is a \emph{local} modification: no history is stored, and only the
magnitude of the current gradient changes. This route leads to the
Herrera-type optimizers of
Section~\ref{subsec:herrera_fractional_optimizers}
\cite{wang2017fractional,herrera2022fractional,herrera2023pytorch}.

The second route starts from the \emph{Gr\"unwald--Letnikov} derivative in
its discrete form (Section~\ref{subsec:discrete_fractional_weierstrass_autodiff}).
Applied along the iteration axis, it produces the finite-history fractional
gradient
\begin{equation}
\label{eq:gl_gradient_recall}
g_t^{(\nu)} = \sum_{k=0}^{K-1} c_k^{(\nu)}\, g_{t-k},
\qquad
c_0^{(\nu)}=1,
\qquad
c_k^{(\nu)}=\Bigl(1-\frac{\nu+1}{k}\Bigr)c_{k-1}^{(\nu)}
\quad (k\ge1),
\end{equation}
a weighted sum of the $K$ most recent gradients with the algebraically
decaying kernel $c_k^{(\nu)}$. The result is a \emph{nonlocal}
modification: the update depends on a genuine gradient history. This route
leads to the memory-based optimizers of
Section~\ref{subsec:memory_fractional_optimizers}
\cite{zhou2023gl,chen2024adagl}.

The distinction between the two routes organizes the entire section. Local
rescaling changes how large the current step is; explicit memory changes
which information the step is computed from. The adaptive framework of
Section~\ref{subsec:adaptive_memory_optimizers} then adds a third element:
it does not modify the fractional operator further, but adapts how much the
optimizer trusts the memory-based gradient relative to the ordinary one.
For both fractional gradients we use the conventions of the notation
appendix: $f_\nu(g_t)$ always denotes the local Caputo-type scaling
factor, and $g_t^{(\nu)}$ always denotes the history-weighted
Gr\"unwald--Letnikov gradient. The two are conceptually different objects
and are never interchanged, even though both are informally called
``the fractional gradient'' in parts of the literature.

\subsection{Herrera-Type Fractional Gradient-Scaled Optimizers}
\label{subsec:herrera_fractional_optimizers}

This subsection introduces Herrera-type fractional optimizers, which modify classical update rules by rescaling the current gradient with a Caputo-inspired factor. These methods use only the gradient at the current iteration and do not introduce an explicit gradient-history term.

\,\par\noindent\textbf{Derivation of the scaling factor. }
The construction of Herrera-Alc\'antara and collaborators
\cite{herrera2022fractional,herrera2023pytorch} is motivated by the Caputo
derivative applied to the identity map $\theta\mapsto\theta$
(componentwise). For $0<\nu<1$, the classical Caputo derivative satisfies
the closed-form identity
\begin{equation}
\label{eq:caputo_identity_map}
D^\nu_\theta\,\theta
=
\frac{\theta^{\,1-\nu}}{\Gamma(2-\nu)}.
\end{equation}
Treating the loss as a function of the parameters and applying a
chain-rule-type approximation motivates the expression
\begin{equation}
\label{eq:herrera_approximation}
D^\nu_\theta L(\theta)
\;\approx\;
\frac{\partial L}{\partial \theta}\cdot
\frac{\theta^{\,1-\nu}}{\Gamma(2-\nu)}.
\end{equation}
Two remarks are required for correctness. First,
\eqref{eq:herrera_approximation} is a deliberate simplification, not an
exact identity; the exact fractional chain rule is substantially more
complicated. Second, the cited optimizer implementations use the
corresponding power law as a Caputo-inspired gradient-scaling heuristic,
evaluate it on the magnitude of the current gradient, and extend it to
orders $0<\nu<2$, $\nu\neq1$
\cite{herrera2022fractional,herrera2023pytorch}. The stabilized,
componentwise \emph{Caputo-type scaling factor} is therefore
\begin{equation}
\label{eq:herrera_factor}
f_\nu(g_t)
=
\frac{\bigl(|g_t|+\varepsilon\bigr)^{1-\nu}}{\Gamma(2-\nu)},
\qquad \varepsilon>0,
\end{equation}
where the constant $\Gamma(2-\nu)$ is fixed for the whole run and
precomputed. The \emph{scaled gradient} is then defined as
\begin{equation}
\label{eq:herrera_scaled_gradient}
g_t^{\mathrm{sc}}
=
f_\nu(g_t)\odot g_t,
\end{equation}
where $\odot$ is the componentwise product. Note that
$g_t^{\mathrm{sc}}$ is a rescaling of the single current gradient; it is
conceptually distinct from the history sum $g_t^{(\nu)}$ of
\eqref{eq:gl_gradient_recall}.

\,\par\noindent\textbf{Resulting optimizers. }
A Herrera-type fractional optimizer is obtained by passing
$g_t^{\mathrm{sc}}$, instead of $g_t$, into the unchanged update rule of a
base optimizer. Writing the base update abstractly as
$\theta_{t+1}=\mathcal{U}\bigl(\theta_t,\,g_t,\,\Xi_t\bigr)$, where $\Xi_t$
denotes the internal state of the base optimizer (velocity, moments,
accumulators), the fractional variant is
\begin{equation}
\label{eq:herrera_generic}
\theta_{t+1}
=
\mathcal{U}\bigl(\theta_t,\;g_t^{\mathrm{sc}},\;\Xi_t\bigr).
\end{equation}
Applying \eqref{eq:herrera_generic} to the four baselines of
Section~\ref{subsec:classical_baselines} yields the four optimizers of
this group. For FSGD with momentum, for example,
\begin{equation}
\label{eq:fsgd}
v_t^{(\mathrm{mom})}
= \beta_m\, v_{t-1}^{(\mathrm{mom})} - \eta\, g_t^{\mathrm{sc}},
\qquad
\theta_{t+1} = \theta_t + v_t^{(\mathrm{mom})},
\end{equation}
and for FAdam both moment estimates are computed from the scaled gradient,
\begin{equation}
\label{eq:fadam}
m_t = \beta_1 m_{t-1} + (1-\beta_1)\, g_t^{\mathrm{sc}},
\qquad
v_t^{(2)} = \beta_2 v_{t-1}^{(2)} + (1-\beta_2)\,
\bigl(g_t^{\mathrm{sc}}\bigr)^2,
\end{equation}
followed by the unchanged bias correction and update of \eqref{eq:adam}.
FRMSprop and FAdadelta are obtained in the same way by substituting
$g_t\to g_t^{\mathrm{sc}}$ in \eqref{eq:rmsprop} and \eqref{eq:adadelta}.
In the experiments this group is evaluated at every order of the sweep
$\nu\in\{0.75,1.25,1.50\}$; the order is held fixed within each run.

\,\par\noindent\textbf{Properties. }
Four properties characterize this group. \emph{Locality:} the update
depends only on the current gradient; no gradient history is stored.
\emph{Compatibility:} the construction applies to any gradient-based
optimizer, since only the gradient input changes. \emph{Low overhead:} the
cost is one componentwise power and one multiplication per step.
\emph{Classical limit:} for $\nu=1$ the exponent in
\eqref{eq:herrera_factor} is zero and $\Gamma(1)=1$, so $f_1(g_t)=1$
identically and every optimizer of this group reduces exactly to its base
optimizer. The interpretation is that the gradient direction is preserved
while its componentwise magnitude is modulated by a power law of the
gradient magnitude: for $\nu<1$, components with large $|g_t|$ receive
relatively amplified updates; for $\nu>1$, they receive relatively damped
updates. In the terminology of Section~\ref{sec:weierstrass_fractional},
these methods realize fractional \emph{scaling} but not fractional
\emph{memory}; they correspond to the non-hereditary surrogate discussed
there.

\subsection{Memory-Based Fractional Optimizers}
\label{subsec:memory_fractional_optimizers}

This subsection introduces memory-based fractional optimizers that replace the current gradient with a finite-history Grünwald--Letnikov gradient. The same memory-based direction is then integrated into the update rules of the corresponding classical optimizers.

\,\par\noindent\textbf{Construction. }
The second group preserves the defining property of fractional calculus,
nonlocality, by building the update direction from the finite-history
Gr\"unwald--Letnikov gradient $g_t^{(\nu)}$ of
\eqref{eq:gl_gradient_recall}. The optimizer maintains a buffer of the $K$
most recent gradients,
\begin{equation}
\label{eq:gradient_buffer}
\mathcal{H}_t=\{g_t,\,g_{t-1},\,\dots,\,g_{t-K+1}\},
\end{equation}
and at every iteration replaces the oldest entry by the new gradient and
computes $g_t^{(\nu)}$ as the weighted sum \eqref{eq:gl_gradient_recall};
this is a discrete convolution of the buffer with the kernel $c_k^{(\nu)}$.
(The implementation stores the buffer as a circular array and writes one
entry per step, so the per-step cost of the buffer update is independent
of $K$.) Optionally the coefficients are normalized,
\begin{equation}
\label{eq:normalized_coefficients}
\tilde c_k^{(\nu)}
=
\frac{c_k^{(\nu)}}{\sum_{j=0}^{K-1}\bigl|c_j^{(\nu)}\bigr|},
\end{equation}
which fixes the total absolute weight of the kernel; the experiments in
this article use the unnormalized coefficients for this group. Because
truncated discrete fractional operators do not reduce to the classical
case under every coefficient convention, the implementations impose the
explicit classical fallback $g_t^{(\nu)}=g_t$ whenever
$|\nu-1|\le\tau_\nu$ for a small tolerance $\tau_\nu>0$, or when $K=1$;
this preserves an exact classical limit at $\nu=1$. The implementations
use $\tau_\nu=10^{-12}$, and the fallback bypasses the descent safeguard
and the norm matching described below, so that the recovery of the base
optimizer at $\nu=1$ is exact rather than approximate. Since the swept
orders of this group are $\nu\in\{0.75,1.25,1.50\}$, the fallback is never
active in the reported experiments; it exists to make the classical limit
well defined.

\,\par\noindent\textbf{Two stabilizing mechanisms. }
The raw substitution $g_t\to g_t^{(\nu)}$ has a structural weakness that
follows directly from the kernel. The full Gr\"unwald--Letnikov
coefficient sequence sums to zero, $\sum_{k=0}^{\infty}c_k^{(\nu)}=0$ for
every $\nu>0$ \cite{podlubny1999fractional}, so the truncated sums
$\sum_{k=0}^{K-1}c_k^{(\nu)}$ are small (for example $+0.018$ for
$\nu=0.9$, $K=8$) and can even be negative for $\nu\in(1,2)$ (for example
$-0.016$ for $\nu=1.5$, $K=8$). In a phase of training where consecutive
gradients are similar, the history terms therefore nearly cancel the
current gradient: the response of the raw filter to a persistent descent
direction is strongly attenuated for $\nu<1$ and points uphill for
$\nu>1$. The implementations counter this with two mechanisms, both
enabled in all experiments. First, a \emph{descent safeguard} discards the
memory combination whenever it is not descent-compatible with the current
gradient,
\begin{equation}
\label{eq:memory_descent_safeguard}
\tilde g_t^{(\nu)}
=
\begin{cases}
g_t^{(\nu)}, & \bigl\langle g_t^{(\nu)},\,g_t\bigr\rangle > 0,\\[0.2em]
g_t, & \text{otherwise},
\end{cases}
\end{equation}
evaluated per parameter tensor. Second, \emph{norm matching} rescales the
result to the current gradient norm,
\begin{equation}
\label{eq:memory_norm_matching}
\hat g_t^{(\nu)}
=
\tilde g_t^{(\nu)}\,
\frac{\|g_t\|}{\|\tilde g_t^{(\nu)}\|+\varepsilon},
\end{equation}
so that the memory term contributes the \emph{direction} assembled from
the gradient history, while the step magnitude stays on the scale of the
current gradient, for which the base-optimizer hyperparameters are tuned.
This is the same norm-matching construction used in the adaptive group
below and listed in the notation appendix.

\,\par\noindent\textbf{Resulting optimizers. }
A memory-based fractional optimizer is obtained by substituting
$g_t\to \hat g_t^{(\nu)}$ in the \emph{directional} part of the base
update, while every squared-gradient accumulator is computed from the raw
gradient $g_t$, which keeps the per-coordinate variance estimates stable.
MemoryFSGD reads
\begin{equation}
\label{eq:memory_fsgd}
v_t^{(\mathrm{mom})}
= \beta_m\, v_{t-1}^{(\mathrm{mom})} - \eta\, \hat g_t^{(\nu)},
\qquad
\theta_{t+1} = \theta_t + v_t^{(\mathrm{mom})},
\end{equation}
which for $\beta_m=0$ is the hereditary fractional gradient descent of
Section~\ref{sec:weierstrass_fractional} with the two safeguards above.
MemoryFAdam computes the first moment from the safeguarded fractional
gradient and the second moment from the raw gradient,
\begin{equation}
\label{eq:memory_fadam}
m_t = \beta_1 m_{t-1} + (1-\beta_1)\, \hat g_t^{(\nu)},
\qquad
v_t^{(2)} = \beta_2 v_{t-1}^{(2)} + (1-\beta_2)\, g_t^2,
\end{equation}
followed by the unchanged bias correction and update of \eqref{eq:adam}.
MemoryFRMSprop and MemoryFAdadelta are obtained by the same substitution
in \eqref{eq:rmsprop} and \eqref{eq:adadelta}: $\hat g_t^{(\nu)}$ replaces
$g_t$ in the numerator and in the Adadelta update increment, and $g_t$
drives the accumulators. (Feeding the fractional gradient into the
squared-gradient accumulators as well remains available as a
configuration switch but is not used in the experiments.)

\,\par\noindent\textbf{Relation to the other groups. }
The relation to the baselines is a substitution plus safeguards: for
$\nu=1$ the kernel collapses ($c_0^{(1)}=1$, $c_k^{(1)}=0$ for $k\ge1$
under the fallback convention), the safeguards act as identities, and each
method reduces exactly to its base optimizer. The relation to the Herrera
group is a change of mechanism: FAdam rescales the current gradient
componentwise by $f_\nu(g_t)$ and stores nothing, while MemoryFAdam
leaves each individual gradient unchanged and instead combines $K$ of them
with the power-law kernel. The two groups therefore probe the two routes
of Section~\ref{subsec:from_integer_to_fractional} separately: fractional
scaling versus fractional memory. The price of memory is state: each
parameter tensor carries a history buffer of $K$ gradient copies, so
memory grows linearly in $K$. Following the short-memory principle
\cite{podlubny1999fractional} and the practice of Zhou et al.
\cite{zhou2023gl}, the experiments use small buffers (history length six
in the neural-network experiments).

\,\par\noindent\textbf{Interaction with multiscale structure. }
This group is the direct optimizer-side counterpart of the fractal objects
in this article. As derived in Section~\ref{sec:weierstrass_fractional},
gradients produced by Weierstrass-type activations, and gradients on
surfaces carrying the fractal perturbation ladder, contain persistent
oscillatory components across scales. The kernel $c_k^{(\nu)}$ retains such
components over several iterations instead of reacting only to the newest
one; whether this helps or hurts depends on whether the retained structure
carries stable directional information, which is exactly what the surface
and classification experiments measure.

\subsection{Adaptive Memory-Based Fractional Optimizers}
\label{subsec:adaptive_memory_optimizers}

This subsection introduces adaptive memory-based fractional optimizers that combine the ordinary gradient with a finite-history fractional gradient. The fractional order remains fixed, while a bounded mixing coefficient is adapted during optimization to control the contribution of the memory term.

\,\par\noindent\textbf{Motivation. }
A fixed order $\nu$ fixes the memory behaviour for the whole run, which may
be too rigid: memory can be useful in one phase of training and harmful in
another. The literature addresses this through variable-order or
adaptive-exponent schemes
\cite{xiang2025aofgd,partohaghighi2025twoscale,chen2024foadam}. Related
fractional methods instead adapt the learning rate or other optimizer
parameters while retaining a fixed fractional-derivative construction
\cite{ma2025apfogdl}. However, adapting the order introduces an additional
time-varying control variable and associated design choices. The broader
literature identifies stability, oscillation, implementation, and
convergence challenges for adaptive and variable-order fractional methods
\cite{elnady2025survey}. The framework of this group avoids this additional
time-varying order variable entirely. The order $\nu$ stays fixed; what is
adapted is a single bounded scalar, the \emph{mixing coefficient} $\lambda_t\in[0,1]$, which
controls how much the optimizer trusts the memory-based gradient relative
to the ordinary one.

\,\par\noindent\textbf{Step 1: fractional gradient, descent safeguard, and norm
matching. }
At each iteration the optimizer computes the finite-history fractional
gradient $g_t^{(\nu)}$ of \eqref{eq:gl_gradient_recall}, using the
normalized coefficients \eqref{eq:normalized_coefficients} in this group.
The order is held fixed at $\nu=1.25$ throughout, with memory length
$K=4$; unlike the two preceding groups, this group is not swept over the
order, because its adaptive quantity is the mixing coefficient rather than
$\nu$. The fixed order is deliberately kept away from the classical value,
so the fallback of
Section~\ref{subsec:memory_fractional_optimizers} never activates and the
memory term is a genuine fractional object at every step.
The same two safeguards as in
Section~\ref{subsec:memory_fractional_optimizers} are applied: the descent
safeguard \eqref{eq:memory_descent_safeguard} falls back to $g_t$ whenever
the memory combination is not descent-compatible, and norm matching
rescales the result to the current gradient norm,
\begin{equation}
\label{eq:norm_matching}
\hat g_t^{(\nu)}
=
\tilde g_t^{(\nu)}\,
\frac{\|g_t\|}{\|\tilde g_t^{(\nu)}\|+\varepsilon},
\qquad\text{so that}\qquad
\|\hat g_t^{(\nu)}\|\approx\|g_t\|.
\end{equation}
Norm matching separates direction from magnitude: the memory term
contributes only directional information, while the step size remains
governed by the current gradient and the base optimizer.

\,\par\noindent\textbf{Step 2: adaptive mixing. }
The effective gradient is the convex combination
\begin{equation}
\label{eq:effective_gradient}
g_t^{\mathrm{eff}}
=
(1-\lambda_t)\, g_t + \lambda_t\, \hat g_t^{(\nu)}.
\end{equation}
The two endpoints are exact: $\lambda_t=0$ reproduces the base optimizer,
and $\lambda_t=1$ reproduces a purely fractional-memory optimizer. In
between, $\lambda_t$ interpolates continuously between local information
and optimization history. In the experiments the coefficient is
initialized at $\lambda_0=0.10$ and clipped to
$[\lambda_{\min},\lambda_{\max}]=[0,0.30]$, so the memory term never
contributes more than $30\%$ of the effective gradient: the configuration
is deliberately conservative, and the upper endpoint $\lambda_t=1$ is a
property of the construction rather than a setting used here.

\,\par\noindent\textbf{Step 3: stability-driven adaptation of $\lambda_t$. }
The signal that drives $\lambda_t$ is a gradient-stability estimate. The
raw variability ratio and its exponentially smoothed version are
\begin{equation}
\label{eq:stability_score}
\rho_t
=
\min\!\left(
\frac{\|g_t-g_{t-1}\|}{\|g_{t-1}\|+\varepsilon},\;
\rho_{\max}\right),
\qquad
\bar\rho_t
=
\gamma_\rho\,\bar\rho_{t-1} + (1-\gamma_\rho)\,\rho_t,
\end{equation}
where the clip value $\rho_{\max}$ guards against outliers and
$\gamma_\rho\in[0,1)$ is the EMA decay rate. Small $\bar\rho_t$ indicates
consistent gradients; large $\bar\rho_t$ indicates rapidly changing
gradients. The mixing coefficient follows a bounded ratchet rule with
stability threshold $\tau_s$ and increment $\eta_\lambda>0$: first a
target value is formed,
\begin{equation}
\label{eq:lambda_target}
\lambda_t^{\mathrm{tar}}
=
\operatorname{clip}\!\Bigl(
\lambda_{t-1}
+ \eta_\lambda\,\operatorname{sign}(\tau_s-\bar\rho_t)
- \mathbf{1}_t^{\mathrm{worse}}\,\Delta_\lambda,\;
\lambda_{\min},\,\lambda_{\max}^{(t)}
\Bigr),
\end{equation}
where $\mathbf{1}_t^{\mathrm{worse}}\in\{0,1\}$ is the loss-aware
indicator of Step~4 and $\Delta_\lambda$ its decrement; then the stored
coefficient is moved toward this target by EMA smoothing with decay rate
$\gamma_\lambda$ and clipped once more,
\begin{equation}
\label{eq:lambda_ratchet}
\lambda_t
=
\operatorname{clip}\!\Bigl(
\gamma_\lambda\,\lambda_{t-1}
+(1-\gamma_\lambda)\,\lambda_t^{\mathrm{tar}},\;
\lambda_{\min},\,\lambda_{\max}^{(t)}
\Bigr),
\end{equation}
and this value is used in \eqref{eq:effective_gradient}. The upper bound
is optionally ramped up linearly over a warm-up horizon of $T_w$ steps,
$\lambda_{\max}^{(t)}=\lambda_{\max}\min(t/T_w,1)$ (with
$\lambda_{\max}^{(t)}=\lambda_{\max}$ when $T_w=0$), so that the memory
contribution grows only after the first gradients have been observed. In
plain terms: when training is stable, the optimizer trusts its
accumulated experience a little more; when gradients become inconsistent,
it falls back toward the classical method. The experiments use
$\eta_\lambda=0.01$, $\gamma_\lambda=0.90$, $\gamma_\rho=0.95$,
$\tau_s=0.25$, and $\rho_{\max}=10$, with the warm-up ramp disabled
($T_w=0$), so that $\lambda_{\max}^{(t)}=\lambda_{\max}=0.30$ from the
first step onward.

\,\par\noindent\textbf{Step 4: loss-aware override. }
When loss values are available, a second safeguard is applied. The
smoothed loss $\bar L_t=\gamma_L\bar L_{t-1}+(1-\gamma_L)L_t$ is tracked
together with its best value $\bar L_{\mathrm{best}}$. The indicator in
\eqref{eq:lambda_target} is set to
$\mathbf{1}_t^{\mathrm{worse}}=1$ whenever the smoothed loss is
significantly worse than the best observed value,
$\bar L_t>\bar L_{\mathrm{best}}+\delta_L$, and to $0$ otherwise; the
decrement $\Delta_\lambda$ is therefore applied at every step for which
the condition holds. This allows the optimizer to withdraw the memory
contribution whenever it is demonstrably harmful, independently of the
gradient-stability signal. The experiments use $\gamma_L=0.95$,
$\delta_L=10^{-4}$, and $\Delta_\lambda=0.02$; the decrement is twice the
stability increment $\eta_\lambda$, so a worsening loss withdraws memory
faster than a stable gradient sequence restores it.

\,\par\noindent\textbf{Resulting optimizers. }
The effective gradient enters only the \emph{directional} part of each
base update; every squared-gradient accumulator is computed from the raw
gradient $g_t$, which keeps the variance estimates stable. The four
instances are:
\begin{align}
\text{AdaptiveMemoryFSGD:}\quad
& v_t^{(\mathrm{mom})} = \beta_m v_{t-1}^{(\mathrm{mom})}
  - \eta\, g_t^{\mathrm{eff}},
\qquad
\theta_{t+1}=\theta_t+v_t^{(\mathrm{mom})};
\label{eq:amfsgd}\\[0.4em]
\text{AdaptiveMemoryFRMSprop:}\quad
& v_t^{(2)} = \gamma_v v_{t-1}^{(2)} + (1-\gamma_v)\,g_t^2,
\qquad
\theta_{t+1}=\theta_t-\eta\,
\frac{g_t^{\mathrm{eff}}}{\sqrt{v_t^{(2)}+\varepsilon}};
\label{eq:amfrmsprop}\\[0.4em]
\text{AdaptiveMemoryFAdam:}\quad
& m_t = \beta_1 m_{t-1} + (1-\beta_1)\,g_t^{\mathrm{eff}},
\qquad
v_t^{(2)} = \beta_2 v_{t-1}^{(2)} + (1-\beta_2)\,g_t^2,
\label{eq:amfadam}\\
& \hat m_t = \frac{m_t}{1-\beta_1^t},
\quad
\hat v_t^{(2)} = \frac{v_t^{(2)}}{1-\beta_2^t},
\quad
\theta_{t+1}=\theta_t-\eta\,
\frac{\hat m_t}{\sqrt{\hat v_t^{(2)}}+\varepsilon};
\nonumber\\[0.4em]
\text{AdaptiveMemoryFAdadelta:}\quad
& v_t^{(2)} = \gamma_v v_{t-1}^{(2)} + (1-\gamma_v)\,g_t^2,
\qquad
\Delta\theta_t = -\,
\frac{\sqrt{v_{t-1}^{(\Delta)}+\varepsilon}}
     {\sqrt{v_t^{(2)}+\varepsilon}}\;
g_t^{\mathrm{eff}},
\label{eq:amfadadelta}
\end{align}
with the Adadelta update accumulator and parameter step as in
\eqref{eq:adadelta}.

\,\par\noindent\textbf{Interpretation. }
The framework can be read as a continuously self-adjusting interpolation
between classical and fractional optimization. The ordinary gradient is
the optimizer's immediate observation of the landscape; the norm-matched
fractional gradient is its accumulated experience; $\lambda_t$ decides how
strongly the two are combined, based on whether recent experience has been
consistent and whether the loss is improving. Unlike adaptive-order
methods, no second optimization problem is solved inside the optimizer,
and unlike the substitution-based methods of the previous two groups, the
classical optimizer is exactly recovered at $\lambda_t=0$ at any point
during training. The methods are therefore conservative, memory-aware
extensions of SGD, RMSprop, Adam, and Adadelta. The conservatism is also
quantitative: with $\lambda_{\max}=0.30$ the update direction is at least
$70\%$ ordinary gradient at every step, which is the intended design point
and should be kept in mind when the measured differences to the
corresponding baselines are small.

\subsection{Related Fractional Optimizers}
\label{subsec:related_work_optimizers}

The fifth group contains five optimizers from the recent literature. They
are included because each realizes one of the design principles above in a
different way, which allows the experiments to compare design principles
rather than isolated methods. All five are restated here in the canonical
notation of this article; the fractional order is written $\nu$ (or
$\nu_t$ when it varies), regardless of the symbol used in the original
works. Like the adaptive-memory group, these methods are evaluated at a
single fixed internal configuration and are not swept over the order; the
values used are stated with each method.

\,\par\noindent\textbf{FCSGD\_GL and FCAdam\_GL (Zhou, Zhao, and Huang
\cite{zhou2023gl}). }
These methods substitute a Gr\"unwald--Letnikov gradient for the ordinary
gradient, as in Section~\ref{subsec:memory_fractional_optimizers} but
without safeguards, with one additional element: a stochastic perturbation
of the history terms that is intended to reduce premature convergence to
local optima. With independent Bernoulli variables $b_k\in\{0,1\}$ satisfying
$\Pr(b_k=1)=p$ and $\Pr(b_k=0)=1-p$, the perturbed fractional gradient is
\begin{equation}
\label{eq:fcgl_gradient}
g_t^{\mathrm{FC}}
=
g_t + \sum_{k=1}^{K-1} b_k\, c_k^{(\nu)}\, g_{t-k},
\end{equation}
where the original formulation uses ten history terms ($K=11$). FCSGD\_GL
applies plain SGD to this gradient,
$\theta_{t+1}=\theta_t-\eta\,g_t^{\mathrm{FC}}$, and FCAdam\_GL feeds it
into \emph{both} Adam moments, exactly as in \eqref{eq:adam} with
$g_t$ replaced by $g_t^{\mathrm{FC}}$. Replacing the Bernoulli variables
by $b_k=1$ deterministically recovers an unsafeguarded MemoryF-type
optimizer. There is no mechanism to fall back to the classical optimizer:
the order is fixed and the substitution is permanent. The implementation
used here runs both methods with a fixed order $\nu=0.9$, the buffer
length $K=11$ of the original formulation, and retention probability
$p=0.5$, so that each history term is kept or dropped with equal
probability at every step.

\,\par\noindent\textbf{AdaGL (Chen, Zhang, and Mu \cite{chen2024adagl}). }
AdaGL also replaces the gradient inside both Adam moments by the
Gr\"unwald--Letnikov gradient $g_t^{(\nu)}$ (ten history terms), and adds
a short-term step-size control coefficient computed from the instantaneous
gradient change,
\begin{equation}
\label{eq:adagl_coefficient}
C_t
=
1.1-\frac{1}{2\bigl(1+|g_t-g_{t-1}|\bigr)},
\qquad
C_t\in[0.6,\,1.1),
\end{equation}
applied componentwise, which rescales the Adam step:
\begin{equation}
\label{eq:adagl_update}
\theta_{t+1}
=
\theta_t-\eta\, C_t\,
\frac{\hat m_t}{\sqrt{\hat v_t^{(2)}}+\varepsilon},
\end{equation}
with $\hat m_t$ and $\hat v_t^{(2)}$ built from $g_t^{(\nu)}$. The
implementation used here runs AdaGL with a fixed order $\nu=0.9$ and
memory length $K=10$. AdaGL is
the closest published method to the adaptive framework of
Section~\ref{subsec:adaptive_memory_optimizers}, since it also combines a
long-memory fractional component with a short-term correction. Two
structural differences remain. First, $C_t$ is a closed-form multiplier on
the learning rate recomputed at every step, whereas $\lambda_t$ is an
EMA-smoothed, hysteresis-controlled, clipped trust coefficient with an
additional loss-aware override. Second, AdaGL's moments are always
computed from the fractional gradient: $C_t$ can shrink or slightly
enlarge the step, but it cannot disable the fractional component, so
there is no counterpart to the exact reduction $\lambda_t=0$.

\,\par\noindent\textbf{AOFGD\_SGD and AOFGD\_Adam (Xiang et al.
\cite{xiang2025aofgd}). }
The AOFGD preprint proposes a Herrera-type local scaling method with a
time-varying order \cite{xiang2025aofgd}. In our implementation of AOFGD,
a convergence evaluation factor compares consecutive gradient norms,
\begin{equation}
\label{eq:aofgd_cef}
\mathrm{CEF}_t
=
\frac{\|g_t\|}{\|g_{t-1}\|+\varepsilon},
\end{equation}
and the order is instantiated by a bounded ratchet rule with increment
$\Delta_\nu$, threshold $\tau_{\mathrm{CEF}}$, clipping to
$[\nu_{\min},\nu_{\max}]$, and EMA smoothing with decay rate $\gamma_\nu$:
\begin{equation}
\label{eq:aofgd_order_rule}
\nu_t^{\mathrm{tar}}
=
\operatorname{clip}\!\Bigl(
\nu_{t-1}
- \Delta_\nu\,\operatorname{sign}\!\bigl(\mathrm{CEF}_t-1\bigr)
\cdot\mathbf{1}\bigl[\,|\mathrm{CEF}_t-1|>\tau_{\mathrm{CEF}}\,\bigr],
\;\nu_{\min},\,\nu_{\max}\Bigr),
\qquad
\nu_t
=
\gamma_\nu\,\nu_{t-1}+(1-\gamma_\nu)\,\nu_t^{\mathrm{tar}},
\end{equation}
so that a growing gradient norm decreases the order and increases the
power-law exponent, whereas a shrinking gradient norm increases the order
and, for orders below one, moves the scaling closer to its classical
limit. The gradient is then
scaled as in \eqref{eq:herrera_scaled_gradient} with the current smoothed
order, $g_t^{\mathrm{sc}}=f_{\nu_t}(g_t)\odot g_t$. AOFGD\_SGD applies plain
SGD to this gradient. In the implementation used here, AOFGD\_Adam feeds it
into the first moment while computing the second moment from the raw
gradient, which stabilizes the variance estimate when $\nu_t$ changes,
\begin{equation}
\label{eq:aofgd_adam}
m_t=\beta_1 m_{t-1}+(1-\beta_1)\,g_t^{\mathrm{sc}},
\qquad
v_t^{(2)}=\beta_2 v_{t-1}^{(2)}+(1-\beta_2)\,g_t^{2},
\end{equation}
followed by the standard bias-corrected Adam step. The order rule is run
with the initial value $\nu_0=0.9$, bounds
$[\nu_{\min},\nu_{\max}]=[0.5,1.5]$, increment $\Delta_\nu=0.02$,
threshold $\tau_{\mathrm{CEF}}=0.05$, and EMA decay $\gamma_\nu=0.95$;
these methods therefore store no gradient history, but they do carry the
scalar state $\nu_t$ from step to step. In contrast to the
adaptive framework of
Section~\ref{subsec:adaptive_memory_optimizers}, AOFGD adapts the
derivative order itself, always uses the Caputo-inspired scaled gradient,
and has no exact classical fallback.

\,\par\noindent\textbf{Positioning. }
Across the five related methods, fractional structure is introduced either
by permanent gradient substitution with fixed order (FCSGD\_GL, FCAdam\_GL;
similarly FracM at the level of momentum \cite{yu2022fracm}), by rescaling
the step size of an always-fractional update (AdaGL), or by adapting the
order of a local scaling law (AOFGD). The adaptive memory framework of
this article differs from all three: it keeps $\nu$ fixed, treats the
memory contribution as an explicitly bounded, hysteresis-controlled trust
coefficient applied as a convex combination of two gradient candidates,
and reduces exactly to the underlying classical optimizer at one boundary
of that coefficient.
\subsection{Summary of the 21 Optimizers}
\label{subsec:optimizer_summary}

Table~\ref{tab:optimizer_summary} summarizes the five optimizer groups.
Reading the table by group: the four baselines use the raw gradient $g_t$
and exponential forgetting. The four Herrera-type optimizers rescale $g_t$
componentwise by the Caputo-type factor $f_\nu(g_t)$; they add
fractional scaling at negligible cost, store no history, and recover
their baselines at $\nu=1$. The four memory-based optimizers build their
update direction from the finite-history Gr\"unwald--Letnikov gradient,
safeguarded by the descent check \eqref{eq:memory_descent_safeguard} and
norm matching \eqref{eq:memory_norm_matching}, keep all squared-gradient
accumulators on the raw gradient, and recover their baselines at $\nu=1$
via the explicit fallback convention. The four adaptive memory-based
optimizers mix $g_t$ with the same safeguarded, norm-matched memory
gradient $\hat g_t^{(\nu)}$ through the bounded, stability- and
loss-controlled coefficient $\lambda_t$, keep all variance estimates on
the raw gradient, and reduce exactly to their baselines at $\lambda_t=0$.
Among the related-work methods, FCSGD\_GL and FCAdam\_GL are
permanent-substitution memory methods with a Bernoulli-perturbed history;
AdaGL is a permanent-substitution memory method with an additional
short-term step-size coefficient $C_t$; and AOFGD\_SGD and AOFGD\_Adam are
Herrera-type scaling methods with an adaptively ratcheted order $\nu_t$
in the implementation used here.
The first two fractional groups are swept over
$\nu\in\{0.75,1.25,1.50\}$, with memory length $K=6$ for the memory-based
group; the adaptive memory-based group is run at the fixed order
$\nu=1.25$ with $K=4$, and the related-work methods at their own fixed
internal orders ($\nu=0.9$ for FCSGD\_GL, FCAdam\_GL, and AdaGL; an
adapted $\nu_t$ initialized at $0.9$ for the two AOFGD variants), with
$K=11$, $K=11$, and $K=10$ respectively and no gradient history for
AOFGD.
Extensions of both memory-based families to further base
optimizers---Adagrad, AdamW, Nadam, and Adamax---built on the same
safeguards and notation, are given in
Appendix~\ref{app:additional_memory_optimizers}.

\begin{table}[H]
\caption{The five optimizer groups compared in this article (21
optimizers in total). ``Update direction'' states which gradient object
enters the update; $\hat g_t^{(\nu)}$ is the descent-safeguarded,
norm-matched Gr\"unwald--Letnikov gradient of
\eqref{eq:memory_descent_safeguard}--\eqref{eq:memory_norm_matching}.
``Fallback'' states when the method reduces exactly to its base
optimizer.\label{tab:optimizer_summary}}
\small
\begin{tabularx}{\textwidth}{lXXcc}
\toprule
\textbf{Group} & \textbf{Optimizers} & \textbf{Update direction} &
\textbf{Memory} & \textbf{Fallback} \\
\midrule
Baselines &
SGD, RMSprop, Adam, Adadelta &
$g_t$ &
no & -- \\ \hline
\addlinespace[0.3em]
Herrera-type &
FSGD, FRMSprop, FAdam, FAdadelta &
$f_\nu(g_t)\odot g_t$, swept $\nu\in\{0.75,1.25,1.50\}$ &
no & $\nu=1$ \\ \hline
\addlinespace[0.3em]
Memory-based &
MemoryFSGD, MemoryFRMSprop, MemoryFAdam, MemoryFAdadelta &
$\hat g_t^{(\nu)}$, swept $\nu\in\{0.75,1.25,1.50\}$; accumulators from
$g_t$ &
$K=6$ & $\nu=1$ \\ \hline
\addlinespace[0.3em]
Adaptive memory &
AdaptiveMemoryFSGD, AdaptiveMemoryFRMSprop, AdaptiveMemoryFAdam,
AdaptiveMemoryFAdadelta &
$(1-\lambda_t)\,g_t+\lambda_t\,\hat g_t^{(\nu)}$, fixed $\nu=1.25$,
$\lambda_t\in[0,0.30]$; accumulators from $g_t$ &
$K=4$ & $\lambda_t=0$ \\ \hline
\addlinespace[0.3em]
Related work &
FCSGD\_GL, FCAdam\_GL, AdaGL, AOFGD\_SGD, AOFGD\_Adam &
GL substitution with fixed $\nu=0.9$ (Bernoulli-masked with $p=0.5$ for
FC*, step scaled by $C_t$ for
AdaGL); AOFGD: $f_{\nu_t}(g_t)\odot g_t$ with adaptive $\nu_t$ &
\begin{tabular}[c]{@{}l@{}}$K=11$ (FC*);\\ $K=10$ (AdaGL);\\ AOFGD: no\end{tabular} & none \\
\bottomrule
\end{tabularx}
\end{table}

Two observations conclude the section. First, the five groups decompose
the design space along three axes that the experiments can then evaluate
separately: local fractional scaling versus explicit fractional memory
(Herrera versus Memory groups), fixed versus adaptive control (fixed
$\nu$ versus adaptive $\nu_t$ or adaptive $\lambda_t$), and permanent
substitution versus safeguarded use with exact classical fallback (related
work versus the Memory and AdaptiveMemory groups). Second, all
memory-carrying methods share the same kernel $c_k^{(\nu)}$ derived in
Section~\ref{sec:weierstrass_fractional}, so differences in their
behaviour on fractal surfaces and with fractal activations are
attributable to how the kernel output is used, not to the kernel itself.

\begin{takeawaybox}[yellowA]{Main Takeaways --- Fractional Optimizers}
  \begin{itemize}[
    label={},
    leftmargin=0em,
    itemindent=0em,
    itemsep=3pt,
    topsep=2pt
  ]
    \item \textbf{The comparison covers five optimizer groups.}
          The 21 methods include classical baselines, Herrera-type
          optimizers, memory-based variants, adaptive-memory variants, and
          published fractional alternatives.

    \item \textbf{Fractional updates follow two main mechanisms.}
          Herrera-type methods rescale the current gradient locally, whereas
          Gr\"unwald--Letnikov methods combine past gradients through an
          algebraically decaying memory kernel.

    \item \textbf{Memory must be controlled to be useful.}
          Descent safeguards, norm matching, and the adaptive coefficient
          \(\lambda_t\) regulate how strongly history is trusted and allow
          exact recovery of the base optimizer.
  \end{itemize}
\end{takeawaybox}

\section{Surface Optimization Experiments}
\label{sec:surface_experiments}

The first experimental block studies optimizer behaviour on controlled two-dimensional benchmark surfaces. The goal is not to solve a practical classification task, but to observe how different optimizer families move on known objective landscapes whose minima, local structure, and perturbations can be inspected directly.

\subsection{Benchmark Surfaces, Fractal Perturbations, and Experimental Design}
\label{subsec:surface_design}
\begin{figure}[H]
    \centering

    \begin{minipage}{0.32\textwidth}
        \centering
        \includegraphics[width=\linewidth]{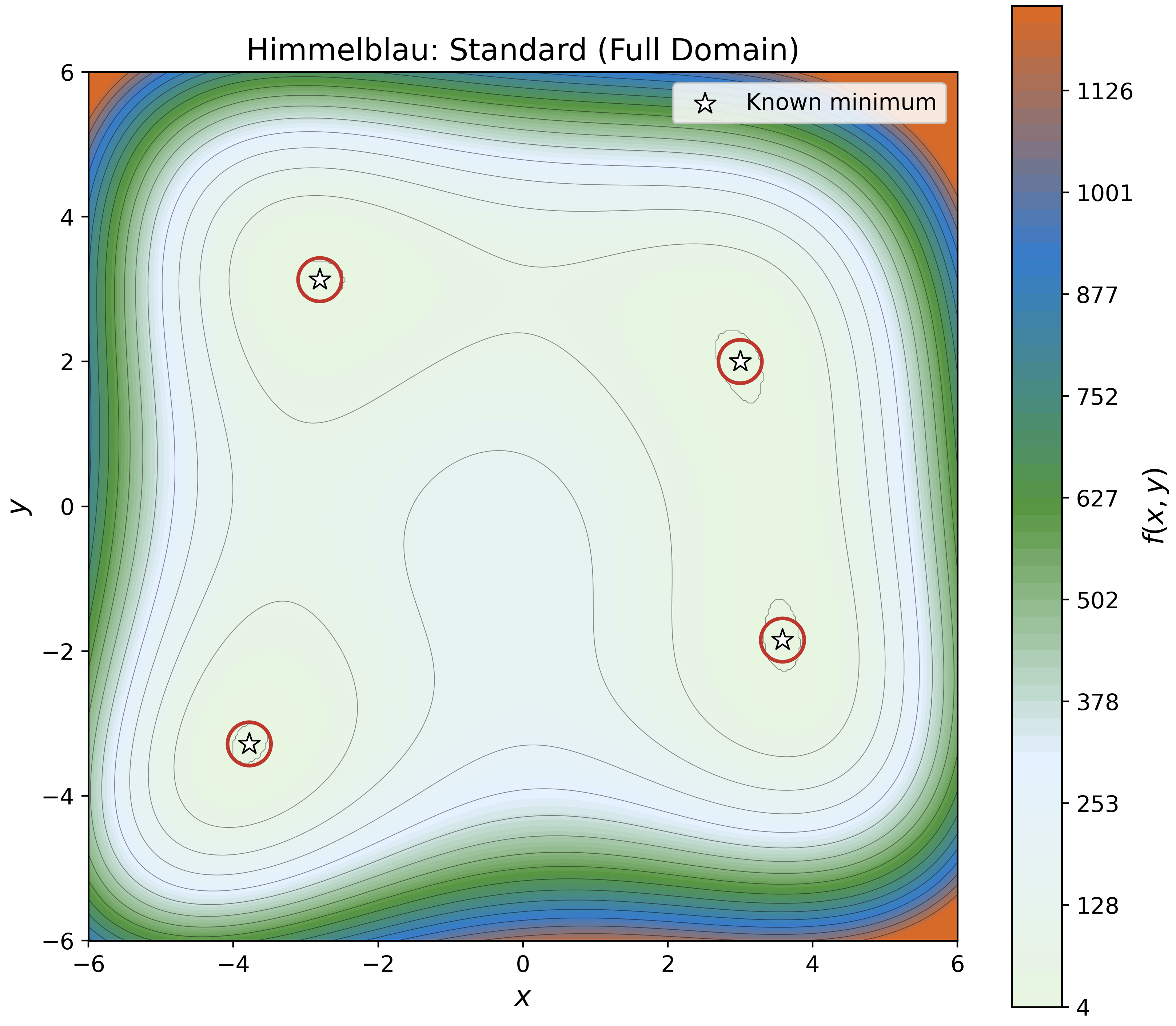}
    \end{minipage}
    \hfill
    \begin{minipage}{0.32\textwidth}
        \centering
        \includegraphics[width=\linewidth]{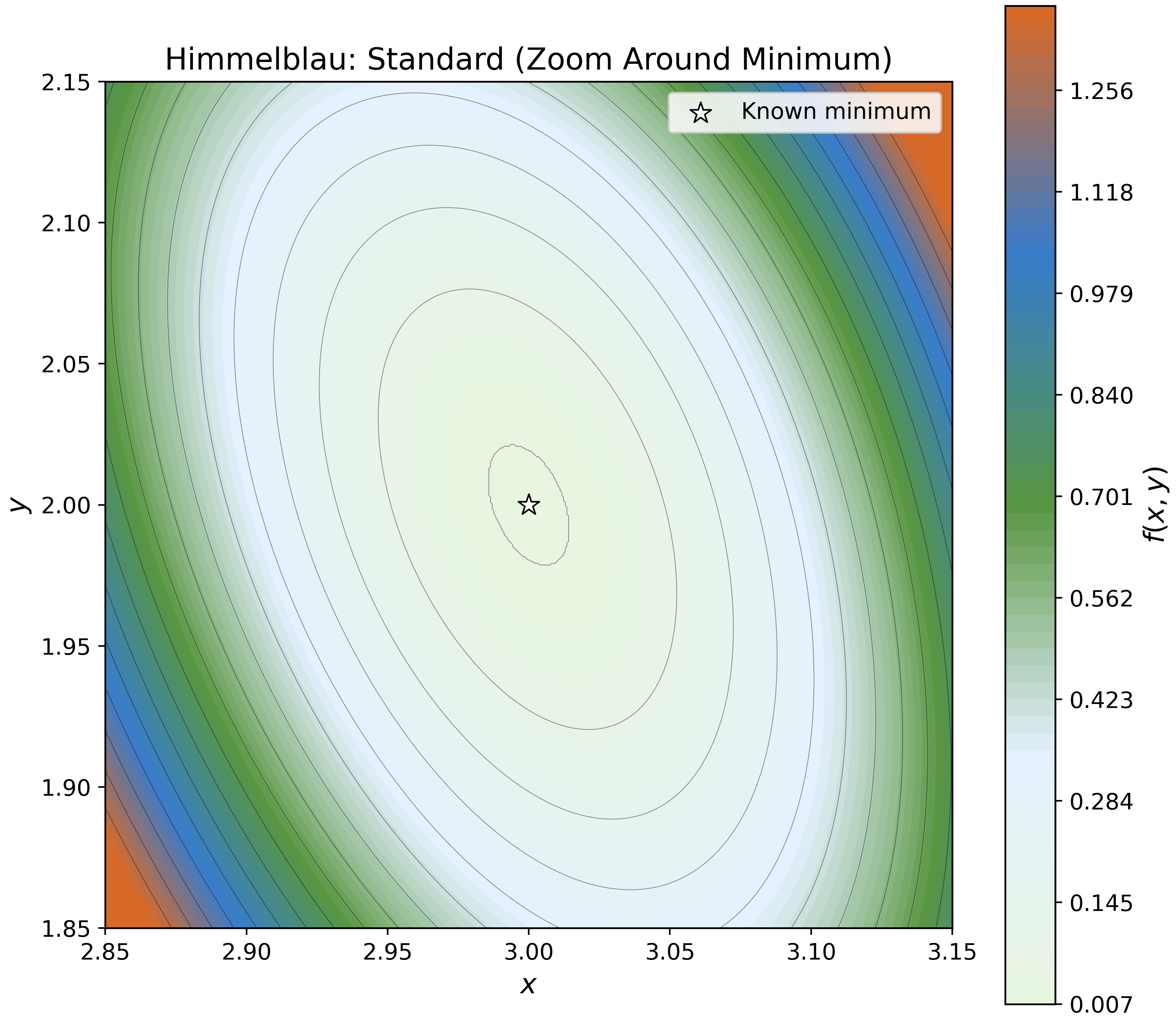}
    \end{minipage}
    \hfill
    \begin{minipage}{0.32\textwidth}
        \centering
        \includegraphics[width=\linewidth]{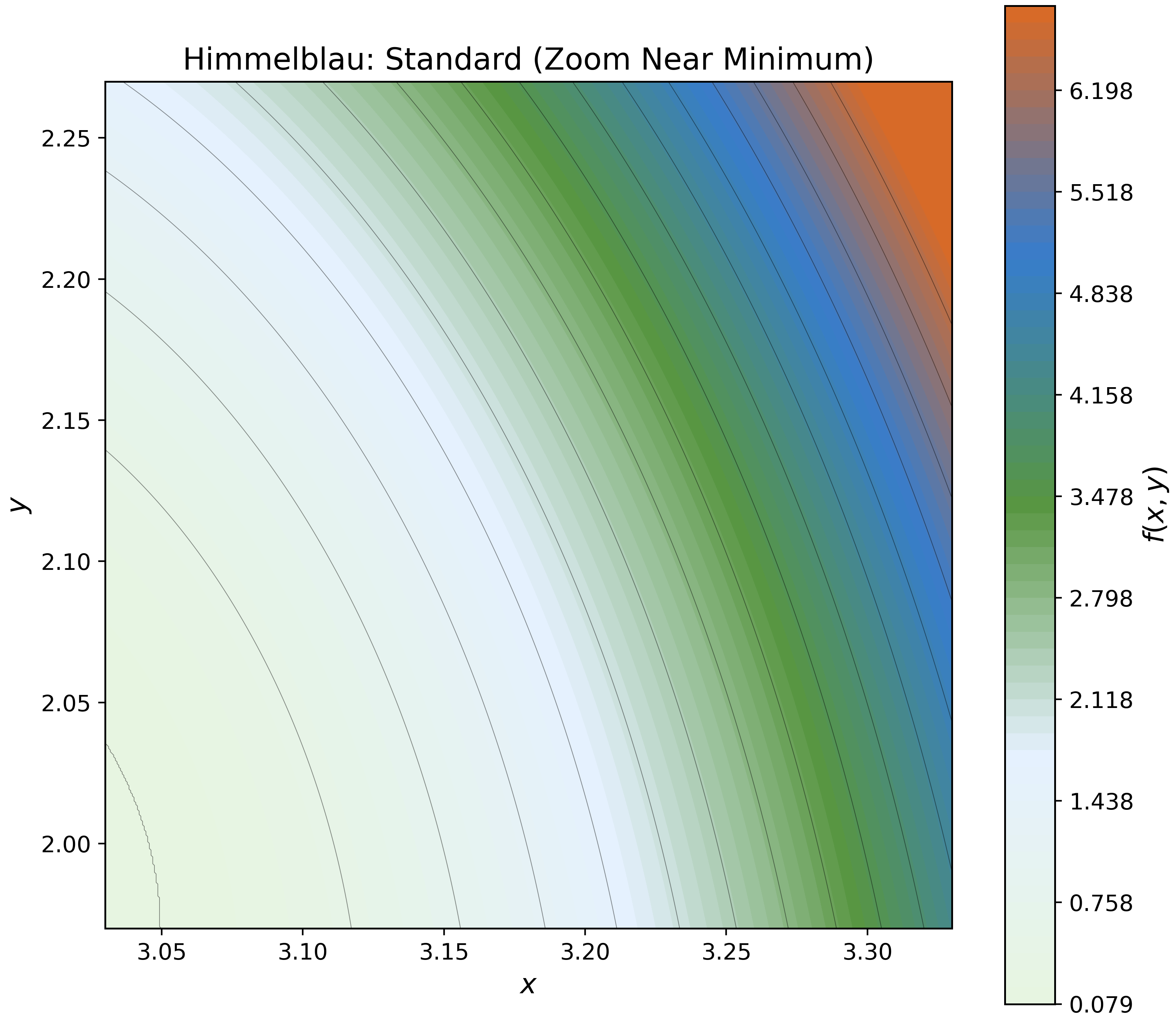}
    \end{minipage}

    \vspace{0.5em}

    \begin{minipage}{0.32\textwidth}
        \centering
        \includegraphics[width=\linewidth]{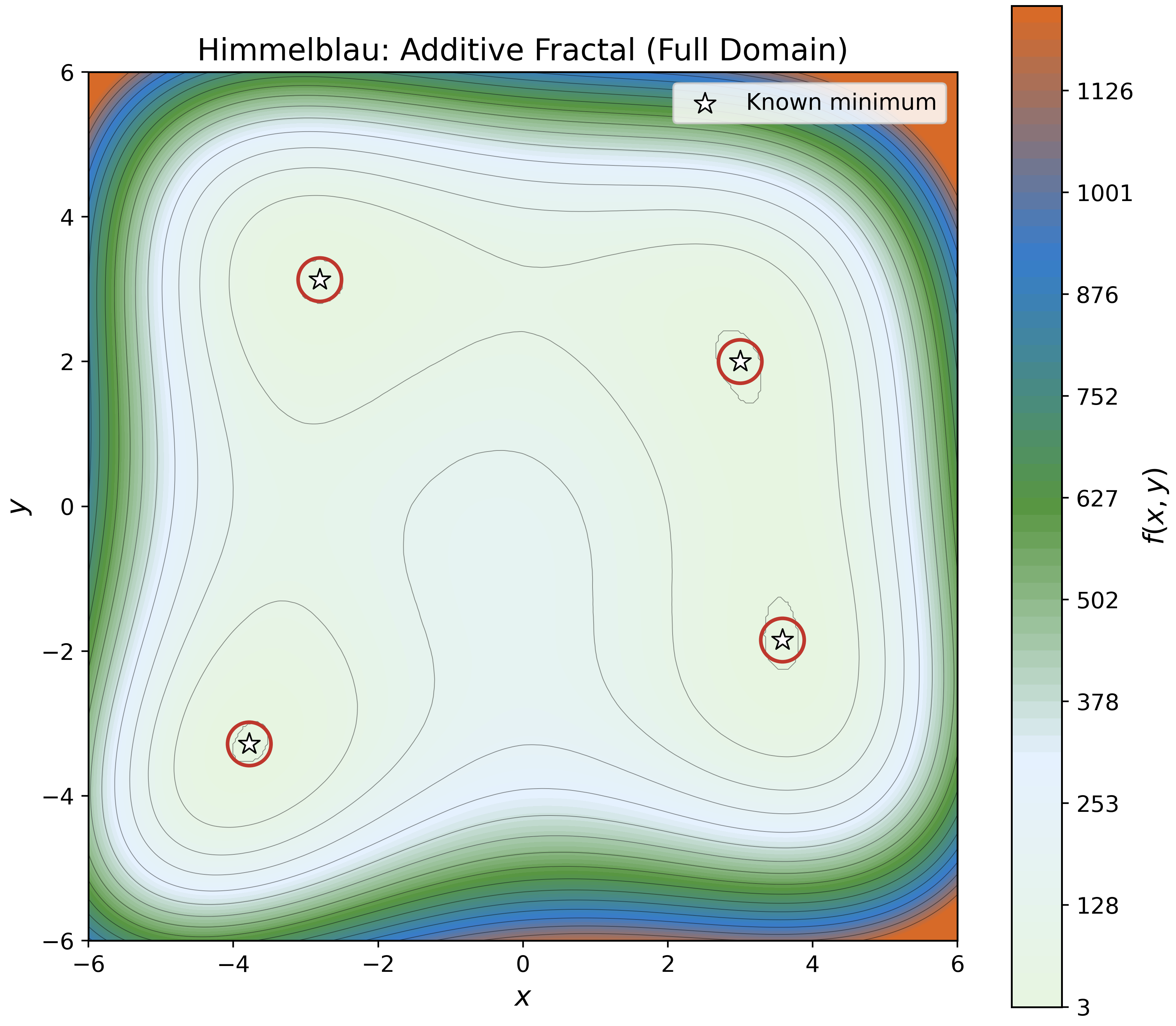}
    \end{minipage}
    \hfill
    \begin{minipage}{0.32\textwidth}
        \centering
        \includegraphics[width=\linewidth]{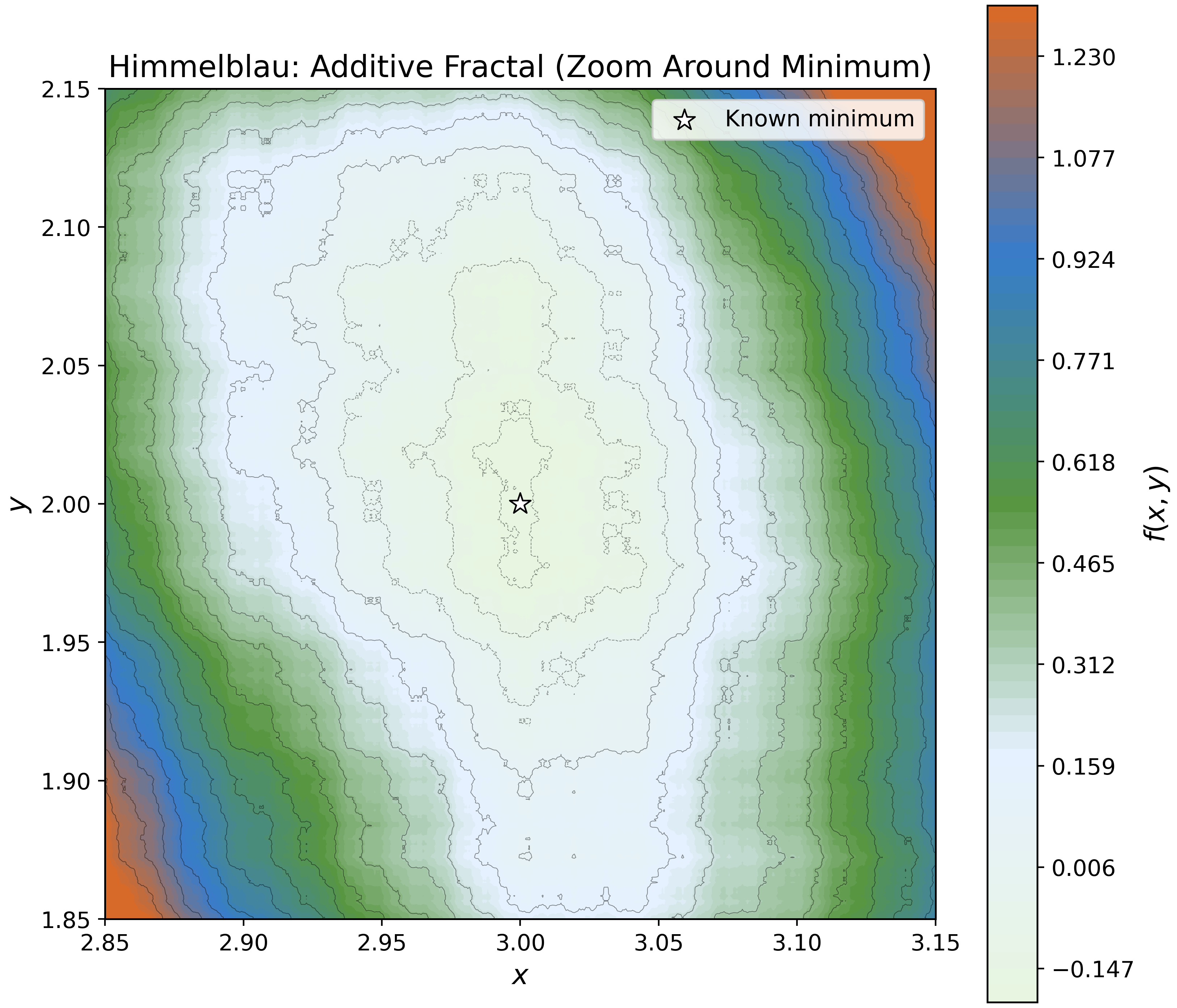}
    \end{minipage}
    \hfill
    \begin{minipage}{0.32\textwidth}
        \centering
        \includegraphics[width=\linewidth]{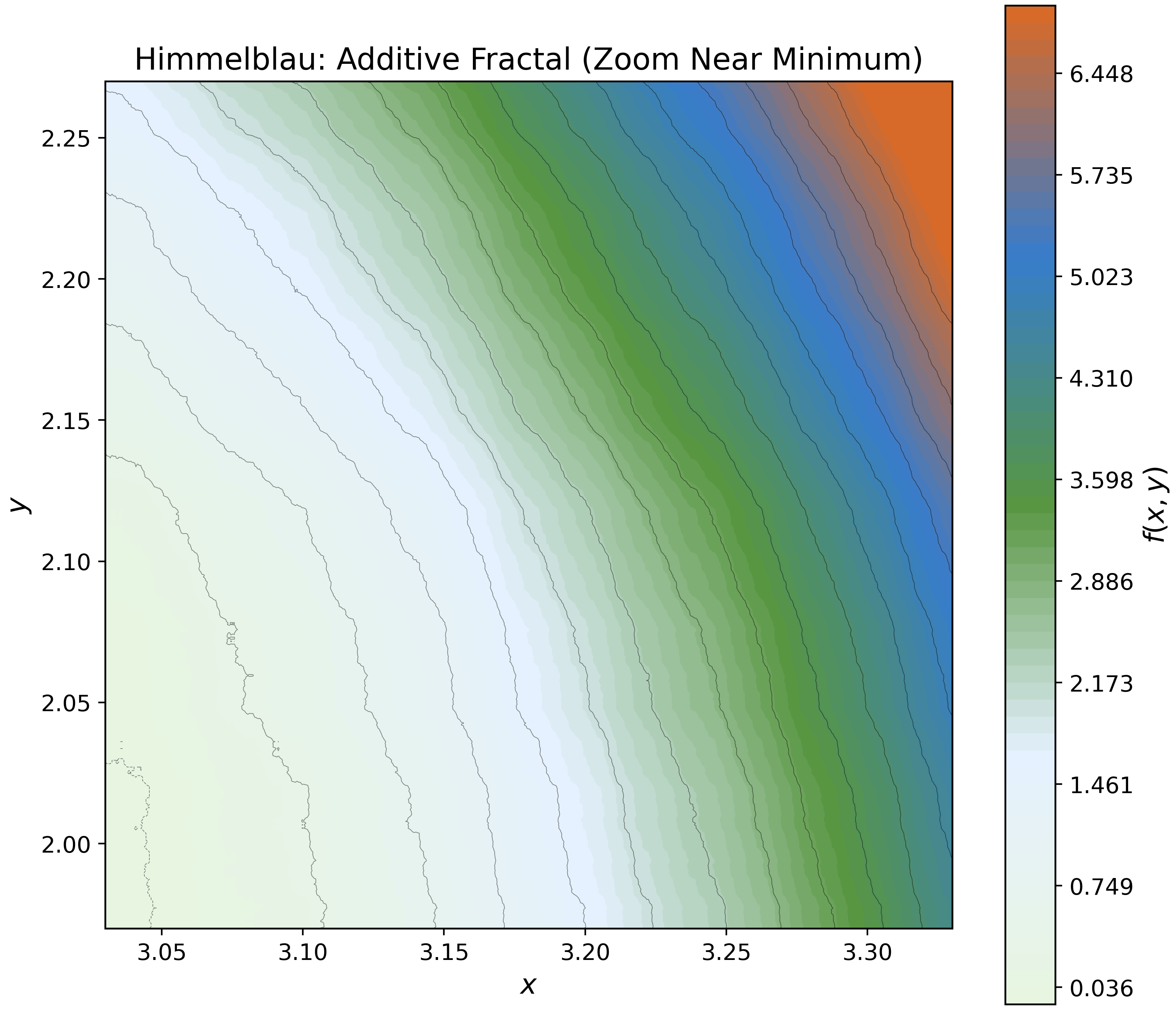}
    \end{minipage}

    \caption{Comparison of the standard Himmelblau surface with the additive fractal perturbation. Each row shows one surface variant. The columns show the full domain, a zoom around a known minimum, and a nearby zoomed region. The additive perturbation introduces local roughness directly into the basin.}
    \label{fig:himmelblau_standard_additive}
\end{figure}

\begin{figure}[H]
    \centering

    \begin{minipage}{0.32\textwidth}
        \centering
        \includegraphics[width=\linewidth]{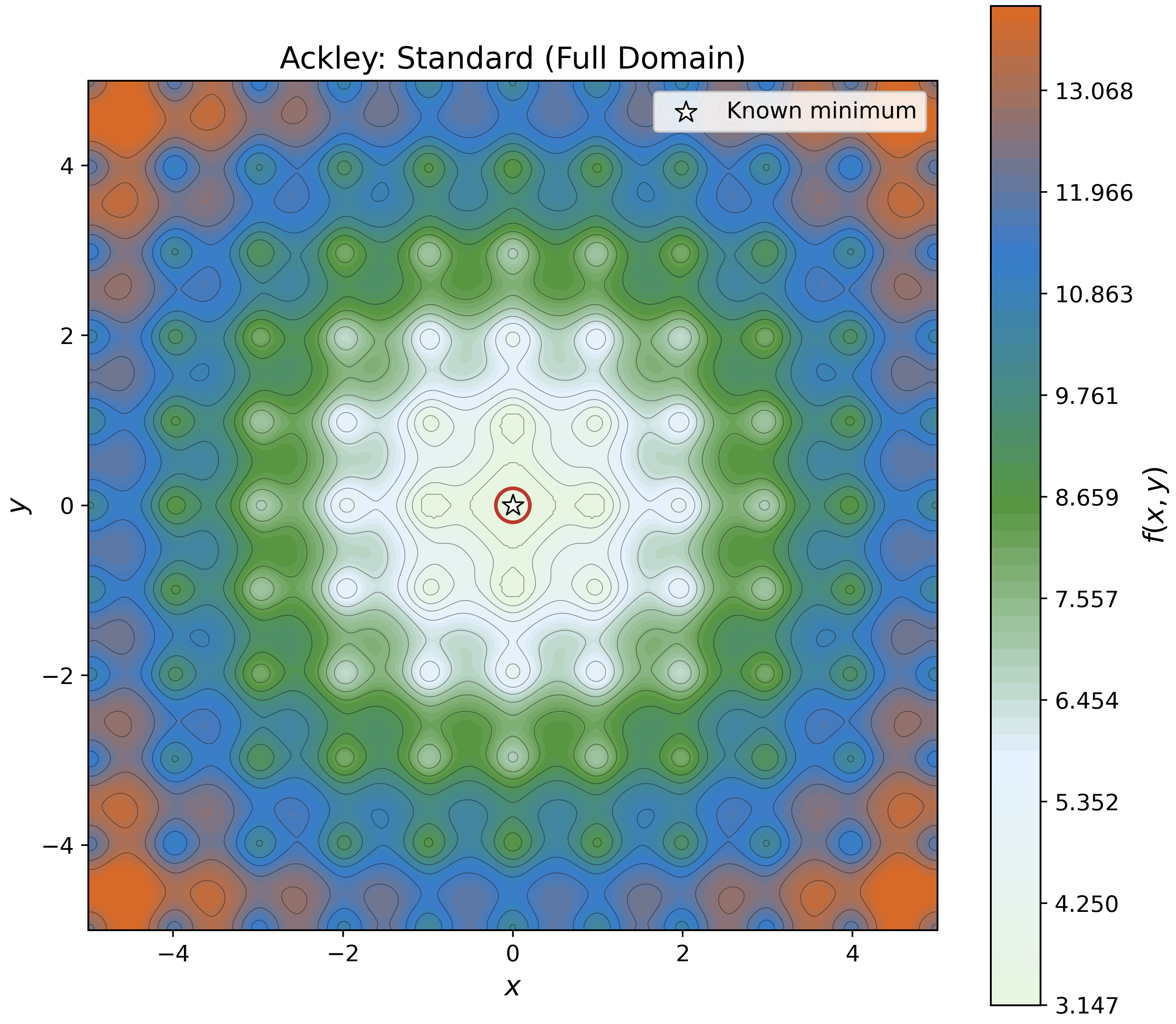}
    \end{minipage}
    \hfill
    \begin{minipage}{0.32\textwidth}
        \centering
        \includegraphics[width=\linewidth]{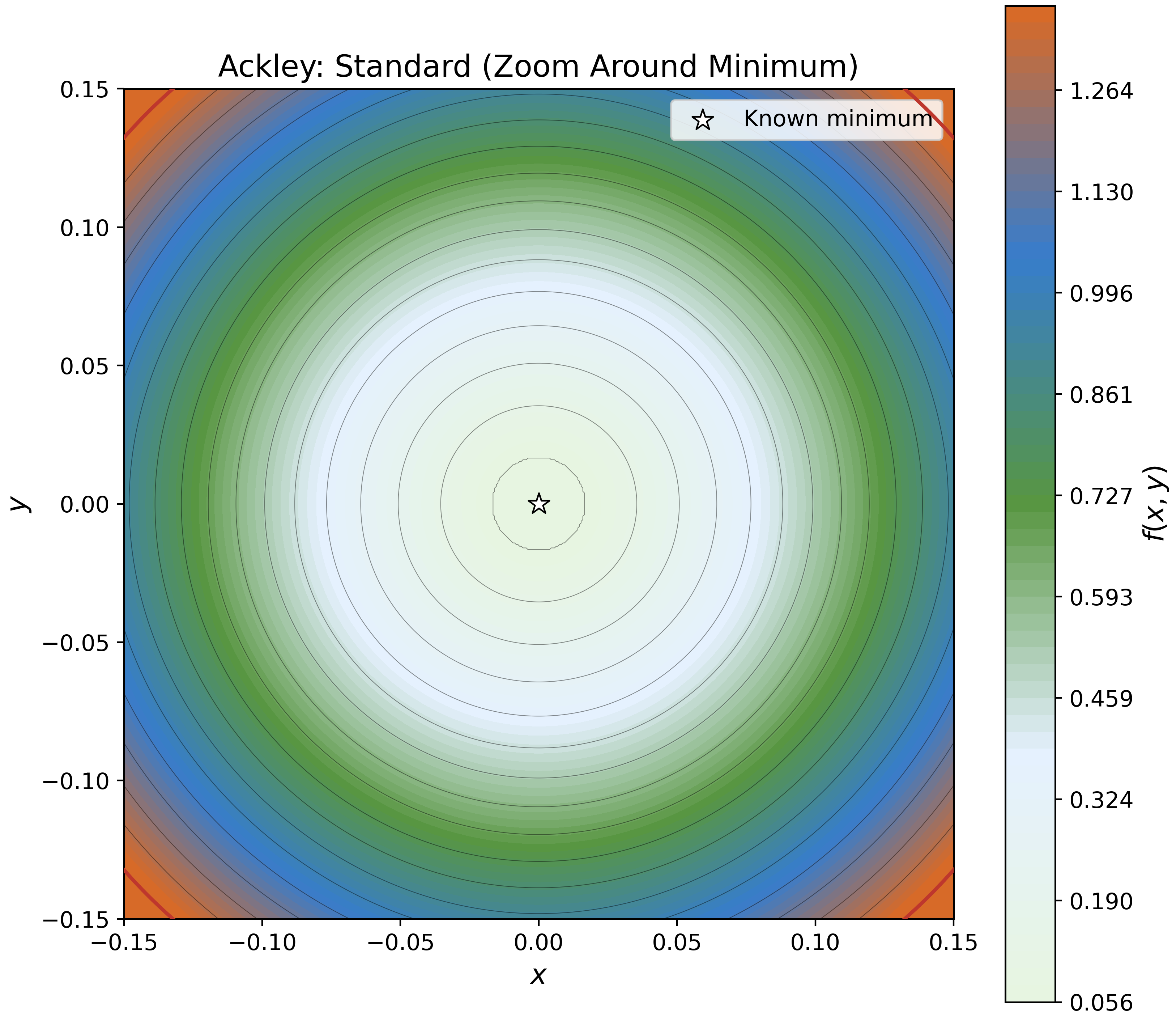}
    \end{minipage}
    \hfill
    \begin{minipage}{0.32\textwidth}
        \centering
        \includegraphics[width=\linewidth]{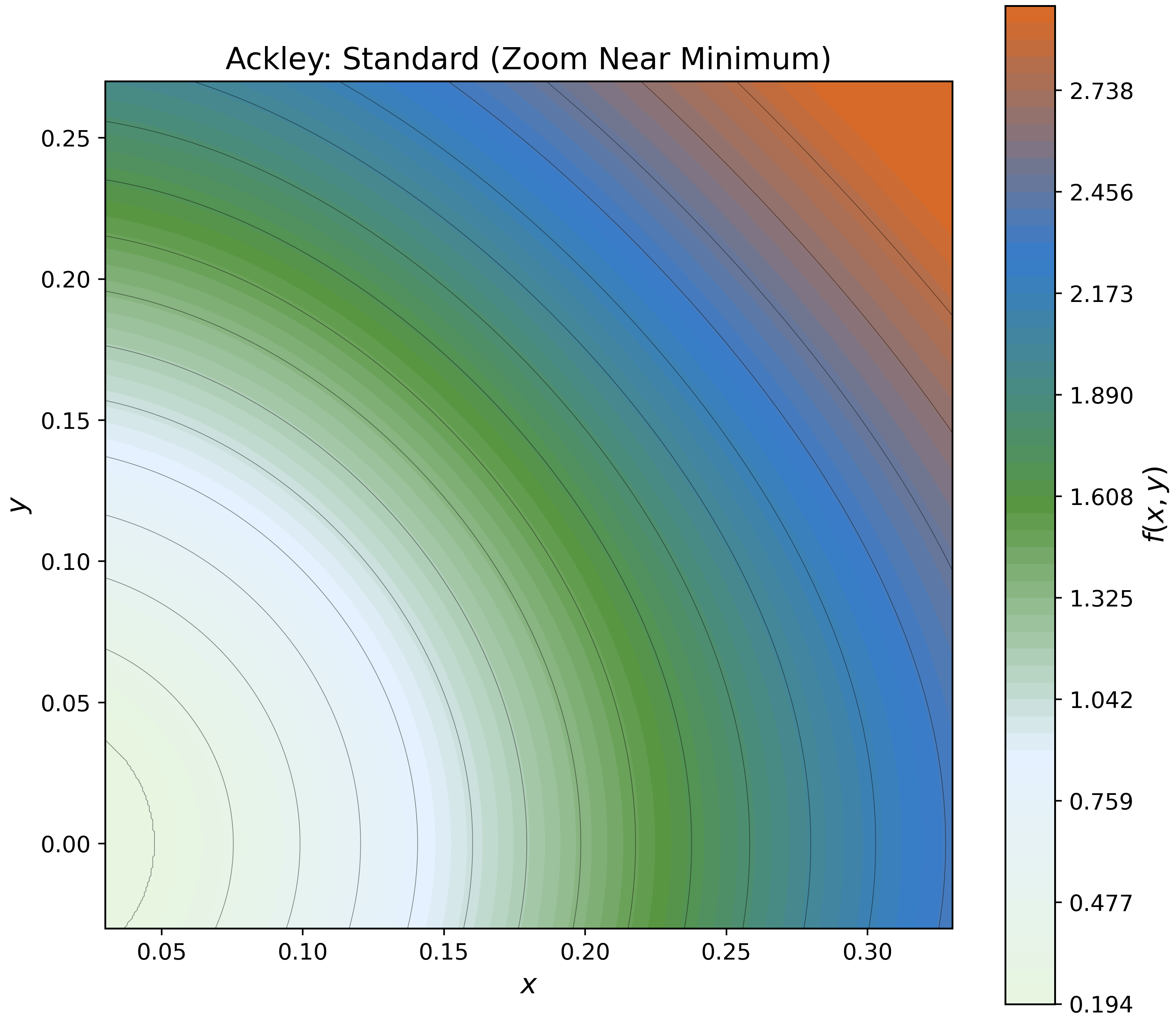}
    \end{minipage}

    \vspace{0.5em}

    \begin{minipage}{0.32\textwidth}
        \centering
        \includegraphics[width=\linewidth]{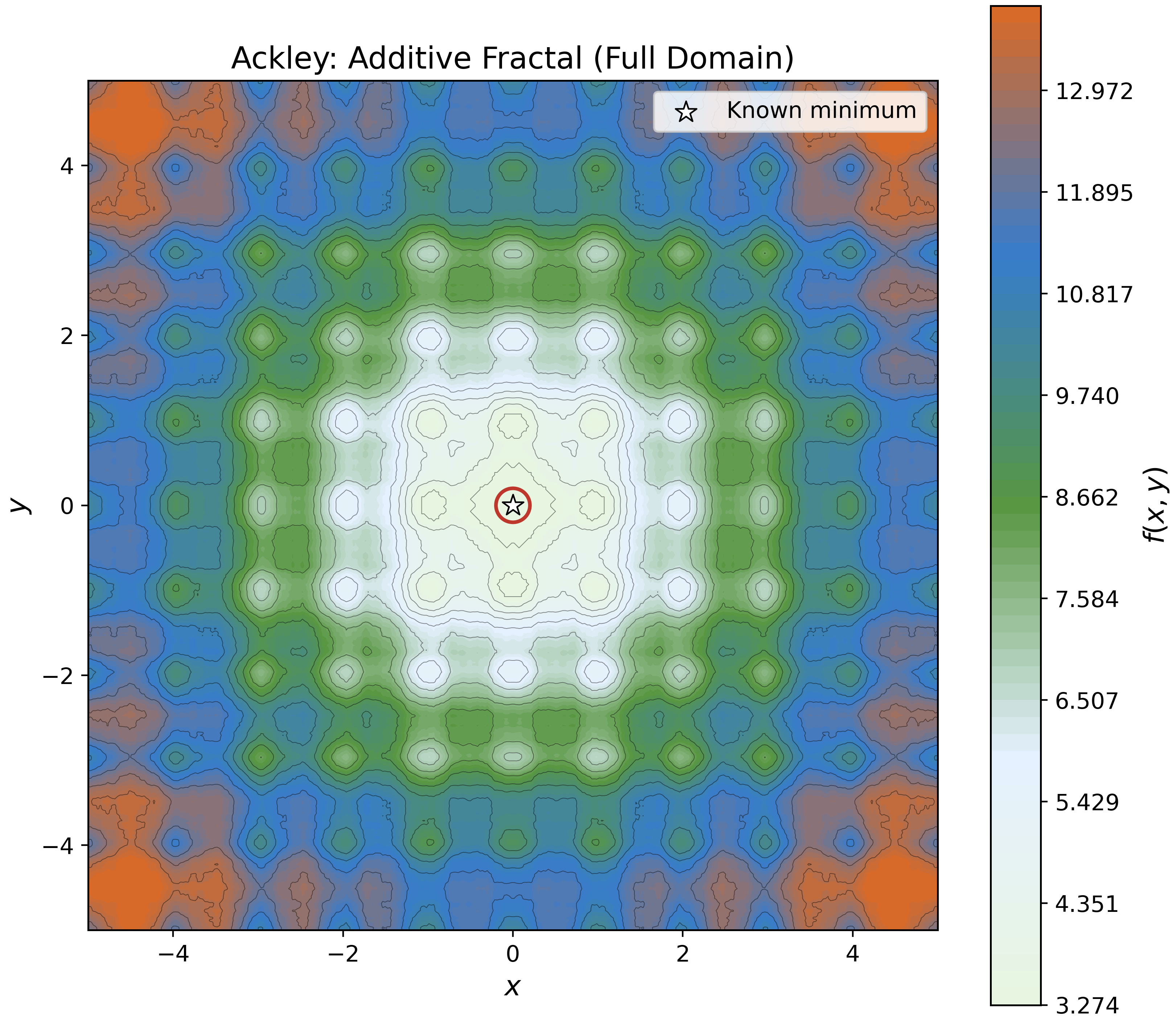}
    \end{minipage}
    \hfill
    \begin{minipage}{0.32\textwidth}
        \centering
        \includegraphics[width=\linewidth]{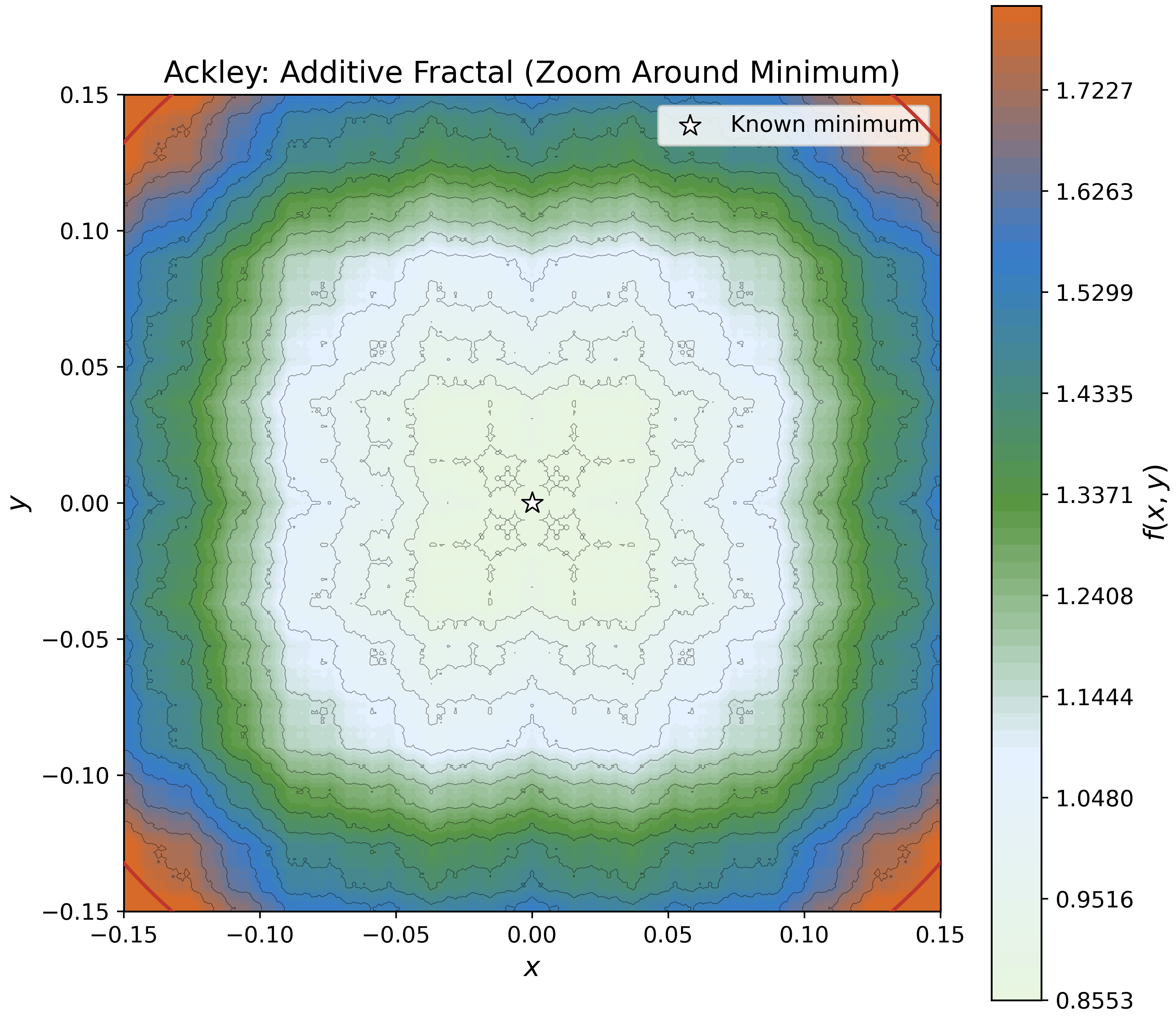}
    \end{minipage}
    \hfill
    \begin{minipage}{0.32\textwidth}
        \centering
        \includegraphics[width=\linewidth]{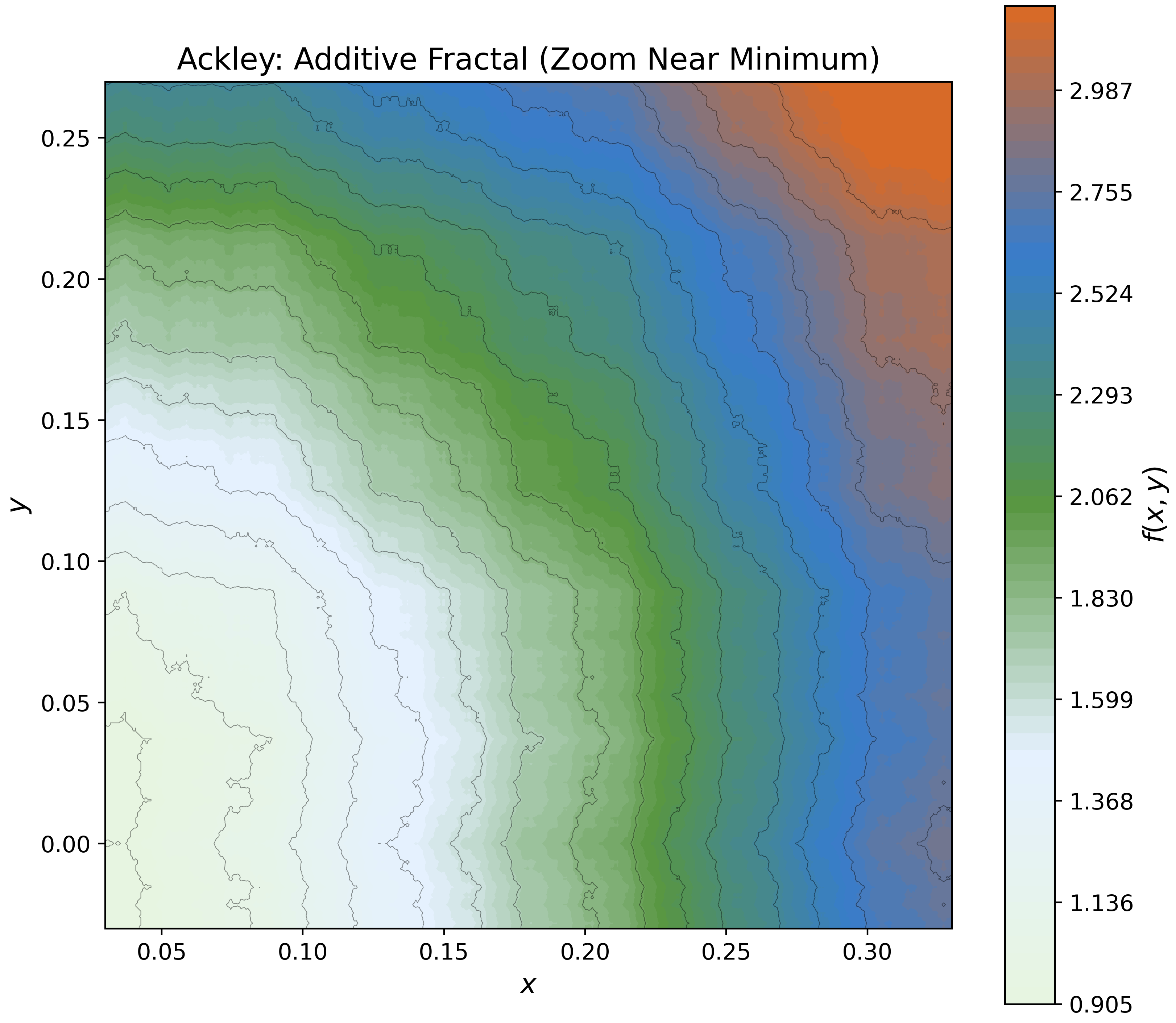}
    \end{minipage}

    \caption{Comparison of the standard Ackley surface with additive fractal perturbations.}
    \label{fig:ackley_standard_additive}
\end{figure}

We evaluate the optimizers on two standard minimization surfaces: Ackley and Himmelblau. The two surfaces expose different types of difficulty. Ackley has a broad, almost flat outer region with small oscillations and a single narrow central minimum, so progress depends on whether an optimizer can traverse the outer plateau. Himmelblau has four equivalent global minima in well-formed basins, so the task is mainly to descend reliably into one of several attainable targets. Together, the two surfaces cover one hard-to-enter landscape and one comparatively tractable landscape, which is sufficient to separate the optimizer families before the neural-network experiments.
For the two-dimensional Ackley function we use
\begin{equation}
A(x,y)
=
-20 \exp\left(-0.2 \sqrt{0.5(x^2+y^2)}\right)
-
\exp\left(0.5\left[\cos(2\pi x)+\cos(2\pi y)\right]\right)
+20+e,
\end{equation}
with the standard global minimum $A(0,0)=0$ \cite{ackley1987connectionist,sfu_ackley}. The Himmelblau function is
\begin{equation}
H(x,y)
=
(x^2+y-11)^2+(x+y^2-7)^2,
\end{equation}
with four known global minima at approximately $(3,2)$, $(-2.805118,3.131312)$,\\ $(-3.779310,-3.283186)$, and $(3.584428,-1.848126)$ \cite{himmelblau1972applied,jamil2013survey}.
To introduce controlled fractal structure, each base surface $f(x,y)$ is evaluated together with a perturbed variant. The perturbation term is
\begin{equation}
P_J(x,y)
=
\sum_{k=0}^{J} a_P^{\,k}
\left[
\cos(b_P^{\,k} \pi x)+\cos(b_P^{\,k} \pi y)
\right],
\end{equation}
where $a_P$ controls amplitude decay, $b_P$ controls frequency growth, and $J$ determines the number of scales. The subscript $P$ and the scale index $J$ are used deliberately: the surface perturbation shares the geometric-ladder form of the activation functions and of the optimizer kernel, but its parameters are chosen independently, so they are kept notationally distinct from the amplitude and frequency parameters $a$ and $b$ of the activation ladders and from the memory length $K$ of the fractional optimizers. This term adds repeated oscillatory structure at several resolutions and is the two-dimensional analogue of the Weierstrass-type ladders used elsewhere in this article. The additive variant is
\begin{equation}
f_{\mathrm{add}}(x,y)
=
f(x,y)+\alpha_{\mathrm{add}}P_J(x,y),
\end{equation}
where $\alpha_{\mathrm{add}}>0$ weights the perturbation against the base objective. Both surfaces use the same weight $\alpha_{\mathrm{add}}=0.2$, so that the Ackley and Himmelblau additive variants are perturbed at an identical relative strength and remain directly comparable. The remaining perturbation parameters are $a_P=0.55$ and $J=8$ on both surfaces, with the frequency base set to $b_P=2.4$ on Ackley and $b_P=2.5$ on Himmelblau. In both cases the ladder is supercritical in the sense of Section~\ref{sec:fractal_activations}, since $a_Pb_P>1$ and the corresponding roughness exponent $\gamma_H=-\ln a_P/\ln b_P$ is below one ($\gamma_H\approx0.68$ for Ackley and $\gamma_H\approx0.65$ for Himmelblau), so the perturbation carries genuine multi-scale roughness rather than a smooth oscillatory correction. 
It changes the objective by adding local oscillations directly to the base loss, which makes the surface less smooth while keeping the construction explicit and reproducible. Each surface is therefore evaluated in two variants: the standard surface without perturbation and the additive fractal variant.
For each surface and each variant, the same set of 21 optimizers is compared: standard optimizers, Herrera-style fractional optimizers, memory-based fractional optimizers, adaptive memory-based fractional optimizers, and related fractional optimization methods from the literature. Each optimizer is run for a fixed number of steps from controlled starting points. In each repeated run, a reference minimum is selected, and each optimizer starts at the same fixed radius from this minimum but with an optimizer-specific angle. This gives comparable initial difficulty while avoiding identical trajectories. Each surface variant is evaluated over 40 repeated runs per optimizer, giving $21 \times 2 \times 40 = 1680$ individual optimizer runs per surface. The evaluation records final loss, best loss reached during the trajectory, distance to the nearest known minimum, success rate under a fixed distance threshold, number of optimization steps, and runtime.
The Himmelblau surface shows the expected four-basin structure in the standard case, with smooth contour lines around the known minima and gradually increasing function values away from the basins, Figure~\ref{fig:himmelblau_standard_additive}. The additive fractal perturbation preserves the global layout of the Himmelblau landscape, but it introduces small-scale oscillatory structure directly into the objective values. This is most visible in the zoom around the minimum, where the basin remains centred near the known minimum but the contour lines become irregular and locally fragmented. The perturbation therefore changes the local basin geometry directly, without moving the basins themselves. Similar plots for the Ackley surface are collected in Figure~\ref{fig:ackley_standard_additive}.

\subsection{Surface Optimization Results}
\label{subsec:surface_results}

This subsection reports the aggregate results of the surface optimization experiments. Each surface was evaluated in two variants: the standard surface without fractal perturbation and the additive fractal perturbation. Each configuration was evaluated over 40 repeated runs for each of the 21 optimizers. The main reported quantities are the mean final loss, the mean best loss observed during optimization, the success rate, the mean distance to the nearest target minimum, and the mean runtime.

\begin{table}[H]
\centering
\caption{Best optimizer per surface and perturbation variant, ranked by mean final loss over 40 runs. The standard variant corresponds to the original surface without fractal perturbation.}
\label{tab:surface_actual_winners}
\small
\begin{tabular}{lllrrrr}
\hline
Surface & Variant & Best optimizer & Final loss & Best loss & Success rate & Runtime (s) \\
\hline
Ackley      & Standard & FSGD         & 5.842 & 5.838 & 0.15  & 0.850 \\
Ackley      & Additive & FCSGD\_GL    & 4.591 & 1.209 & 0.08  & 3.017 \\
\hline
Himmelblau  & Standard & FCSGD\_GL    & 0.000 & 0.000 & 1.00  & 1.582 \\
Himmelblau  & Additive & AOFGD\_Adam  & 1.978 & 1.988 & 0.73  & 2.671 \\
\hline
\end{tabular}
\end{table}

The two surfaces separate clearly in difficulty. Himmelblau was largely tractable: the standard surface was solved completely by FCSGD\_GL with a mean final loss of zero and a success rate of 1.00, and even the additive variant was solved in the majority of runs by the best method. Ackley was difficult in both variants: no optimizer exceeded a success rate of 0.15, and the mean final losses of the best methods remained between 4.6 and 5.8. This reflects the flat outer region of the Ackley surface, in which gradient information is weak and most trajectories terminate before reaching the central basin.

On the standard Himmelblau surface, four optimizers solved every run: FCSGD\_GL, MemoryFSGD, AdaptiveMemoryFSGD, and SGD all reached a success rate of 1.00, followed by AOFGD\_SGD with 0.88 and AOFGD\_Adam with 0.83. The additive Himmelblau variant reduced the success rates of all methods but preserved the overall ordering of the families: AOFGD\_Adam was the best method with a success rate of 0.73 and a mean final loss of 1.98, followed by FCSGD\_GL and MemoryFSGD with 0.65 each. The additive perturbation therefore made the task harder without changing which optimizer families are competitive. It is also visible that the Adam-based methods gained ground on the perturbed surface: AOFGD\_Adam, AdaptiveMemoryFAdam, and Adam all ranked among the top five by final loss on the additive variant, whereas the standard variant was dominated by SGD-type methods.

On Ackley, the picture is different. The standard surface was led by FSGD with a success rate of 0.15, ahead of SGD and AOFGD\_SGD with 0.08 each; all remaining methods reached the target in at most one run out of 40. On the additive variant, FCSGD\_GL obtained the lowest mean final loss (4.59) and, together with the memory-based methods, also the lowest mean best loss (1.21): the additive oscillations create local regions of low objective value that the stochastically perturbed and memory-based SGD variants exploit during the trajectory. The highest success rate on the additive variant was reached by MemoryFSGD with 0.13, although it ranked second by mean final loss. Final loss and success rate are therefore not equivalent rankings, and both are reported. The negative mean best losses of several methods on the additive variant (for example $-0.10$ for FCSGD\_GL on Himmelblau additive) are expected, because the additive perturbation can shift the objective below the zero level of the unperturbed base function.

\begin{table}[H]
\centering
\caption{Optimizers that reached the target at least once. The entries report the number of optimizers with a nonzero success rate and the five highest success rates for each surface variant, over 40 runs.}
\label{tab:surface_actual_success}
\small
\begin{tabular}{lllp{5.2cm}}
\hline
Surface & Variant & Successful optimizers & Highest success rates \\
\hline
Ackley & Standard & 7 & FSGD (0.15), SGD (0.08), AOFGD\_SGD (0.08), FCSGD\_GL (0.03), MemoryFSGD (0.03) \\
Ackley & Additive & 6 & MemoryFSGD (0.13), FCSGD\_GL (0.08), FSGD (0.08), AOFGD\_SGD (0.05), AdaptiveMemoryFSGD (0.03) \\
\hline
Himmelblau & Standard & 17 & FCSGD\_GL (1.00), MemoryFSGD (1.00), AdaptiveMemoryFSGD (1.00), SGD (1.00), AOFGD\_SGD (0.88) \\
Himmelblau & Additive & 14 & AOFGD\_Adam (0.73), FCSGD\_GL (0.65), MemoryFSGD (0.65), AdaptiveMemoryFAdam (0.58), Adam (0.58) \\
\hline
\end{tabular}
\end{table}
Table~\ref{tab:surface_actual_success} shows that reaching the target was not evenly distributed across surfaces or optimizer families. On Himmelblau, most SGD-like and Adam-like methods reached the target regularly, while the RMSprop- and Adadelta-based methods rarely did: on the standard surface, the best RMSprop-type result was FRMSprop with 0.10, and no Adadelta-type method exceeded 0.05 on either variant. On Ackley, success was confined to the SGD family in both variants; no RMSprop-, Adam-, or Adadelta-based method except MemoryFAdadelta (one successful run) reached the target at all. Across all four configurations, the leading positions were occupied by FCSGD\_GL, MemoryFSGD, AdaptiveMemoryFSGD, FSGD, AOFGD\_SGD, and, on the additive Himmelblau variant, AOFGD\_Adam. Fractional and memory-based SGD variants therefore consistently matched or exceeded their classical counterparts, whereas the corresponding RMSprop and Adadelta variants inherited the weakness of their base optimizers.

\begin{figure}[H]
    \centering

    \begin{minipage}{0.49\textwidth}
        \centering
        \includegraphics[width=\linewidth]{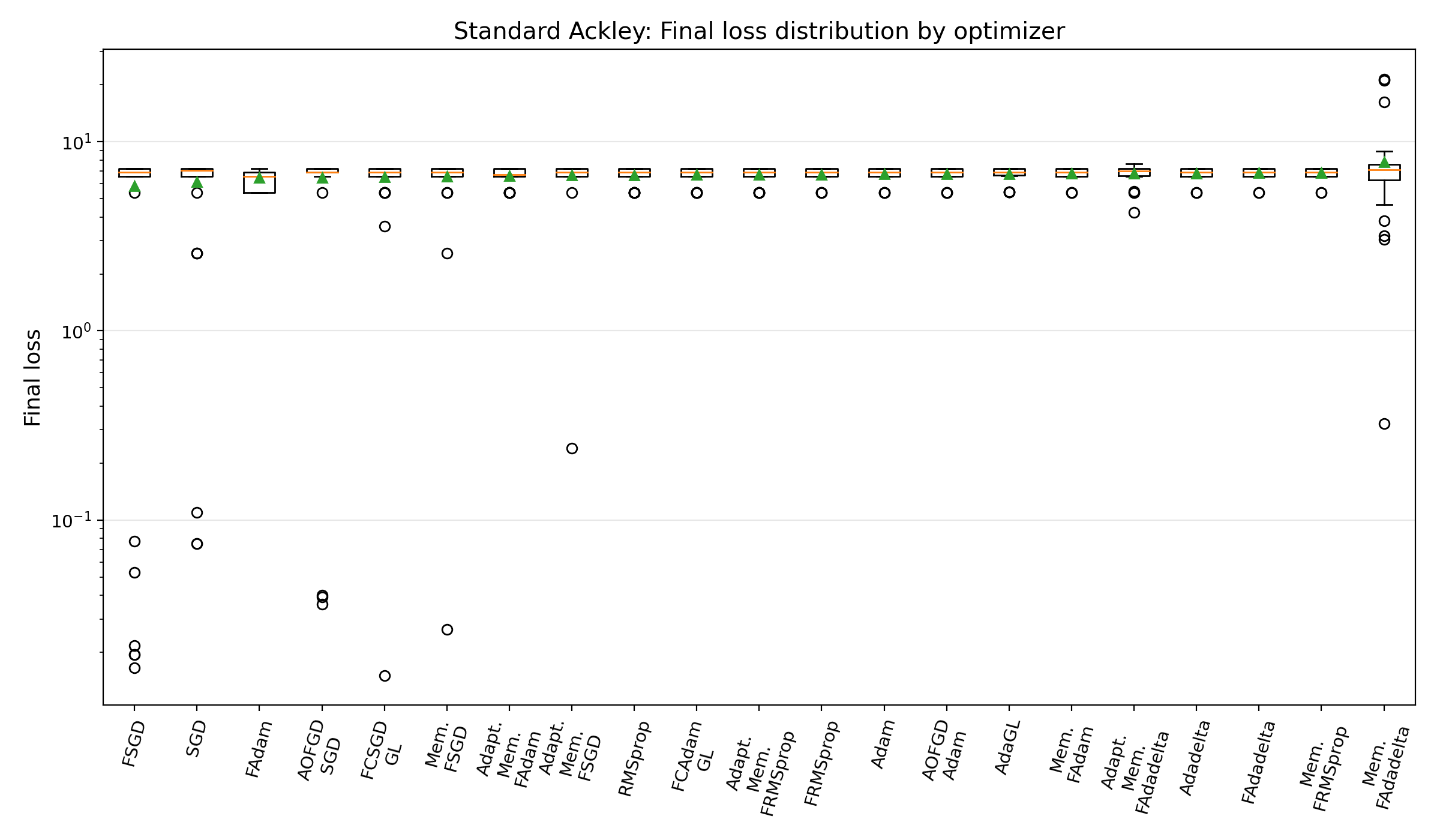}
    \end{minipage}
    \begin{minipage}{0.49\textwidth}
        \centering
        \includegraphics[width=\linewidth]{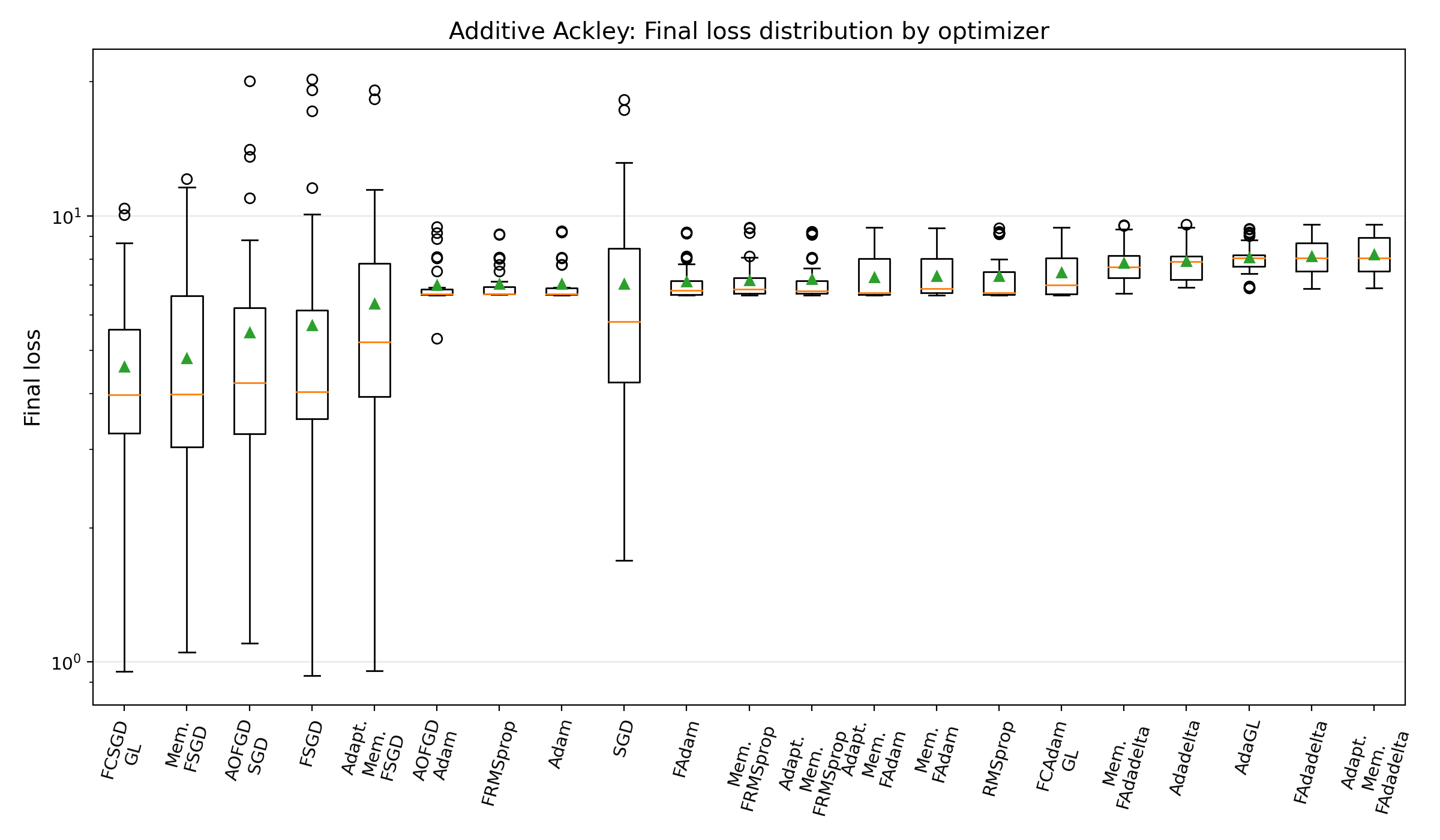}
    \end{minipage}

    \caption{Final-loss distributions on the Ackley surface over 40 runs per optimizer. Left: standard Ackley surface. Right: additive fractal Ackley variant. The final losses are shown on a logarithmic scale and the optimizers are ordered by mean final loss. On the standard surface, almost all runs terminate on the outer plateau at a final loss near 7, and only the SGD family reaches the central basin in isolated runs. On the additive variant, the final-loss distributions of the SGD-type methods widen towards lower values, which reflects the locally lowered objective regions created by the perturbation.}
    \label{fig:ackley_final_loss_boxplots}
\end{figure}

\begin{figure}[H]
    \centering

    \begin{minipage}{0.49\textwidth}
        \centering
        \includegraphics[width=\linewidth]{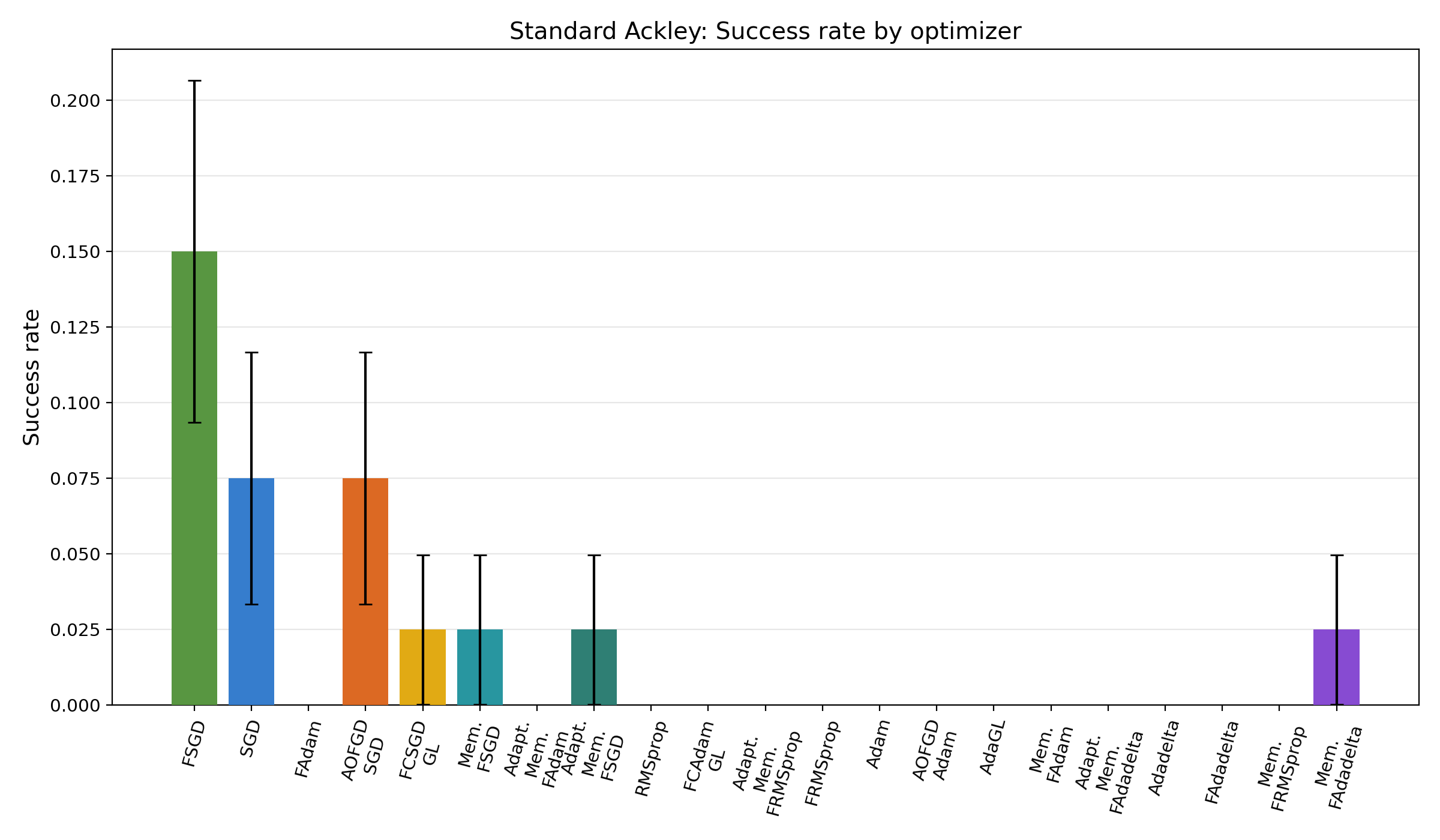}
    \end{minipage}
    \begin{minipage}{0.49\textwidth}
        \centering
        \includegraphics[width=\linewidth]{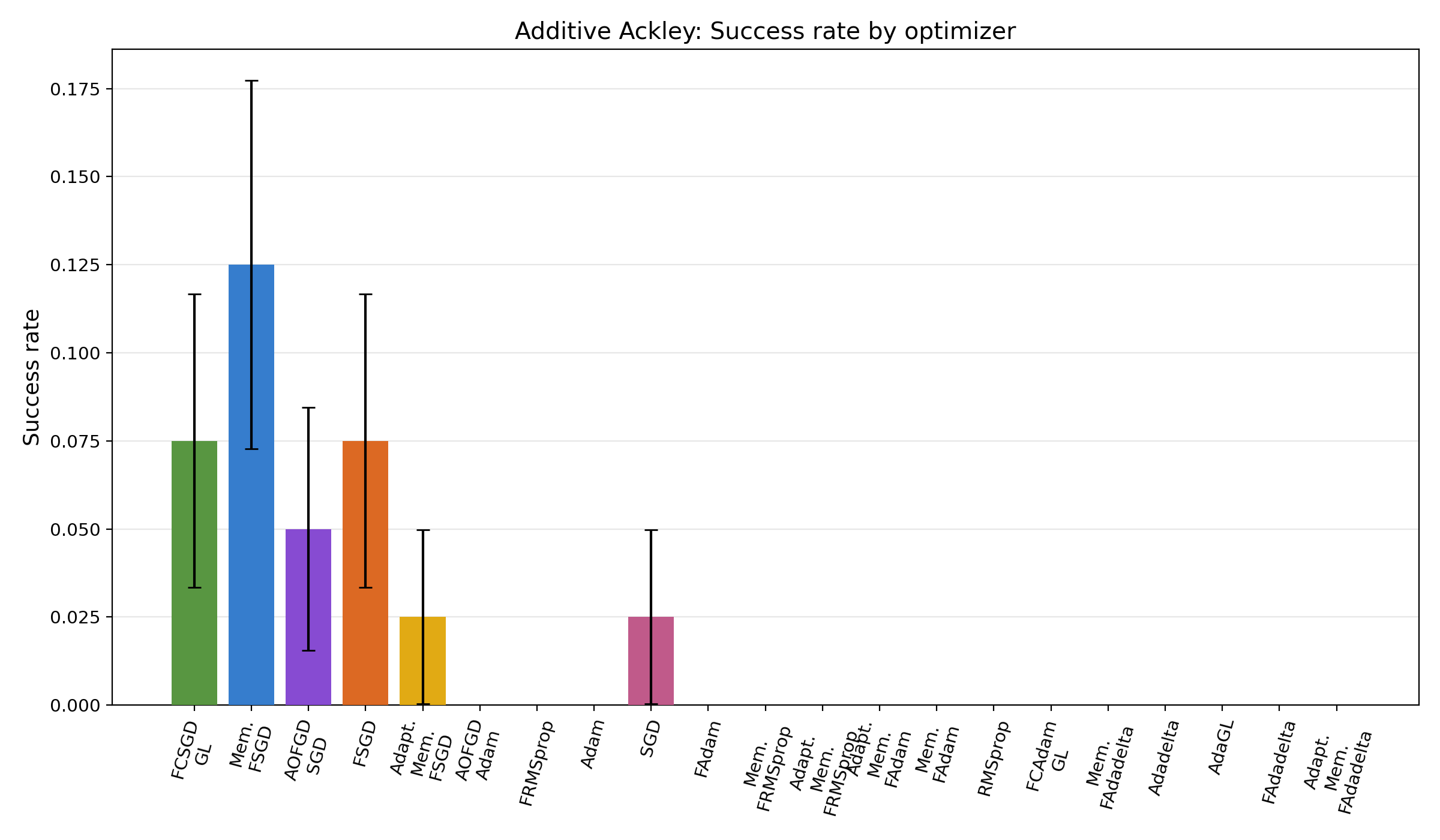}
    \end{minipage}
   
    \caption{Success rates on the Ackley surface over 40 runs per optimizer. Left: standard Ackley surface. Right: additive fractal Ackley variant. The bars show the fraction of runs that reached the target, with 95\% confidence intervals. Success on Ackley is confined almost entirely to the SGD family in both variants, while RMSprop-, Adam-, and Adadelta-based methods largely fail to reach the target.}
    \label{fig:ackley_success_rate_bars}
\end{figure}

\begin{figure}[H]
    \centering

    \begin{minipage}{0.49\textwidth}
        \centering
        \includegraphics[width=\linewidth]{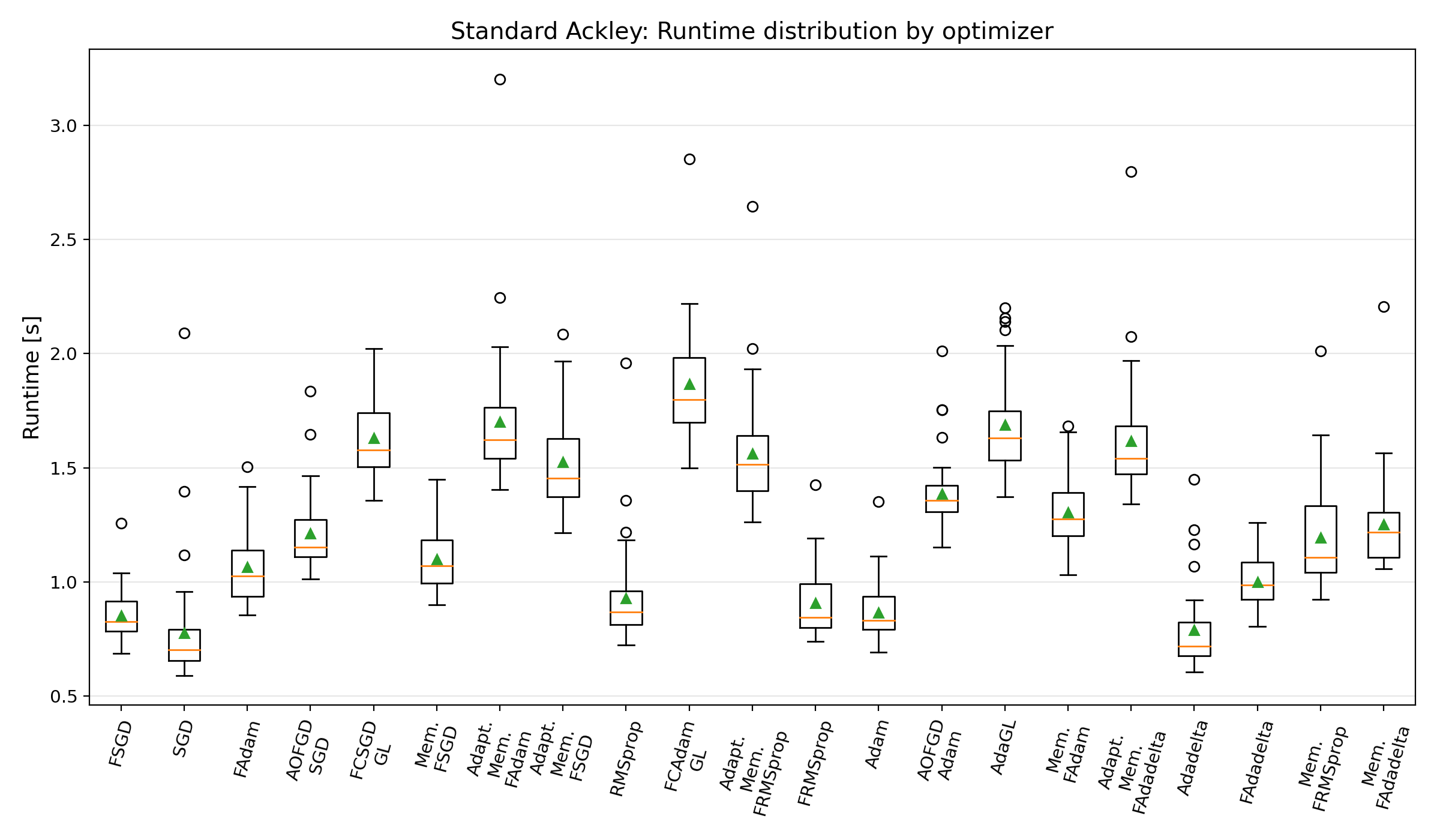}
    \end{minipage}
    \begin{minipage}{0.49\textwidth}
        \centering
        \includegraphics[width=\linewidth]{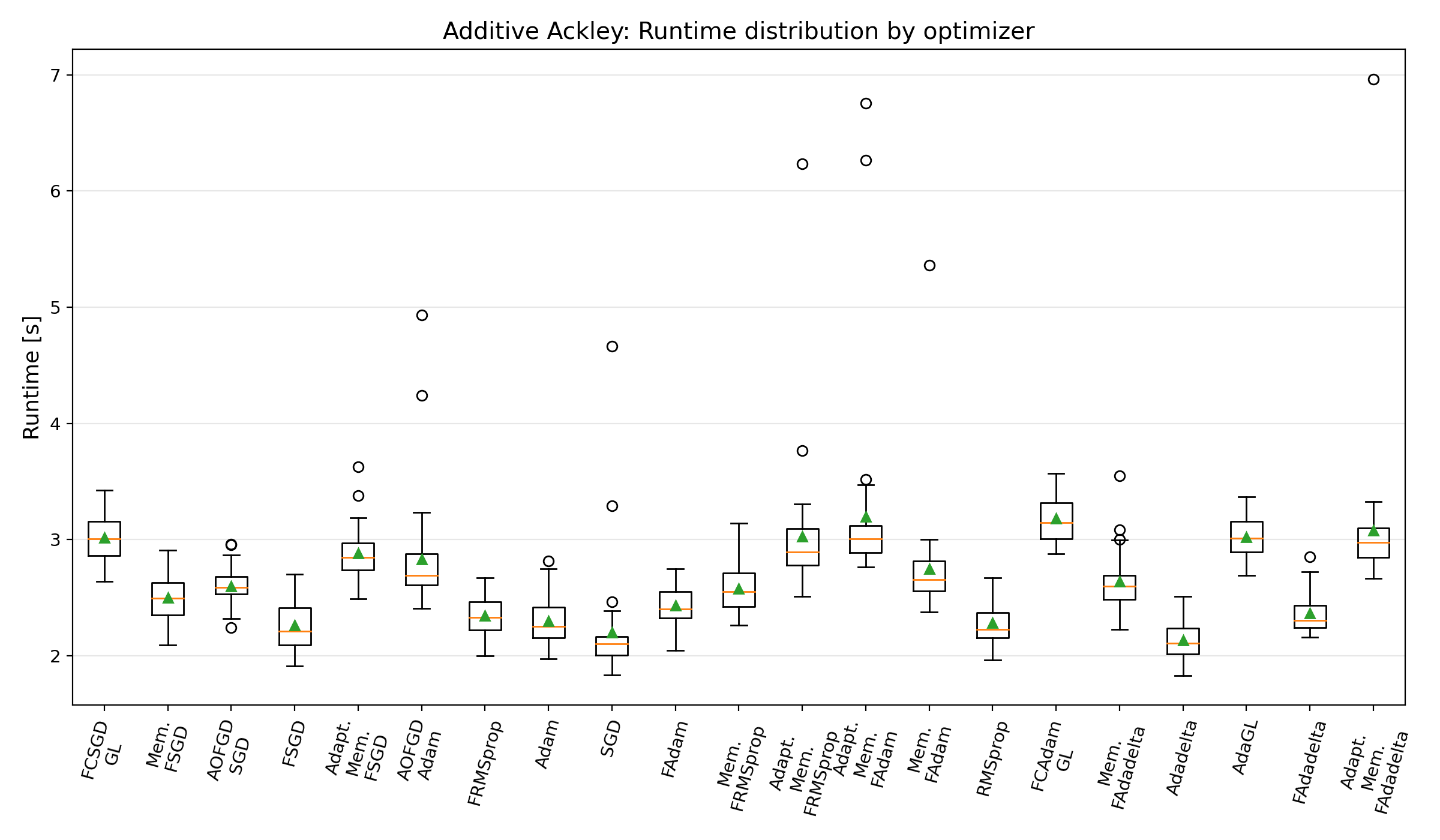}
    \end{minipage}

    \caption{Runtime distributions on the Ackley surface over 40 runs per optimizer. Left: standard Ackley surface. Right: additive fractal Ackley variant. The boxplots show the runtime per run for each optimizer under the same experimental setting. These runtime distributions complement the final-loss and success-rate plots by showing that the better-performing SGD-type methods do not owe their advantage to a fundamentally different evaluation setting.}
    \label{fig:ackley_runtime_boxplots}
\end{figure}

\begin{figure}[H]
    \centering

    \begin{minipage}{0.49\textwidth}
        \centering
        \includegraphics[width=\linewidth]{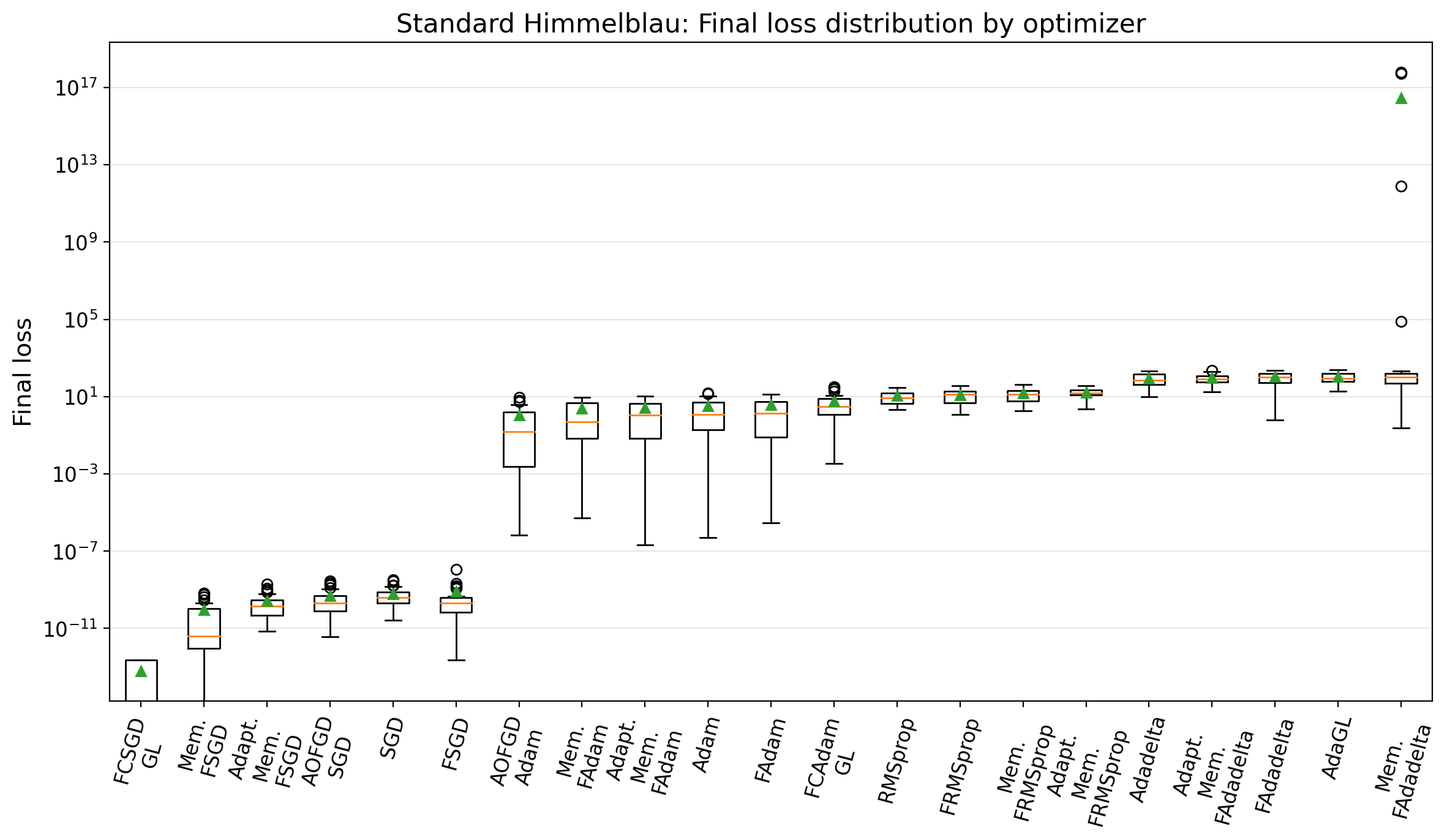}
    \end{minipage}
    \begin{minipage}{0.49\textwidth}
        \centering
        \includegraphics[width=\linewidth]{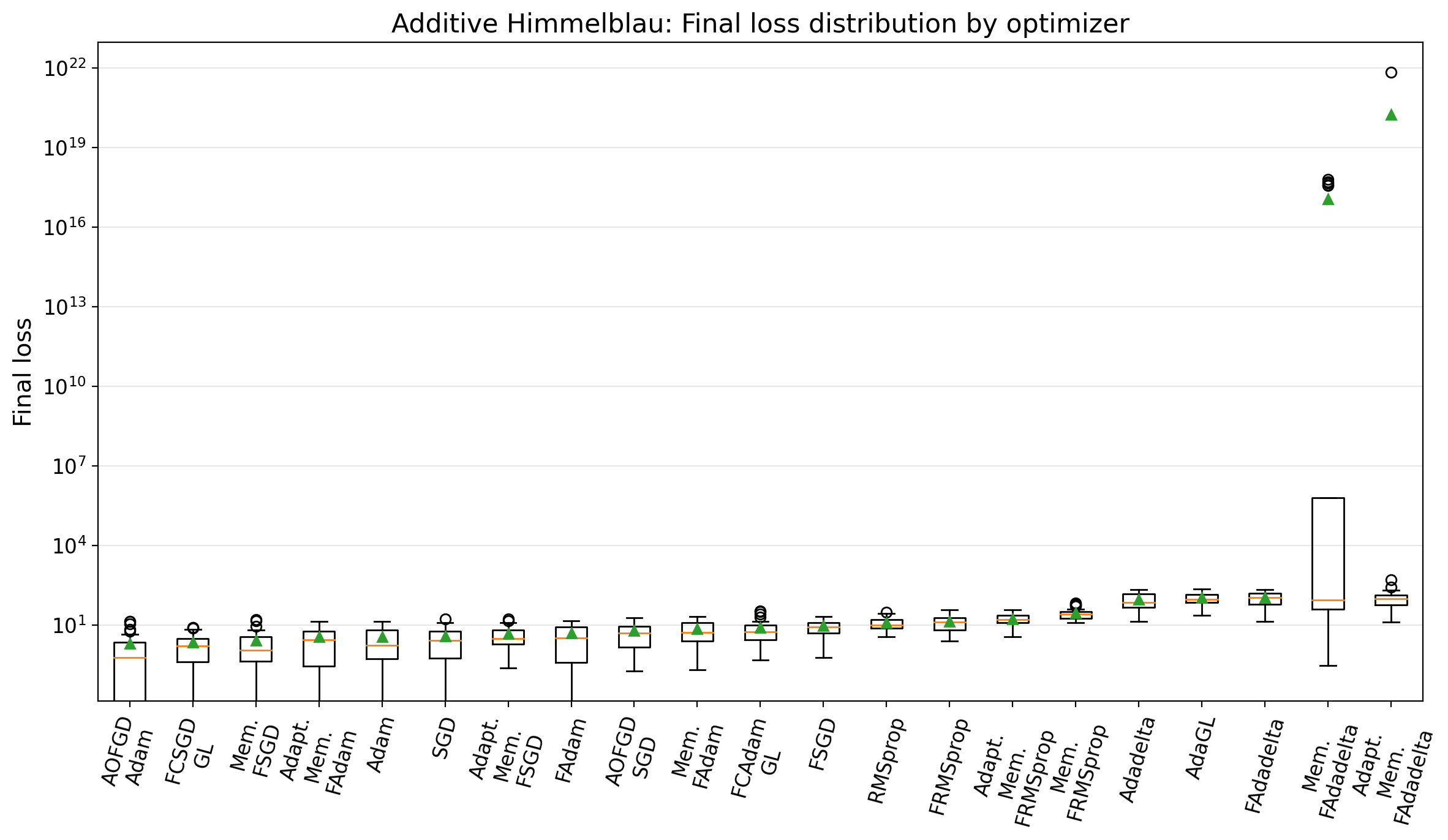}
    \end{minipage}

    \caption{Final-loss distributions on the Himmelblau surface over 40 runs per optimizer. Left: standard Himmelblau surface. Right: additive fractal Himmelblau variant. The final losses are shown on a logarithmic scale and the optimizers are ordered by mean final loss. On the standard surface, the SGD family converges to final losses around $10^{-10}$ and below, the Adam family reaches intermediate values, and the RMSprop and Adadelta families remain near or above 10. The additive variant compresses these differences but keeps the ordering of the families.}
    \label{fig:himmelblau_final_loss_boxplots}
\end{figure}

\begin{figure}[H]
    \centering

    \begin{minipage}{0.49\textwidth}
        \centering
        \includegraphics[width=\linewidth]{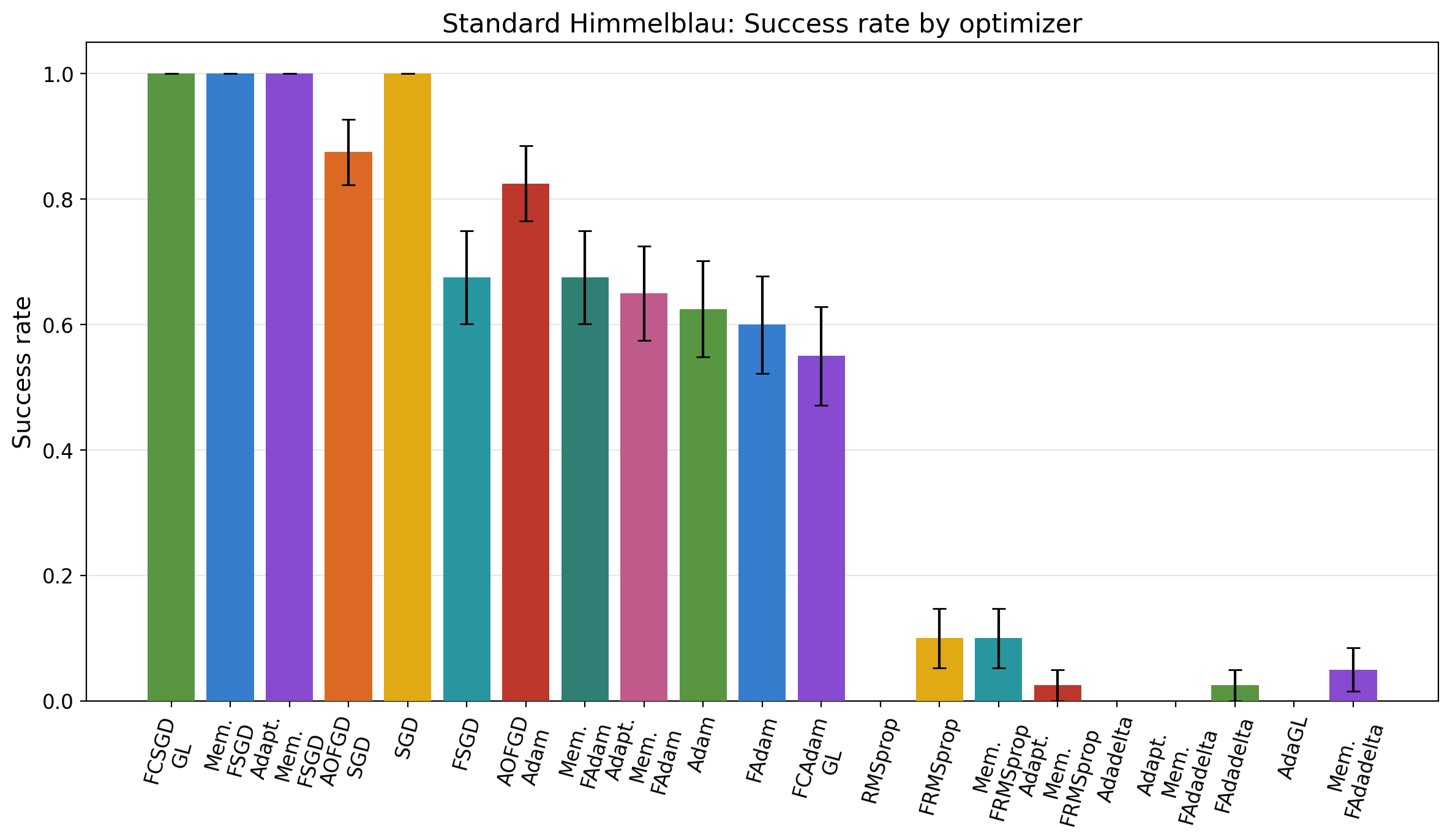}
    \end{minipage}
    \begin{minipage}{0.49\textwidth}
        \centering
        \includegraphics[width=\linewidth]{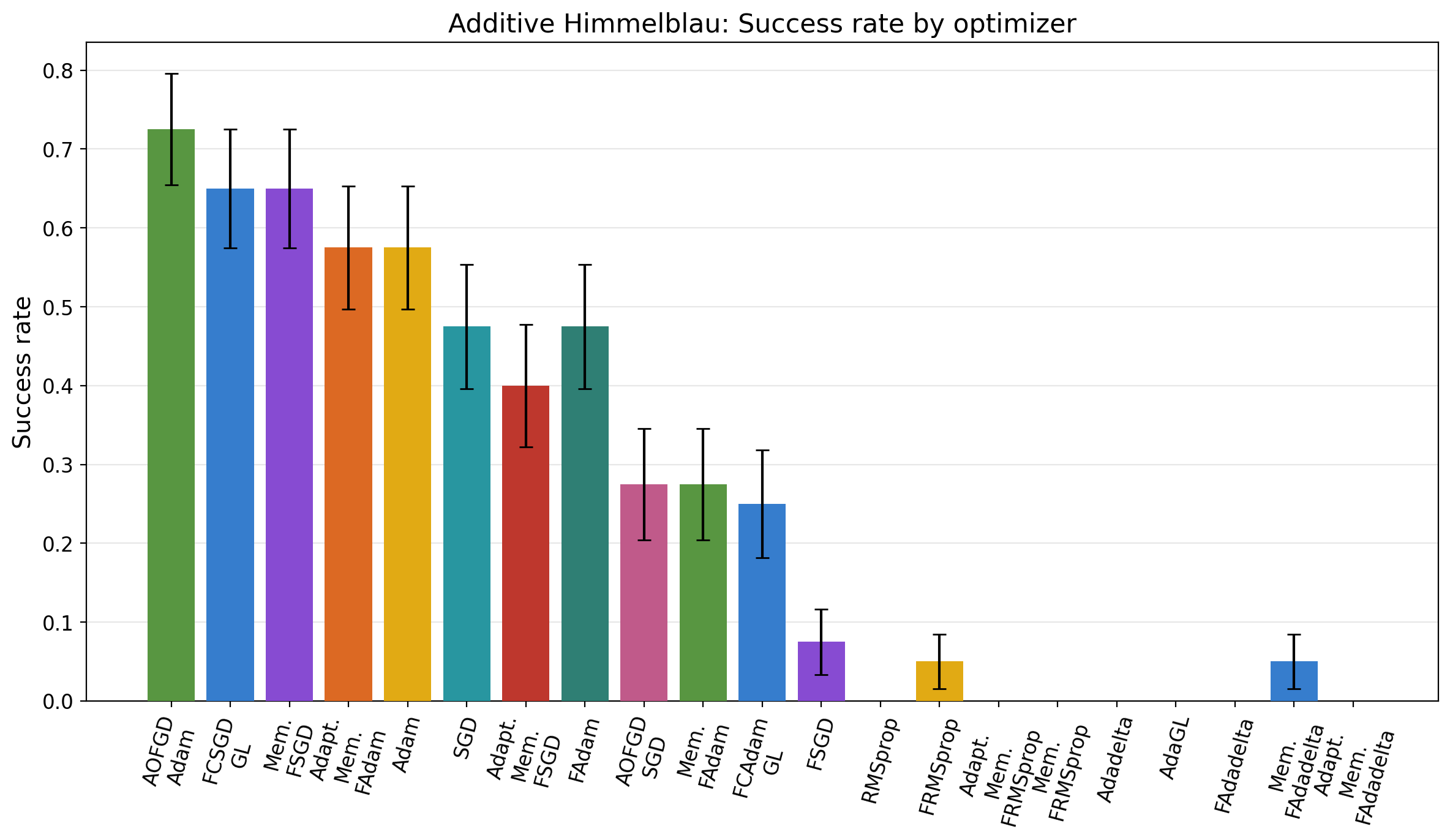}
    \end{minipage}

    \caption{Success rates on the Himmelblau surface over 40 runs per optimizer. Left: standard Himmelblau surface. Right: additive fractal Himmelblau variant. The bars show the fraction of runs that reached the target, with 95\% confidence intervals. Most SGD-like and Adam-like methods reach the target regularly, while RMSprop- and Adadelta-based methods rarely do. The additive variant preserves the broad family-level separation, while slightly changing the relative success rates of individual optimizers.}
    \label{fig:himmelblau_success_rate_bars}
\end{figure}

\begin{figure}[H]
    \centering

    \begin{minipage}{0.49\textwidth}
        \centering
        \includegraphics[width=\linewidth]{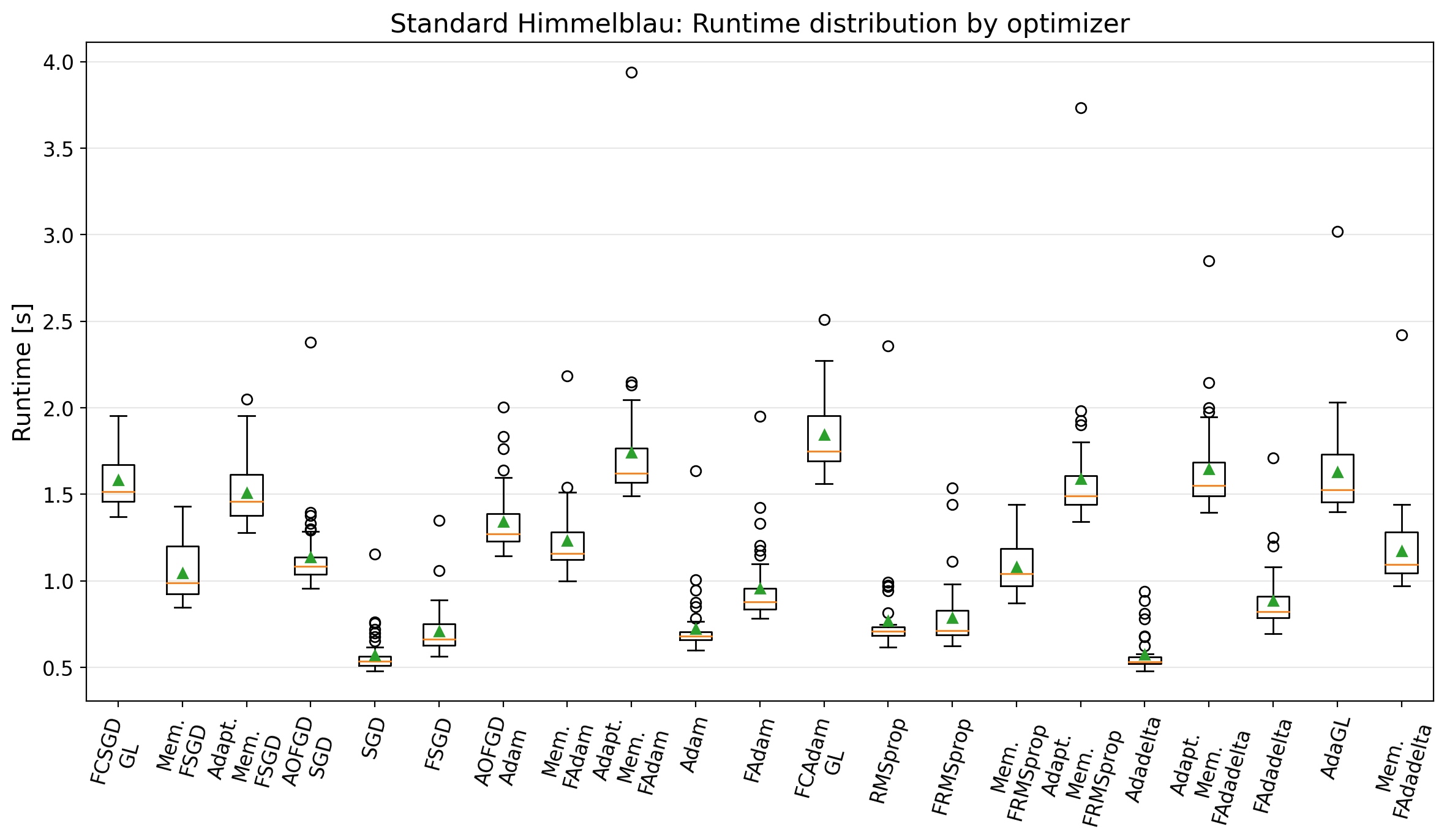}
    \end{minipage}
    \begin{minipage}{0.49\textwidth}
        \centering
        \includegraphics[width=\linewidth]{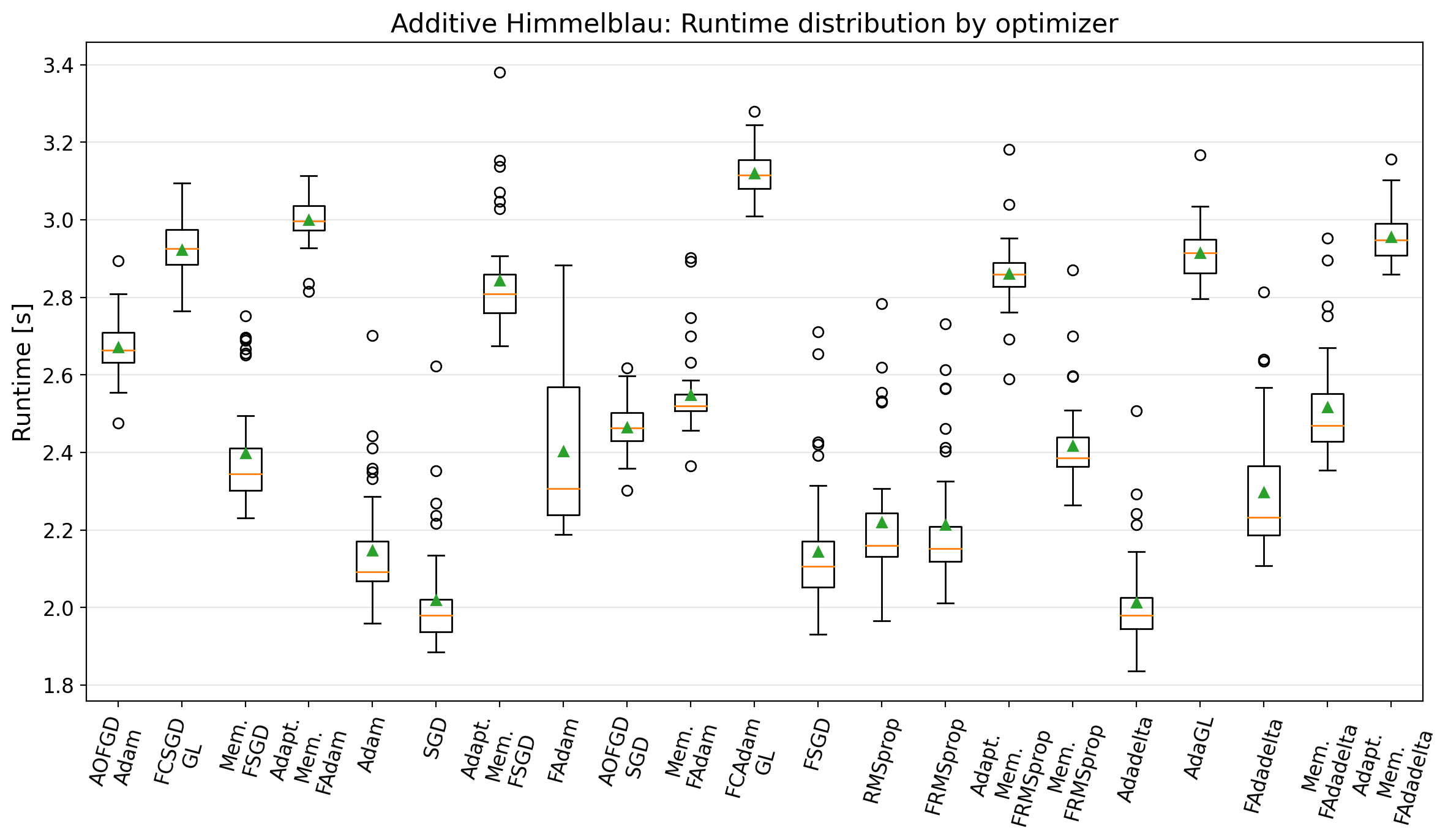}
    \end{minipage}

    \caption{Runtime distributions on the Himmelblau surface over 40 runs per optimizer. Left: standard Himmelblau surface. Right: additive fractal Himmelblau variant. The boxplots show the runtime per run for each optimizer under the same experimental setting. Together with the final-loss and success-rate plots, these runtime distributions show that the observed differences are mainly optimization effects rather than artifacts of different computational budgets.}
    \label{fig:himmelblau_runtime_boxplots}
\end{figure}

Figures~\ref{fig:ackley_final_loss_boxplots}, \ref{fig:ackley_success_rate_bars}, and~\ref{fig:ackley_runtime_boxplots} show the Ackley distributions behind the aggregate numbers: final loss per run, success rate with confidence intervals, and runtime per run, for both the standard and additive variants. Figures~\ref{fig:himmelblau_final_loss_boxplots}, \ref{fig:himmelblau_success_rate_bars}, and~\ref{fig:himmelblau_runtime_boxplots} show the corresponding results for Himmelblau. The panels are ordered by mean final loss.

The distribution plots also expose an effect that the mean values alone would misrepresent. On Himmelblau, the mean final losses of MemoryFAdadelta and AdaptiveMemoryFAdadelta are extremely large (of the order $10^{16}$ and above on both variants), although their median final losses are moderate. A small number of runs of these two optimizers diverged, and the diverged runs dominate the mean. The same optimizers behave unremarkably on Ackley, where the bounded structure of the surface limits the damage of an unstable trajectory. This instability of the Adadelta-based fractional variants on Himmelblau is a genuine finding of the surface experiments, and it is the reason why the tables report distributions and success rates in addition to means.

\begin{table}[H]
\centering
\caption{Fastest optimizer per surface variant. Runtime is the mean runtime over 40 repeated runs; the last column reports the runtime of the best-performing optimizer from Table~\ref{tab:surface_actual_winners} for comparison.}
\label{tab:surface_actual_fastest}
\small
\begin{tabular}{lllrr}
\hline
Surface & Variant & Fastest optimizer & Runtime (s) & Best optimizer runtime (s) \\
\hline
Ackley      & Standard & SGD      & 0.775 & 0.850 \\
Ackley      & Additive & Adadelta & 2.132 & 3.017 \\
\hline
Himmelblau  & Standard & SGD      & 0.571 & 1.582 \\
Himmelblau  & Additive & Adadelta & 2.013 & 2.671 \\
\hline
\end{tabular}
\end{table}

The runtime results show a consistent pattern. SGD and Adadelta were the fastest methods on every variant, but they were usually not the best-performing methods. The best-performing optimizers required additional runtime because they use fractional updates, memory terms, or adaptive update mechanisms: measured per step, FCSGD\_GL needed roughly $1.9$--$2.8$ times the step time of SGD on the standard surfaces, and the memory- and adaptive-memory methods lay between these two extremes. Evaluating the additive variant is itself more expensive than evaluating the standard surface, because the perturbation sum $P_J$ is computed at every step; this adds an almost constant offset of roughly $5$--$6$\,ms per step to every optimizer, so it affects all methods equally in absolute terms. The additional runtime of the fractional and memory-based methods should therefore be read as the computational cost of more complex update rules, not of the perturbation.

Runtime alone is not sufficient to evaluate an optimizer. On the standard Himmelblau surface, SGD was both fast and fully successful, so the more expensive methods offer no advantage there. On the additive Himmelblau variant, however, the fastest methods were unreliable, and the best success rates required the more expensive AOFGD\_Adam, FCSGD\_GL, and MemoryFSGD updates. On Ackley, no method was reliable, and the moderate success-rate advantages of FSGD and MemoryFSGD came at step costs comparable to plain SGD (FSGD) or roughly 15--40\% above it (MemoryFSGD).

Overall, the surface experiments show that the usefulness of fractional and memory-based optimizers depends on the structure of the objective landscape and on the base optimizer they extend. On Himmelblau, fractional and memory-based SGD variants matched plain SGD on the standard surface and were among the most reliable methods once the additive fractal perturbation was introduced, with AOFGD\_Adam as the strongest method on the perturbed surface. On Ackley, only SGD-type methods made progress at all, and the fractional variants FSGD and FCSGD\_GL led both variants. The RMSprop- and Adadelta-based methods were weak throughout, independently of whether they were classical, fractional, or memory-based, and the Adadelta-based memory variants were additionally unstable on Himmelblau. These results support the use of the surface experiments as diagnostic tests: they reveal not only which optimizer has the lowest final loss, but also whether the optimizer actually reaches the intended target region, how consistently it does so across runs, and how much runtime this requires.

\subsection{Discussion and Summary}
\label{subsec:surface_discussion}

The surface experiments provide a controlled setting for studying optimizer behaviour. They are not meant to replace classification experiments, but they help explain why a method may behave differently once the loss geometry becomes more irregular. Because the minima of the base surfaces are known, the distance-to-minimum metric gives an interpretable complement to final loss. This is important for the perturbed surfaces, where a low objective value alone may not always mean that the optimizer moved toward the intended basin, and where the objective can locally fall below the zero level of the base function.

The additive fractal variant serves as a controlled stress test: it tests sensitivity to direct high-frequency loss perturbations while keeping the global basin layout intact. The comparison between the standard and additive variants of the same surface therefore isolates the effect of multi-scale roughness from the effect of the base geometry. On Himmelblau, this comparison shows that the perturbation reduces reliability for all methods but preserves the ranking of the optimizer families and favours the adaptive fractional Adam variant. On Ackley, it shows that the perturbation reshapes the reachable objective values without solving the underlying plateau problem.

The main limitation is that these are two-dimensional artificial surfaces. They are useful because trajectories can be inspected and the minima are known, but their behaviour cannot be transferred directly to neural network training. A second limitation is the fixed step budget and initialization protocol, which affects all optimizers equally but caps the attainable success rates on the harder Ackley surface. The results should therefore be interpreted as mechanistic evidence rather than final performance evidence. The later classification experiments are needed to test whether the same optimizer families also provide practical advantages on learned models.

Overall, these experiments establish the loss-surface part of the study. It compares the same 21 optimizers across a smooth multi-basin surface, a plateau-dominated surface, and their additively perturbed fractal counterparts, under repeated and reproducible starting conditions with 40 runs per configuration. The reported quantities cover mean performance, variability, success rate, and runtime rather than only the best observed run, which makes the comparison reliable and directly connectable to the neural-network experiments in the following section.

\begin{takeawaybox}[cyanA]{Main Takeaways --- Surface Optimization Experiments}
  \begin{itemize}[
    label={},
    leftmargin=0em,
    itemindent=0em,
    itemsep=3pt,
    topsep=2pt
  ]
    \item \textbf{The surfaces provide controlled optimizer diagnostics.}
          Ackley and Himmelblau are evaluated in standard and additively
          perturbed forms using 21 optimizers and 40 repeated runs per
          configuration, with final loss, success rate, distance, and
          runtime recorded.

    \item \textbf{Landscape geometry determines the difficulty of the task.}
          Himmelblau is largely tractable because its global minima lie in
          well-formed basins, whereas Ackley's outer plateau prevents most
          methods from reaching the narrow central minimum.

    \item \textbf{Fractional and memory-based SGD variants are most reliable,
          but performance must be read beyond mean final loss.}
          FSGD, FCSGD\_GL, MemoryFSGD, and related SGD-type methods lead on
          the harder cases, while success rates, distributions, target
          distance, and runtime reveal instability and computational cost.
  \end{itemize}
\end{takeawaybox}

\section{Neural Network Experiments with Fractal Activations and Fractional Optimizers}
\label{sec:nn_experiments}

The second experimental block studies whether fractal activation functions and fractional optimizer variants affect neural network training on classification tasks. The goal is to evaluate these methods in a standard supervised learning setting after first studying their behaviour on controlled optimization surfaces.

\subsection{Experimental Pipeline, Architectures, Activations, and Optimizers}
\label{subsec:nn_pipeline}

We evaluate feed-forward neural networks on ten publicly available OpenML classification datasets. The dataset set follows the classification benchmark setting used in the fractal activation function study by Raubitzek et al.~\cite{raubitzek_fractals_2026}. The datasets are climate-model-simulation-crashes, diabetes, ionosphere, tic-tac-toe, vertebra-column, glass, iris, seeds, vehicle, and wine. These datasets cover binary and multi-class classification tasks and provide a compact but varied testbed for optimizer--activation interactions.

All datasets are loaded through OpenML and preprocessed in a common pipeline. Categorical features are ordinally encoded, missing values are replaced by zero, and target labels are integer encoded. Each dataset is split into training, validation, and test partitions using stratified sampling. The training features are standardized with a scaler fitted only on the training partition, and the same transformation is then applied to validation and test data.

The neural network architecture is dataset-specific. Binary datasets use one hidden layer, while multi-class datasets use two hidden layers with widths $(n,2n)$. The width $n$, batch size, and number of training epochs follow the earlier fractal activation study. Table~\ref{tab:nn_dataset_design} summarizes the dataset-level settings.

\begin{table}[htbp]
\centering
\caption{Dataset-specific neural network settings.}
\label{tab:nn_dataset_design}
\begin{tabular}{llrrrr}
\hline
Dataset & Task & Width $n$ & Hidden layers & Batch size & Epochs \\
\hline
climate-model. & binary      & 32  & 1 & 32 & 30 \\
diabetes                         & binary      & 64  & 1 & 32 & 30 \\
ionosphere                       & binary      & 128 & 1 & 32 & 30 \\
tic-tac-toe                      & binary      & 64  & 1 & 32 & 25 \\
vertebra-column                  & binary      & 32  & 1 & 16 & 30 \\
glass                            & multi-class & 64  & 2 & 32 & 30 \\
iris                             & multi-class & 32  & 2 & 16 & 30 \\
seeds                            & multi-class & 64  & 2 & 32 & 30 \\
vehicle                          & multi-class & 128 & 2 & 32 & 25 \\
wine                             & multi-class & 64  & 2 & 32 & 30 \\
\hline
\end{tabular}
\end{table}

The activation comparison includes standard activations and selected fractal activations. The standard activations are ReLU and tanh. The fractal activations are taken from the earlier fractal activation function work by Raubitzek et al.~\cite{raubitzek_fractals_2026}. To keep the computational cost manageable, we do not evaluate the full activation catalogue in this experiment. Instead, we use four selected fractal activations that showed strong behaviour in the earlier study: the modulated Blancmange curve, the decaying cosine activation, the modified Weierstrass--tanh activation, and the Weierstrass--Mandelbrot $x+\sin$ activation. This keeps the comparison focused while still covering different types of multi-scale nonlinearities.

The optimizer comparison contains 21 optimizers divided into five groups. The first group contains four standard baselines: SGD, Adam, RMSprop, and Adadelta. These methods provide reference points for first-order optimization and adaptive gradient scaling~\cite{kingma2015adam,zeiler2012adadelta,tieleman2012rmsprop}. The second group contains four Herrera-style fractional optimizers: FSGD, FAdam, FRMSprop, and FAdadelta. These variants modify standard optimizers by introducing a fractional derivative order. The third group contains four explicit memory-based fractional optimizers: MemoryFSGD, MemoryFRMSprop, MemoryFAdam, and MemoryFAdadelta. These methods use a finite history of previous gradients to approximate Grünwald--Letnikov-type memory effects. The fourth group contains four adaptive memory-based fractional optimizers: AdaptiveMemoryFSGD, AdaptiveMemoryFRMSprop, AdaptiveMemoryFAdam, and AdaptiveMemoryFAdadelta. These methods mix ordinary and fractional update components and adapt the contribution of the memory term during training. The fifth group contains five related-work fractional optimizers: AdaGL, FCSGD\_GL, FCAdam\_GL, AOFGD\_SGD, and AOFGD\_Adam.

For the Herrera-style and memory-based optimizers, the fractional derivative order is swept over
\begin{equation}
\nu \in \{0.75, 1.25, 1.5\}.
\end{equation}
The memory-based methods use a history size of six. All configurations are repeated over 40 random seeds. For each run we record accuracy, macro-F1, precision, recall, training time, test time, number of completed epochs, and best validation loss. The analysis script then aggregates results by configuration, optimizer, activation, optimizer group, activation family, and fractional order. It also produces ranked accuracy plots, boxplots, macro-F1 plots, training-time plots, optimizer--activation heatmaps, optimizer-group summaries, fractional-order sensitivity plots, and accuracy-versus-runtime plots.

\subsection{Classification Results}
\label{subsec:nn_results}

The classification results are reported separately for each dataset. For
each optimizer, the configuration with the highest mean test accuracy is
used in the optimizer-level ranking. The activation-level figures instead
aggregate all evaluated optimizer configurations and derivative-order
settings. These two views therefore answer different questions: the first
shows the best configuration available to each optimizer, whereas the
second measures the average behaviour of an activation across the complete
experimental grid.

\subsubsection{Climate-Model Simulation Crashes}
\label{subsubsec:nn_results_climate}

The Climate-Model Simulation Crashes dataset
(\href{https://www.openml.org/d/1467}{OpenML ID 1467}) contains 540
climate-model simulations described by 18 numerical configuration
variables. The binary target indicates whether the corresponding
simulation completes or crashes. The experiment comprised 222 unique
optimizer--activation--order configurations, each evaluated over 40
repeated runs, resulting in 8880 individual runs.

Table~\ref{tab:nn_climate_top15} reports the 15 highest-ranked
optimizer-specific configurations, retaining only the best
activation--order setting for each optimizer.

\begin{table}[H]
\caption{Top 15 optimizer-specific configurations on the Climate-Model
Simulation Crashes dataset, ranked by mean test accuracy over 40 repeated
runs. For each optimizer, only its highest-ranked activation--order
configuration is shown. Standard deviation quantifies variation across
the repeated runs. For optimizers without a swept fractional-order
parameter, $\nu=1.00$ denotes the recorded default or integer-order
setting.}
\label{tab:nn_climate_top15}
\small
\begingroup
\setlength{\tabcolsep}{4pt}
\begin{tabularx}{\textwidth}{
    >{\centering\arraybackslash}p{0.8cm}
    >{\raggedright\arraybackslash}X
    l
    l
    c
    >{\raggedleft\arraybackslash}p{1.25cm}
    r
}
\toprule
\textbf{Rank} &
\textbf{Optimizer} &
\textbf{Group} &
\textbf{Activation} &
\boldmath$\boldsymbol{\nu}$ &
\textbf{Avg. Acc.} &
\textbf{Std.} \\
\midrule
\textbf{1} &
\textbf{AdaptiveMemory\allowbreak FAdadelta} &
\textbf{Adaptive-Memory} &
\textbf{mod.\ wei.\ tanh} &
\textbf{1.00} &
\textbf{0.91250} &
\textbf{0.01740} \\

2  & Adadelta                     & Standard        & mod.\ wei.\ tanh & 1.00 & 0.91204 & 0.01789 \\
3  & MemoryFSGD                   & GL-Memory       & mod.\ wei.\ tanh & 1.50 & 0.91204 & 0.01789 \\
4  & FAdam                        & Herrera         & mod.\ wei.\ tanh & 1.50 & 0.91188 & 0.01751 \\
5  & FAdadelta                    & Herrera         & mod.\ wei.\ tanh & 0.75 & 0.91173 & 0.01797 \\
6  & FCSGD\_GL                    & Related-Work    & blancmange       & 1.00 & 0.91173 & 0.01719 \\
7  & AdaptiveMemory\allowbreak FAdam
                                  & Adaptive-Memory & mod.\ wei.\ tanh & 1.00 & 0.91157 & 0.01792 \\
8  & MemoryFRMSprop               & GL-Memory       & mod.\ wei.\ tanh & 0.75 & 0.91157 & 0.01754 \\
9  & MemoryFAdam                  & GL-Memory       & mod.\ wei.\ tanh & 0.75 & 0.91157 & 0.01754 \\
10 & FCAdam\_GL                   & Related-Work    & mod.\ wei.\ tanh & 1.00 & 0.91142 & 0.01760 \\
11 & MemoryFAdadelta              & GL-Memory       & ReLU              & 1.50 & 0.91142 & 0.01727 \\
12 & Adam                         & Standard        & blancmange       & 1.00 & 0.91142 & 0.01727 \\
13 & FRMSprop                     & Herrera         & blancmange       & 0.75 & 0.91142 & 0.01727 \\
14 & RMSprop                      & Standard        & blancmange       & 1.00 & 0.91142 & 0.01727 \\
15 & AdaptiveMemory\allowbreak FRMSprop
                                  & Adaptive-Memory & blancmange       & 1.00 & 0.91142 & 0.01727 \\
\bottomrule
\end{tabularx}
\endgroup
\end{table}

The highest mean accuracy was obtained by AdaptiveMemoryFAdadelta with
the modified Weierstrass--tanh activation and the recorded order
$\nu=1.00$, reaching $0.91250 \pm 0.01740$. Standard Adadelta and
MemoryFSGD followed with the same mean accuracy of $0.91204$ and standard
deviation of $0.01789$. The difference to the leading configuration was
only 0.046 percentage points, small relative to the run-to-run variation,
with substantially overlapping confidence intervals.

All five optimizer groups appeared among the top six positions: Adadelta
was the strongest standard optimizer, MemoryFSGD the strongest GL-memory
method, FAdam the highest Herrera-type optimizer, and FCSGD\_GL the
strongest related-work method. Across the top 15, four entries belonged
to GL-Memory, three each to Standard, Herrera, and Adaptive-Memory, and
two to Related-Work. The complete top-15 range was only 0.108 percentage
points, so the leading methods were closely grouped.

The best standard-activation entry was MemoryFAdadelta with ReLU at
$0.91142 \pm 0.01727$, while the best fractal-activation configuration
was the overall winner. Accuracy and macro-F1 did not produce the same
ordering: AdaptiveMemoryFAdadelta obtained a mean macro-F1 of $0.53004$,
whereas FAdadelta and standard Adadelta reached $0.53419$ and $0.53361$.
The winning configuration also required a mean training time of
$5.66$~s, compared with $5.15$~s for standard Adadelta, so the small
accuracy gain came with approximately $0.51$~s of additional mean
training time.

\begin{figure}[H]
\centering
\includegraphics[width=\textwidth]{
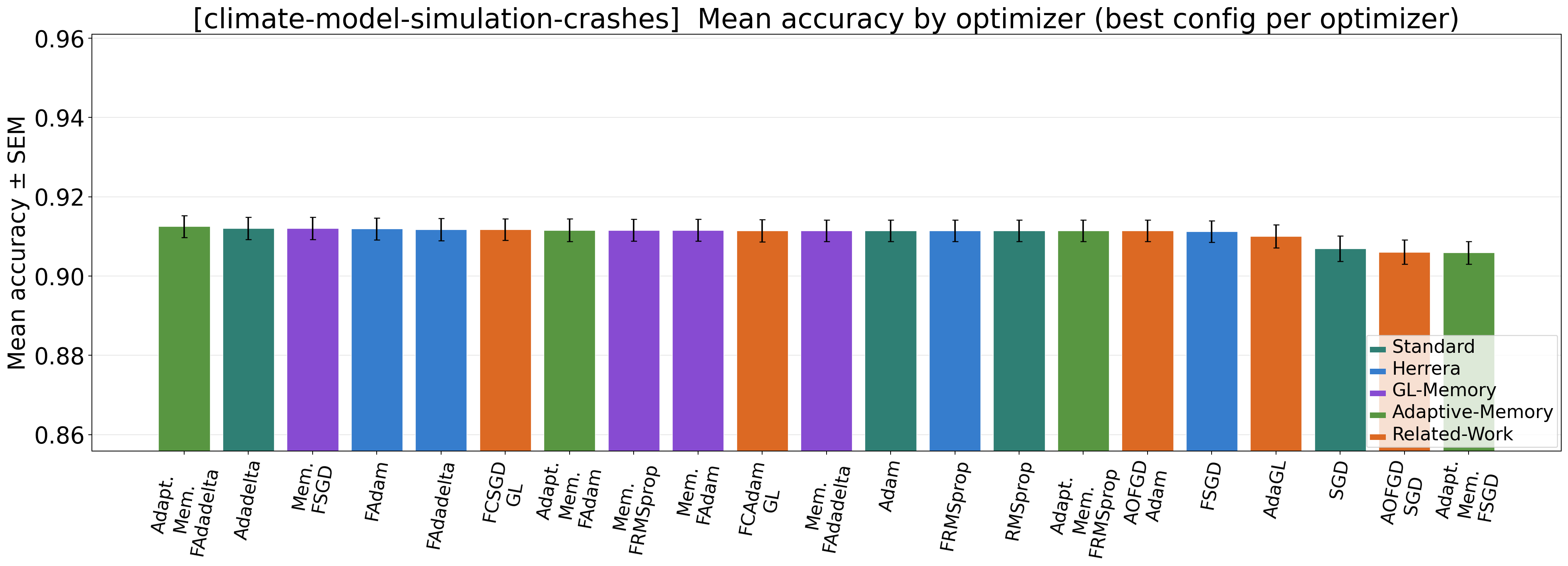}
\caption{Mean test accuracy by optimizer on the Climate-Model Simulation
Crashes dataset. Each bar represents the highest-performing
activation--derivative-order configuration identified for that optimizer.
Optimizers are ordered by mean accuracy, error bars show the standard
error of the mean over 40 repeated runs, and bar colours indicate the
optimizer groups.}
\label{fig:nn_climate_optimizer_accuracy}
\end{figure}

Figure~\ref{fig:nn_climate_optimizer_accuracy} presents the complete
optimizer-level ranking, including all 21 optimizers.

Figure~\ref{fig:nn_climate_optimizer_accuracy} confirms that the leading
optimizer groups cannot be separated by a large accuracy margin. The best
configuration from each group achieved a mean accuracy between $0.91173$
and $0.91250$, and the full optimizer range extended from $0.91250$ for
AdaptiveMemoryFAdadelta to $0.90586$ for AdaptiveMemoryFSGD, a difference
of 0.664 percentage points. Error bars overlap across most of this range.

The lower end contained SGD, AOFGD\_SGD, and AdaptiveMemoryFSGD. This
does not indicate a general limitation of SGD-derived approaches, since
MemoryFSGD ranked third and FSGD remained close to the leading methods.
Rather, performance depended on how the update rule was combined with
the selected activation function.

\begin{figure}[H]
\centering
\includegraphics[width=\textwidth]{
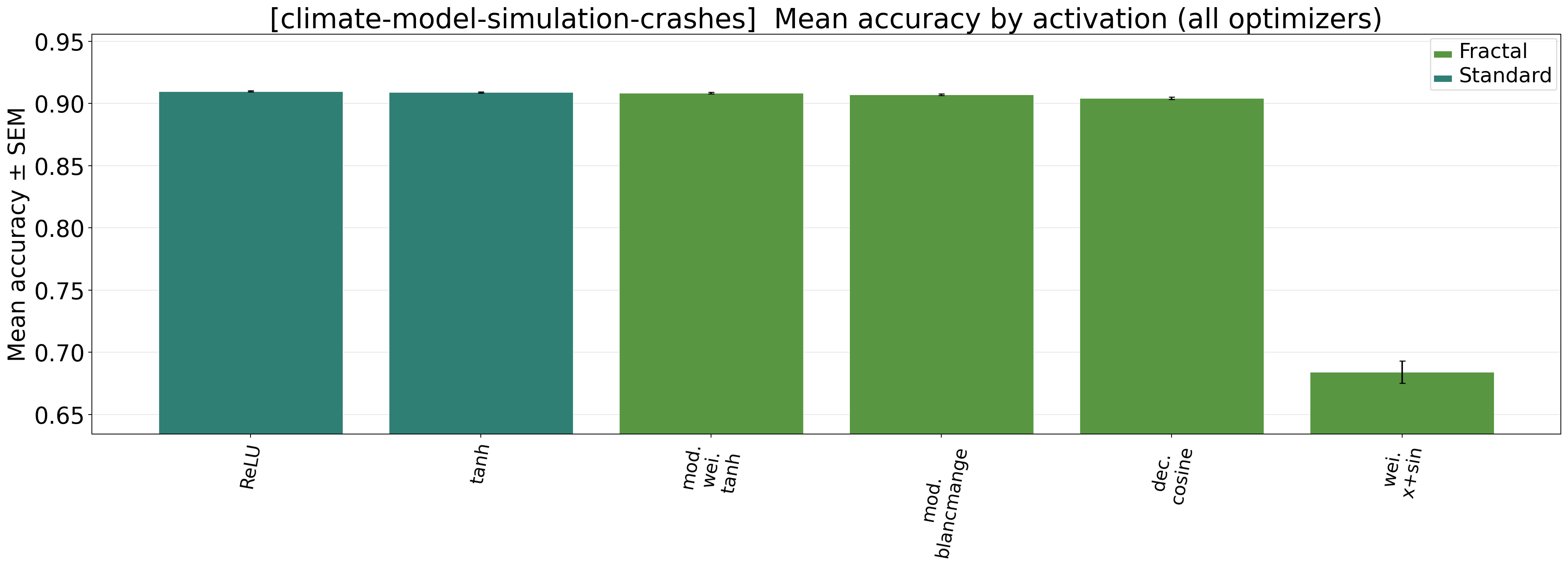}
\caption{Mean test accuracy by activation function on the Climate-Model
Simulation Crashes dataset, aggregated over all evaluated optimizer
configurations and derivative-order settings. Error bars show the
standard error of the mean for the aggregated run-level accuracies.
Colours distinguish standard and fractal activation functions.}
\label{fig:nn_climate_activation_accuracy}
\end{figure}

Figure~\ref{fig:nn_climate_activation_accuracy} aggregates the results by
activation function across all optimizer configurations and
derivative-order settings.

At activation level, ReLU achieved the highest aggregated mean accuracy
at approximately $0.9098$, followed by tanh at approximately $0.9089$.
The modified Weierstrass--tanh activation was the strongest fractal
activation at approximately $0.9084$, about 0.14 percentage points below
ReLU. The modulated Blancmange and decaying cosine activations followed
at approximately $0.9071$ and $0.9040$, with small error bars across
optimizer configurations.

The Weierstrass--Mandelbrot $x+\sin$ activation was the outlier, with an
aggregated mean accuracy of approximately $0.6844$ and a visibly larger
uncertainty range. Consequently, the mean across all standard-activation
runs was $0.90934$, compared with $0.85095$ across all fractal-activation
runs. The 5.84 percentage-point difference was primarily caused by this
low mean, while the remaining three fractal activations stayed within
approximately 0.58 percentage points of ReLU.

The configuration-level and activation-level results are compatible but
different. The best individual configuration used the modified
Weierstrass--tanh activation, whereas ReLU had the highest mean over the
complete optimizer grid. The modified Weierstrass--tanh activation
therefore appears to require a suitable optimizer pairing rather than
providing a uniform advantage across all update rules.

For this dataset, the best observed configuration combined a fractal
activation with an adaptive-memory optimizer, but its advantage over
standard Adadelta was small relative to the variability across runs and
was obtained with the recorded order $\nu=1.00$. The results therefore do
not establish a general fractional-order advantage on this task. They
instead show a dataset-specific optimizer--activation interaction:
modified Weierstrass--tanh was competitive and produced the best tuned
configuration, while the Weierstrass--Mandelbrot $x+\sin$ activation was
not robust across the broader optimizer comparison.

\subsubsection{Diabetes}
\label{subsubsec:nn_results_diabetes}

The Diabetes dataset
(\href{https://www.openml.org/d/37}{OpenML ID 37}) contains 768 patient
records described by eight numerical diagnostic variables. The binary
target indicates whether diabetes is present or absent. The experiment
comprised 222 unique optimizer--activation--order configurations, each
evaluated over 40 repeated runs, resulting in 8880 individual runs.

Table~\ref{tab:nn_diabetes_top15} reports the 15 highest-ranked
optimizer-specific configurations, retaining only the best
activation--order setting for each optimizer.

\begin{table}[H]
\caption{Top 15 optimizer-specific configurations on the Diabetes dataset,
ranked by mean test accuracy over 40 repeated runs. For each optimizer,
only its highest-ranked activation--order configuration is shown.
Standard deviation quantifies variation across the repeated runs. For
optimizers without a swept fractional-order parameter, $\nu=1.00$ denotes
the recorded default or integer-order setting.}
\label{tab:nn_diabetes_top15}
\small
\begingroup
\setlength{\tabcolsep}{4pt}
\begin{tabularx}{\textwidth}{
    >{\centering\arraybackslash}p{0.8cm}
    >{\raggedright\arraybackslash}X
    l
    l
    c
    >{\raggedleft\arraybackslash}p{1.25cm}
    r
}
\toprule
\textbf{Rank} &
\textbf{Optimizer} &
\textbf{Group} &
\textbf{Activation} &
\boldmath$\boldsymbol{\nu}$ &
\textbf{Avg. Acc.} &
\textbf{Std.} \\
\midrule
\textbf{1} &
\textbf{FSGD} &
\textbf{Herrera} &
\textbf{ReLU} &
\textbf{1.25} &
\textbf{0.77197} &
\textbf{0.02215} \\

2  & AdaptiveMemory\allowbreak FAdam
   & Adaptive-Memory & mod.\ wei.\ tanh & 1.00 & 0.76981 & 0.02222 \\
3  & FAdam
   & Herrera         & mod.\ wei.\ tanh & 1.50 & 0.76970 & 0.02162 \\
4  & Adam
   & Standard        & mod.\ wei.\ tanh & 1.00 & 0.76959 & 0.02224 \\
5  & FCSGD\_GL
   & Related-Work    & mod.\ wei.\ tanh & 1.00 & 0.76916 & 0.02251 \\
6  & AOFGD\_Adam
   & Related-Work    & mod.\ wei.\ tanh & 1.00 & 0.76840 & 0.02326 \\
7  & AdaptiveMemory\allowbreak FSGD
   & Adaptive-Memory & tanh             & 1.00 & 0.76786 & 0.02350 \\
8  & AdaptiveMemory\allowbreak FAdadelta
   & Adaptive-Memory & mod.\ wei.\ tanh & 1.00 & 0.76764 & 0.02366 \\
9  & AdaptiveMemory\allowbreak FRMSprop
   & Adaptive-Memory & mod.\ wei.\ tanh & 1.00 & 0.76764 & 0.02186 \\
10 & FRMSprop
   & Herrera         & mod.\ wei.\ tanh & 1.50 & 0.76753 & 0.02199 \\
11 & SGD
   & Standard        & tanh             & 1.00 & 0.76753 & 0.02374 \\
12 & RMSprop
   & Standard        & mod.\ wei.\ tanh & 1.00 & 0.76721 & 0.02217 \\
13 & FAdadelta
   & Herrera         & mod.\ wei.\ tanh & 0.75 & 0.76667 & 0.02136 \\
14 & AOFGD\_SGD
   & Related-Work    & mod.\ wei.\ tanh & 1.00 & 0.76667 & 0.02297 \\
15 & Adadelta
   & Standard        & mod.\ wei.\ tanh & 1.00 & 0.76656 & 0.02343 \\
\bottomrule
\end{tabularx}
\endgroup
\end{table}

The highest mean accuracy was obtained by FSGD with ReLU and fractional
order $\nu=1.25$, reaching $0.77197 \pm 0.02215$. AdaptiveMemoryFAdam
with modified Weierstrass--tanh ranked second at
$0.76981 \pm 0.02222$, and FAdam with the same activation and
$\nu=1.50$ followed at $0.76970 \pm 0.02162$. The 0.216-percentage-point
gap between the first two configurations was small compared with the
standard deviations of approximately 2.2 percentage points, and the
confidence intervals overlapped.

Adam was the strongest standard optimizer at $0.76959 \pm 0.02224$ with
modified Weierstrass--tanh. The best standard-activation configuration
was the overall winner, while the best fractal-activation configuration
was AdaptiveMemoryFAdam with modified Weierstrass--tanh. The top 15
contained four standard optimizers, all four Herrera-type optimizers,
all four adaptive-memory optimizers, and three related-work methods;
none of the explicit GL-memory optimizers entered this group.

The range within Table~\ref{tab:nn_diabetes_top15} was narrow: FSGD in
first place and Adadelta in fifteenth differed by only
0.541 percentage points. The represented optimizer families therefore
produced similar best-case accuracies when paired with suitable
activations, and the ranking does not indicate a uniformly dominant
family.

Macro-F1 gave a similar but not identical ordering. FSGD obtained
$0.73161$, whereas FCSGD\_GL reached a slightly higher value of
$0.73282$ despite ranking fifth by accuracy. The FSGD configuration also
required only $1.36$~s of mean training time, compared with $7.59$~s for
the second-ranked AdaptiveMemoryFAdam configuration, combining the
highest mean accuracy with substantially lower cost than most leading
modified-Weierstrass--tanh configurations.

\begin{figure}[H]
\centering
\includegraphics[width=\textwidth]{
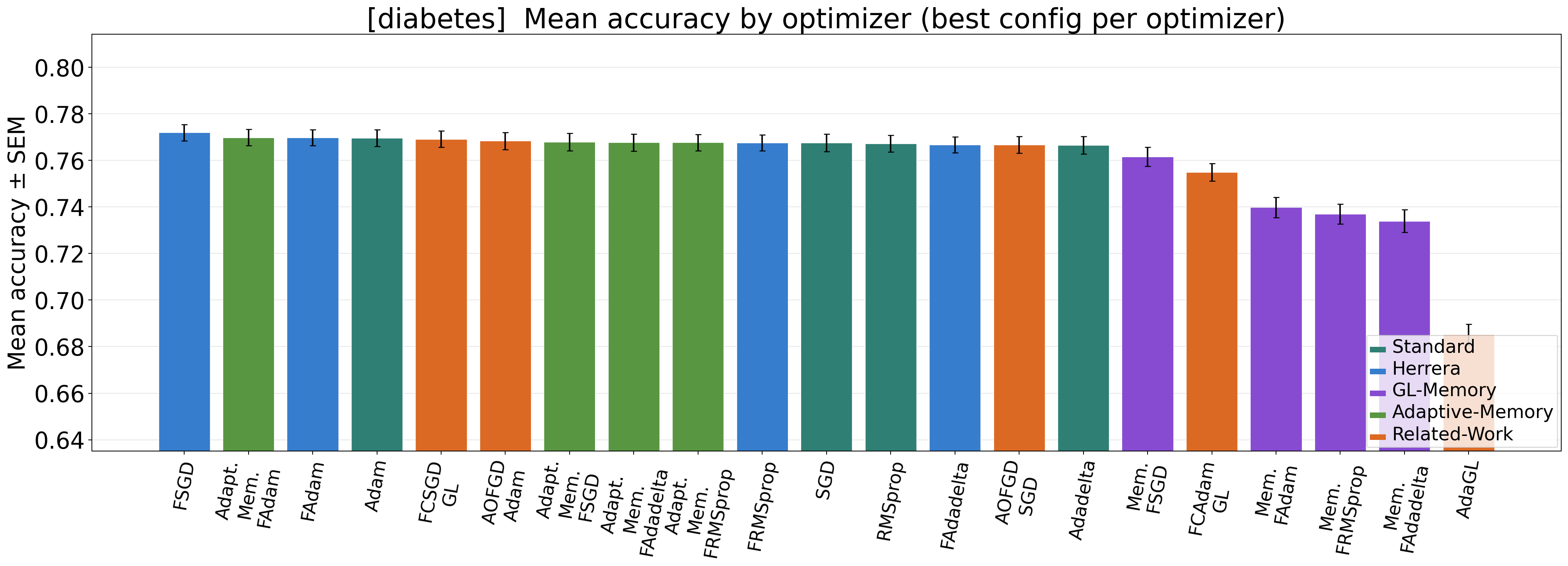}
\caption{Mean test accuracy by optimizer on the Diabetes dataset. Each
bar represents the highest-performing activation--fractional-order
configuration identified for that optimizer. Optimizers are ordered by
mean accuracy, error bars show the standard error of the mean over
40 repeated runs, and bar colours indicate the optimizer groups.}
\label{fig:nn_diabetes_optimizer_accuracy}
\end{figure}

Figure~\ref{fig:nn_diabetes_optimizer_accuracy} presents the complete
optimizer-level ranking, including all 21 optimizers.

Figure~\ref{fig:nn_diabetes_optimizer_accuracy} shows a closely grouped
upper ranking: FSGD, AdaptiveMemoryFAdam, FAdam, Adam, FCSGD\_GL, and
AOFGD\_Adam all reached mean accuracies between $0.76840$ and
$0.77197$. The best result from each of the Standard, Herrera,
Adaptive-Memory, and Related-Work groups therefore differed by less than
0.3 percentage points.

The explicit GL-memory methods formed a lower-performing part of the
ranking. MemoryFSGD was strongest in this group at
$0.76158 \pm 0.02585$, while MemoryFAdam, MemoryFRMSprop, and
MemoryFAdadelta reached $0.73983$, $0.73701$, and $0.73398$,
respectively, with generally larger error bars than the leading methods.
Explicit gradient memory was therefore not consistently beneficial for
this dataset, although MemoryFSGD remained closer to the main optimizer
cluster than the other GL-memory variants.

Across all 21 optimizers, mean accuracy ranged from $0.77197$ for FSGD
to $0.68517$ for AdaGL, a difference of 8.680 percentage points. AdaGL
was separated from the other related-work methods, since FCSGD\_GL and
AOFGD\_Adam both ranked within the leading six configurations. Its lower
result should therefore be interpreted as method-specific rather than as
a common property of the Related-Work group.

\begin{figure}[H]
\centering
\includegraphics[width=\textwidth]{
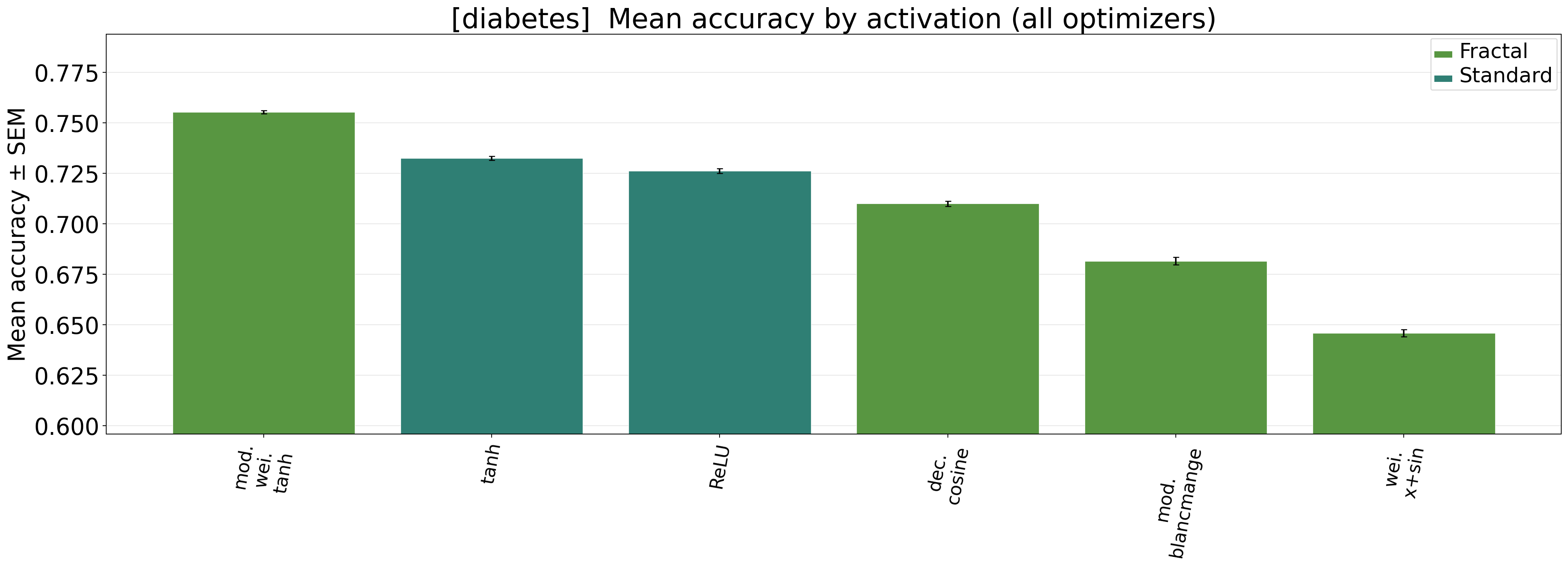}
\caption{Mean test accuracy by activation function on the Diabetes
dataset, aggregated over all evaluated optimizer configurations and
fractional-order settings. Error bars show the standard error of the mean
for the aggregated run-level accuracies. Colours distinguish standard and
fractal activation functions.}
\label{fig:nn_diabetes_activation_accuracy}
\end{figure}

Figure~\ref{fig:nn_diabetes_activation_accuracy} aggregates the results
by activation function across all optimizer configurations and
fractional-order settings.

At activation level, modified Weierstrass--tanh obtained the highest
aggregated mean accuracy at approximately $0.755$. Tanh was the strongest
standard activation at approximately $0.732$, followed by ReLU at
approximately $0.726$, so modified Weierstrass--tanh exceeded tanh by
approximately 2.3 percentage points. This indicates that its performance
was not limited to one selected configuration.

The remaining fractal activations did not show the same behaviour:
decaying cosine reached approximately $0.710$, modulated Blancmange
approximately $0.682$, and Weierstrass--Mandelbrot $x+\sin$
approximately $0.646$. The comparatively small error bars indicate that
these activation-level differences persisted after aggregation over the
optimizer and derivative-order settings.

When combined into families, standard activations obtained a mean
accuracy of $0.72942$, compared with $0.69828$ for fractal activations.
This does not contradict the first-place result of modified
Weierstrass--tanh, because the fractal-family average combines one
strong activation with three lower-performing variants, in particular
Weierstrass--Mandelbrot $x+\sin$. The individual activation ranking is
therefore more informative than the broad family mean for this dataset.

The configuration-level and activation-level analyses are complementary.
The best individual configuration used FSGD with ReLU, whereas modified
Weierstrass--tanh achieved the highest mean accuracy over the complete
optimizer grid; the best standard optimizer, Adam, also used modified
Weierstrass--tanh. The results therefore indicate broad compatibility of
this fractal activation with adaptive optimizers, while the highest
single accuracy came from a comparatively simple and inexpensive
FSGD--ReLU pairing.

For this dataset, the results do not support a general advantage for
explicit GL-memory optimization. They instead show that direct
fractional gradient modification can provide a competitive result and
that the effect of a fractal activation depends strongly on its specific
construction. Modified Weierstrass--tanh performed well both in selected
configurations and after aggregation, whereas the other fractal
activations produced lower average accuracies. The Diabetes results
therefore support analysing individual optimizer--activation pairings
rather than treating either fractional optimizers or fractal activations
as homogeneous method families.

\subsubsection{Glass Identification}
\label{subsubsec:nn_results_glass}

The Glass Identification dataset
(\href{https://www.openml.org/d/41}{OpenML ID 41}) contains 214 glass
samples described by nine continuous chemical and physical measurements.
The target distinguishes six glass types, resulting in a small
multi-class classification problem. The experiment comprised 222 unique
optimizer--activation--order configurations, each evaluated over 40
repeated runs, resulting in 8880 individual runs.

Table~\ref{tab:nn_glass_top15} reports the 15 highest-ranked
optimizer-specific configurations, retaining only the best
activation--order setting for each optimizer.

\begin{table}[H]
\caption{Top 15 optimizer-specific configurations on the Glass
Identification dataset, ranked by mean test accuracy over 40 repeated
runs. For each optimizer, only its highest-ranked activation--order
configuration is shown. Standard deviation quantifies variation across
the repeated runs. For optimizers without a swept fractional-order
parameter, $\nu=1.00$ denotes the recorded default or integer-order
setting.}
\label{tab:nn_glass_top15}
\small
\begingroup
\setlength{\tabcolsep}{4pt}
\begin{tabularx}{\textwidth}{
    >{\centering\arraybackslash}p{0.8cm}
    >{\raggedright\arraybackslash}X
    l
    l
    c
    >{\raggedleft\arraybackslash}p{1.25cm}
    r
}
\toprule
\textbf{Rank} &
\textbf{Optimizer} &
\textbf{Group} &
\textbf{Activation} &
\boldmath$\boldsymbol{\nu}$ &
\textbf{Avg. Acc.} &
\textbf{Std.} \\
\midrule
\textbf{1} &
\textbf{AdaptiveMemory\allowbreak FAdadelta} &
\textbf{Adaptive-Memory} &
\textbf{dec.\ cosine} &
\textbf{1.00} &
\textbf{0.59538} &
\textbf{0.06739} \\

2  & FAdadelta
   & Herrera         & tanh             & 1.50 & 0.58846 & 0.07388 \\
3  & AdaptiveMemory\allowbreak FAdam
   & Adaptive-Memory & dec.\ cosine     & 1.00 & 0.58654 & 0.05051 \\
4  & FAdam
   & Herrera         & tanh             & 1.50 & 0.58615 & 0.07981 \\
5  & RMSprop
   & Standard        & dec.\ cosine     & 1.00 & 0.58385 & 0.06222 \\
6  & FRMSprop
   & Herrera         & dec.\ cosine     & 1.25 & 0.58038 & 0.06074 \\
7  & AdaptiveMemory\allowbreak FRMSprop
   & Adaptive-Memory & dec.\ cosine     & 1.00 & 0.58038 & 0.07316 \\
8  & Adam
   & Standard        & dec.\ cosine     & 1.00 & 0.57731 & 0.06976 \\
9  & Adadelta
   & Standard        & dec.\ cosine     & 1.00 & 0.57346 & 0.06862 \\
10 & AOFGD\_Adam
   & Related-Work    & dec.\ cosine     & 1.00 & 0.56692 & 0.06276 \\
11 & FCAdam\_GL
   & Related-Work    & dec.\ cosine     & 1.00 & 0.56385 & 0.06504 \\
12 & AdaptiveMemory\allowbreak FSGD
   & Adaptive-Memory & mod.\ wei.\ tanh & 1.00 & 0.53885 & 0.08706 \\
13 & AOFGD\_SGD
   & Related-Work    & mod.\ wei.\ tanh & 1.00 & 0.52769 & 0.06224 \\
14 & FSGD
   & Herrera         & mod.\ wei.\ tanh & 1.25 & 0.52423 & 0.06786 \\
15 & SGD
   & Standard        & mod.\ wei.\ tanh & 1.00 & 0.52231 & 0.08462 \\
\bottomrule
\end{tabularx}
\endgroup
\end{table}

The highest mean accuracy was obtained by AdaptiveMemoryFAdadelta with
the decaying cosine activation and the recorded order $\nu=1.00$,
reaching $0.59538 \pm 0.06739$. FAdadelta with tanh and $\nu=1.50$
ranked second at $0.58846 \pm 0.07388$, while AdaptiveMemoryFAdam with
decaying cosine ranked third at $0.58654 \pm 0.05051$. The
0.692-percentage-point gap between the first two configurations was
small relative to the run-to-run standard deviations, and the 95\%
confidence intervals overlapped.

RMSprop was the strongest standard optimizer at $0.58385 \pm 0.06222$
with decaying cosine. The best standard-activation configuration was
FAdadelta with tanh, while the best fractal-activation configuration was
the overall winner. The top 15 contained four Standard, four Herrera,
four Adaptive-Memory, and three Related-Work optimizers; none of the
explicit GL-Memory optimizers entered this group.

The decaying cosine activation appeared in nine of the first 11
positions, including the best Standard, Adaptive-Memory, and Related-Work
configurations, whereas tanh was used by the second- and fourth-ranked
Herrera-type methods. Modified Weierstrass--tanh appeared only from rank
12 onward. The activation choice therefore contributed substantially to
the optimizer-level ordering on this dataset.

The range within Table~\ref{tab:nn_glass_top15} was larger than for the
previous binary datasets. AdaptiveMemoryFAdadelta in first place and SGD
in fifteenth differed by 7.307 percentage points, although the first
11 configurations formed a more compact group between $0.56385$ and
$0.59538$. A larger decrease occurred between FCAdam\_GL in eleventh
place and AdaptiveMemoryFSGD in twelfth place.

Accuracy and macro-F1 were similar at the top but not identical. The
winning AdaptiveMemoryFAdadelta configuration obtained a mean macro-F1 of
$0.49421$, whereas FAdadelta reached $0.49777$ and required only
$1.49$~s of mean training time, compared with $19.22$~s for the winner.
The third-ranked AdaptiveMemoryFAdam configuration had the lowest
accuracy standard deviation among the first four entries, but also
required approximately $19.27$~s per run.

\begin{figure}[H]
\centering
\includegraphics[width=\textwidth]{
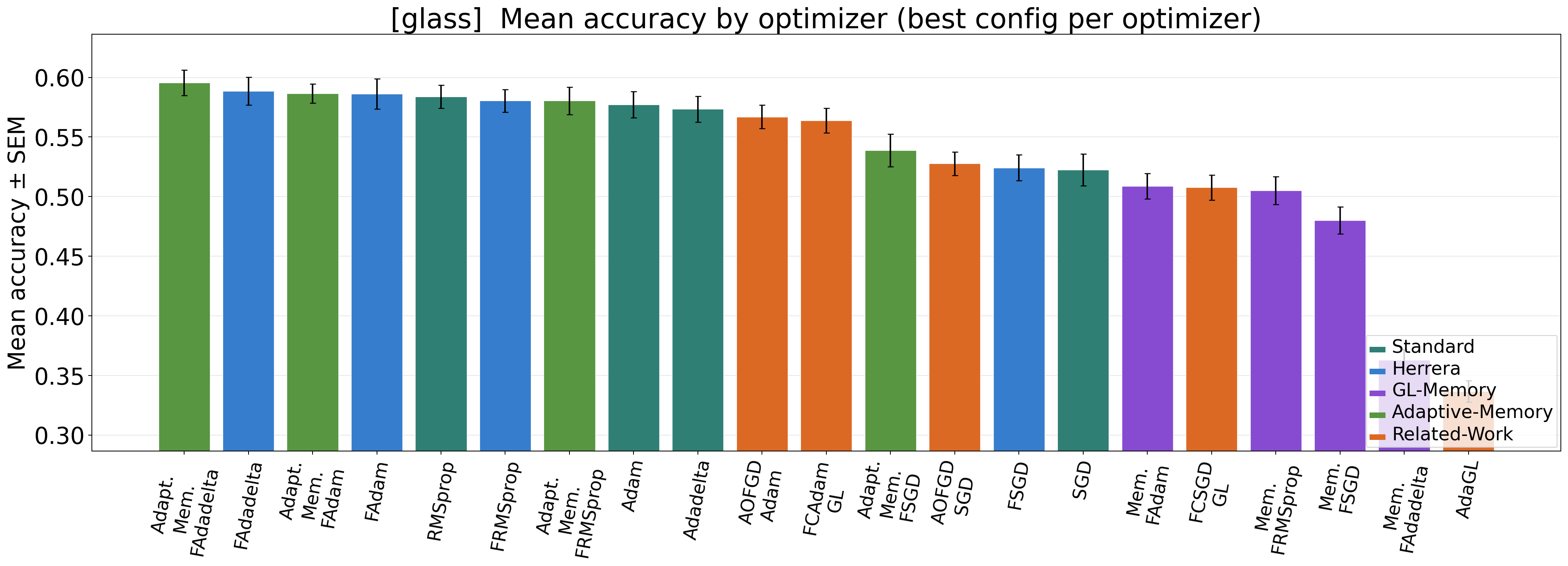}
\caption{Mean test accuracy by optimizer on the Glass Identification
dataset. Each bar represents the highest-performing
activation--fractional-order configuration identified for that optimizer.
Optimizers are ordered by mean accuracy, error bars show the standard
error of the mean over 40 repeated runs, and bar colours indicate the
optimizer groups.}
\label{fig:nn_glass_optimizer_accuracy}
\end{figure}

Figure~\ref{fig:nn_glass_optimizer_accuracy} presents the complete
optimizer-level ranking, including all 21 optimizers.

Figure~\ref{fig:nn_glass_optimizer_accuracy} shows substantial
differences across optimizer families. The strongest Adaptive-Memory,
Herrera, and Standard configurations reached $0.59538$, $0.58846$, and
$0.58385$, respectively, while AOFGD\_Adam was the strongest
Related-Work optimizer at $0.56692$. In contrast, the best explicit
GL-Memory method, MemoryFAdam, reached only $0.50885 \pm 0.06741$.

The four GL-Memory methods occupied ranks 16, 18, 19, and 20.
MemoryFAdam and MemoryFRMSprop reached approximately $0.509$ and
$0.505$, while MemoryFSGD reached $0.48000$ and MemoryFAdadelta reached
$0.36308$. Explicit gradient memory was therefore not well suited to this
dataset under the evaluated settings. The sensitivity results further
show that the GL-Memory variants generally performed best at $\nu=0.75$,
whereas orders of $1.25$ and $1.50$ frequently reduced their mean
accuracy.

Across all 21 optimizers, mean accuracy ranged from $0.59538$ for
AdaptiveMemoryFAdadelta to $0.33654$ for AdaGL, a difference of
25.884 percentage points. AdaGL was separated from the other
Related-Work methods, since AOFGD\_Adam and FCAdam\_GL remained within
the upper half of the ranking; its low result should therefore be read as
method-specific.

\begin{figure}[H]
\centering
\includegraphics[width=\textwidth]{
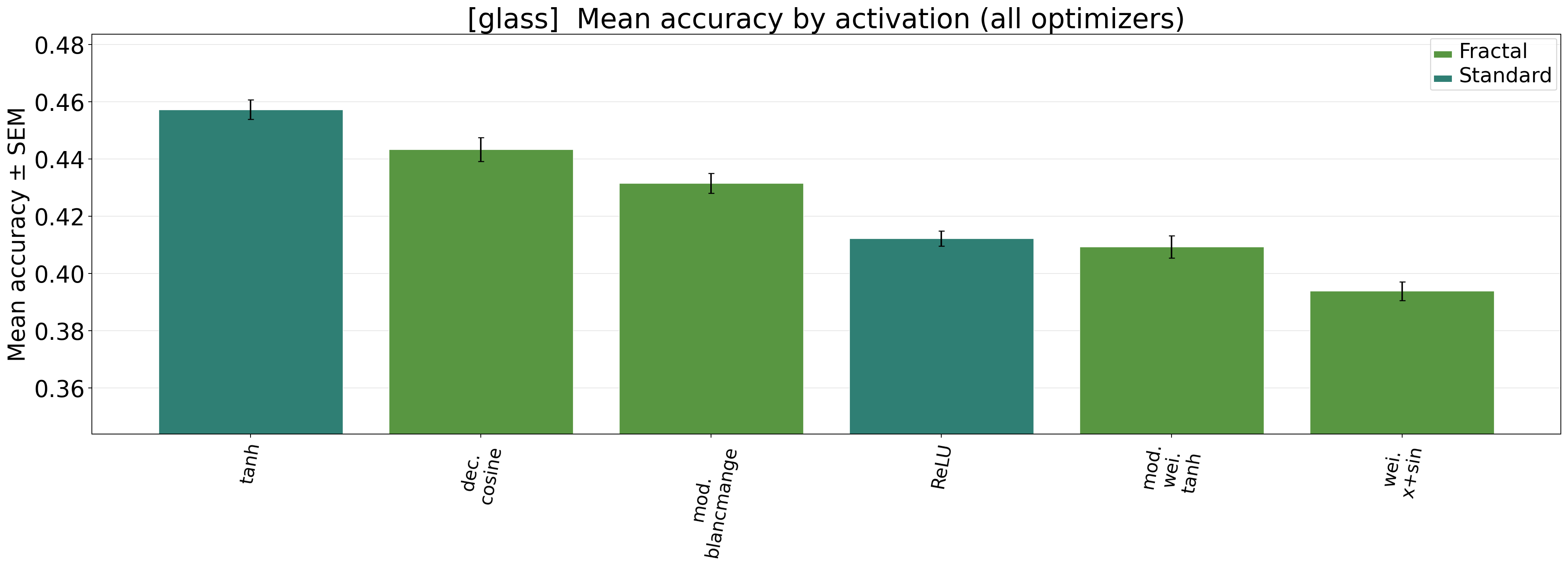}
\caption{Mean test accuracy by activation function on the Glass
Identification dataset, aggregated over all evaluated optimizer
configurations and fractional-order settings. Error bars show the standard
error of the mean for the aggregated run-level accuracies. Colours
distinguish standard and fractal activation functions.}
\label{fig:nn_glass_activation_accuracy}
\end{figure}

Figure~\ref{fig:nn_glass_activation_accuracy} aggregates the results by
activation function across all optimizer configurations and
fractional-order settings.

At activation level, tanh obtained the highest aggregated mean accuracy
at approximately $0.457$. Decaying cosine was the strongest fractal
activation at approximately $0.443$, about 1.4 percentage points below
tanh, followed by modulated Blancmange at approximately $0.431$. ReLU
reached approximately $0.412$, while modified Weierstrass--tanh and
Weierstrass--Mandelbrot $x+\sin$ obtained approximately $0.409$ and
$0.394$.

The activation-level ranking differs from the best-configuration
ranking. Decaying cosine occurred in the overall winning configuration
and in most of the highest-ranked optimizer-specific configurations, but
tanh achieved the highest mean across the complete optimizer grid. This
indicates that decaying cosine performed particularly well with selected
adaptive and RMSprop-type optimizers, whereas tanh was more robust across
the full set of update rules.

When combined into families, standard activations reached a mean accuracy
of $0.43471$, compared with $0.41951$ for fractal activations, a
difference of 1.520 percentage points. The macro-F1 family summary showed
the opposite ordering: fractal activations obtained $0.26190$, compared
with $0.24427$ for the standard functions. Broad family averages
therefore depend on the selected evaluation measure and conceal
differences between individual fractal activations.

For this dataset, the highest-ranked configuration combined the decaying
cosine activation with an adaptive-memory optimizer. Its accuracy
advantage over the much faster FAdadelta--tanh configuration was small
relative to the variation across runs, and the activation aggregation
placed tanh ahead of decaying cosine on average. The results therefore
suggest a specific interaction between decaying cosine and selected
adaptive optimizers rather than a uniform fractal-activation advantage.
At the same time, the weak results of the explicit GL-Memory methods show
that introducing gradient history did not generally improve performance
on this small six-class problem.

\subsubsection{Ionosphere}
\label{subsubsec:nn_results_ionosphere}

The Ionosphere dataset
(\href{https://www.openml.org/d/59}{OpenML ID 59}) contains 351 radar
returns described by 34 real-valued attributes derived from complex radar
signals. The binary target distinguishes between good and bad radar
returns. The experiment comprised 222 unique
optimizer--activation--order configurations, each evaluated over 40
repeated runs, resulting in 8880 individual runs.

Table~\ref{tab:nn_ionosphere_top15} reports the 15 highest-ranked
optimizer-specific configurations, retaining only the best
activation--order setting for each optimizer.

\begin{table}[H]
\caption{Top 15 optimizer-specific configurations on the Ionosphere
dataset, ranked by mean test accuracy over 40 repeated runs. For each
optimizer, only its highest-ranked activation--order configuration is
shown. Standard deviation quantifies variation across the repeated runs.
For optimizers without a swept fractional-order parameter, $\nu=1.00$
denotes the recorded default or integer-order setting.}
\label{tab:nn_ionosphere_top15}
\small
\begingroup
\setlength{\tabcolsep}{4pt}
\begin{tabularx}{\textwidth}{
    >{\centering\arraybackslash}p{0.8cm}
    >{\raggedright\arraybackslash}X
    l
    l
    c
    >{\raggedleft\arraybackslash}p{1.25cm}
    r
}
\toprule
\textbf{Rank} &
\textbf{Optimizer} &
\textbf{Group} &
\textbf{Activation} &
\boldmath$\boldsymbol{\nu}$ &
\textbf{Avg. Acc.} &
\textbf{Std.} \\
\midrule
\textbf{1} &
\textbf{FSGD} &
\textbf{Herrera} &
\textbf{mod.\ wei.\ tanh} &
\textbf{1.50} &
\textbf{0.91745} &
\textbf{0.03502} \\

2  & FAdam
   & Herrera         & ReLU             & 0.75 & 0.91108 & 0.03051 \\
3  & Adam
   & Standard        & ReLU             & 1.00 & 0.90873 & 0.03077 \\
4  & AOFGD\_SGD
   & Related-Work    & mod.\ wei.\ tanh & 1.00 & 0.90849 & 0.03248 \\
5  & SGD
   & Standard        & mod.\ wei.\ tanh & 1.00 & 0.90778 & 0.03501 \\
6  & AdaptiveMemory\allowbreak FSGD
   & Adaptive-Memory & mod.\ wei.\ tanh & 1.00 & 0.90755 & 0.03479 \\
7  & AdaptiveMemory\allowbreak FAdam
   & Adaptive-Memory & mod.\ wei.\ tanh & 1.00 & 0.90637 & 0.02764 \\
8  & FCSGD\_GL
   & Related-Work    & mod.\ wei.\ tanh & 1.00 & 0.90472 & 0.03430 \\
9  & AOFGD\_Adam
   & Related-Work    & mod.\ wei.\ tanh & 1.00 & 0.90212 & 0.02614 \\
10 & FRMSprop
   & Herrera         & mod.\ wei.\ tanh & 1.25 & 0.89858 & 0.03329 \\
11 & RMSprop
   & Standard        & mod.\ wei.\ tanh & 1.00 & 0.89811 & 0.03446 \\
12 & AdaptiveMemory\allowbreak FRMSprop
   & Adaptive-Memory & mod.\ wei.\ tanh & 1.00 & 0.89764 & 0.03610 \\
13 & FAdadelta
   & Herrera         & mod.\ wei.\ tanh & 0.75 & 0.89458 & 0.03281 \\
14 & Adadelta
   & Standard        & mod.\ wei.\ tanh & 1.00 & 0.89387 & 0.03116 \\
15 & AdaptiveMemory\allowbreak FAdadelta
   & Adaptive-Memory & mod.\ wei.\ tanh & 1.00 & 0.89198 & 0.03204 \\
\bottomrule
\end{tabularx}
\endgroup
\end{table}

The highest mean accuracy was obtained by FSGD with the modified
Weierstrass--tanh activation and fractional order $\nu=1.50$, reaching
$0.91745 \pm 0.03502$. FAdam with ReLU and $\nu=0.75$ ranked second at
$0.91108 \pm 0.03051$, followed by standard Adam with ReLU at
$0.90873 \pm 0.03077$. The 0.637-percentage-point gap between the first
two configurations was small relative to the run-to-run standard
deviations, and their 95\% confidence intervals,
$[0.90660,0.92831]$ and $[0.90163,0.92054]$, overlapped.

Adam was the strongest standard optimizer, AdaptiveMemoryFSGD the
highest-ranked Adaptive-Memory method, AOFGD\_SGD the strongest
Related-Work method, and MemoryFSGD the best GL-Memory optimizer. The
best standard-activation configuration was FAdam with ReLU, whereas the
best fractal-activation configuration was the overall winner.

The top 15 contained four Standard, four Herrera, four Adaptive-Memory,
and three Related-Work optimizers; none of the explicit GL-Memory methods
entered this group. Modified Weierstrass--tanh appeared in 13 of the
15 configurations, including the best configuration from every optimizer
family except the Standard group. ReLU appeared in the second- and
third-ranked configurations, and no other tested activation occurred in
the top 15.

The first- and fifteenth-ranked configurations differed by
2.547 percentage points. The first nine configurations all achieved mean
accuracies above $0.90$, and their error ranges overlapped substantially,
so the ordering should be read as a ranking of mean values rather than as
evidence of statistically distinct performance.

The macro-F1 results supported the accuracy ranking. The winning FSGD
configuration obtained a mean macro-F1 of $0.90634$, compared with
$0.89639$ for FAdam and $0.89333$ for Adam. It required a mean training
time of $5.50$~s, whereas FAdam and Adam required approximately $1.20$~s
and $0.97$~s; the 0.637-percentage-point accuracy increase over FAdam was
therefore accompanied by an approximately $4.30$~s runtime increase.

\begin{figure}[H]
\centering
\includegraphics[width=\textwidth]{
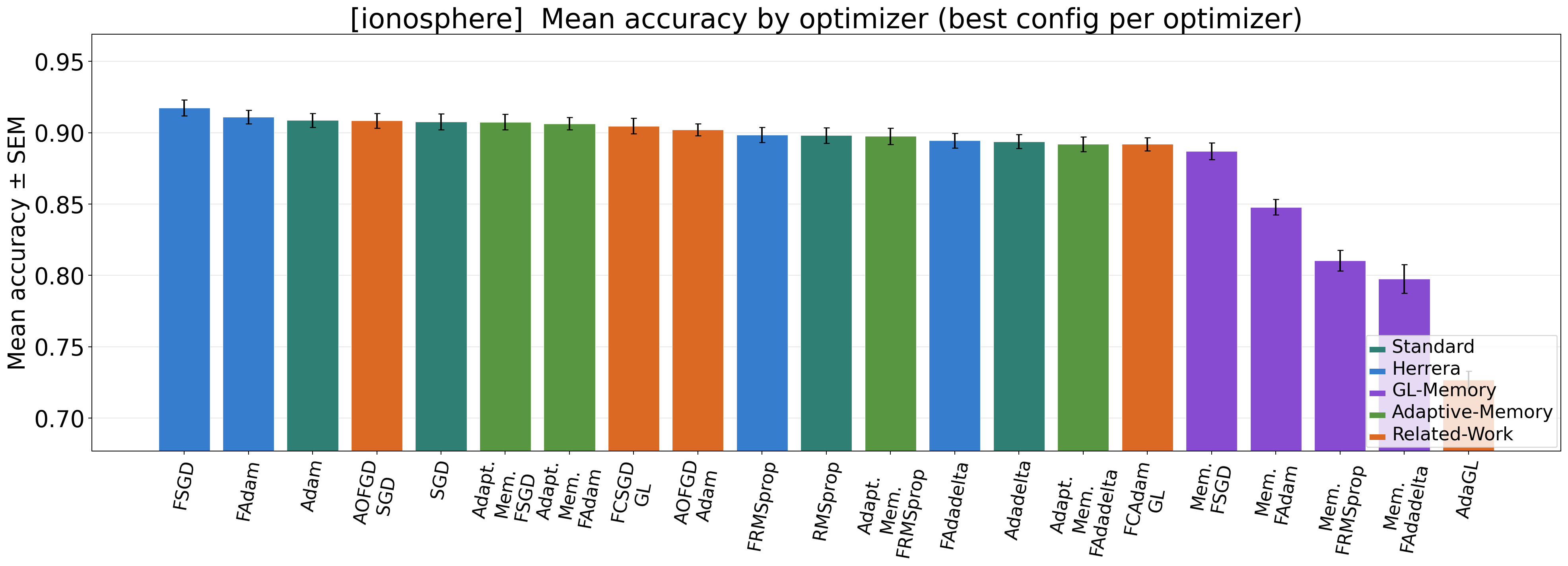}
\caption{Mean test accuracy by optimizer on the Ionosphere dataset. Each
bar represents the highest-performing activation--fractional-order
configuration identified for that optimizer. Optimizers are ordered by
mean accuracy, error bars show the standard error of the mean over
40 repeated runs, and bar colours indicate the optimizer groups.}
\label{fig:nn_ionosphere_optimizer_accuracy}
\end{figure}

Figure~\ref{fig:nn_ionosphere_optimizer_accuracy} presents the complete
optimizer-level ranking, including all 21 optimizers.

Figure~\ref{fig:nn_ionosphere_optimizer_accuracy} shows that the strongest
Herrera, Standard, Adaptive-Memory, and Related-Work configurations
occupied a common upper range, with best mean accuracies of $0.91745$,
$0.90873$, $0.90755$, and $0.90849$. The latter three group leaders
differed by less than 0.12 percentage points. FSGD ranked approximately
0.9 percentage points above this cluster, but its uncertainty interval
still overlapped those of the following methods.

The explicit GL-Memory optimizers occupied the lower part of the ranking.
MemoryFSGD was strongest at $0.88703 \pm 0.03728$, followed by
MemoryFAdam at $0.84788 \pm 0.03379$, MemoryFRMSprop at
$0.81038 \pm 0.04571$, and MemoryFAdadelta at
$0.79764 \pm 0.06391$. The larger standard deviation of MemoryFAdadelta
indicates additional instability across the repeated splits, and the
results do not indicate a general benefit from explicit gradient-history
aggregation on this dataset.

Across all 21 optimizers, mean accuracy ranged from $0.91745$ for FSGD
to $0.72689$ for AdaGL, a difference of 19.056 percentage points. AdaGL
was separated from the other Related-Work methods, since AOFGD\_SGD,
FCSGD\_GL, and AOFGD\_Adam all achieved mean accuracies above $0.90$;
its low result should therefore be treated as method-specific.

The broader optimizer-group aggregation supports this distinction:
Herrera, Standard, and Adaptive-Memory methods obtained mean accuracies
of $0.82390$, $0.82336$, and $0.82278$, differing by less than
0.12 percentage points, whereas Related-Work reached $0.78025$ and
GL-Memory $0.73442$. Both the best-configuration and full-group views
therefore place explicit GL-Memory below the other main optimizer
families on this task.

\begin{figure}[H]
\centering
\includegraphics[width=\textwidth]{
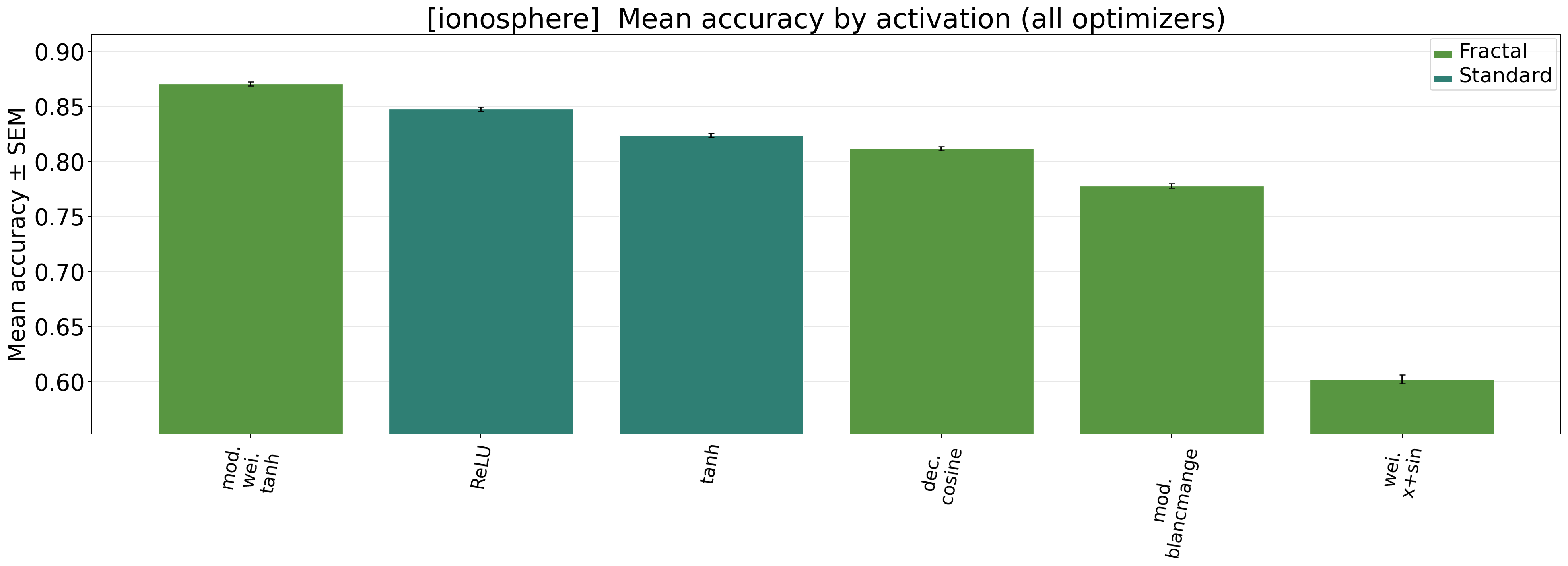}
\caption{Mean test accuracy by activation function on the Ionosphere
dataset, aggregated over all evaluated optimizer configurations and
fractional-order settings. Error bars show the standard error of the mean
for the aggregated run-level accuracies. Colours distinguish standard and
fractal activation functions.}
\label{fig:nn_ionosphere_activation_accuracy}
\end{figure}

Figure~\ref{fig:nn_ionosphere_activation_accuracy} aggregates the results
by activation function across all optimizer configurations and
fractional-order settings.

At activation level, modified Weierstrass--tanh obtained the highest
aggregated mean accuracy at approximately $0.869$. ReLU was the strongest
standard activation at approximately $0.846$, about 2.24 percentage
points below modified Weierstrass--tanh. Tanh followed at approximately
$0.822$, decaying cosine at approximately $0.810$, and modulated
Blancmange and Weierstrass--Mandelbrot $x+\sin$ at approximately $0.777$
and $0.601$.

The small error bars in
Figure~\ref{fig:nn_ionosphere_activation_accuracy} indicate that these
activation-level differences persisted after aggregation across optimizer
configurations and fractional orders. Modified Weierstrass--tanh was
therefore not limited to one favourable optimizer pairing, but produced
the highest average accuracy over the full grid and also appeared in the
overall best configuration.

When combined into families, the two standard activations obtained a mean
accuracy of $0.83559$, compared with $0.76540$ for the four fractal
activations. This family-level difference of 7.019 percentage points
does not contradict the first-place result of modified
Weierstrass--tanh, because the fractal-family average includes the much
lower Weierstrass--Mandelbrot $x+\sin$ result and the lower modulated
Blancmange and decaying cosine results. The individual activation ranking
is therefore more informative than the binary family division for this
dataset.

For the Ionosphere dataset, the highest-ranked configuration combined a
Herrera-type fractional optimizer with the modified Weierstrass--tanh
activation. The same activation also achieved the highest aggregated
accuracy across all optimizers, indicating broader compatibility than a
single tuned pairing. The fractional FSGD configuration nevertheless
provided only a small mean improvement over faster FAdam and Adam
configurations relative to the observed variation. The results therefore
support the use of modified Weierstrass--tanh on this task, but they do
not show a general advantage for every fractional optimizer design.
In particular, the explicit GL-Memory methods performed below the direct
fractional, standard, and adaptive-memory alternatives.

\subsubsection{Iris}
\label{subsubsec:nn_results_iris}

The Iris dataset
(\href{https://www.openml.org/d/61}{OpenML ID 61}) contains 150 flower
samples described by four continuous measurements: sepal length, sepal
width, petal length, and petal width. The target distinguishes the three
species \textit{Iris setosa}, \textit{Iris versicolor}, and
\textit{Iris virginica}. The experiment comprised 222 unique
optimizer--activation--order configurations, each evaluated over 40
repeated runs, resulting in 8880 individual runs.

Table~\ref{tab:nn_iris_top15} reports the 15 highest-ranked
optimizer-specific configurations, retaining only the best
activation--order setting for each optimizer.

\begin{table}[H]
\caption{Top 15 optimizer-specific configurations on the Iris dataset,
ranked by mean test accuracy over 40 repeated runs. For each optimizer,
only its highest-ranked activation--order configuration is shown.
Standard deviation quantifies variation across the repeated runs. For
optimizers without a swept fractional-order parameter, $\nu=1.00$
denotes the recorded default or integer-order setting.}
\label{tab:nn_iris_top15}
\small
\begingroup
\setlength{\tabcolsep}{4pt}
\begin{tabularx}{\textwidth}{
    >{\centering\arraybackslash}p{0.8cm}
    >{\raggedright\arraybackslash}X
    l
    l
    c
    >{\raggedleft\arraybackslash}p{1.25cm}
    r
}
\toprule
\textbf{Rank} &
\textbf{Optimizer} &
\textbf{Group} &
\textbf{Activation} &
\boldmath$\boldsymbol{\nu}$ &
\textbf{Avg. Acc.} &
\textbf{Std.} \\
\midrule
\textbf{1} &
\textbf{FAdam} &
\textbf{Herrera} &
\textbf{dec.\ cosine} &
\textbf{1.25} &
\textbf{0.93444} &
\textbf{0.03374} \\

2  & FRMSprop
   & Herrera         & dec.\ cosine     & 1.50 & 0.93333 & 0.04905 \\
3  & AOFGD\_Adam
   & Related-Work    & dec.\ cosine     & 1.00 & 0.93278 & 0.03782 \\
4  & RMSprop
   & Standard        & dec.\ cosine     & 1.00 & 0.93222 & 0.04087 \\
5  & AdaptiveMemory\allowbreak FRMSprop
   & Adaptive-Memory & dec.\ cosine     & 1.00 & 0.93222 & 0.03336 \\
6  & Adadelta
   & Standard        & dec.\ cosine     & 1.00 & 0.93222 & 0.03411 \\
7  & FAdadelta
   & Herrera         & dec.\ cosine     & 1.25 & 0.93167 & 0.03464 \\
8  & AdaptiveMemory\allowbreak FAdam
   & Adaptive-Memory & dec.\ cosine     & 1.00 & 0.92944 & 0.03482 \\
9  & AdaptiveMemory\allowbreak FAdadelta
   & Adaptive-Memory & dec.\ cosine     & 1.00 & 0.92722 & 0.03299 \\
10 & Adam
   & Standard        & dec.\ cosine     & 1.00 & 0.92389 & 0.04771 \\
11 & SGD
   & Standard        & mod.\ wei.\ tanh & 1.00 & 0.92167 & 0.07941 \\
12 & FSGD
   & Herrera         & mod.\ wei.\ tanh & 1.25 & 0.92000 & 0.06180 \\
13 & AOFGD\_SGD
   & Related-Work    & mod.\ wei.\ tanh & 1.00 & 0.91556 & 0.06623 \\
14 & FCAdam\_GL
   & Related-Work    & dec.\ cosine     & 1.00 & 0.91389 & 0.04672 \\
15 & AdaptiveMemory\allowbreak FSGD
   & Adaptive-Memory & mod.\ wei.\ tanh & 1.00 & 0.91389 & 0.07767 \\
\bottomrule
\end{tabularx}
\endgroup
\end{table}

The highest mean accuracy was obtained by FAdam with the decaying cosine
activation and fractional order $\nu=1.25$, reaching
$0.93444 \pm 0.03374$. FRMSprop with the same activation and
$\nu=1.50$ ranked second at $0.93333 \pm 0.04905$, while AOFGD\_Adam
ranked third at $0.93278 \pm 0.03782$. The 0.111-percentage-point
difference between the first two configurations was small relative to the
run-to-run standard deviations, and their 95\% confidence intervals,
$[0.92399,0.94490]$ and $[0.91813,0.94853]$, overlapped substantially.

RMSprop was the strongest standard optimizer at $0.93222 \pm 0.04087$
with decaying cosine, only 0.222 percentage points below FAdam.
AdaptiveMemoryFRMSprop and Adadelta obtained the same displayed mean
accuracy of $0.93222$, with standard deviations of $0.03336$ and
$0.03411$. The first six configurations therefore represented the
Herrera, Related-Work, Standard, and Adaptive-Memory groups within a
0.222-percentage-point interval.

The top 15 contained four Standard, four Herrera, four Adaptive-Memory,
and three Related-Work optimizers; none of the explicit GL-Memory
optimizers entered this group. Decaying cosine occurred in 11 of the
15 configurations, including the best configuration from each of the four
represented optimizer groups. The remaining four entries used modified
Weierstrass--tanh, and no standard activation appeared among the top 15.

The first- and fifteenth-ranked configurations differed by
2.055 percentage points. The first nine configurations all exceeded
$0.927$ mean accuracy and had overlapping uncertainty intervals. A more
visible decrease occurred for the SGD-type configurations from rank 11
onward, which also showed larger standard deviations, especially SGD
($0.07941$) and AdaptiveMemoryFSGD ($0.07767$), compared with values near
$0.033$ for several leading configurations.

The macro-F1 results closely followed the accuracy ranking: FAdam
obtained $0.93306$, followed by FRMSprop at $0.93220$, AOFGD\_Adam at
$0.93129$, and RMSprop at $0.93112$. The leading FAdam configuration
required $21.97$~s of mean training time, compared with $22.15$~s for
FRMSprop and $21.34$~s for RMSprop, so runtime differences among the
leading decaying-cosine configurations were moderate. SGD with modified
Weierstrass--tanh was faster at $10.40$~s, but its mean accuracy was
1.277 percentage points below FAdam and its variation across runs was
larger.

\begin{figure}[H]
\centering
\includegraphics[width=\textwidth]{
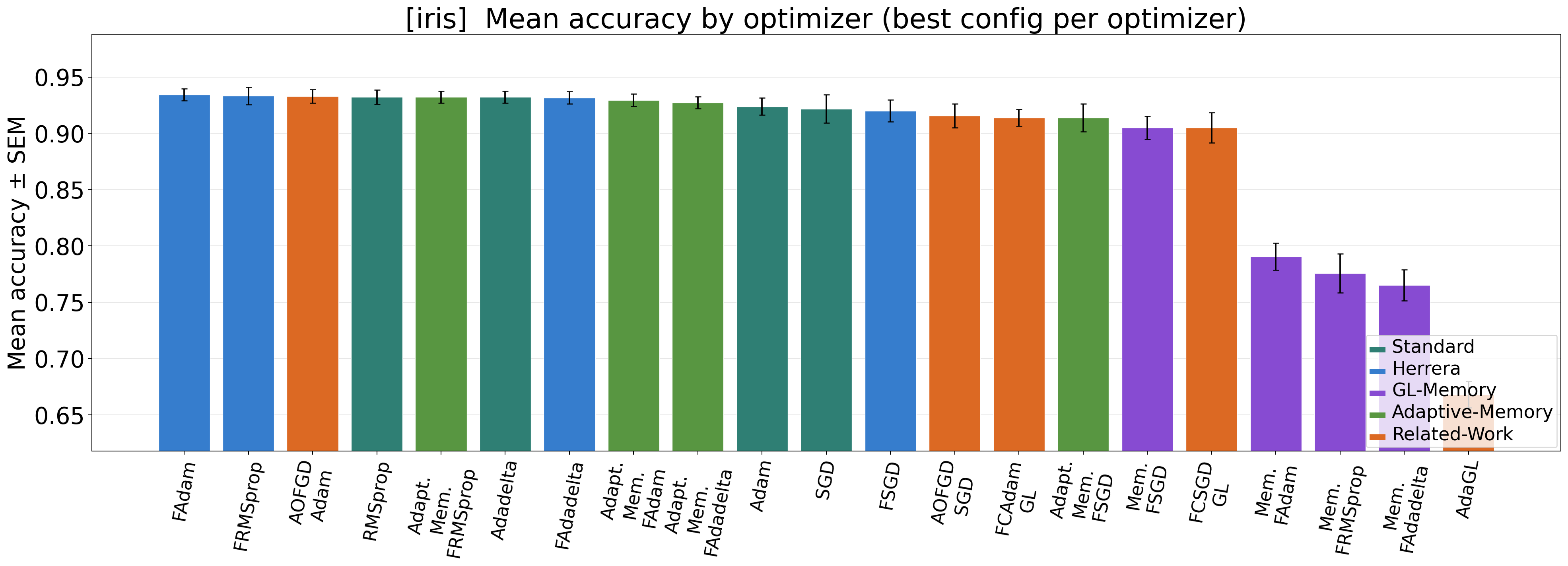}
\caption{Mean test accuracy by optimizer on the Iris dataset. Each bar
represents the highest-performing activation--fractional-order
configuration identified for that optimizer. Optimizers are ordered by
mean accuracy, error bars show the standard error of the mean over
40 repeated runs, and bar colours indicate the optimizer groups.}
\label{fig:nn_iris_optimizer_accuracy}
\end{figure}

Figure~\ref{fig:nn_iris_optimizer_accuracy} presents the complete
optimizer-level ranking, including all 21 optimizers.

Figure~\ref{fig:nn_iris_optimizer_accuracy} shows that the strongest
Herrera, Standard, Adaptive-Memory, and Related-Work configurations
occupied a common upper range, with best mean accuracies of $0.93444$,
$0.93222$, $0.93222$, and $0.93278$. The difference between these four
group leaders was at most 0.222 percentage points, and the overlapping
error bars indicate closely grouped optimizer-specific results across
the repeated splits.

The explicit GL-Memory methods occupied the lower part of the ranking.
MemoryFSGD was strongest in this group at $0.90500 \pm 0.06483$, while
MemoryFAdam, MemoryFRMSprop, and MemoryFAdadelta obtained mean accuracies
of $0.79056$, $0.77556$, and $0.76500$. The latter methods also showed
comparatively large run-to-run variation, particularly MemoryFRMSprop
with a standard deviation of $0.10942$. These results do not indicate a
general benefit from explicit gradient-history aggregation on Iris.

Across the complete optimizer ranking, mean accuracy ranged from
$0.93444$ for FAdam to $0.66778$ for AdaGL, a difference of
26.666 percentage points. AdaGL was separated from the other Related-Work
methods, since AOFGD\_Adam ranked third and both AOFGD\_SGD and
FCAdam\_GL remained above $0.91$; its low result should therefore be read
as method-specific.

The optimizer-group aggregation gives a broader view. Standard and
Herrera optimizers achieved almost identical aggregate mean accuracies of
$0.76514$ and $0.76513$, differing by only 0.001 percentage points.
Adaptive-Memory and Related-Work followed at $0.71963$ and $0.71278$,
while GL-Memory reached $0.57952$. The Herrera group obtained a slightly
higher aggregate macro-F1 than the Standard group, $0.72828$ versus
$0.72503$, but the first-place FAdam configuration did not translate into
a broad accuracy advantage for the complete Herrera family.

\begin{figure}[H]
\centering
\includegraphics[width=\textwidth]{
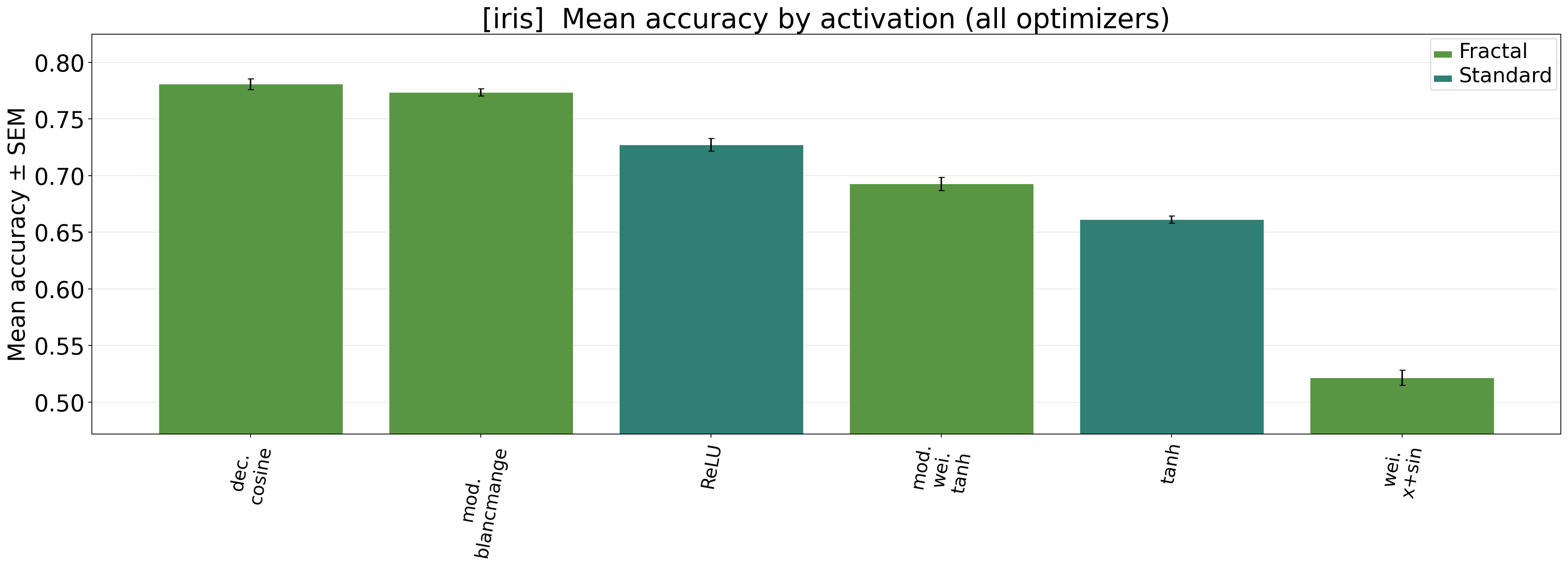}
\caption{Mean test accuracy by activation function on the Iris dataset,
aggregated over all evaluated optimizer configurations and
fractional-order settings. Error bars show the standard error of the mean
for the aggregated run-level accuracies. Colours distinguish standard
and fractal activation functions.}
\label{fig:nn_iris_activation_accuracy}
\end{figure}

Figure~\ref{fig:nn_iris_activation_accuracy} aggregates the results by
activation function across all optimizer configurations and
fractional-order settings.

At activation level, decaying cosine obtained the highest aggregated mean
accuracy at approximately $0.779$, followed by modulated Blancmange at
approximately $0.772$. ReLU was the strongest standard activation at
approximately $0.726$, about 5.3 percentage points below decaying
cosine. Modified Weierstrass--tanh reached approximately $0.691$,
followed by tanh at approximately $0.660$ and Weierstrass--Mandelbrot
$x+\sin$ at approximately $0.520$.

The small difference of approximately 0.7 percentage points between
decaying cosine and modulated Blancmange indicates that both functions
performed well across many optimizer configurations. Decaying cosine
nevertheless connected the configuration-level and activation-level
results most strongly: it appeared in the overall winning configuration,
in most leading optimizer-specific configurations, and achieved the
highest aggregated activation accuracy.

When combined into families, standard activations obtained a mean
accuracy of $0.69432$, compared with $0.69225$ for fractal activations, a
difference of only 0.207 percentage points. The fractal family obtained a
slightly higher mean macro-F1 of $0.64962$, compared with $0.64380$ for
the standard family. These broad averages conceal substantial variation:
the strong decaying cosine and modulated Blancmange results were offset
by the lower Weierstrass--Mandelbrot $x+\sin$ result, while the standard
average combined stronger ReLU with lower tanh.

For the Iris dataset, the highest-ranked configuration combined a
Herrera-type fractional optimizer with the decaying cosine activation.
The same activation also produced the highest aggregated accuracy across
the complete optimizer grid, and modulated Blancmange ranked second. This
indicates that the leading fractal activations were not restricted to a
single tuned optimizer pairing. However, the nearly identical aggregate
accuracy of the Standard and Herrera groups shows that the winning
fractional configuration did not establish a general family-level
optimizer advantage. The explicit GL-Memory methods performed
substantially below the direct fractional and standard alternatives,
demonstrating that the form in which fractional structure is introduced
remains important for this dataset.

\subsubsection{Seeds}
\label{subsubsec:nn_results_seeds}

The Seeds dataset
(\href{https://www.openml.org/d/1499}{OpenML ID 1499}) contains 210
wheat-kernel samples described by seven numerical geometric
measurements. The target distinguishes three wheat varieties, resulting
in a small three-class classification problem. The experiment comprised
222 unique optimizer--activation--order configurations, each evaluated
over 40 repeated runs, resulting in 8880 individual runs.

Table~\ref{tab:nn_seeds_top15} reports the 15 highest-ranked
optimizer-specific configurations, retaining only the best
activation--order setting for each optimizer.

\begin{table}[H]
\caption{Top 15 optimizer-specific configurations on the Seeds dataset,
ranked by mean test accuracy over 40 repeated runs. For each optimizer,
only its highest-ranked activation--order configuration is shown.
Standard deviation quantifies variation across the repeated runs. For
optimizers without a swept fractional-order parameter, $\nu=1.00$
denotes the recorded default or integer-order setting.}
\label{tab:nn_seeds_top15}
\small
\begingroup
\setlength{\tabcolsep}{4pt}
\begin{tabularx}{\textwidth}{
    >{\centering\arraybackslash}p{0.8cm}
    >{\raggedright\arraybackslash}X
    l
    l
    c
    >{\raggedleft\arraybackslash}p{1.25cm}
    r
}
\toprule
\textbf{Rank} &
\textbf{Optimizer} &
\textbf{Group} &
\textbf{Activation} &
\boldmath$\boldsymbol{\nu}$ &
\textbf{Avg. Acc.} &
\textbf{Std.} \\
\midrule
\textbf{1} &
\textbf{FSGD} &
\textbf{Herrera} &
\textbf{tanh} &
\textbf{0.75} &
\textbf{0.91230} &
\textbf{0.04222} \\

2  & AdaptiveMemory\allowbreak FSGD
   & Adaptive-Memory & mod.\ wei.\ tanh & 1.00 & 0.90595 & 0.04104 \\
3  & FAdam
   & Herrera         & tanh             & 1.50 & 0.90476 & 0.04915 \\
4  & FAdadelta
   & Herrera         & tanh             & 1.50 & 0.90397 & 0.04503 \\
5  & SGD
   & Standard        & mod.\ wei.\ tanh & 1.00 & 0.89722 & 0.04941 \\
6  & FRMSprop
   & Herrera         & dec.\ cosine     & 0.75 & 0.89683 & 0.03938 \\
7  & Adam
   & Standard        & dec.\ cosine     & 1.00 & 0.89524 & 0.05119 \\
8  & AdaptiveMemory\allowbreak FAdam
   & Adaptive-Memory & dec.\ cosine     & 1.00 & 0.89444 & 0.04277 \\
9  & RMSprop
   & Standard        & dec.\ cosine     & 1.00 & 0.89405 & 0.04349 \\
10 & AdaptiveMemory\allowbreak FRMSprop
   & Adaptive-Memory & dec.\ cosine     & 1.00 & 0.89167 & 0.04874 \\
11 & AdaptiveMemory\allowbreak FAdadelta
   & Adaptive-Memory & dec.\ cosine     & 1.00 & 0.89167 & 0.04297 \\
12 & AOFGD\_Adam
   & Related-Work    & dec.\ cosine     & 1.00 & 0.89127 & 0.04625 \\
13 & Adadelta
   & Standard        & dec.\ cosine     & 1.00 & 0.88968 & 0.04503 \\
14 & FCAdam\_GL
   & Related-Work    & dec.\ cosine     & 1.00 & 0.87302 & 0.04942 \\
15 & AOFGD\_SGD
   & Related-Work    & mod.\ wei.\ tanh & 1.00 & 0.87024 & 0.08634 \\
\bottomrule
\end{tabularx}
\endgroup
\end{table}

The highest mean accuracy was obtained by FSGD with tanh and fractional
order $\nu=0.75$, reaching $0.91230 \pm 0.04222$.
AdaptiveMemoryFSGD with modified Weierstrass--tanh ranked second at
$0.90595 \pm 0.04104$, followed by FAdam with tanh and $\nu=1.50$ at
$0.90476 \pm 0.04915$. The 0.635-percentage-point gap between the first
two configurations was small relative to the run-to-run standard
deviations, and their 95\% confidence intervals,
$[0.89922,0.92539]$ and $[0.89323,0.91867]$, overlapped.

SGD was the strongest standard optimizer at $0.89722 \pm 0.04941$ with
modified Weierstrass--tanh. AdaptiveMemoryFSGD was the strongest
Adaptive-Memory method, AOFGD\_Adam the highest-ranked Related-Work
method, and MemoryFSGD the strongest GL-Memory method. The best
standard-activation configuration was the overall winner, whereas the
best fractal-activation configuration was AdaptiveMemoryFSGD in second
place.

The top 15 contained four Standard, four Herrera, four Adaptive-Memory,
and three Related-Work optimizers; none of the explicit GL-Memory methods
entered this group. Fractal activations nevertheless formed the majority
of the top 15: decaying cosine appeared in nine entries, and modified
Weierstrass--tanh in ranks two, five, and fifteen. Tanh appeared in the
first, third, and fourth positions, so the leading configuration did not
use the activation that occurred most frequently in the upper ranking.

The first- and fifteenth-ranked configurations differed by
4.206 percentage points, but the first 13 formed a compact range between
$0.88968$ and $0.91230$. A larger decrease occurred for FCAdam\_GL and
AOFGD\_SGD in ranks 14 and 15. The standard deviation of AOFGD\_SGD was
also substantially larger than those of most preceding configurations,
indicating greater sensitivity to the repeated train--test partitions.

The macro-F1 results closely followed the accuracy ranking. The winning
FSGD configuration obtained $0.91112$, followed by AdaptiveMemoryFSGD at
$0.90509$, FAdam at $0.90324$, and FAdadelta at $0.90265$. FSGD required
only $1.29$~s of mean training time, compared with $9.91$~s for
AdaptiveMemoryFSGD, giving a training-time reduction of approximately
$8.63$~s relative to the second-ranked method.

\begin{figure}[H]
\centering
\includegraphics[width=\textwidth]{
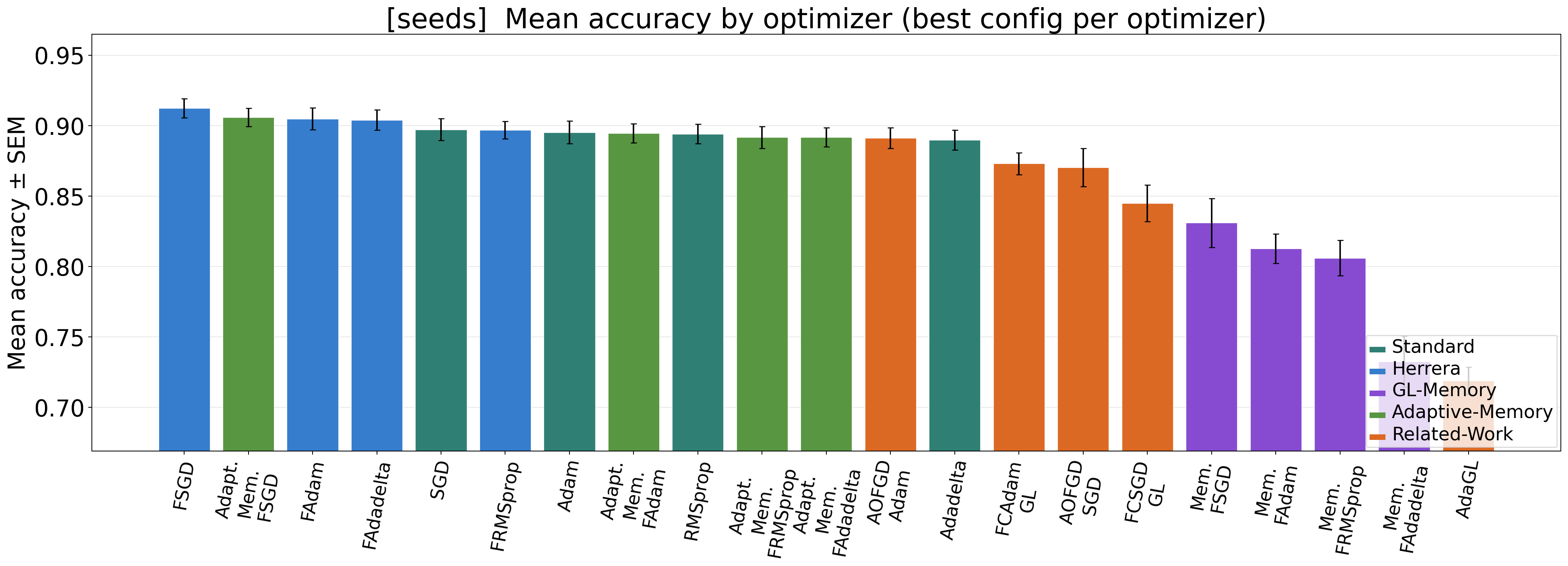}
\caption{Mean test accuracy by optimizer on the Seeds dataset. Each bar
represents the highest-performing activation--fractional-order
configuration identified for that optimizer. Optimizers are ordered by
mean accuracy, error bars show the standard error of the mean over
40 repeated runs, and bar colours indicate the optimizer groups.}
\label{fig:nn_seeds_optimizer_accuracy}
\end{figure}

Figure~\ref{fig:nn_seeds_optimizer_accuracy} presents the complete
optimizer-level ranking, including all 21 optimizers.

Figure~\ref{fig:nn_seeds_optimizer_accuracy} shows that the strongest
Herrera, Adaptive-Memory, Standard, and Related-Work configurations
reached $0.91230$, $0.90595$, $0.89722$, and $0.89127$, respectively.
The Herrera result was 0.635 percentage points above the strongest
Adaptive-Memory configuration and 1.508 percentage points above the
strongest Standard configuration, but the uncertainty intervals
overlapped.

The explicit GL-Memory optimizers occupied the lower part of the ranking.
MemoryFSGD was strongest at $0.83095 \pm 0.10965$, followed by
MemoryFAdam at $0.81270 \pm 0.06522$, MemoryFRMSprop at
$0.80595 \pm 0.07945$, and MemoryFAdadelta at
$0.73254 \pm 0.11466$. These methods also showed greater run-to-run
variation than most leading configurations, so the results do not
indicate a general benefit from explicit gradient-history aggregation.

Across all 21 optimizers, mean accuracy ranged from $0.91230$ for FSGD
to $0.71905$ for AdaGL, a difference of 19.325 percentage points. AdaGL
was separated from the other Related-Work methods, since AOFGD\_Adam
reached $0.89127$ and both FCAdam\_GL and AOFGD\_SGD remained above
$0.87$; its lower result should therefore be read as method-specific.

The optimizer-group aggregation gives a broader view. Herrera optimizers
obtained the highest aggregate mean accuracy at $0.77962$, followed by
the Standard group at $0.76968$, a difference of 0.994 percentage
points. Related-Work and Adaptive-Memory reached $0.72771$ and $0.72409$,
while GL-Memory reached $0.58161$. The aggregate macro-F1 ranking
followed the same pattern, with $0.75643$ for Herrera, $0.74547$ for
Standard, and $0.51984$ for GL-Memory.

\begin{figure}[H]
\centering
\includegraphics[width=\textwidth]{
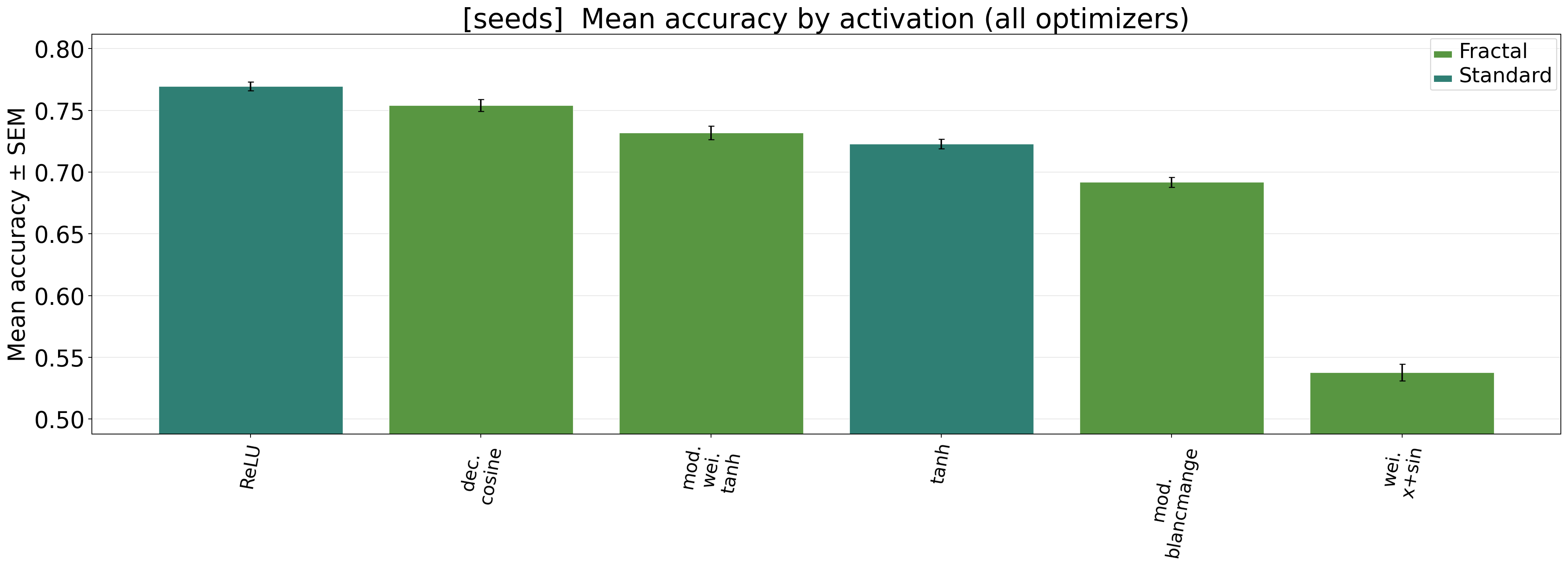}
\caption{Mean test accuracy by activation function on the Seeds dataset,
aggregated over all evaluated optimizer configurations and
fractional-order settings. Error bars show the standard error of the
mean for the aggregated run-level accuracies. Colours distinguish
standard and fractal activation functions.}
\label{fig:nn_seeds_activation_accuracy}
\end{figure}

Figure~\ref{fig:nn_seeds_activation_accuracy} aggregates the results by
activation function across all optimizer configurations and
fractional-order settings.

At activation level, ReLU obtained the highest aggregated mean accuracy
at approximately $0.769$. Decaying cosine was the strongest fractal
activation at approximately $0.754$, about 1.5 percentage points below
ReLU. Modified Weierstrass--tanh followed at approximately $0.731$, tanh
at approximately $0.722$, and modulated Blancmange and
Weierstrass--Mandelbrot $x+\sin$ at approximately $0.691$ and $0.538$.

The activation-level ranking differs from the optimizer-specific
ranking. Tanh appeared in the overall winning FSGD configuration, but
ReLU obtained the highest mean over the complete optimizer grid.
Similarly, decaying cosine appeared in most of the middle and upper
optimizer-specific configurations and was the strongest fractal
activation, but it did not occur in the first five positions. Activation
performance therefore depended on the optimizer pairing.

When combined into families, standard activations obtained a mean
accuracy of $0.74617$, compared with $0.67888$ for fractal activations, a
difference of 6.729 percentage points. The standard family also obtained
a higher mean macro-F1 of $0.71755$, compared with $0.63553$ for the
fractal family. This broad difference was partly caused by the lower
Weierstrass--Mandelbrot $x+\sin$ result; decaying cosine and modified
Weierstrass--tanh remained considerably closer to the standard functions
than the family mean suggests.

For the Seeds dataset, the highest-ranked configuration combined a
Herrera-type fractional optimizer with the standard tanh activation. The
same optimizer family also achieved the highest aggregate accuracy,
although its advantage over the Standard group remained below one
percentage point. ReLU was the strongest activation after aggregation,
while decaying cosine was the strongest fractal alternative. The results
therefore do not indicate a general fractal-activation advantage on this
dataset. They instead show that direct fractional optimization can
produce a competitive and computationally efficient result when paired
with a suitable standard activation, whereas the explicit GL-Memory
methods performed below the direct fractional, adaptive-memory, and
standard alternatives.

\subsubsection{Vertebral Column}
\label{subsubsec:nn_results_vertebra}

The Vertebral Column dataset
(\href{https://www.openml.org/d/1524}{OpenML ID 1524}) contains 310
samples described by six numerical biomechanical attributes. The binary
target distinguishes normal from abnormal spinal conditions. The
experiment comprised 222 unique optimizer--activation--order
configurations, each evaluated over 40 repeated runs, resulting in
8880 individual runs.

Table~\ref{tab:nn_vertebra_top15} reports the 15 highest-ranked
optimizer-specific configurations, retaining only the best
activation--order setting for each optimizer.

\begin{table}[H]
\caption{Top 15 optimizer-specific configurations on the Vertebral Column
dataset, ranked by mean test accuracy over 40 repeated runs. For each
optimizer, only its highest-ranked activation--order configuration is
shown. Standard deviation quantifies variation across the repeated runs.
For optimizers without a swept fractional-order parameter, $\nu=1.00$
denotes the recorded default or integer-order setting.}
\label{tab:nn_vertebra_top15}
\small
\begingroup
\setlength{\tabcolsep}{4pt}
\begin{tabularx}{\textwidth}{
    >{\centering\arraybackslash}p{0.8cm}
    >{\raggedright\arraybackslash}X
    l
    l
    c
    >{\raggedleft\arraybackslash}p{1.25cm}
    r
}
\toprule
\textbf{Rank} &
\textbf{Optimizer} &
\textbf{Group} &
\textbf{Activation} &
\boldmath$\boldsymbol{\nu}$ &
\textbf{Avg. Acc.} &
\textbf{Std.} \\
\midrule
\textbf{1} &
\textbf{FAdam} &
\textbf{Herrera} &
\textbf{tanh} &
\textbf{1.50} &
\textbf{0.83763} &
\textbf{0.03527} \\

2  & FAdadelta
   & Herrera         & tanh             & 1.50 & 0.83360 & 0.02932 \\
3  & AOFGD\_Adam
   & Related-Work    & tanh             & 1.00 & 0.82366 & 0.03355 \\
4  & FCAdam\_GL
   & Related-Work    & tanh             & 1.00 & 0.82124 & 0.03526 \\
5  & AdaptiveMemory\allowbreak FRMSprop
   & Adaptive-Memory & mod.\ wei.\ tanh & 1.00 & 0.82016 & 0.04536 \\
6  & FSGD
   & Herrera         & tanh             & 0.75 & 0.81935 & 0.04583 \\
7  & AdaptiveMemory\allowbreak FAdadelta
   & Adaptive-Memory & tanh             & 1.00 & 0.81774 & 0.03732 \\
8  & RMSprop
   & Standard        & mod.\ wei.\ tanh & 1.00 & 0.81425 & 0.04168 \\
9  & Adadelta
   & Standard        & tanh             & 1.00 & 0.81398 & 0.03943 \\
10 & AOFGD\_SGD
   & Related-Work    & tanh             & 1.00 & 0.81398 & 0.04908 \\
11 & AdaptiveMemory\allowbreak FAdam
   & Adaptive-Memory & dec.\ cosine     & 1.00 & 0.81210 & 0.04349 \\
12 & FRMSprop
   & Herrera         & mod.\ wei.\ tanh & 1.25 & 0.81183 & 0.04665 \\
13 & Adam
   & Standard        & tanh             & 1.00 & 0.81075 & 0.04341 \\
14 & SGD
   & Standard        & tanh             & 1.00 & 0.81048 & 0.05187 \\
15 & AdaptiveMemory\allowbreak FSGD
   & Adaptive-Memory & tanh             & 1.00 & 0.80941 & 0.05307 \\
\bottomrule
\end{tabularx}
\endgroup
\end{table}

The highest mean accuracy was obtained by FAdam with tanh and fractional
order $\nu=1.50$, reaching $0.83763 \pm 0.03527$. FAdadelta with the
same activation and order ranked second at $0.83360 \pm 0.02932$, while
AOFGD\_Adam with tanh ranked third at $0.82366 \pm 0.03355$. The
0.403-percentage-point difference between the first two configurations
was small relative to run-to-run variation, and their 95\% confidence
intervals, $[0.82670,0.84856]$ and $[0.82452,0.84269]$, overlapped
substantially.

RMSprop was the strongest standard optimizer at $0.81425 \pm 0.04168$
with modified Weierstrass--tanh. The best standard-activation
configuration was the overall winner, while the best fractal-activation
configuration was AdaptiveMemoryFRMSprop with modified Weierstrass--tanh
in fifth place at $0.82016 \pm 0.04536$, 1.747 percentage points below
FAdam.

The top 15 contained four Standard, four Herrera, four Adaptive-Memory,
and three Related-Work optimizers; none of the explicit GL-Memory
optimizers entered this group. Standard activations dominated the table:
tanh appeared in 11 of the 15 configurations, including the four highest
entries and the best Herrera and Related-Work results. Modified
Weierstrass--tanh occurred in three entries, decaying cosine once, and
ReLU, modulated Blancmange, and Weierstrass--Mandelbrot $x+\sin$ did not
appear.

The first- and fifteenth-ranked configurations differed by
2.822 percentage points. The first seven configurations reached mean
accuracies above $0.817$, while ranks eight to fifteen formed a second
closely grouped range between $0.80941$ and $0.81425$. Because the
uncertainty intervals overlapped throughout much of the table, the
ranking shows small differences among leading configurations rather than
clearly separated performance levels.

The macro-F1 results generally supported the accuracy ranking. FAdam
obtained the highest mean macro-F1 of $0.79949$, while
AdaptiveMemoryFRMSprop reached $0.78617$, slightly above the $0.78576$
of second-ranked FAdadelta. The winning FAdam configuration required
$1.81$~s of mean training time, compared with $1.65$~s for FAdadelta,
$2.62$~s for AOFGD\_Adam, and $9.61$~s for RMSprop, combining the
highest mean accuracy and macro-F1 with a moderate training time.

\begin{figure}[H]
\centering
\includegraphics[width=\textwidth]{
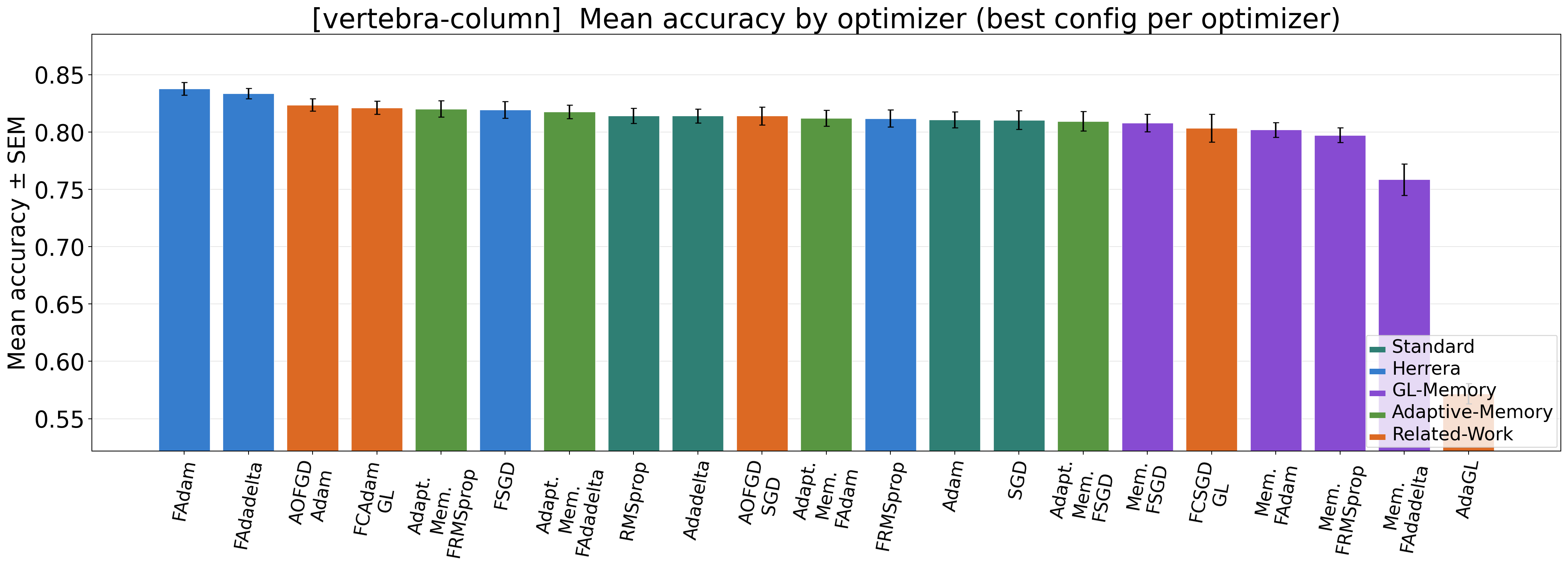}
\caption{Mean test accuracy by optimizer on the Vertebral Column
dataset. Each bar represents the highest-performing
activation--fractional-order configuration identified for that
optimizer. Optimizers are ordered by mean accuracy, error bars show the
standard error of the mean over 40 repeated runs, and bar colours
indicate the optimizer groups.}
\label{fig:nn_vertebra_optimizer_accuracy}
\end{figure}

Figure~\ref{fig:nn_vertebra_optimizer_accuracy} presents the complete
optimizer-level ranking, including all 21 optimizers.

Figure~\ref{fig:nn_vertebra_optimizer_accuracy} shows that the strongest
Herrera, Related-Work, Adaptive-Memory, Standard, and GL-Memory
configurations reached $0.83763$, $0.82366$, $0.82016$, $0.81425$, and
$0.80780$, respectively. FAdam therefore exceeded the best configuration
from every other optimizer group, but the error bars of the leading
methods still overlapped.

The explicit GL-Memory optimizers occupied the lower part of the ranking.
MemoryFSGD was strongest in this group at $0.80780 \pm 0.04828$,
followed by MemoryFAdam at $0.80188 \pm 0.04117$, MemoryFRMSprop at
$0.79731 \pm 0.04079$, and MemoryFAdadelta at
$0.75860 \pm 0.08688$. MemoryFAdadelta had the largest standard
deviation within this group, indicating greater sensitivity to the
repeated data partitions, and the results do not indicate a general
benefit from explicit gradient-history aggregation.

Across all 21 optimizers, mean accuracy ranged from $0.83763$ for FAdam
to $0.57177$ for AdaGL, a difference of 26.586 percentage points. AdaGL
was separated from the other Related-Work methods, since AOFGD\_Adam and
FCAdam\_GL ranked third and fourth, while AOFGD\_SGD also remained above
$0.81$; its low result should therefore be read as method-specific.

The optimizer-group aggregation gives a broader view. Herrera optimizers
obtained the highest aggregate mean accuracy at $0.72565$, followed by
the Standard group at $0.71041$, a difference of 1.524 percentage
points. Adaptive-Memory and Related-Work reached $0.66943$ and
$0.63028$, while GL-Memory reached $0.53332$. The aggregate macro-F1
ranking followed the same pattern, with $0.65777$ for Herrera, $0.63982$
for Standard, and $0.44991$ for GL-Memory.

\begin{figure}[H]
\centering
\includegraphics[width=\textwidth]{
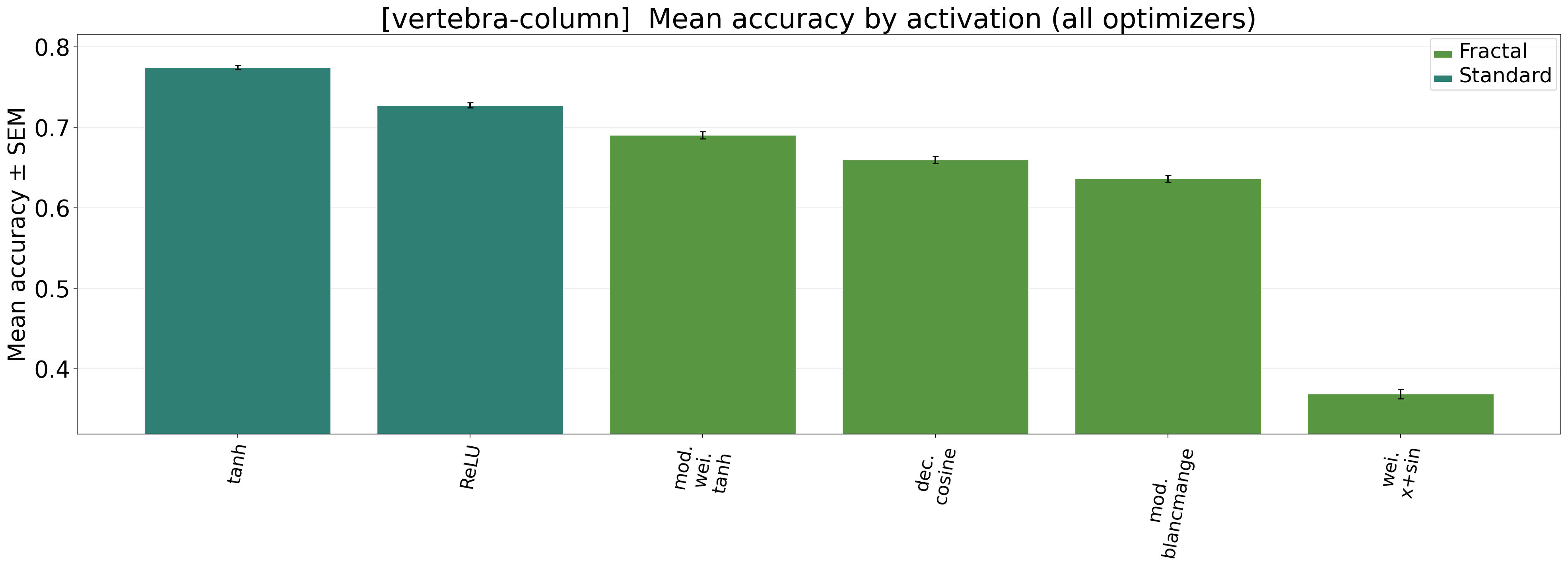}
\caption{Mean test accuracy by activation function on the Vertebral
Column dataset, aggregated over all evaluated optimizer configurations
and fractional-order settings. Error bars show the standard error of
the mean for the aggregated run-level accuracies. Colours distinguish
standard and fractal activation functions.}
\label{fig:nn_vertebra_activation_accuracy}
\end{figure}

Figure~\ref{fig:nn_vertebra_activation_accuracy} aggregates the results
by activation function across all optimizer configurations and
fractional-order settings.

At activation level, tanh obtained the highest aggregated mean accuracy
at approximately $0.773$, followed by ReLU at approximately $0.726$.
Modified Weierstrass--tanh was the strongest fractal function at
approximately $0.689$, about 8.4 percentage points below tanh. Decaying
cosine and modulated Blancmange reached approximately $0.659$ and
$0.635$, while Weierstrass--Mandelbrot $x+\sin$ obtained approximately
$0.369$.

The activation-level ranking is consistent with the configuration-level
result: tanh appeared in the overall winning configuration and achieved
the highest mean across the complete optimizer grid. Modified
Weierstrass--tanh produced several competitive individual configurations,
including the best Adaptive-Memory and Standard optimizer results, but
its lower aggregate mean indicates less consistent performance across
all evaluated update rules.

When combined into families, standard activations obtained a mean
accuracy of $0.75078$, compared with $0.58860$ for fractal activations, a
difference of 16.218 percentage points. The standard family also achieved
a higher mean macro-F1 of $0.66925$, compared with $0.51956$ for the
fractal family. The low Weierstrass--Mandelbrot $x+\sin$ result
contributed substantially to this difference, but all four fractal
activations remained below tanh in the individual activation ranking.

For the Vertebral Column dataset, the highest-ranked configuration
combined a Herrera-type fractional optimizer with the standard tanh
activation. The Herrera group also achieved the highest aggregate
accuracy, although its advantage over the Standard group was smaller than
the variation observed across individual runs. The activation aggregation
showed a broader advantage for standard activations, particularly tanh.
Fractal activations remained competitive in selected optimizer pairings
but did not provide a general average improvement. The explicit
GL-Memory methods also performed below the direct fractional and standard
alternatives, indicating that the effectiveness of fractional structure
depended on the specific update formulation used for this task.

\subsubsection{Wine Recognition}
\label{subsubsec:nn_results_wine}

The Wine Recognition dataset
(\href{https://www.openml.org/d/187}{OpenML ID 187}) contains 178 wine
samples described by 13 numerical chemical measurements. The target
distinguishes three grape cultivars, resulting in a small three-class
classification problem. The experiment comprised 222 unique
optimizer--activation--order configurations, each evaluated over 40
repeated runs, resulting in 8880 individual runs.

Table~\ref{tab:nn_wine_top15} reports the 15 highest-ranked
optimizer-specific configurations, retaining only the best
activation--order setting for each optimizer.

\begin{table}[H]
\caption{Top 15 optimizer-specific configurations on the Wine Recognition
dataset, ranked by mean test accuracy over 40 repeated runs. For each
optimizer, only its highest-ranked activation--order configuration is
shown. Standard deviation quantifies variation across the repeated runs.
For optimizers without a swept fractional-order parameter, $\nu=1.00$
denotes the recorded default or integer-order setting.}
\label{tab:nn_wine_top15}
\small
\begingroup
\setlength{\tabcolsep}{4pt}
\begin{tabularx}{\textwidth}{
    >{\centering\arraybackslash}p{0.8cm}
    >{\raggedright\arraybackslash}X
    l
    l
    c
    >{\raggedleft\arraybackslash}p{1.25cm}
    r
}
\toprule
\textbf{Rank} &
\textbf{Optimizer} &
\textbf{Group} &
\textbf{Activation} &
\boldmath$\boldsymbol{\nu}$ &
\textbf{Avg. Acc.} &
\textbf{Std.} \\
\midrule
\textbf{1} &
\textbf{AOFGD\_Adam} &
\textbf{Related-Work} &
\textbf{blancmange} &
\textbf{1.00} &
\textbf{0.96111} &
\textbf{0.02807} \\

2  & Adadelta
   & Standard        & blancmange       & 1.00 & 0.95972 & 0.02810 \\
3  & RMSprop
   & Standard        & blancmange       & 1.00 & 0.95694 & 0.03006 \\
4  & AdaptiveMemory\allowbreak FAdadelta
   & Adaptive-Memory & blancmange       & 1.00 & 0.95556 & 0.03478 \\
5  & AdaptiveMemory\allowbreak FAdam
   & Adaptive-Memory & blancmange       & 1.00 & 0.95509 & 0.02932 \\
6  & SGD
   & Standard        & mod.\ wei.\ tanh & 1.00 & 0.95417 & 0.03407 \\
7  & AdaptiveMemory\allowbreak FRMSprop
   & Adaptive-Memory & blancmange       & 1.00 & 0.95417 & 0.03820 \\
8  & FRMSprop
   & Herrera         & blancmange       & 0.75 & 0.95370 & 0.03866 \\
9  & Adam
   & Standard        & blancmange       & 1.00 & 0.95324 & 0.02965 \\
10 & FSGD
   & Herrera         & mod.\ wei.\ tanh & 0.75 & 0.95324 & 0.03774 \\
11 & FAdam
   & Herrera         & blancmange       & 1.25 & 0.95278 & 0.03354 \\
12 & FCAdam\_GL
   & Related-Work    & blancmange       & 1.00 & 0.95046 & 0.03339 \\
13 & AOFGD\_SGD
   & Related-Work    & mod.\ wei.\ tanh & 1.00 & 0.95046 & 0.03149 \\
14 & FAdadelta
   & Herrera         & blancmange       & 1.25 & 0.94630 & 0.03504 \\
15 & AdaptiveMemory\allowbreak FSGD
   & Adaptive-Memory & mod.\ wei.\ tanh & 1.00 & 0.94398 & 0.05041 \\
\bottomrule
\end{tabularx}
\endgroup
\end{table}

The highest mean accuracy was obtained by AOFGD\_Adam with the modulated
Blancmange activation and the recorded order $\nu=1.00$, reaching
$0.96111 \pm 0.02807$. Standard Adadelta with the same activation ranked
second at $0.95972 \pm 0.02810$, and RMSprop with the same activation
ranked third at $0.95694 \pm 0.03006$. The 0.139-percentage-point gap
between the first two configurations was small relative to the
run-to-run standard deviations, and their 95\% confidence intervals,
$[0.95241,0.96981]$ and $[0.95102,0.96843]$, overlapped substantially.

Adadelta was the strongest standard optimizer, AdaptiveMemoryFAdadelta
the strongest Adaptive-Memory method, FRMSprop the highest-ranked
Herrera-type optimizer, and MemoryFSGD the best GL-Memory method. The
best fractal-activation configuration was the overall winner, whereas
the strongest standard-activation configuration was MemoryFSGD with tanh
in sixteenth place at $0.93981 \pm 0.05234$, 2.130 percentage points
below AOFGD\_Adam.

The top 15 contained four Standard, four Herrera, four Adaptive-Memory,
and three Related-Work optimizers; none of the explicit GL-Memory
optimizers entered this group. All top-15 entries used fractal
activations: modulated Blancmange appeared in 11 configurations,
including the five highest-ranked entries and the best configurations of
four optimizer groups, while the remaining four entries used modified
Weierstrass--tanh. Neither ReLU nor tanh appeared in the table.

The first- and fifteenth-ranked configurations differed by
1.713 percentage points, and the first 13 were contained within only
1.065 percentage points. Several optimizer families therefore achieved
similar best-case performance when paired with a suitable fractal
activation, and the overlapping uncertainty intervals mean that the
ranking should be read as an ordering of mean values rather than as
statistically distinct performance levels.

The macro-F1 results closely followed the accuracy ranking. AOFGD\_Adam
obtained $0.96231$, followed by Adadelta at $0.96056$ and RMSprop at
$0.95747$. The winning configuration required $7.19$~s of mean training
time, compared with $5.44$~s for Adadelta and $5.49$~s for RMSprop, so
its small accuracy advantage over Adadelta came with approximately
$1.75$~s of additional mean training time.

\begin{figure}[H]
\centering
\includegraphics[width=\textwidth]{
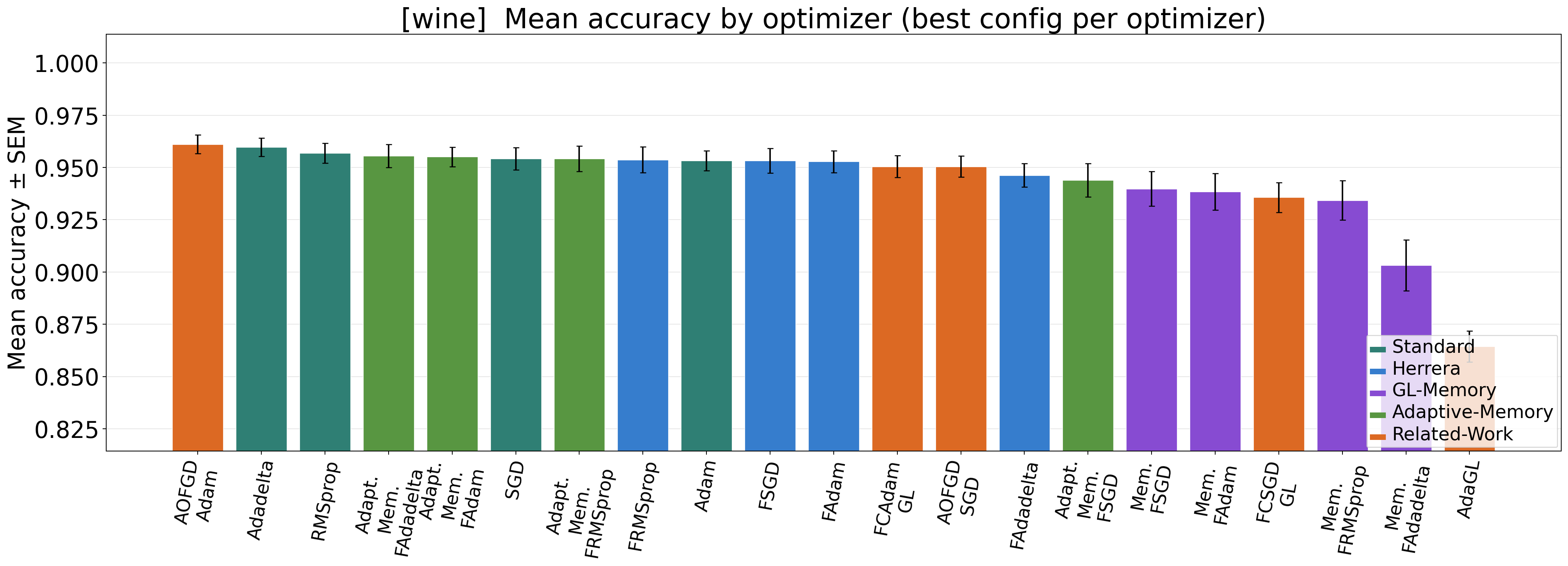}
\caption{Mean test accuracy by optimizer on the Wine Recognition
dataset. Each bar represents the highest-performing
activation--fractional-order configuration identified for that
optimizer. Optimizers are ordered by mean accuracy, error bars show the
standard error of the mean over 40 repeated runs, and bar colours
indicate the optimizer groups.}
\label{fig:nn_wine_optimizer_accuracy}
\end{figure}

Figure~\ref{fig:nn_wine_optimizer_accuracy} presents the complete
optimizer-level ranking, including all 21 optimizers.

Figure~\ref{fig:nn_wine_optimizer_accuracy} shows that the strongest
Related-Work, Standard, Adaptive-Memory, Herrera, and GL-Memory
configurations reached $0.96111$, $0.95972$, $0.95556$, $0.95370$, and
$0.93981$, respectively. The best Related-Work and Standard
configurations differed by only 0.139 percentage points, and the
strongest configurations from the first four groups all exceeded $0.95$,
with overlapping error bars.

The explicit GL-Memory methods occupied a lower part of the ranking.
MemoryFSGD was strongest at $0.93981 \pm 0.05234$, followed by
MemoryFAdam at $0.93843 \pm 0.05506$ and MemoryFRMSprop at
$0.93426 \pm 0.05945$. MemoryFAdadelta reached
$0.90324 \pm 0.07704$ and showed greater run-to-run variation than the
other memory-based variants, so the results do not indicate a general
advantage from explicit gradient-history aggregation.

Across all 21 optimizers, mean accuracy ranged from $0.96111$ for
AOFGD\_Adam to $0.86435$ for AdaGL, a difference of
9.676 percentage points. AdaGL was separated from the other Related-Work
methods, since AOFGD\_Adam ranked first, FCAdam\_GL and AOFGD\_SGD both
exceeded $0.95$, and FCSGD\_GL reached $0.93565$. Its lower result should
therefore be read as method-specific.

The optimizer-group aggregation gives a broader view. Standard optimizers
obtained the highest aggregate mean accuracy at $0.83181$, followed by
Herrera at $0.82683$, a difference of 0.498 percentage points.
Adaptive-Memory and Related-Work reached $0.78823$ and $0.76761$, while
GL-Memory reached $0.58176$. The first-place AOFGD\_Adam configuration
therefore did not translate into a general aggregate advantage for the
complete Related-Work family.

\begin{figure}[H]
\centering
\includegraphics[width=\textwidth]{
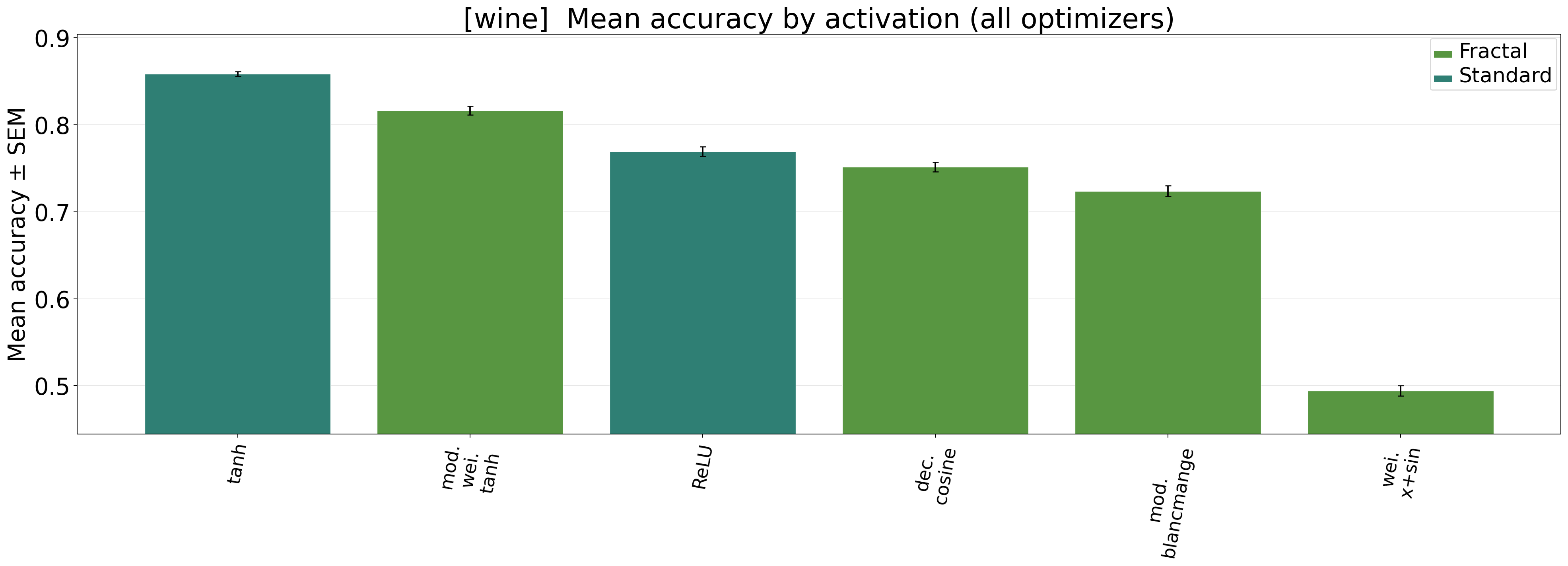}
\caption{Mean test accuracy by activation function on the Wine
Recognition dataset, aggregated over all evaluated optimizer
configurations and fractional-order settings. Error bars show the
standard error of the mean for the aggregated run-level accuracies.
Colours distinguish standard and fractal activation functions.}
\label{fig:nn_wine_activation_accuracy}
\end{figure}

Figure~\ref{fig:nn_wine_activation_accuracy} aggregates the results by
activation function across all optimizer configurations and
fractional-order settings.

At activation level, tanh obtained the highest aggregated mean accuracy
at approximately $0.857$. Modified Weierstrass--tanh was the strongest
fractal function at approximately $0.815$, about 4.2 percentage points
below tanh. ReLU followed at approximately $0.768$, decaying cosine at
approximately $0.751$, and modulated Blancmange and
Weierstrass--Mandelbrot $x+\sin$ at approximately $0.723$ and $0.493$.

The activation-level ranking differs substantially from the
optimizer-specific ranking. Modulated Blancmange appeared in the overall
winning configuration and in 11 of the 15 leading optimizer-specific
configurations, but its aggregate mean across the complete optimizer grid
ranked only fifth among the six activations. Tanh, by contrast, produced
the highest aggregated mean despite not appearing in the top-15 table.
This indicates a strong optimizer--activation interaction: modulated
Blancmange performed well with selected optimizers but was less robust
when averaged across all update rules.

Modified Weierstrass--tanh showed broader average performance than
modulated Blancmange, although its best optimizer-specific result
remained below the leading Blancmange configurations. ReLU and tanh also
differed substantially, showing that the broad Standard category
conceals relevant differences between its two activation functions. The
small SEM error bars indicate that the aggregate activation ranking was
stable across the large set of included runs.

When combined into families, standard activations obtained a mean
accuracy of $0.81398$, compared with $0.69657$ for fractal activations, a
difference of 11.741 percentage points. The standard family also achieved
a higher mean macro-F1 of $0.79321$, compared with $0.65478$ for the
fractal family. This family-level difference was influenced by the low
aggregate result of Weierstrass--Mandelbrot $x+\sin$ and by the limited
cross-optimizer robustness of modulated Blancmange, but it does not
contradict the strong configuration-level results of selected fractal
activation pairings.

For the Wine Recognition dataset, the highest-ranked configuration
combined the modulated Blancmange activation with AOFGD\_Adam at the
recorded order $\nu=1.00$. Several other optimizer families also
obtained their best results with the same activation, indicating that it
can support high accuracy under suitable update rules. Its lower
activation-wide mean nevertheless shows that this performance was not
uniform across the complete optimizer grid. Tanh provided the strongest
average activation behaviour, while the Standard optimizer family
achieved the highest aggregate optimizer accuracy. The results therefore
show a dataset-specific interaction in which a fractal activation
produced the best tuned configurations, but standard activations and
optimizers were more robust when performance was averaged over the full
experimental design.

\subsubsection{Tic-Tac-Toe Endgame}
\label{subsubsec:nn_results_tic_tac_toe}

The Tic-Tac-Toe Endgame dataset
(\href{https://www.openml.org/d/50}{OpenML ID 50}) contains 958 completed
game states described by nine categorical board-position attributes. The
binary target distinguishes positive from negative endgame outcomes. The
experiment comprised 222 unique optimizer--activation--order
configurations, each evaluated over 40 repeated runs, resulting in 8880
individual runs.

Table~\ref{tab:nn_tic_tac_toe_top15} reports the 15 highest-ranked
optimizer-specific configurations, retaining only the best
activation--order setting for each optimizer.

\begin{table}[H]
\caption{Top 15 optimizer-specific configurations on the Tic-Tac-Toe
Endgame dataset, ranked by mean test accuracy over 40 repeated runs. For
each optimizer, only its highest-ranked activation--order configuration
is shown. Standard deviation quantifies variation across the repeated
runs. For optimizers without a swept fractional-order parameter,
$\nu=1.00$ denotes the recorded default or integer-order setting.}
\label{tab:nn_tic_tac_toe_top15}
\small
\begingroup
\setlength{\tabcolsep}{4pt}
\begin{tabularx}{\textwidth}{
    >{\centering\arraybackslash}p{0.8cm}
    >{\raggedright\arraybackslash}X
    l
    l
    c
    >{\raggedleft\arraybackslash}p{1.25cm}
    r
}
\toprule
\textbf{Rank} &
\textbf{Optimizer} &
\textbf{Group} &
\textbf{Activation} &
\boldmath$\boldsymbol{\nu}$ &
\textbf{Avg. Acc.} &
\textbf{Std.} \\
\midrule
\textbf{1} &
\textbf{FSGD} &
\textbf{Herrera} &
\textbf{mod.\ wei.\ tanh} &
\textbf{1.50} &
\textbf{0.78411} &
\textbf{0.02705} \\

2  & AdaptiveMemory\allowbreak FSGD
   & Adaptive-Memory & mod.\ wei.\ tanh & 1.00 & 0.76337 & 0.02729 \\
3  & SGD
   & Standard        & mod.\ wei.\ tanh & 1.00 & 0.76293 & 0.02733 \\
4  & AOFGD\_SGD
   & Related-Work    & mod.\ wei.\ tanh & 1.00 & 0.76259 & 0.02863 \\
5  & FAdadelta
   & Herrera         & mod.\ wei.\ tanh & 0.75 & 0.75547 & 0.02754 \\
6  & Adadelta
   & Standard        & mod.\ wei.\ tanh & 1.00 & 0.75538 & 0.02662 \\
7  & FCSGD\_GL
   & Related-Work    & mod.\ wei.\ tanh & 1.00 & 0.75417 & 0.02800 \\
8  & FAdam
   & Herrera         & mod.\ wei.\ tanh & 1.25 & 0.75399 & 0.02188 \\
9  & AdaptiveMemory\allowbreak FAdadelta
   & Adaptive-Memory & mod.\ wei.\ tanh & 1.00 & 0.75339 & 0.02706 \\
10 & AdaptiveMemory\allowbreak FAdam
   & Adaptive-Memory & mod.\ wei.\ tanh & 1.00 & 0.75312 & 0.02211 \\
11 & Adam
   & Standard        & mod.\ wei.\ tanh & 1.00 & 0.75243 & 0.02287 \\
12 & FRMSprop
   & Herrera         & mod.\ wei.\ tanh & 0.75 & 0.75165 & 0.02381 \\
13 & AdaptiveMemory\allowbreak FRMSprop
   & Adaptive-Memory & mod.\ wei.\ tanh & 1.00 & 0.75139 & 0.02364 \\
14 & RMSprop
   & Standard        & mod.\ wei.\ tanh & 1.00 & 0.75122 & 0.02379 \\
15 & AOFGD\_Adam
   & Related-Work    & mod.\ wei.\ tanh & 1.00 & 0.74740 & 0.02021 \\
\bottomrule
\end{tabularx}
\endgroup
\end{table}

The highest mean accuracy was obtained by FSGD with the modified
Weierstrass--tanh activation and fractional order $\nu=1.50$, reaching
$0.78411 \pm 0.02705$. AdaptiveMemoryFSGD with the same activation
ranked second at $0.76337 \pm 0.02729$, followed by standard SGD at
$0.76293 \pm 0.02733$. The difference between the first- and
second-ranked configurations was 2.074 percentage points.

The 95\% confidence interval of FSGD was $[0.77573,0.79250]$, whereas
that of AdaptiveMemoryFSGD was $[0.75491,0.77183]$. These intervals did
not overlap, making the separation larger than for most previous
datasets, although no claim of statistical significance is made without a
direct paired comparison of the repeated runs.

SGD was the strongest standard optimizer, AdaptiveMemoryFSGD the
strongest Adaptive-Memory method, AOFGD\_SGD the highest-ranked
Related-Work method, and MemoryFSGD the best GL-Memory optimizer. The
best fractal-activation configuration was the overall winner. The
strongest standard-activation configuration was MemoryFAdam with tanh in
eighteenth place at $0.70755 \pm 0.02626$, 7.656 percentage points below
FSGD.

The top 15 contained four Standard, four Herrera, four Adaptive-Memory,
and three Related-Work optimizers; none of the explicit GL-Memory
optimizers entered this group. All 15 configurations used modified
Weierstrass--tanh, indicating a consistent preference for this activation
among the best configurations of the leading optimizers.

The first- and fifteenth-ranked configurations differed by
3.671 percentage points. A distinct decrease occurred after FSGD: ranks
two to four were closely grouped between $0.76259$ and $0.76337$,
approximately 2.1 percentage points below FSGD, while ranks five to
fourteen formed another compact range between $0.75122$ and $0.75547$.
The winner was therefore more clearly separated from the remaining
top-15 methods than those methods were from one another.

The macro-F1 results followed the same general ordering. FSGD obtained
$0.74823$, compared with $0.71647$ for AdaptiveMemoryFSGD and $0.71644$
for standard SGD. FSGD required $6.52$~s of mean training time, compared
with $6.37$~s for SGD and $7.11$~s for AdaptiveMemoryFSGD, so its higher
mean accuracy was not associated with a substantial runtime increase
relative to the other leading SGD-derived methods.

\begin{figure}[H]
\centering
\includegraphics[width=\textwidth]{
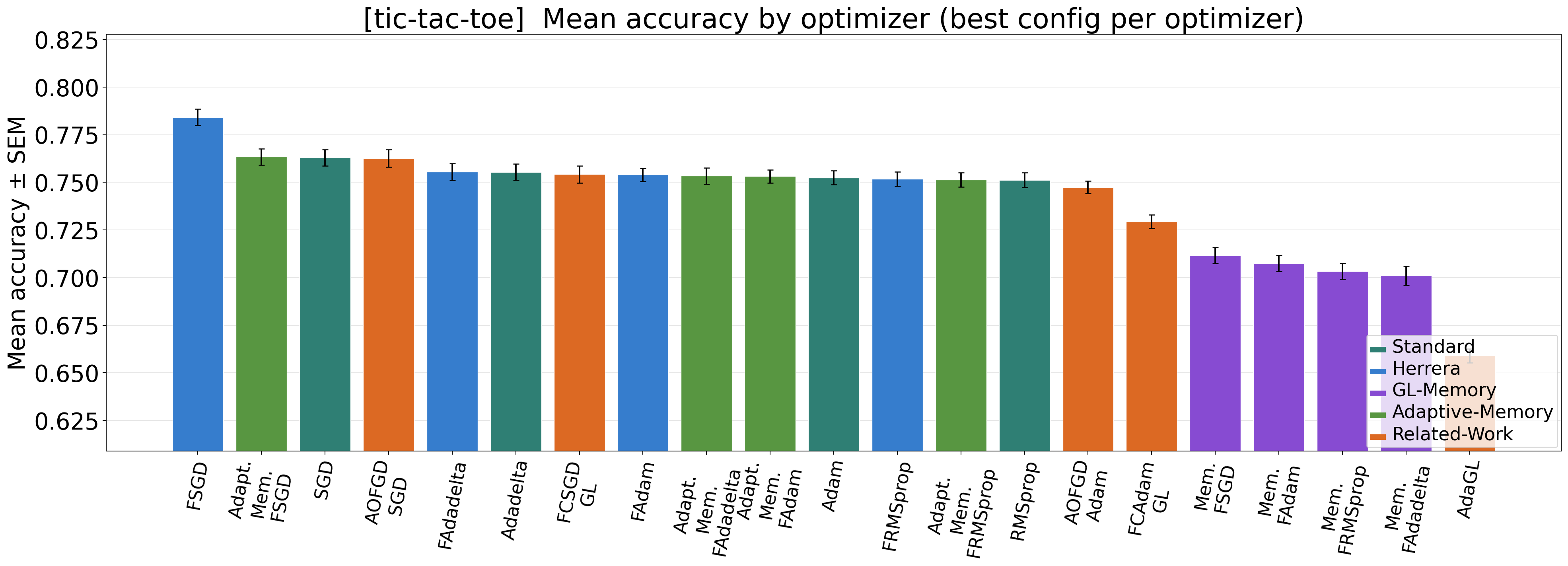}
\caption{Mean test accuracy by optimizer on the Tic-Tac-Toe Endgame
dataset. Each bar represents the highest-performing
activation--fractional-order configuration identified for that
optimizer. Optimizers are ordered by mean accuracy, error bars show the
standard error of the mean over 40 repeated runs, and bar colours
indicate the optimizer groups.}
\label{fig:nn_tic_tac_toe_optimizer_accuracy}
\end{figure}

Figure~\ref{fig:nn_tic_tac_toe_optimizer_accuracy} presents the complete
optimizer-level ranking, including all 21 optimizers.

Figure~\ref{fig:nn_tic_tac_toe_optimizer_accuracy} confirms that FSGD
was separated from the remaining optimizer-specific results. The best
Herrera, Adaptive-Memory, Standard, Related-Work, and GL-Memory
configurations reached $0.78411$, $0.76337$, $0.76293$, $0.76259$, and
$0.71163$, respectively. The best Standard, Adaptive-Memory, and
Related-Work configurations differed by less than 0.08 percentage
points, whereas FSGD exceeded each by more than two percentage points.

The explicit GL-Memory methods occupied the lower part of the ranking.
MemoryFSGD was strongest at $0.71163 \pm 0.02611$, followed by
MemoryFAdam at $0.70755 \pm 0.02626$, MemoryFRMSprop at
$0.70330 \pm 0.02687$, and MemoryFAdadelta at
$0.70104 \pm 0.03180$. These methods formed a compact group, but their
mean accuracies remained below those of the direct fractional, standard,
and adaptive-memory alternatives.

The derivative-order analysis shows different patterns for the direct
fractional optimizers. FAdadelta, FAdam, and FRMSprop changed only
slightly when aggregated across activations and orders. FSGD was more
sensitive: its aggregate mean accuracy increased from $0.66584$ at
$\nu=0.75$ and $0.67164$ at $\nu=1.25$ to $0.70791$ at $\nu=1.50$,
consistent with the configuration-level result.

Across all 21 optimizers, mean accuracy ranged from $0.78411$ for FSGD
to $0.65894$ for AdaGL, a difference of 12.517 percentage points. AdaGL
was separated from the other Related-Work methods, since AOFGD\_SGD
reached $0.76259$, FCSGD\_GL reached $0.75417$, and AOFGD\_Adam reached
$0.74740$; its lower result should therefore be read as method-specific.

The optimizer-group aggregation gives a broader view. Herrera optimizers
obtained the highest aggregate mean accuracy at $0.70695$, followed by
Standard and Adaptive-Memory at $0.70424$ and $0.70417$, respectively.
The difference between Herrera and Standard was only
0.271 percentage points, while Standard and Adaptive-Memory differed by
0.007 percentage points. Related-Work and GL-Memory reached aggregate
means of $0.67832$ and $0.65837$.

The macro-F1 group ranking was slightly different. Standard optimizers
obtained the highest aggregate macro-F1 at $0.64258$, followed closely by
Adaptive-Memory at $0.64236$ and Herrera at $0.64087$. The strong
configuration-level result of FSGD therefore did not produce a broad
macro-F1 advantage for the complete Herrera family.

\begin{figure}[H]
\centering
\includegraphics[width=\textwidth]{
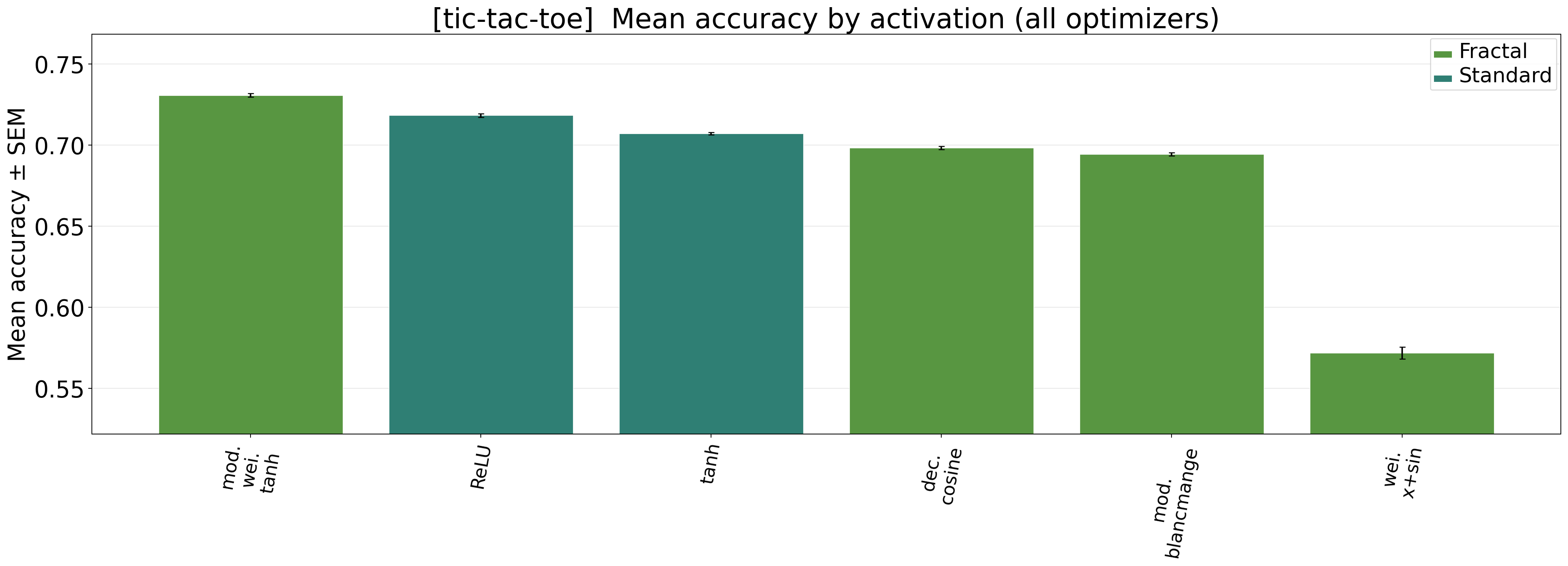}
\caption{Mean test accuracy by activation function on the Tic-Tac-Toe
Endgame dataset, aggregated over all evaluated optimizer configurations
and fractional-order settings. Error bars show the standard error of
the mean for the aggregated run-level accuracies. Colours distinguish
standard and fractal activation functions.}
\label{fig:nn_tic_tac_toe_activation_accuracy}
\end{figure}

Figure~\ref{fig:nn_tic_tac_toe_activation_accuracy} aggregates the
results by activation function across all optimizer configurations and
fractional-order settings.

At activation level, modified Weierstrass--tanh obtained the highest
aggregated mean accuracy at approximately $0.730$. ReLU was the strongest
standard activation at approximately $0.718$, about 1.2 percentage points
below modified Weierstrass--tanh. Tanh followed at approximately
$0.707$, while decaying cosine and modulated Blancmange reached
approximately $0.698$ and $0.694$. Weierstrass--Mandelbrot $x+\sin$
obtained the lowest aggregated mean at approximately $0.571$.

The activation-level result is consistent with the optimizer-specific
ranking. Modified Weierstrass--tanh appeared in the overall winning
configuration, in each of the top 15 optimizer-specific configurations,
and obtained the highest mean over the complete optimizer grid. Its
performance was therefore not restricted to one favourable optimizer
pairing.

The decaying cosine and modulated Blancmange activations remained closer
to tanh than to Weierstrass--Mandelbrot $x+\sin$, showing that the four
fractal functions did not behave as a homogeneous group. The low
Weierstrass--Mandelbrot $x+\sin$ result reduced the broad fractal-family
average despite the first-place result of modified Weierstrass--tanh.

When combined into families, standard activations obtained a mean
accuracy of $0.71260$, compared with $0.67380$ for fractal activations, a
difference of 3.880 percentage points. The standard family also achieved
a higher mean macro-F1 of $0.63126$, compared with $0.59501$ for the
fractal family. These family-level results do not contradict the
individual activation ranking: modified Weierstrass--tanh was the
strongest individual activation, but its advantage was offset by the
lower performance of the other fractal functions.

For the Tic-Tac-Toe Endgame dataset, the highest-ranked configuration
combined a Herrera-type fractional optimizer with the modified
Weierstrass--tanh activation. This pairing was more clearly separated
from the following configurations than the leading methods were on most
other datasets. The same activation also obtained the highest aggregate
accuracy and was selected in every top-15 optimizer configuration,
indicating broad compatibility with the evaluated update rules. The
overall Herrera-family advantage was nevertheless small when all
configurations were aggregated, and the standard activation family
retained the higher broad mean because the remaining fractal functions
performed less consistently. The results therefore indicate a specific
advantage for modified Weierstrass--tanh together with FSGD at
$\nu=1.50$, rather than a general advantage for every fractal activation
or every fractional optimizer formulation.

\subsubsection{Vehicle Silhouettes}
\label{subsubsec:nn_results_vehicle}

The Vehicle Silhouettes dataset
(\href{https://www.openml.org/d/54}{OpenML ID 54}) contains 846 vehicle
silhouettes described by 18 numerical shape descriptors. The target
distinguishes four vehicle classes, resulting in a four-class
classification problem. The experiment comprised 222 unique
optimizer--activation--order configurations, each evaluated over 40
repeated runs, resulting in 8880 individual runs.

Table~\ref{tab:nn_vehicle_top15} reports the 15 highest-ranked
optimizer-specific configurations, retaining only the best
activation--order setting for each optimizer.

\begin{table}[H]
\caption{Top 15 optimizer-specific configurations on the Vehicle
Silhouettes dataset, ranked by mean test accuracy over 40 repeated runs.
For each optimizer, only its highest-ranked activation--order
configuration is shown. Standard deviation quantifies variation across
the repeated runs. For optimizers without a swept fractional-order
parameter, $\nu=1.00$ denotes the recorded default or integer-order
setting.}
\label{tab:nn_vehicle_top15}
\small
\begingroup
\setlength{\tabcolsep}{4pt}
\begin{tabularx}{\textwidth}{
    >{\centering\arraybackslash}p{0.8cm}
    >{\raggedright\arraybackslash}X
    l
    l
    c
    >{\raggedleft\arraybackslash}p{1.25cm}
    r
}
\toprule
\textbf{Rank} &
\textbf{Optimizer} &
\textbf{Group} &
\textbf{Activation} &
\boldmath$\boldsymbol{\nu}$ &
\textbf{Avg. Acc.} &
\textbf{Std.} \\
\midrule
\textbf{1} &
\textbf{FAdadelta} &
\textbf{Herrera} &
\textbf{ReLU} &
\textbf{0.75} &
\textbf{0.79163} &
\textbf{0.03624} \\

2  & AOFGD\_Adam
   & Related-Work    & ReLU & 1.00 & 0.79163 & 0.03144 \\
3  & FAdam
   & Herrera         & ReLU & 0.75 & 0.78917 & 0.03043 \\
4  & RMSprop
   & Standard        & ReLU & 1.00 & 0.78858 & 0.02748 \\
5  & FRMSprop
   & Herrera         & ReLU & 1.50 & 0.78730 & 0.03478 \\
6  & Adam
   & Standard        & ReLU & 1.00 & 0.78711 & 0.03351 \\
7  & AdaptiveMemory\allowbreak FRMSprop
   & Adaptive-Memory & ReLU & 1.00 & 0.78681 & 0.03032 \\
8  & AdaptiveMemory\allowbreak FAdam
   & Adaptive-Memory & ReLU & 1.00 & 0.78504 & 0.02880 \\
9  & Adadelta
   & Standard        & ReLU & 1.00 & 0.78425 & 0.03209 \\
10 & AdaptiveMemory\allowbreak FAdadelta
   & Adaptive-Memory & ReLU & 1.00 & 0.78415 & 0.03134 \\
11 & FCAdam\_GL
   & Related-Work    & ReLU & 1.00 & 0.77825 & 0.03344 \\
12 & FSGD
   & Herrera         & ReLU & 0.75 & 0.77520 & 0.03965 \\
13 & SGD
   & Standard        & ReLU & 1.00 & 0.76024 & 0.03995 \\
14 & AdaptiveMemory\allowbreak FSGD
   & Adaptive-Memory & ReLU & 1.00 & 0.75797 & 0.04717 \\
15 & AOFGD\_SGD
   & Related-Work    & ReLU & 1.00 & 0.75531 & 0.05646 \\
\bottomrule
\end{tabularx}
\endgroup
\end{table}

FAdadelta with ReLU and fractional order $\nu=0.75$ was placed first
with $0.79163 \pm 0.03624$. AOFGD\_Adam with ReLU obtained the same mean
accuracy at the reported precision and a lower standard deviation of
$0.03144$. The numerical difference was therefore 0.000 percentage
points, and the 95\% confidence intervals, $[0.78040,0.80287]$ and
$[0.78189,0.80138]$, overlapped almost completely. The first two methods
should consequently be treated as tied by mean accuracy rather than as
clearly separated results.

FAdam with ReLU and $\nu=0.75$ ranked third at
$0.78917 \pm 0.03043$, followed by standard RMSprop at
$0.78858 \pm 0.02748$. RMSprop was the strongest standard optimizer and
had the lowest standard deviation among the first four configurations.
Its difference from the tied leading mean was only 0.305 percentage
points, still small relative to run-to-run variation.

The top 15 contained four Standard, four Herrera, four Adaptive-Memory,
and three Related-Work optimizers; none of the explicit GL-Memory
optimizers entered this group. All 15 configurations used ReLU, so the
optimizer-level ranking provides no evidence that a fractal activation
was required for any leading optimizer on this dataset.

The first- and fifteenth-ranked configurations differed by
3.632 percentage points, but the first ten formed a closely grouped range
between $0.78415$ and $0.79163$. A larger decrease appeared among the
SGD-based configurations from rank 12 onward. AOFGD\_SGD also had the
largest standard deviation in the top 15, indicating greater sensitivity
to the repeated data partitions.

The macro-F1 results broadly followed the accuracy ranking. FAdadelta
obtained $0.78569$, compared with $0.78416$ for AOFGD\_Adam and
$0.78200$ for RMSprop. FAdadelta required $2.18$~s of mean training time,
AOFGD\_Adam $3.74$~s, and RMSprop $1.83$~s. RMSprop therefore provided
almost the same mean accuracy with the shortest mean training time among
these three configurations.

\begin{figure}[H]
\centering
\includegraphics[width=\textwidth]{
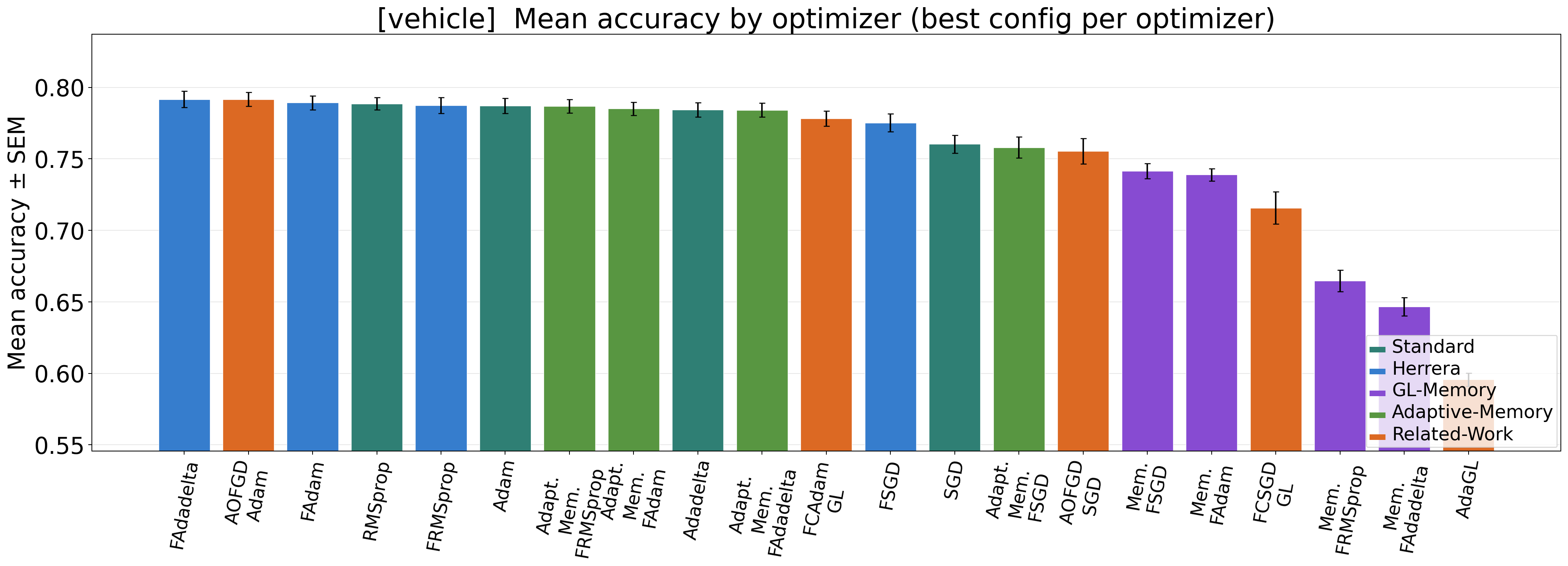}
\caption{Mean test accuracy by optimizer on the Vehicle Silhouettes
dataset. Each bar represents the highest-performing
activation--fractional-order configuration identified for that optimizer.
Optimizers are ordered by mean accuracy, error bars show the standard
error of the mean over 40 repeated runs, and bar colours indicate the
optimizer groups.}
\label{fig:nn_vehicle_optimizer_accuracy}
\end{figure}

Figure~\ref{fig:nn_vehicle_optimizer_accuracy} presents the complete
optimizer-level ranking, including all 21 optimizers.

Figure~\ref{fig:nn_vehicle_optimizer_accuracy} shows that the strongest
Herrera and Related-Work configurations obtained the same mean accuracy
of $0.79163$, followed by Standard and Adaptive-Memory at $0.78858$ and
$0.78681$. The best results from these four groups were contained within
only 0.482 percentage points, and their uncertainty intervals overlapped.

The explicit GL-Memory methods occupied a lower part of the ranking.
MemoryFSGD was strongest in this group at $0.74154 \pm 0.03328$,
followed by MemoryFAdam at $0.73888 \pm 0.02763$. MemoryFRMSprop and
MemoryFAdadelta reached $0.66467 \pm 0.04684$ and
$0.64665 \pm 0.03989$. These results do not indicate a general benefit
from explicit gradient-history aggregation on this dataset.

The derivative-order analysis further shows that the direct fractional
Adam-, Adadelta-, and RMSprop-type methods were relatively stable across
the evaluated orders, while FSGD was more sensitive and declined from
$0.54162$ at $\nu=0.75$ to $0.47293$ at $\nu=1.50$ when all activations
were aggregated. The explicit GL-Memory methods showed stronger order
dependence, with highest aggregate accuracies at $\nu=0.75$ and sharp
decreases at $\nu=1.25$ and $\nu=1.50$.

Across all 21 optimizers, mean accuracy ranged from $0.79163$ for
FAdadelta and AOFGD\_Adam to $0.59567$ for AdaGL, a difference of
19.596 percentage points. AdaGL was separated from the other Related-Work
methods, since AOFGD\_Adam ranked jointly first, FCAdam\_GL reached
$0.77825$, and AOFGD\_SGD reached $0.75531$. Its lower result should be
read as method-specific.

The optimizer-group aggregation gives a broader view. Herrera optimizers
obtained the highest aggregate mean accuracy at $0.65710$, followed by
the Standard group at $0.64406$, a difference of 1.304 percentage
points. Adaptive-Memory and Related-Work reached $0.61168$ and
$0.57925$, while GL-Memory reached $0.34358$. The aggregate macro-F1
values followed the same ordering, ranging from $0.63068$ for Herrera to
$0.29508$ for GL-Memory.

\begin{figure}[H]
\centering
\includegraphics[width=\textwidth]{
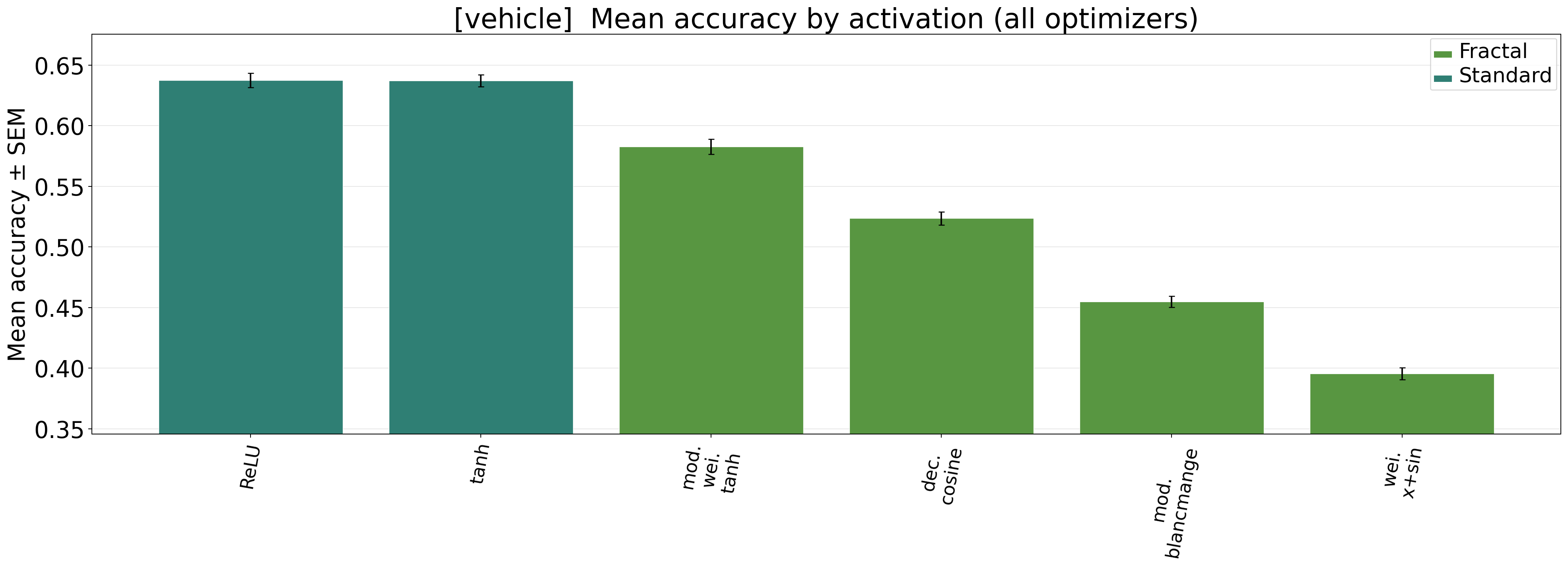}
\caption{Mean test accuracy by activation function on the Vehicle
Silhouettes dataset, aggregated over all evaluated optimizer
configurations and fractional-order settings. Error bars show the
standard error of the mean for the aggregated run-level accuracies.
Colours distinguish standard and fractal activation functions.}
\label{fig:nn_vehicle_activation_accuracy}
\end{figure}

Figure~\ref{fig:nn_vehicle_activation_accuracy} aggregates the results by
activation function across all optimizer configurations and
fractional-order settings.

At activation level, ReLU and tanh obtained almost identical aggregated
mean accuracies of approximately $0.637$, with ReLU ranked slightly
first. Modified Weierstrass--tanh was the strongest fractal function at
approximately $0.582$, about 5.5 percentage points below ReLU. Decaying
cosine followed at approximately $0.523$, while modulated Blancmange and
Weierstrass--Mandelbrot $x+\sin$ reached approximately $0.454$ and
$0.395$.

The activation-level ranking is consistent with the optimizer-specific
results. ReLU appeared in all 15 leading optimizer configurations and
also achieved the highest aggregated mean across the complete optimizer
grid. Tanh produced a similar aggregate mean but appeared only in some
lower-ranked optimizer-specific configurations. Modified
Weierstrass--tanh was the strongest fractal alternative, but remained
below both standard activations on average.

When combined into families, standard activations obtained a mean
accuracy of $0.63741$, compared with $0.48916$ for fractal activations, a
difference of 14.825 percentage points. The standard family also
obtained a higher mean macro-F1 of $0.61700$, compared with $0.44452$ for
the fractal family. Although the lower Weierstrass--Mandelbrot
$x+\sin$ and modulated Blancmange results contributed to this
family-level difference, even the strongest fractal activation remained
below ReLU and tanh in the individual ranking.

For the Vehicle Silhouettes dataset, the highest-ranked configurations
used the standard ReLU activation. A Herrera-type fractional optimizer
and a Related-Work optimizer were tied for the highest mean accuracy,
while standard RMSprop followed closely and showed lower variability and
training time. The Herrera group also obtained the highest aggregate
optimizer accuracy, although its advantage over the Standard group was
moderate. Fractal activations did not provide an advantage either in the
best-per-optimizer ranking or in the activation-wide aggregation. The
results therefore indicate that direct fractional gradient modification
can remain competitive on this task, but its benefit depends on the
optimizer formulation rather than on a simultaneous use of fractal
activation functions. The explicit GL-Memory methods performed below the
direct fractional, standard, and adaptive-memory alternatives.

\subsubsection{Summary of the Classification Results}
\label{subsubsec:nn_results_summary}

The classification experiments covered ten datasets with different
sample sizes, feature types, and numbers of target classes. For each
dataset, 222 optimizer--activation--order configurations were evaluated
over 40 repeated runs, corresponding to 8880 runs per dataset and
88\,800 runs in total. The results therefore provide a broader assessment
than a comparison based only on individual train--test partitions or a
single optimizer. In particular, the optimizer-specific rankings show
the best configuration available to each optimizer, whereas the
activation-level results measure average behaviour over the complete
optimizer and fractional-order grid. Both views are required because the
best tuned configuration and the most robust activation were frequently
not the same.

The present experiments extend the preceding study of fractal activation
functions \cite{raubitzek_fractals_2026}. That study provided the basis
for selecting the fractal activations considered here. The current work
retains the most relevant functions from that comparison and evaluates
them in a substantially broader optimization setting. Instead of
studying the activation functions under a restricted set of conventional
update rules, the present design combines them with standard optimizers,
direct Herrera-type fractional optimizers, explicit GL-memory methods,
adaptive-memory variants, and related fractional approaches. The
fractional derivative order is also varied where applicable. The main
extension is therefore not only the addition of further optimizers, but
the systematic analysis of the interaction among the activation
function, optimizer formulation, and derivative order.

At the level of the 15 best optimizer-specific configurations per
dataset, fractal activations showed a clear presence. They formed the
majority of the top-15 entries on eight of the ten datasets. This pattern
occurred for Climate-Model Simulation Crashes, Diabetes, Glass
Identification, Ionosphere, Iris, Seeds, Wine Recognition, and
Tic-Tac-Toe Endgame. Only the Vertebral Column and Vehicle Silhouettes
datasets showed a clear dominance of standard activations within their
top-15 rankings. The pattern was particularly pronounced for Iris, Wine
Recognition, and Tic-Tac-Toe Endgame, where every top-15 configuration
used a fractal activation. Similarly, almost all leading configurations
for Climate-Model Simulation Crashes and Ionosphere used the modified
Weierstrass--tanh activation, while the decaying cosine activation was
frequent among the leading Glass Identification and Iris configurations.

The single best configuration on each dataset provides a more
restrictive comparison. A fractal activation occurred in the winning
configuration on six of the ten datasets: Climate-Model Simulation
Crashes, Glass Identification, Ionosphere, Iris, Wine Recognition, and
Tic-Tac-Toe Endgame. Standard activations produced the highest-ranked
configuration on Diabetes, Seeds, Vertebral Column, and Vehicle
Silhouettes. This result shows that the dominance of fractal activations
is strongest when the full upper part of the optimizer ranking is
considered. It should not be interpreted as a claim that a fractal
activation produced the single highest mean on every dataset.

The activation-wide aggregation provides a further distinction. When
accuracy was averaged over all optimizer configurations and
fractional-order settings, a fractal activation ranked first on four
datasets: modified Weierstrass--tanh on Diabetes, Ionosphere, and
Tic-Tac-Toe Endgame, and decaying cosine on Iris. A standard activation
ranked first on the remaining six datasets. Consequently, the principal
benefit of the selected fractal functions was often configuration
dependent. They regularly produced the strongest or most frequent
top-ranked optimizer pairings, but their advantage did not always remain
when weak and incompatible optimizer combinations were included in the
average. Wine Recognition provides a direct example: the modulated
Blancmange activation occurred in the overall winning configuration and
in most of the top-15 entries, whereas tanh achieved the highest
activation-wide mean. A similar difference was observed for Glass
Identification, where decaying cosine produced the best individual
configuration but tanh was more robust across the complete optimizer
grid.

Among the fractal functions, modified Weierstrass--tanh showed the most
consistent behaviour across datasets. It produced the winning
configuration for Climate-Model Simulation Crashes, Ionosphere, and
Tic-Tac-Toe Endgame, achieved the highest activation-wide mean on
Diabetes, Ionosphere, and Tic-Tac-Toe Endgame, and appeared repeatedly
in the leading configurations of several other datasets. The decaying
cosine activation was particularly effective on Glass Identification and
Iris, while the modulated Blancmange activation produced the strongest
Wine Recognition configuration. The results therefore confirm that the
selected fractal functions should not be treated as one homogeneous
activation family. Their behaviour depended on both the dataset and the
optimizer. In particular, the Weierstrass--Mandelbrot $x+\sin$
activation generally obtained the lowest activation-wide mean and
frequently reduced the average of the complete fractal family. Broad
comparisons between ``standard'' and ``fractal'' activations therefore
conceal substantial differences among the individual functions.

The optimizer results show a similarly strong interaction. A
Herrera-type optimizer produced the sole highest-ranked result on six
datasets: Diabetes, Ionosphere, Iris, Seeds, Vertebral Column, and
Tic-Tac-Toe Endgame. FAdadelta also shared the highest mean accuracy with
AOFGD\_Adam on Vehicle Silhouettes. Adaptive-memory optimizers ranked
first on Climate-Model Simulation Crashes and Glass Identification,
while AOFGD\_Adam ranked first on Wine Recognition. Neither a standard
optimizer nor an explicit GL-Memory optimizer produced an unshared
first-place result. Direct fractional gradient modification was therefore
frequently associated with the strongest dataset-specific configuration,
including cases in which the selected activation was standard.

The combination of fractional optimization and fractal activations was
especially effective on Ionosphere, Iris, and Tic-Tac-Toe Endgame. On
these datasets, the winning configuration combined a Herrera-type
fractional optimizer with a fractal activation. The same fractal
activations also remained strong in the corresponding activation-wide
aggregations. These cases provide the clearest evidence of a useful
interaction between the two components. On other datasets, fractional
optimizers improved or matched the strongest results while using a
standard activation. FSGD with ReLU ranked first on Diabetes, FSGD with
tanh ranked first on Seeds, FAdam with tanh ranked first on Vertebral
Column, and FAdadelta with ReLU shared the highest mean on Vehicle
Silhouettes. Fractional optimization should therefore not be understood
only as a mechanism for supporting fractal activations. Its effect also
depends directly on the loss landscape generated by the dataset and the
selected standard activation.

The explicit GL-Memory optimizers showed a less favourable overall
pattern. Climate-Model Simulation Crashes was a notable exception, since
several GL-Memory configurations were included in the leading group.
Across most other datasets, however, the GL-Memory variants occupied the
lower part of the optimizer ranking and were absent from the top 15.
Their standard deviations and SEM values were also often larger,
particularly on Iris, Seeds, Vertebral Column, Glass Identification, and
Wine Recognition. The group-level averages consequently placed
GL-Memory below the direct Herrera, Standard, and Adaptive-Memory
families on most datasets. These results indicate that adding an explicit
history of fractional gradients does not automatically improve
classification performance and can increase sensitivity to data
partitioning and hyperparameter selection.

Part of this behaviour may be related to the selected fractional orders.
The best direct fractional configurations used different values across
the datasets, including $\nu=0.75$, $\nu=1.25$, and $\nu=1.50$. No
single order was consistently optimal. The order sensitivity was
particularly visible for FSGD and for several GL-Memory variants, while
the fractional Adam-, RMSprop-, and Adadelta-type methods were often more
stable across the tested values. The lower group averages of the
memory-based methods may therefore partly reflect configurations in
which the derivative order was unsuitable for the specific
optimizer--activation--dataset combination. This is a plausible
interpretation rather than a confirmed cause, since memory length,
learning rate, activation geometry, and derivative order were varied
only within the defined experimental grid. A more targeted optimization
of these parameters would be required to separate their individual
effects.

The uncertainty estimates also limit how strongly small differences
between the leading configurations should be interpreted. On most
datasets, the difference between the first few mean accuracies was small
relative to the standard deviation across the 40 runs, and their
confidence intervals overlapped. Tic-Tac-Toe Endgame provided one of the
clearest separations, where FSGD with modified Weierstrass--tanh and
$\nu=1.50$ exceeded the following optimizer-specific configurations by
more than two percentage points. On several other datasets, including
Climate-Model Simulation Crashes, Diabetes, Iris, Wine Recognition, and
Vehicle Silhouettes, multiple optimizer families achieved closely
grouped results. Macro-F1 generally supported the accuracy rankings, but
it did not always preserve their exact order. The main finding is
therefore the repeated occurrence of effective optimizer--activation
pairings rather than a uniform advantage of one method under every
evaluation measure.

Overall, the experiments strengthen and refine the findings of the
preceding activation-focused study. The selected fractal activations
remained highly competitive and dominated the upper optimizer-specific
rankings on eight of the ten datasets. Their strongest results were often
obtained together with fractional or adaptive-memory optimizers,
demonstrating that the optimizer can increase the practical benefit of
the activation. At the same time, activation-wide and optimizer-group
averages show that this improvement is not uniform. Standard activations
were more robust on several datasets, explicit GL-Memory methods often
introduced additional variability, and the performance of the fractal
functions differed substantially. The classification results therefore
support a joint selection of the activation function, optimizer
formulation, and fractional order rather than the independent use of
either fractal activations or fractional optimizers as fixed method
families.

\begin{takeawaybox}[tealA]{Main Takeaways --- Classification Experiments}
  \begin{itemize}[    label={},
    leftmargin=0em,
    itemindent=0em,
    itemsep=3pt,
    topsep=2pt]
    \item \textbf{Fractal activations dominate the leading configurations.}
          They form the majority of the top-15 results on eight datasets
          and provide the best configuration on six. Standard activations
          dominate only on \texttt{vertebra-column} and \texttt{vehicle}.

    \item \textbf{The strongest gains depend on optimizer--activation
          interactions.} Modified Weierstrass--tanh performs best on
          \texttt{ionosphere}, \texttt{diabetes}, and
          \texttt{tic-tac-toe}; decaying cosine leads on \texttt{iris}
          and \texttt{glass}, and Blancmange on \texttt{wine}.

    \item \textbf{Direct fractional optimizers are consistently
          competitive.} Herrera-type methods lead on six datasets and
          share first place on \texttt{vehicle}, using both fractal and
          standard activations.

    \item \textbf{Explicit gradient-memory methods are less stable.}
          GL-Memory optimizers usually remain outside the top 15 and show
          larger variation, possibly due in part to sensitivity to the
          selected fractional order \(\nu\).

    \item \textbf{Repeated-run evaluation is essential.} The study covers
          222 configurations over 40 runs on ten datasets, yielding
          88\,800 runs. Overlapping error ranges show that single-run
          rankings would be unreliable.
  \end{itemize}
\end{takeawaybox}

\section{Discussion}
\label{sec:discussion}

This work investigated the interaction between fractional optimization
methods and fractal activation functions for neural network training. The
study combines concepts from fractional calculus, fractal geometry, and
gradient-based optimization into a common experimental framework. Rather
than evaluating only prediction performance, the work first studies
optimizer behaviour on controlled benchmark surfaces and subsequently
evaluates the same 21 optimizers in neural network classification tasks.
This two-stage design allows observations made under controlled
optimization conditions to be compared with the behaviour observed during
neural network training. The discussion below follows the three
objectives stated in the introduction: optimizer behaviour on fractally
perturbed landscapes, the joint evaluation of fractional optimizers and
fractal activations in neural networks, and the assessment of the
proposed adaptive memory-fractional framework.

The conceptual link between the two components is the one developed in
Sections~\ref{sec:weierstrass_fractional}--\ref{sec:fractional_optimizers}:
fractal activations and fractal surface perturbations insert
Weierstrass-type multi-scale structure into the objective, while the
memory-carrying fractional optimizers apply a discrete
Gr\"unwald--Letnikov fractional-difference kernel along the iteration
axis. The mathematical connection between fractional differentiation and
Weierstrass-type functions is established in
\cite{zahle1996fractional}, whereas the general
Gr\"unwald--Letnikov construction is described in
\cite{podlubny1999fractional}. The activation and the perturbation change
the geometry of the objective; the optimizer changes how this geometry is
traversed. The experiments were designed to test whether this mathematical
connection translates into a practical advantage when both components are
used together.

\,\par\noindent\textbf{Optimizer behaviour on fractally perturbed surfaces.}
The surface experiments answer the first objective with a differentiated
result. The dominant factor was not the fractional mechanism but the base
optimizer family. On both Ackley and Himmelblau, in the standard and the
additive variant alike, the leading positions were occupied almost
exclusively by SGD-type methods, whether classical, fractional, or
memory-based, while the RMSprop- and Adadelta-based methods performed
poorly regardless of the fractional machinery attached to them. Within
the SGD family, the fractional variants consistently matched or exceeded
their classical counterparts: FSGD led both Ackley variants, FCSGD\_GL
\cite{zhou2023gl} solved the standard Himmelblau surface exactly, and
MemoryFSGD reached the highest success rate on the additive Ackley
variant. The additive perturbation itself changed the ranking in an
interpretable way. On Himmelblau it reduced the reliability of all
methods but promoted the adaptive-order method AOFGD\_Adam
\cite{xiang2025aofgd} to the best result and moved the Adam-based methods
up the ranking; on Ackley it lowered the reachable objective values
without solving the underlying plateau problem. Two further observations
are relevant for practice. First, low mean final loss and high target
success rate were not equivalent rankings, which justifies reporting
both. Second, the Adadelta-based memory variants diverged in a small
number of Himmelblau runs, with means dominated by these outliers; this
instability is a genuine property of combining an update-ratio method
with a history-based gradient on a polynomial-growth surface, and it is
the reason the manuscript reports distributions rather than means alone.

\,\par\noindent\textbf{Fractal activations in the classification experiments.}
The second objective concerns the neural network experiments, and here
the activation-side finding of the predecessor study
\cite{raubitzek_fractals_2026} was confirmed and refined. The four
selected fractal activations formed the majority of the top-15
optimizer-specific configurations on eight of the ten datasets and
provided the single best configuration on six of them, while
activation-wide averages over the complete optimizer grid favoured a
fractal activation on only four datasets. The advantage of the fractal
activations is therefore configuration-dependent: they supply the
strongest optimizer pairings, but not uniformly better averages once
weak optimizer combinations are included. The individual functions also
behaved differently, in line with the regularity classification of
Section~\ref{sec:fractal_activations}: 
the subcritical modified Weierstrass--tanh activation was the most
consistent function across datasets, the two critical activations
(decaying cosine, modulated Blancmange) were strong on specific datasets
(Glass and Iris, and Wine, respectively), and the supercritical
Weierstrass--Mandelbrot $x{+}\sin$ activation, the roughest of the four,
generally obtained the lowest activation-wide means. Under equal
training conditions, the mildest multiscale structure was thus the most
broadly useful one, and genuine critical-order roughness was useful only
in particular optimizer pairings. Sweeping statements about ``fractal
versus standard'' activations conceal these differences.

\,\par\noindent\textbf{Fractional optimizers in the classification experiments.}
On the optimizer side, the clearest result is the strength of the
Herrera-type local scaling methods
\cite{herrera2022fractional,herrera2023pytorch}: they produced the sole
best configuration on six of the ten datasets and shared first place on a
seventh, using fractal activations on some datasets (Ionosphere, Iris,
Tic-Tac-Toe) and standard activations on others (Diabetes, Seeds,
Vertebral Column, Vehicle). The adaptive-memory framework ranked first on
Climate-Model Simulation Crashes and Glass Identification, and the
adaptive-order method AOFGD\_Adam ranked first on Wine Recognition.
Neither a standard optimizer nor an explicit GL-memory optimizer produced
an unshared first place. The explicit GL-memory methods, in fact, showed
the least favourable overall pattern: outside the Climate dataset they
were usually absent from the top 15 and exhibited larger run-to-run
variation, and their group averages fell below the Standard, Herrera, and
Adaptive-Memory groups on most datasets. The fractional order showed no
universally optimal value; the best configurations used $\nu=0.75$,
$1.25$, and $1.50$ depending on the dataset, with FSGD and several
GL-memory variants reacting most strongly to the order and the
fractional Adam-, RMSprop-, and Adadelta-type methods being more stable.
Finally, on most datasets
the leading mean accuracies differed by less than the run-to-run
standard deviation over 40 runs, with Tic-Tac-Toe Endgame as the main
exception, where FSGD with the modified Weierstrass--tanh activation and
$\nu=1.50$ exceeded the following configurations by more than two
percentage points. The robust finding is the repeated occurrence of
effective optimizer--activation pairings, not a uniform advantage of any
single method.

\,\par\noindent\textbf{Why the two experimental stages disagree about memory.}
Comparing the two stages yields the most instructive observation of the
study. On the controlled surfaces, explicit gradient memory was an asset:
MemoryFSGD was among the most reliable methods on the perturbed
Himmelblau and Ackley variants. In network training, the same mechanism
was a liability, and the local-scaling methods dominated instead. The
mechanistic analysis of
Section~\ref{sec:fractional_optimizers} 
offers a consistent explanation. The truncated Gr\"unwald--Letnikov
kernel is a difference filter whose coefficients nearly cancel; its
output is informative when successive gradients carry persistent,
deterministic oscillatory structure, which is exactly the situation on a
fixed two-dimensional surface with an additive Weierstrass-type
perturbation evaluated with exact gradients. In minibatch training, by
contrast, successive gradients differ mainly through sampling noise,
batch-to-batch nonstationarity, and evolving network state; a
short-memory difference filter then amplifies noise rather than
extracting structure. The descent safeguard and norm matching introduced
in this work keep the memory methods functional in this regime, but they
cannot create an advantage where the history carries little stable
directional information. The Herrera-type factor
$f_\nu(g_t)$, which uses no history at all, is unaffected by this
problem and inherits the stability of its base optimizer, which matches
its strong classification results. A second cross-stage observation
points in the same direction: the surface experiments diagnose landscape
navigation, not generalization. Adadelta-based methods were among the
weakest and least stable methods on the surfaces, yet an Adadelta-based
method won the Climate dataset. Conclusions from controlled surfaces
therefore transfer to network training only at the level of mechanisms,
not at the level of method rankings.

\,\par\noindent\textbf{Assessment of the adaptive memory-fractional framework.}
The third objective was to test the proposed adaptive framework, which
keeps the order $\nu$ fixed and adapts a bounded trust coefficient
$\lambda_t$ with an exact classical fallback at $\lambda_t=0$. The
results support the design goal of robustness rather than peak
performance. In the classification experiments, the adaptive-memory
group consistently ranked above the plain GL-memory group in the
group-level averages, avoided the large variance of the plain memory
methods on most datasets, and produced two dataset wins; on the surfaces,
AdaptiveMemoryFSGD was among the reliable methods on every variant. The
framework therefore behaves as intended: it retains the memory mechanism
where it is useful and withdraws it where it is not, at the cost of never
committing fully to either regime. Its main failure case, the divergence
of AdaptiveMemoryFAdadelta on the additive Himmelblau surface, is
inherited from the base optimizer rather than from the adaptation rule.
Compared with the related-work alternatives, the framework occupies a
distinct position: FCSGD\_GL, FCAdam\_GL \cite{zhou2023gl}, AdaGL
\cite{chen2024adagl}, and FracM \cite{yu2022fracm} substitute the
fractional object permanently, while AOFGD \cite{xiang2025aofgd},
FOAdam with its fractional-order scheduler \cite{chen2024foadam}, and
2SEDFOSGD \cite{partohaghighi2025twoscale} adapt the fractional order or
exponent itself. Other adaptive fractional methods instead modify the
learning rate or update control while retaining their fractional
construction \cite{ma2025apfogdl,huang2024mffgd}. The experiments show
that both adaptive philosophies can win datasets (AOFGD\_Adam on Wine and on the
additive Himmelblau surface; the $\lambda_t$ framework on Climate and
Glass), while the permanent deterministic substitution without
safeguards was the weakest design. Among the permanent-substitution
methods, the Bernoulli-masked variant of Zhou et al. \cite{zhou2023gl}
performed clearly best. Our kernel-sum analysis in
Section~\ref{subsec:related_work_optimizers} 
offers one possible explanation: because each history term is retained
with probability $p=0.5$, the expected effective kernel sum is
\[
1+0.5\sum_{k=1}^{K-1}c_k^{(\nu)} = 0.5+0.5\sum_{k=0}^{K-1}c_k^{(\nu)},
\]
which partially reduces the near-cancellation of the deterministic
truncated kernel.

\,\par\noindent\textbf{Relation to the literature.}
The results relate to previous work in three ways. First, they confirm,
under matched conditions and repeated runs, the central claim of the
fractional-optimizer literature that non-integer-order modifications of
standard optimizers can improve training
\cite{wang2017fractional,bao2018fractional,herrera2022fractional,
zhou2023gl,shin2023accelerating,naifar2026tempered}, but they locate the
improvement almost entirely in the local-scaling and adaptive designs
and show that it is dataset- and order-dependent. This differentiated
picture responds directly to the observation in recent surveys that
fractional optimizers are rarely compared systematically against each
other under identical protocols
\cite{elnady2025survey,fernandez2025role}. Second, the activation-side
results extend the predecessor study \cite{raubitzek_fractals_2026} from
five standard optimizers to 21 optimizers and a swept fractional order,
and show that the earlier conclusions survive this much broader
optimization setting in refined form. Third, the two-stage design
complements the observation that the boundary between stable and divergent
neural-network training can itself be
fractal in hyperparameter space \cite{sohldickstein2024boundary}. The
present study addresses a different but related question by deliberately
introducing fractal structure into benchmark surfaces and activation
functions and examining how this changes which optimizer family performs
best.

\,\par\noindent\textbf{Practical guidance.}
For practitioners, the results translate into four concrete points.
First, choose the activation and the optimizer jointly; neither the
best activation nor the best optimizer of this study is independent of
the other component. Second, as a default fractional extension, the
Herrera-type scaling is the most attractive option: it has negligible
overhead, an exact classical limit, and produced the most dataset wins.
Third, the fractional order should be treated as a small discrete
hyperparameter (the values $0.75$, $1.25$, and $1.50$ covered all
winning configurations here) rather than as a quantity with a single
correct value. Fourth, explicit gradient memory should be reserved for
settings that resemble the conditions under which it helped: exact or
low-noise gradients on objectives with persistent multi-scale structure;
in stochastic network training it should be used, if at all, only with
the descent safeguard, norm matching, and raw-gradient second moments,
or through the bounded adaptive mixing of the proposed framework.

\,\par\noindent\textbf{Computational cost.}
The accuracy and robustness differences discussed above are accompanied
by runtime differences, which are quantified in
Appendix~\ref{app:runtimes}. 
Measured per optimization step on the surfaces, where the update rule is
the only varying cost, the overhead of the fractional mechanisms is
moderate and follows the amount of per-step work: typically around
10\% and up to 55\% for the Herrera-type scaling, 40--105\% for the
safeguarded Gr\"unwald--Letnikov memory on the standard surfaces, and a
factor of 2--3 for the adaptive-memory and related-work methods,
relative to their baselines.
Two qualifications keep these factors in perspective. First, the
relative overhead shrinks as the objective becomes more expensive: on
the additive surface variants, where evaluating the perturbation ladder
adds a near-constant 5.1--5.8\,ms to every step, the most expensive
update rules cost only about 1.5 times a plain SGD step, and in network
training, where the forward and backward passes dominate, the optimizer
choice matters even less for the total time. Second, in the
classification experiments the training times of the winning
configurations mix the optimizer cost with the cost of the selected
activation, since the fractal activations are truncated series with
30--100 terms; runtime comparisons between method families must
therefore be read together with the accuracy tables. Under this reading,
the cost--benefit balance favours the same designs as the accuracy
results: the Herrera group combines the highest cross-dataset accuracy
with the second-lowest mean training time (6.5\,s; FAdam trains in
5.5\,s on average), whereas the adaptive-memory group buys its
robustness with the highest mean training time (9.9\,s, roughly 20\%
above the standard group). In absolute terms all methods remain
inexpensive at the scale of this study, with the most expensive mean
best-configuration training time at approximately 12\,s.

\,\par\noindent\textbf{Limitations.}
Several limitations qualify these conclusions. The network experiments
use feed-forward multilayer perceptrons on ten tabular OpenML datasets;
convolutional, recurrent, transformer, and graph architectures, as well
as larger-scale benchmarks, remain to be investigated. The surface
experiments cover two surfaces in two variants under a fixed step budget
and a fixed initialization protocol, which caps attainable success rates
on the plateau-dominated Ackley surface. The experimental grid varies
the derivative order, activation, and optimizer, but memory length,
learning rates, and the adaptation hyperparameters were fixed to
representative values, so the reported group differences may partly
reflect these choices; the order-sensitivity of the memory methods makes
this caveat explicit. Only four selected fractal activations from the
predecessor study were evaluated, for computational reasons. Finally,
many of the reported differences between leading configurations are
small relative to run-to-run variation, and the study focuses on
supervised classification; regression, generative modelling, and
scientific machine learning remain open directions.

Despite these limitations, the experiments support a precise version of
the central hypothesis. Fractal activations and fractional optimizers
are mathematically compatible modifications of two different parts of
the training problem, and their combination can be beneficial, but the
benefit is carried by specific pairings rather than by the method
families as a whole: subcritical fractal activations paired with
local-scaling or adaptive fractional optimizers were repeatedly among
the best configurations, while indiscriminate use of explicit gradient
memory was not supported by the data. The two-stage design was essential
for reaching this conclusion, because it separated what the memory
mechanism can do on controlled multi-scale landscapes from what it does
under the stochastic gradients of practical training.

\begin{takeawaybox}[roseA]{Main Takeaways --- Discussion}
  \begin{itemize}[
    label={},
    leftmargin=0em,
    itemindent=0em,
    itemsep=3pt,
    topsep=2pt
  ]
    \item \textbf{SGD-optimizer variants dominate on controlled surfaces.}
          SGD-type methods lead all Ackley and Himmelblau variants;
          fractional and memory-based SGD variants match or exceed plain
          SGD, and the additive perturbation favours the adaptive
          methods on Himmelblau.

    \item \textbf{Fractal activations win through pairings, not
          averages.} They dominate the top-15 configurations on eight of
          ten datasets and win six, but activation-wide averages favour
          them on only four; the subcritical modified Weierstrass--tanh
          function is the most consistent, the supercritical
          $x{+}\sin$ variant the weakest.

    \item \textbf{Local fractional scaling is the strongest optimizer
          design.} Herrera-type methods take six of ten dataset wins at
          negligible cost; no universally optimal order exists, and
          $\nu\in\{0.75,1.25,1.50\}$ covers all winning configurations.

    \item \textbf{Explicit gradient memory helps on deterministic
          multi-scale landscapes but less in minibatch training.} The
          truncated GL kernel extracts persistent oscillatory structure
          on the surfaces and amplifies sampling noise in network
          training; safeguards keep the memory methods functional but
          not superior.

    \item \textbf{The adaptive $\lambda_t$ framework delivers robustness.}
          It ranks above plain GL-memory throughout, wins two datasets,
          and reduces exactly to the classical optimizer when memory is
          not trusted.

    \item \textbf{Joint selection is the practical consequence.}
          Activation, optimizer formulation, and fractional order should
          be chosen together; repeated-run evaluation is required, since
          most leading configurations differ by less than the
          run-to-run variation.
  \end{itemize}
\end{takeawaybox}

\section{Conclusion}
\label{sec:conclusion}

This paper investigated the interaction between fractional optimization
methods and fractal activation functions for neural network training. The
study connects the fractional-derivative analysis of Weierstrass-type
functions with the discrete Gr\"unwald--Letnikov constructions used in
fractional optimizers, and evaluates 21 optimizers from five families in
a two-stage experimental design: controlled optimization on the Ackley
and Himmelblau surfaces with additive Weierstrass-type perturbations, and
neural network classification on ten public OpenML datasets with the four
strongest fractal activations of the predecessor study alongside
conventional activations. All configurations were evaluated over 40
repeated runs, amounting to 1680 optimizer runs per surface and 88\,800
training runs in the classification experiments.

The main findings can be stated briefly. On the controlled surfaces, the
base optimizer family dominated the outcome: SGD-type methods led every
surface variant, and within this family the fractional and memory-based
variants matched or exceeded their classical counterpart, with explicit
gradient memory among the most reliable mechanisms on the perturbed
landscapes. In the classification experiments, the picture shifted. The
Herrera-type local scaling methods were the strongest optimizer design,
producing the sole best configuration on six of the ten datasets, while
the explicit memory methods fell behind and showed larger run-to-run
variation. The fractal activations formed the majority of the leading
configurations on eight of the ten datasets and provided the winning
configuration on six, with the subcritical modified Weierstrass--tanh
function as the most consistent representative; their advantage was
carried by specific optimizer pairings rather than by grid-wide averages.
No universally optimal fractional order emerged, and most differences
between leading configurations were small relative to run-to-run
variation, which underlines the necessity of repeated-run evaluation.

The disagreement between the two stages regarding gradient memory is
itself a result. The truncated Gr\"unwald--Letnikov kernel extracts
persistent oscillatory structure from deterministic gradients on fixed
multi-scale landscapes, but under stochastic minibatch gradients its
differencing character amplifies noise rather than structure. The descent
safeguard, norm matching, and raw-gradient variance estimates introduced
in this work keep the memory methods functional in the stochastic
regime, and the proposed adaptive framework, which controls the memory
contribution through a bounded trust coefficient $\lambda_t$ with an
exact classical fallback, converted this into consistent robustness: it
ranked above the plain memory methods throughout and produced two
dataset wins. Controlled surface experiments and network training are
therefore connected at the level of mechanisms, not at the level of
method rankings, and both are needed to evaluate optimizers of this
kind.

For practice, the results support four recommendations: select
activation function, optimizer formulation, and fractional order
jointly; use Herrera-type scaling as the default fractional extension,
given its negligible cost and exact classical limit; treat the order as
a small discrete hyperparameter, since $\nu\in\{0.75, 1.25, 1.50\}$
covered all winning configurations; and reserve explicit gradient
memory for low-noise settings with persistent multi-scale structure, or
use it through the bounded adaptive mixing.

This article has deliberately adopted a broad scope. Beyond introducing the underlying mathematical concepts and building intuition for fractal activation functions and fractional optimization, it has examined the resulting methods in detail and discussed the nuances, limitations, and variability of the experimental findings. The proposed ideas were evaluated in two distinct but closely related settings: controlled optimization on standard and fractally perturbed benchmark surfaces, and neural network classification with conventional and fractal activation functions. Although many further experiments could be considered, including additional objective functions, architectures, datasets, optimization settings, and application domains, the present study already provides an extensive and coherent view of the investigated design space. The results therefore offer a sufficiently broad foundation for assessing the proposed methods, while the remaining possibilities constitute natural directions for future work.

Several directions follow from this work. The experiments cover
feed-forward networks on tabular classification tasks; convolutional,
recurrent, transformer, and graph architectures, as well as regression and
generative modelling remain to be
examined, and larger benchmarks would strengthen the statistical
conclusions. On the methodological side, the memory length, the
adaptation hyperparameters, and per-layer or scheduled variants of the
trust coefficient were fixed here and deserve systematic study, as does
the combination of order adaptation and trust adaptation, which the
results identify as the two viable adaptive strategies. On the
theoretical side, the observed regularity dependence of the activations
and the noise sensitivity of the Gr\"unwald--Letnikov filter suggest two
concrete questions: how the H\"older exponent of an activation shapes
the gradient statistics that the optimizer receives, and under which
noise conditions a fractional memory filter provably improves over
exponential forgetting. The Z\"ahle--Ziezold analysis of fractional
derivatives at the critical order provides a natural starting point for
the first question, and stochastic-approximation analyses of the
truncated kernel for the second.

In summary, fractal activation functions and fractional optimizers are
mathematically compatible modifications of two different parts of the
training problem, and their combination is beneficial in specific,
identifiable pairings: subcritical fractal activations with
local-scaling or adaptive fractional optimizers were repeatedly among
the best configurations, while indiscriminate use of explicit gradient
memory was not supported by the data. The unified framework, the
corrected and safeguarded optimizer implementations, and the repeated-run
evaluation protocol of this study provide a reproducible basis for the
further development of mathematically structured optimization methods
and activation functions in deep learning.

\section*{Author Contributions}
Conceptualization, S.R.;
methodology, S.R.;
software, S.R.;
validation, S.R., G.G. and P.K.;
formal analysis, S.R. and G.G.;
investigation, S.R.;
resources, S.R. and S.S.;
data curation, S.R.;
writing---original draft preparation, S.R.;
writing---review and editing, S.R., G.G., S.S., P.K. and K.M.;
visualization, S.R.;
supervision, S.R. and K.M.;
project administration, S.R. and K.M.;
funding acquisition, S.R., S.S. and K.M.
All authors have read and agreed to the published version of the
manuscript.

\section*{Funding}
SBA Research
(SBA-K1 NGC) is a COMET Center within the COMET-Competence Centers for Excellent Technologies Programme and is funded by BMIMI, BMWET, and the federal state of Vienna. The COMET Programme is managed by the Austrian Research Promotion Agency (FFG).
The financial support by the Austrian Federal Ministry of Economy, Energy and Tourism, the National Foundation for Research,
Technology and Development and the Christian Doppler Research Association is gratefully acknowledged.

\section*{Acknowledgements}
During the preparation of this manuscript, the authors
used ChatGPT (OpenAI), Claude (Anthropic), and Grammarly, together with
their associated tools, for the purposes of cleaning and refactoring
code, correcting grammar and typographical errors, and improving the
overall clarity and quality of the manuscript. All ideas, concepts,
formulations, and wording remain the authors' own. The authors have
reviewed and edited all output and take full responsibility for the
content of this publication.

\section*{Data Availability Statement}
A corresponding GitHub repository containing the full code for reproducibility, all employed optimizers, all fractal activation functions, the experiment scripts, and the data downloaders is available at \url{https://github.com/Raubkatz/FractalAndFractional2026}.

\section*{Institutional Review Board Statement}
Not applicable.

\section*{Informed Consent Statement}
Not applicable.

\section*{Declaration of Competing Interest}
The authors declare no conflicts of interest.

\pagebreak
\section*{Abbreviations}
The following abbreviations and optimizer names are used in this
manuscript:\\

\noindent
\begin{tabular}{@{}ll}
GL       & Gr\"unwald--Letnikov (fractional derivative) \\
EMA      & Exponential Moving Average \\
CEF      & Convergence Evaluation Factor (AOFGD order rule) \\
MLP      & Multilayer Perceptron \\
DNN      & Deep Neural Network \\
GAN      & Generative Adversarial Network \\
ReLU     & Rectified Linear Unit \\
SEM      & Standard Error of the Mean \\
OpenML   & Open Machine Learning (dataset repository) \\
API      & Application Programming Interface \\
$P_J$    & Additive fractal (Weierstrass-type) surface perturbation term \\
\end{tabular}

\vspace{1em}
\noindent
\begin{tabular}{@{}ll}
SGD       & Stochastic Gradient Descent \\
RMSprop   & Root Mean Square Propagation \\
Adam      & Adaptive Moment Estimation \\
AdamW     & Adam with decoupled Weight decay \\
Nadam     & Nesterov-accelerated Adaptive Moment estimation \\
Adamax    & Adam variant based on the infinity norm \\
Adagrad   & Adaptive Gradient algorithm \\
Adadelta  & Adaptive Delta (learning-rate) method \\
\end{tabular}

\vspace{1em}
\noindent
\begin{tabular}{@{}ll}
FSGD      & Fractional SGD (Herrera-type, Caputo-scaled) \\
FRMSprop  & Fractional RMSprop (Herrera-type) \\
FAdam     & Fractional Adam (Herrera-type) \\
FAdadelta & Fractional Adadelta (Herrera-type) \\
MemoryFSGD      & Memory-based fractional SGD (Gr\"unwald--Letnikov) \\
MemoryFRMSprop  & Memory-based fractional RMSprop \\
MemoryFAdam     & Memory-based fractional Adam \\
MemoryFAdadelta & Memory-based fractional Adadelta \\
MemoryFAdagrad  & Memory-based fractional Adagrad (Appendix~B) \\
MemoryFAdamW    & Memory-based fractional AdamW (Appendix~B) \\
MemoryFNadam    & Memory-based fractional Nadam (Appendix~B) \\
MemoryFAdamax   & Memory-based fractional Adamax (Appendix~B) \\
\end{tabular}

\vspace{1em}
\noindent
\begin{tabular}{@{}ll}
AdaptiveMemoryFSGD      & Adaptive memory-based fractional SGD \\
AdaptiveMemoryFRMSprop  & Adaptive memory-based fractional RMSprop \\
AdaptiveMemoryFAdam     & Adaptive memory-based fractional Adam \\
AdaptiveMemoryFAdadelta & Adaptive memory-based fractional Adadelta \\
AdaptiveMemoryFAdagrad  & Adaptive memory-based fractional Adagrad (Appendix~B) \\
AdaptiveMemoryFAdamW    & Adaptive memory-based fractional AdamW (Appendix~B) \\
AdaptiveMemoryFNadam    & Adaptive memory-based fractional Nadam (Appendix~B) \\
AdaptiveMemoryFAdamax   & Adaptive memory-based fractional Adamax (Appendix~B) \\
\end{tabular}

\vspace{1em}
\noindent
\begin{tabular}{@{}ll}
FCSGD\_GL  & Fractional-Calculus SGD with Gr\"unwald--Letnikov memory \cite{zhou2023gl} \\
FCAdam\_GL & Fractional-Calculus Adam with Gr\"unwald--Letnikov memory \cite{zhou2023gl} \\
AdaGL      & Adaptive Gr\"unwald--Letnikov optimizer \cite{chen2024adagl} \\
AOFGD      & Adaptive-Order Fractional Gradient Descent \cite{xiang2025aofgd} \\
AOFGD\_SGD  & AOFGD applied to SGD \\
AOFGD\_Adam & AOFGD applied to Adam \\
FracM      & Fractional-order Momentum optimizer \cite{yu2022fracm} \\
\end{tabular}

\vspace{1em}
\noindent
\begin{tabular}{@{}ll}
MDPI & Multidisciplinary Digital Publishing Institute \\
DOAJ & Directory of Open Access Journals \\
\end{tabular}


\appendix

\pagebreak

\section{Notation}
\label{app:notation}

This appendix collects, in one place, the mathematical symbols used in
the article. It is intended as a glossary that a reader can consult
without searching through the text; each entry states the symbol, its
definition, and, where helpful, the section in which it is introduced.

Throughout, lower-case letters in normal font denote scalars, lower-case
vector symbols (e.g.\ $g_t$, $\theta_t$) denote vectors in
$\mathbb{R}^d$, and the subscript $t$ (or, in the discrete-operator
subsection, $n$) denotes a discrete iteration or sampling index unless
stated otherwise. Squares, absolute values, and powers of vectors are
taken componentwise.

\subsection{General Conventions}

\begin{table}[H]
\caption{Basic symbols used throughout the article.\label{tab:notation-general}}
\begin{tabularx}{\textwidth}{lX}
\toprule
\textbf{Symbol} & \textbf{Meaning} \\
\midrule
$t$, $n$ & Discrete iteration (time-step) index $t$; the
discrete-fractional-operator subsection
(\ref{subsec:discrete_fractional_weierstrass_autodiff}) uses $n$ for the
sampling index. \\
$d$ & Dimension of the parameter space, $\theta\in\mathbb{R}^d$. \\
$\theta_t$ & Parameter vector at iteration $t$. \\
$L(\theta)$ & Loss (objective) function. \\
$g_t = \nabla_\theta L(\theta_t)$ & Ordinary (first-order) gradient at
iteration $t$. \\
$\eta>0$ & Base learning rate; the outer step-size multiplier in every
update rule in the article. \\
$K$ & Length of the truncated (finite) gradient history used in every
short-memory Gr\"unwald--Letnikov construction in the article.
Distinguished from the number of scales $J$ of the surface perturbation
(Table~\ref{tab:notation-activation}). \\
$\Gamma(\cdot)$ & Gamma function. \\
$\odot$ & Componentwise (Hadamard) product. \\
$\langle\cdot,\cdot\rangle$, $\|\cdot\|$ & Euclidean inner product and
norm, taken per parameter tensor. \\
$\Xi_t$ & Generic internal state of a base optimizer (velocity, moments,
accumulators) in the abstract update
$\theta_{t+1}=\mathcal{U}(\theta_t,g_t,\Xi_t)$ of
Section~\ref{subsec:herrera_fractional_optimizers}. \\
\bottomrule
\end{tabularx}
\end{table}

\subsection{Fractional Order and Gr\"unwald--Letnikov Coefficients}

\begin{table}[H]
\caption{Notation for the fractional order and the finite-memory
fractional gradients.\label{tab:notation-fractional}}
\begin{tabularx}{\textwidth}{lX}
\toprule
\textbf{Symbol} & \textbf{Meaning} \\
\midrule
$\nu\in(0,2)$ & Fractional order. Held fixed within each optimizer run;
the experiments sweep $\nu\in\{0.75, 1.25, 1.50\}$. The Weyl--Marchaud
analysis of Section~\ref{subsec:weierstrass_fractional_discussion} uses
orders $\nu\in(0,1)$. \\
$c_k^{(\nu)}$ & Gr\"unwald--Letnikov coefficient sequence, defined
recursively by $c_0^{(\nu)}=1$, $c_k^{(\nu)}=\bigl(1-\tfrac{\nu+1}{k}\bigr)
c_{k-1}^{(\nu)}$. Used in
Sections~\ref{subsec:discrete_fractional_weierstrass_autodiff}
and~\ref{subsec:memory_fractional_optimizers}--\ref{subsec:related_work_optimizers}. \\
$\binom{\nu}{k}=\Gamma(\nu+1)/[\Gamma(k+1)\Gamma(\nu-k+1)]$ &
Generalized binomial coefficient; $c_k^{(\nu)}=(-1)^k\binom{\nu}{k}$ is
an equivalent closed-form expression for the same coefficient sequence. \\
$\tilde c_k^{(\nu)}$ & Normalized coefficients,
$\tilde c_k^{(\nu)}=c_k^{(\nu)}/\sum_j|c_j^{(\nu)}|$; used by the
adaptive-memory optimizers. \\
$g_t^{(\nu)}$ & Finite-history fractional gradient,
$g_t^{(\nu)}=\sum_{k=0}^{\min(t,\,K-1)}c_k^{(\nu)}g_{t-k}$. \\
$\tilde g_t^{(\nu)}$ & Descent-safeguarded fractional gradient: equal to
$g_t^{(\nu)}$ if $\langle g_t^{(\nu)},g_t\rangle>0$ and to $g_t$
otherwise (Section~\ref{subsec:memory_fractional_optimizers}). \\
$\hat g_t^{(\nu)}$ & Norm-matched, descent-safeguarded fractional
gradient,
$\hat g_t^{(\nu)} = \tilde g_t^{(\nu)}\,
\|g_t\|/(\|\tilde g_t^{(\nu)}\|+\varepsilon)$. \\
$f_\nu(g_t)$ & Caputo-type multiplicative scaling factor,
$f_\nu(g_t)=(|g_t|+\varepsilon)^{1-\nu}/\Gamma(2-\nu)$, used by the
Herrera-type optimizers
(Section~\ref{subsec:herrera_fractional_optimizers}). The underlying
Caputo power law is derived on the parameter $\theta$, but the
implementations evaluate it on the magnitude of the current gradient.
This is a
\emph{rescaling of the single current gradient} and
is conceptually distinct from $g_t^{(\nu)}$, which is a
\emph{history-weighted sum of several gradients}. \\
$g_t^{\mathrm{sc}}$ & Scaled gradient of the Herrera-type optimizers,
$g_t^{\mathrm{sc}}=f_\nu(g_t)\odot g_t$. \\
$D^\nu$, $D_{h_{\mathrm{step}}}^{\nu}$, ${}_{GL}D_t^\nu$ & Notational
variants for the (continuous or discretized) Gr\"unwald--Letnikov
derivative operator; all refer to the same underlying finite-difference
construction. \\
\bottomrule
\end{tabularx}
\end{table}

\subsection{Adaptive Mixing and Related-Work Control Quantities}

\begin{table}[H]
\caption{Notation for the adaptive mixing coefficient of the
adaptive-memory optimizers
(Section~\ref{subsec:adaptive_memory_optimizers}) and for the control
quantities of the related-work optimizers
(Section~\ref{subsec:related_work_optimizers}).\label{tab:notation-adaptive}}
\begin{tabularx}{\textwidth}{lX}
\toprule
\textbf{Symbol} & \textbf{Meaning} \\
\midrule
$\lambda_t\in[0,1]$ & Adaptive mixing (trust) coefficient, controlling
$g_t^{\mathrm{eff}}=(1-\lambda_t)g_t+\lambda_t\hat g_t^{(\nu)}$. This is
the central quantity of the adaptive memory-based framework. \\
$\lambda_t^{\mathrm{tar}}$ & Ratcheted target value toward which
$\lambda_t$ is moved by EMA smoothing. \\
$\lambda_{\min},\lambda_{\max}$ & Clipping bounds for $\lambda_t$. \\
$\lambda_{\max}^{(t)}$, $T_w$ & Warm-up-ramped upper bound,
$\lambda_{\max}^{(t)}=\lambda_{\max}\min(t/T_w,1)$, with warm-up horizon
$T_w$ (no ramp for $T_w=0$). \\
$\eta_\lambda>0$ & Fixed increment/decrement applied to the target
depending on the stability comparison. Despite the shared letter, this
is unrelated to the learning rate $\eta$. \\
$\Delta_\lambda$ & Loss-triggered decrement applied to the target while
the loss-aware indicator is active; distinct from $\eta_\lambda$. \\
$\mathbf{1}_t^{\mathrm{worse}}$ & Loss-aware indicator
(Table~\ref{tab:notation-loss}). \\
$\gamma_\lambda$ & EMA decay rate for the smoothed mixing coefficient. \\
$g_t^{\mathrm{eff}}$ & Effective gradient passed to the base optimizer
after mixing. \\
$\rho_t$ & Raw gradient-variability ratio,
$\rho_t=\min\bigl(\|g_t-g_{t-1}\|/(\|g_{t-1}\|+\varepsilon),\,
\rho_{\max}\bigr)$. \\
$\bar\rho_t$ & EMA-smoothed variability ratio,
$\bar\rho_t=\gamma_\rho\bar\rho_{t-1}+(1-\gamma_\rho)\rho_t$. \\
$\rho_{\max}$ & Clip value guarding the variability ratio against
outliers. \\
$\gamma_\rho$ & EMA decay rate for $\bar\rho_t$. \\
$\tau_s$ & Stability threshold in the $\lambda_t$ ratchet rule
($\bar\rho_t\le\tau_s$ versus $\bar\rho_t>\tau_s$). \\
$b_k$, $p$ & Independent Bernoulli mask variables,
$\Pr(b_k{=}1)=p$, applied to the history terms of FCSGD\_GL and
FCAdam\_GL. \\
$C_t$ & Componentwise short-term step-size coefficient of AdaGL,
$C_t=1.1-\tfrac{1}{2(1+|g_t-g_{t-1}|)}\in[0.6,1.1)$. \\
$\mathrm{CEF}_t$ & Convergence evaluation factor of the AOFGD methods,
$\mathrm{CEF}_t=\|g_t\|/(\|g_{t-1}\|+\varepsilon)$. \\
$\nu_t$, $\nu_{\min}$, $\nu_{\max}$ & Time-varying fractional order of
the AOFGD methods and its clipping bounds. \\
$\Delta_\nu$, $\tau_{\mathrm{CEF}}$, $\gamma_\nu$ & Order increment,
ratchet threshold, and EMA decay rate of the AOFGD order rule. \\
$\tau_\nu$ & Tolerance of the classical fallback
$g_t^{(\nu)}=g_t$ for $|\nu-1|\le\tau_\nu$
(Section~\ref{subsec:memory_fractional_optimizers}). \\
\bottomrule
\end{tabularx}
\end{table}

\subsection{Loss-Aware Control}

\begin{table}[H]
\caption{Notation for the loss-aware override of
$\lambda_t$.\label{tab:notation-loss}}
\begin{tabularx}{\textwidth}{lX}
\toprule
\textbf{Symbol} & \textbf{Meaning} \\
\midrule
$L_t$ & Instantaneous loss value at iteration $t$. \\
$\bar L_t$ & EMA-smoothed loss,
$\bar L_t=\gamma_L\bar L_{t-1}+(1-\gamma_L)L_t$. \\
$\bar L_{\mathrm{best}}$ & Best (lowest) smoothed loss observed so far. \\
$\delta_L$ & Tolerance defining ``significantly worse'' in the comparison
$\bar L_t>\bar L_{\mathrm{best}}+\delta_L$. Not a division guard;
unrelated to the numerical stabilizer $\varepsilon$. \\
$\gamma_L$ & EMA decay rate for $\bar L_t$. \\
\bottomrule
\end{tabularx}
\end{table}

\subsection{Base-Optimizer State Variables}

\begin{table}[H]
\caption{State variables of the base optimizers and their fractional
variants. In the memory-based and adaptive-memory optimizers, the raw
gradient $g_t$, not the fractional or effective gradient, enters every
squared-gradient accumulator listed
below.\label{tab:notation-optimizers}}
\begin{tabularx}{\textwidth}{lX}
\toprule
\textbf{Symbol} & \textbf{Meaning} \\
\midrule
$v_t^{(\mathrm{mom})}$ & SGD momentum/velocity term,
$v_t^{(\mathrm{mom})}=\beta_m v_{t-1}^{(\mathrm{mom})}-\eta\,g_t$ (with
$g_t$ replaced by the respective fractional or effective gradient in the
fractional variants). \\
$\beta_m$ & Momentum decay rate of the SGD-type updates. \\
$v_t^{(2)}$ & Squared-gradient accumulator of RMSprop, Adam, and
Adadelta, $v_t^{(2)}=\gamma_v v_{t-1}^{(2)}+(1-\gamma_v)g_t^2$ (Adam
uses the decay rate $\beta_2$). \\
$\gamma_v$ & Decay rate of the RMSprop and Adadelta accumulators. \\
$m_t$ & Adam first-moment estimate,
$m_t=\beta_1 m_{t-1}+(1-\beta_1)g_t$ (with the respective gradient
object in the fractional variants). \\
$\beta_1,\beta_2$ & Adam decay rates for $m_t$ and $v_t^{(2)}$. \\
$\hat m_t,\hat v_t^{(2)}$ & Bias-corrected moments,
$\hat m_t=m_t/(1-\beta_1^t)$,
$\hat v_t^{(2)}=v_t^{(2)}/(1-\beta_2^t)$. \\
$v_t^{(\Delta)}$ & Adadelta accumulator of squared past updates. \\
$\Delta\theta_t$ & Adadelta update increment,
$\theta_{t+1}=\theta_t+\eta\,\Delta\theta_t$. \\
$\varepsilon>0$ & Numerical-stability constant; used in every optimizer
denominator, in the norm-matching step, in the variability ratio
$\rho_t$, and in the Caputo-type factor $f_\nu(g_t)$. \\
$e_t$ & Backpropagated downstream error term in the single-neuron
gradient decomposition, $\nabla_{w}\ell_t=e_t\,\phi'(z)\,x$
(Section~\ref{subsec:fractional_gd_fractal_activation}). \\
$\delta\to0^+$ & Vanishing scale parameter (limit variable) in the
Weyl--Marchaud gradual-derivative limit,
$d^{\gamma_H}_-f(x)=\lim_{\delta\to0^+}\dots$
(Section~\ref{subsec:weierstrass_fractional_discussion}); unrelated to
the loss tolerance $\delta_L$. \\
\bottomrule
\end{tabularx}
\end{table}

\subsection{Weierstrass-Type Functions and Continuous Fractional Operators}

\begin{table}[H]
\caption{Notation specific to
Section~\ref{subsec:weierstrass_fractional_discussion}.\label{tab:notation-weierstrass}}
\begin{tabularx}{\textwidth}{lX}
\toprule
\textbf{Symbol} & \textbf{Meaning} \\
\midrule
$h:\mathbb{R}\to\mathbb{R}$ & Periodic generator function of period $1$,
$h(0)=0$, assumed H\"older continuous of order $\beta_H$. Distinguished
by name from the discretization step $h_{\mathrm{step}}$
(Table~\ref{tab:notation-discrete}). \\
$b\in\{2,3,\dots\}$ & Geometric base controlling the frequency growth
of the Weierstrass-type sum
$W_h^{\gamma_H}(x)=\sum_k b^{-k\gamma_H} h(b^k x)$; the same symbol is
used for the frequency base of the activation ladders
(Table~\ref{tab:notation-activation}). \\
$\gamma_H\in(0,\beta_H)$ & Roughness (H\"older) exponent of
$W_h^{\gamma_H}$. The bound $\gamma_H<\beta_H\le1$ applies to the
Weierstrass-type series of
Section~\ref{subsec:weierstrass_fractional_discussion}, where it is
required by the Weyl--Marchaud theory; it does not constrain the
activation ladders of Table~\ref{tab:notation-activation}. \\
$\beta_H\in(0,1]$ & H\"older regularity order of the generator $h$. \\
$C>0$ & Generic H\"older constant,
$|f(x)-f(y)|\le C|x-y|^{\gamma_H}$. \\
$M_h^{\gamma_H}(x)$ & Weierstrass--Mandelbrot function,
$\sum_{k=-\infty}^{\infty}b^{-k\gamma_H}h(b^k x)$, satisfying
$M_h^{\gamma_H}(x)=b^{-\gamma_H}M_h^{\gamma_H}(b x)$. \\
$D^\nu_-,D^\nu_+$ & Left/right Weyl--Marchaud fractional derivative of
order $\nu\in(0,1)$. \\
$d^{\gamma_H}_-f$, $|d^{\gamma_H}_-|f$ & Signed and absolute gradual
derivative in the mean at the critical order $\nu=\gamma_H$. \\
\bottomrule
\end{tabularx}
\end{table}

\subsection{Discrete Fractional Operators and Network-Layer Notation}

\begin{table}[H]
\caption{Notation specific to
Section~\ref{subsec:discrete_fractional_weierstrass_autodiff}.\label{tab:notation-discrete}}
\begin{tabularx}{\textwidth}{lX}
\toprule
\textbf{Symbol} & \textbf{Meaning} \\
\midrule
$h_{\mathrm{step}}>0$ & Discretization (sampling) step,
$t_n=n\,h_{\mathrm{step}}$; also the algorithmic step scale in the
update $\theta_{t+1}=\theta_t-\eta\, h_{\mathrm{step}}^{-\nu}
g_t^{(\nu)}$ of
Section~\ref{subsec:fractional_gd_fractal_activation}. \\
$u_n:=u(t_n)$ & Sampled sequence on the discretization grid. \\
$D_{h_{\mathrm{step}}}^{\nu} u_n$ & Discrete backward
Gr\"unwald--Letnikov derivative,
$D_{h_{\mathrm{step}}}^{\nu} u_n=h_{\mathrm{step}}^{-\nu}
\sum_{k=0}^n(-1)^k\binom{\nu}{k}u_{n-k}$. \\
$\mathbf{D}_{h_{\mathrm{step}}}^{(\nu)}$ & Lower-triangular Toeplitz
matrix representation of the discrete derivative acting on the vector
$\mathbf{u}=(u_0,\dots,u_T)^\top$. \\
$D_{h_{\mathrm{step}},K}^{\nu} u_n$ & Truncated (finite-memory) discrete
derivative with memory window $K$. \\
$\Delta x$, $x_n=n\Delta x$ & Spatial grid and step used when sampling
the Weierstrass function, $W_n:=W_h^{\gamma_H}(x_n)$. \\
$\phi_\nu(z)=\mathcal{F}_\nu[\phi_0](z)$ & Fractionalized scalar
activation; $\phi_0$ is a base activation, $z$ the pre-activation, and
$\mathcal{F}_\nu$ the fractional operator acting on $\phi_0$. \\
$a^{(\ell)}$, $\mathbf{W}^{(\ell)}$, $b^{(\ell)}$ & Layer-$\ell$
activation, weight matrix, and bias. \\
\bottomrule
\end{tabularx}
\end{table}

\subsection{Fractal Activation Functions and Single-Neuron Notation}

\begin{table}[H]
\small
\caption{Notation for the fractal activation functions
(Section~\ref{sec:fractal_activations}) 
and the single-neuron analysis of
Section~\ref{subsec:fractional_gd_fractal_activation}.\label{tab:notation-activation}}
\begin{tabularx}{\textwidth}{lX}
\toprule
\textbf{Symbol} & \textbf{Meaning} \\
\midrule
$\phi(x)=\tanh(x)+\sum_{m=0}^{N-1}(-1)^m ...$ & Modified Weierstrass--Tanh activation. \\
$a\in(0,1)$ & Amplitude decay rate of the oscillatory ladder; related to
the roughness exponent by $a=b^{-\gamma_H}$, i.e.\
$\gamma_H=-\ln a/\ln b$. This identity is a translation between the two
parameterizations and is not subject to the bound $\gamma_H\le1$ of
Table~\ref{tab:notation-weierstrass}: since $ab<1$ is equivalent to
$\gamma_H>1$, the representative choice $a=0.5$, $b=1.5$
($ab=0.75$, $\gamma_H\approx1.71$) places the truncated ladder in the
differentiable regime by design, which is what makes the closed-form
derivative \eqref{eq:mw_tanh_derivative} available. \\
$b>1$ & Frequency growth base of the oscillatory ladder (same role as
the Weierstrass base $b$ of Table~\ref{tab:notation-weierstrass}). \\
$\mu_e>0$ & Envelope decay rate of the activation. \\
$N$ & Truncation length of the activation series; plays the same
``truncation length'' role as the memory length $K$, on the activation
side. \\
$\zeta$ & Global scale constant of the decaying-cosine activation,
chosen so that the aggregated backbone is approximately
$\tanh(\pi x)$. \\
$a_{\mathrm{mod}}$ & Gate/offset modulation constant of the modulated
Blancmange activation. \\
$c_{\mathrm{lin}}$ & Aggregated linear-drift coefficient of the
Weierstrass--Mandelbrot $x{+}\sin$ activation,
$c_{\mathrm{lin}}=\sum_{k=1}^{N}2^{-k\gamma_H}
\to 1/(2^{\gamma_H}-1)$. \\
$z=w^\top x+b$ & Pre-activation of a single neuron; here $w$ is the
per-neuron weight vector and $b$ the bias of that neuron
(Section~\ref{subsec:fractional_gd_fractal_activation}). \\
$\Lambda_t^{(\nu)}$, $\theta_{\mathrm{ref}}$ & Non-hereditary fractional
preconditioner,
$\Lambda_t^{(\nu)}=\mathrm{diag}\bigl(\Gamma(2-\nu)^{-1}
(|\theta_t-\theta_{\mathrm{ref}}|+\varepsilon)^{1-\nu}\bigr)$, and its
reference point $\theta_{\mathrm{ref}}$. \\
$u_t$ & Generic combined update vector used to unify the hereditary and
non-hereditary cases in
Equation~\eqref{eq:optimizer_split_notation}. \\
$P_J(x,y)$ & Additive fractal surface perturbation,
$P_J(x,y)=\sum_{k=0}^{J}a_P^{\,k}[\cos(b_P^{\,k}\pi x)+\cos(b_P^{\,k}\pi y)]$
(Section~\ref{subsec:surface_design}). 
The subscripted names keep the perturbation parameters distinct from the
activation-ladder parameters $a$ and $b$ above and from the optimizer
memory length $K$ of Table~\ref{tab:notation-general}; the three
constructions share the geometric-ladder form but are parameterized
independently. \\
$a_P\in(0,1)$, $b_P>1$, $J$ & Amplitude decay rate, frequency base, and
number of scales of the surface perturbation $P_J$. \\
$\alpha_{\mathrm{add}}>0$ & Weight of the additive surface perturbation
in $f_{\mathrm{add}}(x,y)=f(x,y)+\alpha_{\mathrm{add}}P_J(x,y)$; set to
$\alpha_{\mathrm{add}}=0.2$ on both the Ackley and the Himmelblau
surface. Unrelated to the adaptive mixing coefficient $\lambda_t$ of
Table~\ref{tab:notation-adaptive}. \\
\bottomrule
\end{tabularx}
\end{table}

\section{Extensions of the Memory-Based Optimizer Families}
\label{app:additional_memory_optimizers}

This appendix describes how the fixed-memory and adaptive-memory optimizer
constructions used in the main text can be extended to further first-order
optimizers. The main comparison focuses on the four base families SGD,
RMSprop, Adam, and Adadelta. The same constructions also apply to Adagrad,
AdamW, Nadam, and Adamax. These extensions use the same notation and
safeguards as Sections~\ref{subsec:memory_fractional_optimizers}
and~\ref{subsec:adaptive_memory_optimizers}. They are included here to show
that the proposed memory mechanisms are not restricted to the optimizer
families used in the main experiments.

\subsection{Fixed Memory-Based Extensions}
\label{subsec:appendix_additional_memory_optimizers}

\,\par\noindent\textbf{Common construction. }
All fixed memory-based variants in this appendix use the same finite-history
Gr\"unwald--Letnikov gradient as the memory-based optimizers in the main
text. The raw fractional gradient is first passed through the descent
safeguard of \eqref{eq:memory_descent_safeguard} and then through the norm
matching step of \eqref{eq:memory_norm_matching}. The resulting update
direction is therefore $\hat g_t^{(\nu)}$. For optimizers with
coordinatewise scale estimates, the squared-gradient or norm accumulator is
computed from the raw gradient $g_t$ by default. Thus, the
fractional-memory term changes the update direction, while the scale
estimates remain tied to the raw gradient.

\,\par\noindent\textbf{MemoryFAdagrad. }
MemoryFAdagrad applies the fixed-memory direction to Adagrad. With
accumulator $v_t^{(2)}$, the update is
\begin{equation}
\label{eq:appendix_memory_fadagrad}
v_t^{(2)}
=
v_{t-1}^{(2)}+g_t^2,
\qquad
\theta_{t+1}
=
\theta_t
-
\eta\,
\frac{\hat g_t^{(\nu)}}{\sqrt{v_t^{(2)}}+\varepsilon}.
\end{equation}
The method keeps Adagrad's monotone coordinatewise accumulator, but replaces
the numerator by the safeguarded fractional-memory direction. It therefore
inherits Adagrad's decreasing effective learning rates while allowing the
search direction to depend on a short gradient history.

\,\par\noindent\textbf{MemoryFAdamW. }
MemoryFAdamW is the AdamW counterpart of MemoryFAdam. First, the
MemoryFAdam step is computed,
\begin{equation}
\label{eq:appendix_memory_fadamw_inner}
m_t
=
\beta_1 m_{t-1}+(1-\beta_1)\hat g_t^{(\nu)},
\qquad
v_t^{(2)}
=
\beta_2 v_{t-1}^{(2)}+(1-\beta_2)g_t^2,
\end{equation}
followed by the usual bias correction and Adam-type parameter update. If
the intermediate value after this update is denoted by
$\theta_{t+1/2}$, the decoupled weight-decay step is
\begin{equation}
\label{eq:appendix_memory_fadamw_decay}
\theta_{t+1}
=
\theta_{t+1/2}
-
\eta\,\omega\,\theta_{t+1/2},
\end{equation}
where $\omega\ge0$ is the weight-decay coefficient. The weight decay is
therefore separated from the gradient estimate, as in AdamW. The fractional
memory affects the Adam direction, not the definition of the decay term.

\,\par\noindent\textbf{MemoryFNadam. }
MemoryFNadam is the Nesterov-accelerated Adam variant with the same
memory-based direction. The first and second accumulators are
\begin{equation}
\label{eq:appendix_memory_fnadam_moments}
m_t
=
\beta_1 m_{t-1}+(1-\beta_1)\hat g_t^{(\nu)},
\qquad
v_t^{(2)}
=
\beta_2 v_{t-1}^{(2)}+(1-\beta_2)g_t^2 .
\end{equation}
The Nesterov-style first-moment estimate is
\begin{equation}
\label{eq:appendix_memory_fnadam_nesterov}
\bar m_t
=
\frac{\beta_1 m_t}{1-\beta_1^{t+1}}
+
\frac{(1-\beta_1)\hat g_t^{(\nu)}}{1-\beta_1^t},
\qquad
\hat v_t^{(2)}
=
\frac{v_t^{(2)}}{1-\beta_2^t}.
\end{equation}
The parameter update is then
\begin{equation}
\label{eq:appendix_memory_fnadam_update}
\theta_{t+1}
=
\theta_t
-
\eta\,
\frac{\bar m_t}{\sqrt{\hat v_t^{(2)}}+\varepsilon}.
\end{equation}
This variant uses the fractional-memory direction both in the first moment
and in the look-ahead term of the Nesterov correction. The second-moment
estimate remains based on the raw gradient.

\,\par\noindent\textbf{MemoryFAdamax. }
MemoryFAdamax is the Adamax variant, where the second-order scale estimate
is replaced by an exponentially weighted infinity-norm accumulator. The
update is
\begin{equation}
\label{eq:appendix_memory_fadamax}
m_t
=
\beta_1 m_{t-1}+(1-\beta_1)\hat g_t^{(\nu)},
\qquad
v_t^{(\infty)}
=
\max\!\bigl(\beta_2 v_{t-1}^{(\infty)},\, |g_t|\bigr),
\end{equation}
followed by
\begin{equation}
\label{eq:appendix_memory_fadamax_update}
\theta_{t+1}
=
\theta_t
-
\frac{\eta}{1-\beta_1^t}\,
\frac{m_t}{v_t^{(\infty)}+\varepsilon}.
\end{equation}
The method is useful as a memory-based counterpart to Adamax: the numerator
uses the safeguarded fractional-memory direction, while the infinity-norm
scale remains driven by the raw gradient.

\,\par\noindent\textbf{Interpretation. }
The four fixed-memory variants above follow the same design principle as
MemoryFSGD, MemoryFRMSprop, MemoryFAdam, and MemoryFAdadelta. The
Gr\"unwald--Letnikov history modifies only the directional input. The
accumulators that estimate squared gradients or coordinatewise scales remain
classical by default. This separation avoids feeding the cancellation-prone
fractional gradient directly into the scale estimates. At $\nu=1$, or when
the history length is one, the explicit fallback convention reduces these
methods to their corresponding base optimizers.

\subsection{Adaptive Memory-Based Extensions}
\label{subsec:appendix_additional_adaptive_memory_optimizers}

\,\par\noindent\textbf{Common construction. }
The adaptive variants use the same fractional gradient, descent safeguard,
norm matching, and adaptive mixing coefficient as
Section~\ref{subsec:adaptive_memory_optimizers}. We write the effective
gradient as
\begin{equation}
\label{eq:appendix_adaptive_direction}
g_t^{\mathrm{eff}}
=
(1-\lambda_t)g_t+\lambda_t\hat g_t^{(\nu)} .
\end{equation}
The coefficient $\lambda_t$ is adapted by the same stability- and
loss-aware rule described in
\eqref{eq:lambda_target}--\eqref{eq:lambda_ratchet}. The raw gradient
$g_t$ is again used for squared-gradient and norm accumulators by default.
Thus, $\lambda_t=0$ gives the exact base optimizer. The same holds at
$\nu=1$ under the classical fallback convention.

\,\par\noindent\textbf{AdaptiveMemoryFAdagrad. }
AdaptiveMemoryFAdagrad inserts the adaptive effective gradient into the
Adagrad update:
\begin{equation}
\label{eq:appendix_adaptive_fadagrad}
v_t^{(2)}
=
v_{t-1}^{(2)}+g_t^2,
\qquad
\theta_{t+1}
=
\theta_t
-
\eta\,
\frac{g_t^{\mathrm{eff}}}{\sqrt{v_t^{(2)}}+\varepsilon}.
\end{equation}
Compared with MemoryFAdagrad, the contribution of the memory direction is
not fixed. It is increased when recent gradients are stable and decreased
when they become inconsistent or when the smoothed loss worsens.

\,\par\noindent\textbf{AdaptiveMemoryFAdamW. }
AdaptiveMemoryFAdamW combines AdaptiveMemoryFAdam with decoupled weight
decay. The Adam-type moments are
\begin{equation}
\label{eq:appendix_adaptive_fadamw_inner}
m_t
=
\beta_1 m_{t-1}+(1-\beta_1)g_t^{\mathrm{eff}},
\qquad
v_t^{(2)}
=
\beta_2 v_{t-1}^{(2)}+(1-\beta_2)g_t^2 .
\end{equation}
After the bias-corrected Adam update gives $\theta_{t+1/2}$, the decoupled
decay step is
\begin{equation}
\label{eq:appendix_adaptive_fadamw_decay}
\theta_{t+1}
=
\theta_{t+1/2}
-
\eta\,\omega\,\theta_{t+1/2}.
\end{equation}
The adaptive memory mechanism controls the Adam direction through
$g_t^{\mathrm{eff}}$, while the weight decay remains independent of the
gradient and of the fractional-memory kernel.

\,\par\noindent\textbf{AdaptiveMemoryFNadam. }
AdaptiveMemoryFNadam applies the adaptive effective gradient to the Nadam
update. The moments are
\begin{equation}
\label{eq:appendix_adaptive_fnadam_moments}
m_t
=
\beta_1 m_{t-1}+(1-\beta_1)g_t^{\mathrm{eff}},
\qquad
v_t^{(2)}
=
\beta_2 v_{t-1}^{(2)}+(1-\beta_2)g_t^2 .
\end{equation}
The Nesterov-style first-moment estimate is
\begin{equation}
\label{eq:appendix_adaptive_fnadam_nesterov}
\bar m_t
=
\frac{\beta_1 m_t}{1-\beta_1^{t+1}}
+
\frac{(1-\beta_1)g_t^{\mathrm{eff}}}{1-\beta_1^t},
\qquad
\hat v_t^{(2)}
=
\frac{v_t^{(2)}}{1-\beta_2^t},
\end{equation}
and the update is
\begin{equation}
\label{eq:appendix_adaptive_fnadam_update}
\theta_{t+1}
=
\theta_t
-
\eta\,
\frac{\bar m_t}{\sqrt{\hat v_t^{(2)}}+\varepsilon}.
\end{equation}
This gives a conservative Nadam extension: the method is exactly Nadam when
$\lambda_t=0$ or $\nu=1$, and it becomes more memory-aware only when the
adaptive mixing rule permits it.

\,\par\noindent\textbf{AdaptiveMemoryFAdamax. }
AdaptiveMemoryFAdamax uses the adaptive effective gradient in the Adamax
first moment and the raw gradient in the infinity-norm accumulator:
\begin{equation}
\label{eq:appendix_adaptive_fadamax}
m_t
=
\beta_1 m_{t-1}+(1-\beta_1)g_t^{\mathrm{eff}},
\qquad
v_t^{(\infty)}
=
\max\!\bigl(\beta_2 v_{t-1}^{(\infty)},\, |g_t|\bigr).
\end{equation}
The parameter update is
\begin{equation}
\label{eq:appendix_adaptive_fadamax_update}
\theta_{t+1}
=
\theta_t
-
\frac{\eta}{1-\beta_1^t}\,
\frac{m_t}{v_t^{(\infty)}+\varepsilon}.
\end{equation}
As in the other adaptive variants, the memory contribution is bounded by
$\lambda_t$ and can be reduced during unstable phases of training.

\,\par\noindent\textbf{Interpretation. }
The adaptive-memory variants above extend the same mechanism to four
further classical optimizers. They do not introduce an adaptive fractional
order. The order $\nu$ remains fixed, the finite-history kernel remains the
same, and only the scalar mixing coefficient $\lambda_t$ changes during
training. This keeps the interpretation aligned with
Section~\ref{subsec:adaptive_memory_optimizers}: the optimizer interpolates
between a classical update and a safeguarded fractional-memory update.

\subsection{Summary of the Optimizer-Family Extensions}
\label{subsec:appendix_additional_optimizer_summary}

Table~\ref{tab:appendix_additional_optimizer_summary} summarizes the two
optimizer-family extension groups. They are not part of the 21 optimizers
compared in the main experimental tables. Instead, they show how the same
fixed-memory and adaptive-memory principles transfer to further first-order
methods. Reading the table by group: the fixed memory-based extensions use
the safeguarded, norm-matched Gr\"unwald--Letnikov gradient
$\hat g_t^{(\nu)}$ as update direction, keep all scale estimates on the raw
gradient $g_t$, and recover their corresponding base optimizers at
$\nu=1$. The adaptive memory-based extensions mix $g_t$ with the same
safeguarded memory gradient through the bounded coefficient $\lambda_t$,
keep all scale estimates on the raw gradient, and recover their base
optimizers at $\lambda_t=0$ or $\nu=1$.

\begin{table}[H]
\caption{Memory-based and adaptive memory-based extensions to further
first-order optimizers beyond the optimizer families used in the main
comparison. For the fixed memory-based variants, the update direction is
$\hat g_t^{(\nu)}$. For the adaptive memory-based variants, the update
direction is $(1-\lambda_t)g_t+\lambda_t\hat g_t^{(\nu)}$. The scale
estimates are based on the raw gradient $g_t$ by default.
\label{tab:appendix_additional_optimizer_summary}}
\small
\begin{tabularx}{\textwidth}{lXXcc}
\toprule
\textbf{Group} & \textbf{Optimizers} & \textbf{Update direction} &
\textbf{Memory} & \textbf{Fallback} \\
\midrule
Memory-based extensions &
MemoryFAdagrad, MemoryFAdamW, MemoryFNadam, MemoryFAdamax &
$\hat g_t^{(\nu)}$, fixed $\nu$; accumulators and scale estimates from $g_t$ &
$K$ slots & $\nu=1$ \\
\addlinespace[0.3em]
Adaptive memory extensions &
AdaptiveMemory FAdagrad, AdaptiveMemoryFAdamW, AdaptiveMemoryFNadam,
AdaptiveMemoryFAdamax &
$(1-\lambda_t)\,g_t+\lambda_t\,\hat g_t^{(\nu)}$, fixed $\nu$;
accumulators and scale estimates from $g_t$ &
$K$ slots & $\lambda_t=0$ or $\nu=1$ \\
\bottomrule
\end{tabularx}
\end{table}

These extensions show that the proposed memory mechanisms are not tied to
the four optimizer families selected for the main experiments. They can be
transferred to further first-order methods by applying the same rule: the
safeguarded fractional-memory direction is used where the base optimizer
expects a descent direction, while the scale estimates remain based on the
raw gradient unless specified otherwise.

\section{Runtimes of the Experiments}
\label{app:runtimes}

This appendix collects the runtime measurements of both experimental
blocks. Its purpose is to make the computational cost of the compared
optimizer families explicit, since the main text interprets accuracy and
success-rate differences that are, in part, bought with additional
computation. Two different quantities are reported. For the surface
experiments, the natural unit is the mean wall-clock time per
optimization step, because every optimizer executes the same number of
steps on the same objective and the per-step time therefore isolates the
cost of the update rule itself. For the neural-network experiments, the
reported quantity is the mean training time of each optimizer's
best-performing configuration per dataset, averaged over the ten
datasets; this is the cost that accompanies the accuracy values reported
in Section~\ref{sec:nn_experiments}. 
All values are averages over 40 repeated runs per configuration and were
obtained on the same hardware within each experimental block, so the
numbers are comparable within each table but should not be compared
across tables or transferred to other hardware.

\subsection{Per-Step Cost in the Surface Experiments}
\label{app:runtimes_surfaces}

Table~\ref{tab:runtime_surfaces} reports the mean time per optimization
step for all 21 optimizers on both surfaces and both variants. Three
regularities are visible. First, within each variant the ordering follows
the amount of per-step work of the update rule. The four baselines are
the cheapest methods (2.3--3.7\,ms per step on the standard surfaces).
The Herrera-type optimizers add one componentwise power and one
multiplication and cost between roughly $-2$ and $+55\%$ relative to their baselines, with a median overhead near $10\%$. The
memory-based optimizers add the circular-buffer write and the $K$-term
kernel sum and lie roughly 40--105\% above the baselines on the standard surfaces and 9--25\% above them on the additive variants. The
adaptive-memory optimizers additionally compute the stability score, the
mixing update, and the norm matching, and the related-work methods carry
comparable machinery (Bernoulli masking and $K=11$ history terms for
FCSGD\_GL and FCAdam\_GL, the history sum plus the step coefficient for
AdaGL); both groups cost roughly 2--3 times the baseline step. The
adaptive-order AOFGD methods, which store no history, sit between the
Herrera and memory groups.

Second, the additive fractal variant shifts all per-step times upward by
an almost constant amount of approximately 5.1--5.8\,ms on both surfaces.
This offset is the cost of evaluating the perturbation ladder $P_J$ and
its gradient at every step, and it affects every optimizer equally. As a
consequence, the \emph{relative} overhead of the more complex optimizers
shrinks on the perturbed surfaces: FCAdam\_GL costs about 3.2 times as
much per step as SGD on the standard Himmelblau surface, but only about
1.5 times as much on the additive variant, because the objective
evaluation rather than the update rule dominates the step. This is the
relevant regime for practical problems with expensive objectives: the
more elaborate update rules are comparatively cheaper the more expensive
the gradient computation is.

Third, the two surfaces agree closely. For every optimizer, the per-step
times on Ackley and Himmelblau differ by fractions of a millisecond
within the same variant, which confirms that the measured differences
are properties of the update rules and of the perturbation, not of the
specific objective.

\begin{table}[H]
\caption{Mean wall-clock time per optimization step in the surface
experiments, in milliseconds, averaged over 40 runs per configuration.
Columns give the standard and additive variants of both surfaces.
Horizontal blocks correspond to the five optimizer
groups.\label{tab:runtime_surfaces}}
\small
\begin{tabularx}{\textwidth}{lcccc}
\toprule
\textbf{Optimizer} &
\textbf{Ackley std.} & \textbf{Ackley add.} &
\textbf{Himmelblau std.} & \textbf{Himmelblau add.} \\
\midrule
SGD      & 3.10 & 8.79 & 2.28 & 8.08 \\
RMSprop  & 3.72 & 9.14 & 3.08 & 8.88 \\
Adam     & 3.46 & 9.19 & 2.90 & 8.59 \\
Adadelta & 3.15 & 8.53 & 2.29 & 8.05 \\
\midrule
FSGD      & 3.40 & 9.05 & 2.84 & 8.58 \\
FRMSprop  & 3.63 & 9.38 & 3.15 & 8.85 \\
FAdam     & 4.26 & 9.72 & 3.82 & 9.61 \\
FAdadelta & 4.00 & 9.45 & 3.54 & 9.19 \\
\midrule
MemoryFSGD      & 4.39 & 9.99  & 4.18 & 9.59  \\
MemoryFRMSprop  & 4.78 & 10.32 & 4.32 & 9.67  \\
MemoryFAdam     & 5.22 & 10.99 & 4.94 & 10.19 \\
MemoryFAdadelta & 5.01 & 10.56 & 4.69 & 10.07 \\
\midrule
AdaptiveMemoryFSGD      & 6.10 & 11.53 & 6.03 & 11.37 \\
AdaptiveMemoryFRMSprop  & 6.25 & 12.11 & 6.36 & 11.45 \\
AdaptiveMemoryFAdam     & 6.80 & 12.78 & 6.97 & 12.00 \\
AdaptiveMemoryFAdadelta & 6.47 & 12.32 & 6.58 & 11.82 \\
\midrule
FCSGD\_GL   & 6.52 & 12.07 & 6.33 & 11.69 \\
FCAdam\_GL  & 7.47 & 12.72 & 7.38 & 12.48 \\
AdaGL       & 6.75 & 12.08 & 6.51 & 11.66 \\
AOFGD\_SGD  & 4.84 & 10.40 & 4.55 & 9.86  \\
AOFGD\_Adam & 5.54 & 11.32 & 5.37 & 10.68 \\
\bottomrule
\end{tabularx}
\end{table}

\subsection{Training Time in the Classification Experiments}
\label{app:runtimes_classification}

Table~\ref{tab:runtime_classification} reports, for each optimizer, the
mean training time of its best per-dataset configuration, averaged over
the ten datasets, together with the corresponding group means. These
numbers accompany the cross-dataset accuracy ranking of
Section~\ref{sec:nn_experiments} 
and answer the practical question of what the leading configurations
cost. The Herrera group combines the highest cross-dataset accuracy with
the second-lowest mean training time (6.5\,s), and its two strongest
members, FAdam and FAdadelta, train in 5.5\,s and 6.0\,s on average. The
adaptive-memory group is the most expensive on average (9.9\,s), which
reflects the additional per-step machinery quantified in
Table~\ref{tab:runtime_surfaces}; its accuracy advantage over the
standard group is therefore paid for with roughly 20\% more training
time.

One caution applies to reading this table. Unlike the surface
measurements, these times are not pure optimizer overheads. Each row
refers to the optimizer's \emph{best} configuration per dataset, and
these configurations differ in the selected activation function, whose
evaluation cost varies substantially: the fractal activations are
truncated series with 30--100 terms and cost correspondingly more per
forward and backward pass than ReLU or tanh. The training time of a row
therefore mixes the optimizer's per-step cost with the activation choice
of its winning configurations. This explains apparent inversions
relative to Table~\ref{tab:runtime_surfaces}: the GL-Memory group shows
the lowest mean training time (5.3\,s) although its per-step optimizer
cost exceeds that of the baselines, because its winning configurations
frequently used cheaper activations, and the same effect raises the
times of baselines whose best configurations used expensive fractal
activations (for example RMSprop at 9.7\,s). Within-group comparisons of
optimizers with similar winning activations are reliable; across-group
comparisons should be made together with the accuracy tables of the main
text.

\begin{table}[H]
\caption{Mean training time of the best per-dataset configuration for
each optimizer in the classification experiments, averaged over the ten
datasets (40 runs per configuration). The group rows give the mean over
the optimizers of each group. Times include the full training of the
respective network and therefore reflect both the optimizer cost and the
activation function selected by the winning
configuration.\label{tab:runtime_classification}}
\small
\begin{tabularx}{\textwidth}{lXc}
\toprule
\textbf{Optimizer} & \textbf{Group} & \textbf{Mean training time (s)} \\
\midrule
SGD      & Standard & 5.57 \\
RMSprop  & Standard & 9.69 \\
Adam     & Standard & 8.37 \\
Adadelta & Standard & 8.88 \\
\emph{Group mean} & Standard & 8.13 \\
\midrule
FSGD      & Herrera & 4.68 \\
FRMSprop  & Herrera & 9.81 \\
FAdam     & Herrera & 5.51 \\
FAdadelta & Herrera & 5.97 \\
\emph{Group mean} & Herrera & 6.49 \\
\midrule
MemoryFSGD      & GL-Memory & 6.10 \\
MemoryFRMSprop  & GL-Memory & 4.00 \\
MemoryFAdam     & GL-Memory & 6.18 \\
MemoryFAdadelta & GL-Memory & 4.91 \\
\emph{Group mean} & GL-Memory & 5.30 \\
\midrule
AdaptiveMemoryFSGD      & Adaptive-Memory & 6.35 \\
AdaptiveMemoryFRMSprop  & Adaptive-Memory & 10.92 \\
AdaptiveMemoryFAdam     & Adaptive-Memory & 12.09 \\
AdaptiveMemoryFAdadelta & Adaptive-Memory & 10.27 \\
\emph{Group mean} & Adaptive-Memory & 9.91 \\
\midrule
FCSGD\_GL   & Related-Work & 6.52 \\
FCAdam\_GL  & Related-Work & 10.37 \\
AdaGL       & Related-Work & 4.26 \\
AOFGD\_SGD  & Related-Work & 7.21 \\
AOFGD\_Adam & Related-Work & 10.23 \\
\emph{Group mean} & Related-Work & 7.72 \\
\bottomrule
\end{tabularx}
\end{table}

\subsection{Summary}
\label{app:runtimes_summary}
The runtime measurements support three statements. First, the per-step
overhead of the fractional mechanisms is moderate and predictable:
typically around 10\% and up to 55\% for the Herrera-type scaling,
40--105\% for the safeguarded
Gr\"unwald--Letnikov memory on the standard surfaces, and a factor of 2--3 for the adaptive-memory
and related-work methods, measured against the corresponding baselines
on cheap objectives. Second, this relative overhead shrinks as the
objective becomes more expensive, as the additive surface variants show;
in neural-network training, where the forward and backward passes
dominate, the optimizer choice changes the total training time far less
than these per-step factors suggest. Third, the total training times of
the classification experiments are governed at least as much by the
activation choice of the winning configurations as by the optimizer, so
runtime comparisons between method families should always be read
together with the corresponding accuracy tables. In absolute terms, all
compared methods remain inexpensive for the network sizes of this study:
the most expensive mean best-configuration training time was
approximately 12\,s, and the complete experimental programme of
$88\,800$ training runs and $2\times1680$ surface runs was feasible on
standard hardware.




\bibliographystyle{elsarticle-num}
\bibliography{references}


%




\end{document}